\documentclass[letterpaper]{article} 
\usepackage{aaai25}  
\usepackage{times}  
\usepackage{helvet}  
\usepackage{courier}  
\usepackage[hyphens]{url}  
\usepackage{graphicx} 
\usepackage{natbib}  
\usepackage{caption} 
\usepackage{algorithm}
\usepackage{algorithmic}
\usepackage{subcaption}
\usepackage{bm}

\usepackage{newfloat}
\usepackage{listings}
\usepackage{amssymb}
\usepackage{dsfont}
\usepackage{amsmath}
\usepackage{multirow}

\usepackage{xcolor}
\newcommand{\answerYes}[1]{\textcolor{blue}{#1}}

\usepackage{newfloat}
\usepackage{listings}
\DeclareCaptionStyle{ruled}{labelfont=normalfont,labelsep=colon,strut=off} 
\floatstyle{ruled}
\newfloat{listing}{tb}{lst}{}
\floatname{listing}{Listing}

\def\correspondingauthor{%
  \ifnum\value{correspondingfn}=0%
    \footnote{Corresponding authors: Chen Gong, Shuo Chen.}%
    \setcounter{correspondingfn}{\value{footnote}}%
  \else%
    \footnotemark[\value{correspondingfn}]%
  \fi%
}

\newcounter{correspondingfn}

\title{Modeling Inter-Intra Heterogeneity for Graph Federated Learning}
\author{
    Wentao Yu\textsuperscript{\rm 1},
    Shuo Chen\textsuperscript{\rm 2}\correspondingauthor{},
    Yongxin Tong\textsuperscript{\rm 3},
    Tianlong Gu\textsuperscript{\rm 4},
    Chen Gong\textsuperscript{\rm 5}\correspondingauthor{}
}
\affiliations{

    \textsuperscript{\rm 1}School of Computer Science and Engineering, Nanjing University of Science and Technology, China\\
    \textsuperscript{\rm 2}Center for Advanced Intelligence Project, RIKEN, Japan\\
    \textsuperscript{\rm 3}State Key Laboratory of Complex \& Critical Software Environment, Beihang University, China\\
    \textsuperscript{\rm 4}Engineering Research Center of Trustworthy AI (Ministry of Education), Jinan University, China\\
    \textsuperscript{\rm 5}Department of Automation, Institute of Image Processing and Pattern Recognition, Shanghai Jiao Tong University, China\\

    wentao.yu@njust.edu.cn, shuo.chen.ya@riken.jp, yxtong@buaa.edu.cn, gutianlong@jnu.edu.cn, chen.gong@sjtu.edu.cn
}

\usepackage{bibentry}

\begin{document}

\maketitle

\begin{abstract}
Heterogeneity is a fundamental and challenging issue in federated learning, especially for the graph data due to the complex relationships among the graph nodes. To deal with the heterogeneity, lots of existing methods perform the weighted federation based on their calculated similarities between pairwise clients (\textit{i.e.}, subgraphs). However, their inter-subgraph similarities estimated with the outputs of local models are less reliable, because the final outputs of local models may not comprehensively represent the real distribution of subgraph data. In addition, they ignore the critical intra-heterogeneity which usually exists within each subgraph itself. To address these issues, we propose a novel \textbf{Fed}erated learning method by integrally modeling the \textbf{I}nter-\textbf{I}ntra \textbf{H}eterogeneity (FedIIH). For the inter-subgraph relationship, we propose a novel hierarchical variational model to infer the whole distribution of subgraph data in a multi-level form, so that we can accurately characterize the inter-subgraph similarities with the global perspective. For the intra-heterogeneity, we disentangle the subgraph into multiple latent factors and partition the model parameters into multiple parts, where each part corresponds to a single latent factor. Our FedIIH not only properly computes the distribution similarities between subgraphs, but also learns disentangled representations that are robust to irrelevant factors within subgraphs, so that it successfully considers the inter- and intra- heterogeneity simultaneously. Extensive experiments on six homophilic and five heterophilic graph datasets in both non-overlapping and overlapping settings demonstrate the effectiveness of our method when compared with nine state-of-the-art methods. Specifically, FedIIH averagely outperforms the second-best method by a large margin of 5.79\% on all heterophilic datasets. Code is available at \url{https://github.com/blgpb/FedIIH}.
\end{abstract}

%

\section{Introduction}

Graphs are ubiquitous data structures in lots of important domains such as social media, transportation, and recommendation systems. In many real-world scenarios, a global graph is usually made up of multiple subgraphs that are distributed across devices, and subgraphs are only locally accessible due to privacy and regulatory concerns. Recently, Graph Federated Learning (GFL) has received increasing attention~\cite{he2022spreadgnn, wang2023fedgs, tan2023federated, ijcai2023426}, where each client individually trains a local model based on the corresponding subgraph, and a central server adaptively aggregates the models from all clients.

Since subgraphs are different parts of a global graph, there inevitably exists heterogeneity among them, making it difficult to realize federated collaboration. Ignoring this heterogeneity will degrade the performance of several traditional Federated Learning (FL) methods (\textit{e.g.}, FedAvg~\cite{mcmahan2017communication}) which rely heavily on the assumption that all clients have similar data distributions. To deal with this deficiency, recently, a number of personalized FL methods~\cite{MLSYS2020_1f5fe839, pmlr-v162-pillutla22a, 9743558, baek2023personalized, zhang2023fedala} have been proposed. For example, in~\cite{baek2023personalized, li2023fedgta, zhang2024fedgt}, personalized GFL methods estimate the pairwise similarities between clients, and then they perform weighted federations based on the client similarities.

\begin{figure*}[t]
	\centering
	\includegraphics[width=12.3cm]{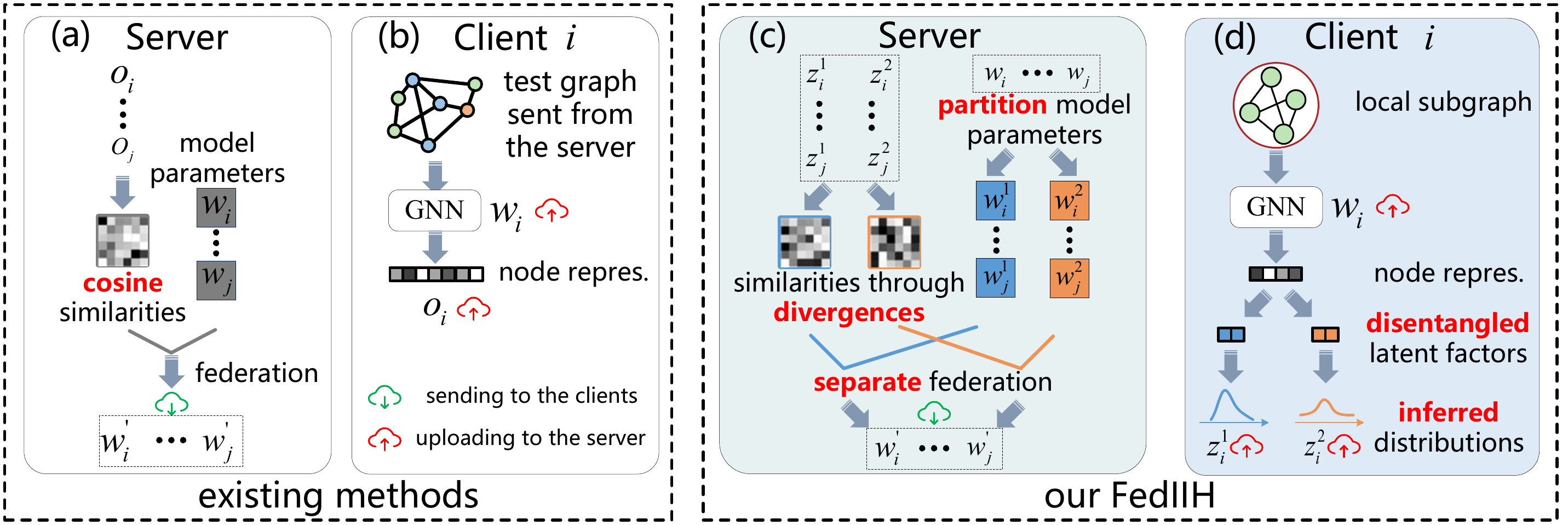}
	\caption{A framework comparison between existing methods and our FedIIH.}
	\label{fig0}
    \vspace{-10pt}
\end{figure*}

However, the performance of most existing personalized GFL methods is still limited due to their improper similarity calculations and ignored intra-heterogeneity. First, most existing methods~\cite{baek2023personalized, li2023fedgta, zhang2024fedgt} compute the inter-subgraph similarities based on the simplex outputs of local models. Since the outputs of local models cannot accurately reveal the whole distribution of subgraph data, the calculated similarities are hardly generalizable, leading to negative impacts on the weighted federation. Moreover, although most existing methods successfully consider inter-heterogeneity, they ignore the crucial intra-heterogeneity (\textit{i.e.}, the heterogeneity of intra-subgraph), which commonly exists in real-world graphs. Intra-heterogeneity can be defined as different types of connections in the subgraph on each client. For example, a user in a social graph is connected to others for various different reasons, such as families, hobbies, studies, and work. Meanwhile, as confirmed by~\cite{NEURIPS2021_9c6947bd}, inter-heterogeneity is defined as divergent distributions of both graph structures and node features among different clients.

To address the above-mentioned issues, we propose a novel \textbf{Fed}erated learning method by integrally modeling the \textbf{I}nter-\textbf{I}ntra \textbf{H}eterogeneity (FedIIH), which is shown in the right panel of Fig.~\ref{fig0}. On one hand, we propose a new Hierarchical Variational Graph AutoEncoder (HVGAE) by using the hierarchical Bayesian model~\cite{gelman2006data, tran2017hierarchical}, so as to build the posterior dependencies between the latent factors of the local subgraphs and those of the global graph. Subsequently, we can successfully infer the subgraph data distribution and compute the similarities between pairwise clients (see Fig.~\ref{fig0}c). On the other hand, we disentangle the local subgraph into multiple latent factors (see Fig.~\ref{fig0}d). After that, the model parameters can be partitioned to correspond exactly to each latent factor, such that we can accordingly perform the separate federation based on our calculated similarities.

Our FedIIH not only properly computes the distribution similarities between subgraphs, but also learns disentangled representations robust to irrelevant factors within each subgraph, effectively alleviating the inter-intra heterogeneity and significantly improving the model performances on different types of graphs. Main contributions of our work are:

\begin{itemize}
 \item For the inter-heterogeneity, we propose a novel method HVGAE to characterize the inter-heterogeneity from a multi-level global perspective, so that we can infer the data distributions of local subgraphs and properly compute the distribution similarities between subgraphs.

 \item For the intra-heterogeneity, we disentangle the subgraph into several latent factors, such that the federations can be separately performed, and this is the first time in GFL that considers the intra-heterogeneity.

 \item Extensive experiments on eleven datasets demonstrate the effectiveness of our proposed FedIIH, where our method averagely outperforms the second-best method by a large margin of 5.79\% on all heterophilic graph data.

\end{itemize}


\section{Related Work}

Here we first describe the preliminaries, and then review the typical work related to this paper, including GFL and personalized FL.

\subsection{Preliminaries}
We focus on the task of node classification and aim to collaboratively train node classifiers with local subgraphs on distributed clients under the control of a server. For given $M$ clients, each of them has a local subgraph $\mathcal{G}_m=\langle\mathcal{V}_m, \mathcal{E}_m\rangle$, where $\mathcal{V}_m$ represents the node set, and $\mathcal{E}_m$ is the edge set ($m = 1, \dots, M$). The node feature matrix and adjacency matrix of $\mathcal{G}_m$ is denoted as $\mathbf{X}_m \in \mathbb{R}^{n_m \times d}$ and $\mathbf{A}_m \in \mathbb{R}^{n_m \times n_m}$, respectively. Here $n_m$ is the number of nodes in $\mathcal{G}_m$ and $d$ is the feature dimension. Due to the privacy constraints, $\mathcal{G}_m$ on each client is inaccessible to the others.

\subsection{Graph Federated Learning}
Since each client owns only a part of the global graph, there inevitably exists heterogeneity. To deal with this issue, FED-PUB~\cite{baek2023personalized} estimates similarities between subgraphs based on the outputs of local models. Then, it performs a weighted averaging of the local models for each client based on the similarities. Similarly, FedGTA~\cite{li2023fedgta} and FedGT~\cite{ zhang2024fedgt} compute the similarities based on the mixed moments of processed neighbor features and embedding vectors, respectively. Then, they both perform the weighted federation, where model parameters with high similarities are assigned with larger weights during the weighted federation. However, the outputs of local models (\textit{e.g.}, embedding vectors) cannot faithfully reveal the whole distribution of subgraph data, such that the computed similarities are improper, which may impact the weighted federation, thus decreasing the performances on the clients. Consequently, we infer the whole distribution of subgraph data in a multi-level global perspective, so as to properly compute the similarities.

\subsection{Personalized Federated Learning}
Heterogeneity is a fundamental and challenging problem in FL~\cite{ye2023heterogeneous}. To deal with the heterogeneity, personalized FL methods~\cite{MLSYS2020_1f5fe839, 9743558, Arivazhagan2019} have obtained increasing attention. Unlike FedAvg, which aims to train a global model collaboratively, personalized FL methods not only pursue commonalities among multiple clients, but also retain the personality of each client as much as possible. Personalized FL methods can be categorized as similarity-based methods~\cite{baek2023personalized, li2023fedgta, zhang2024fedgt}, local customization-based methods~\cite{MLSYS2020_1f5fe839, Arivazhagan2019, NEURIPS2020_f4f1f13c}, and meta-learning-based methods~\cite{chen2018federated, NEURIPS2020_24389bfe}. For similarity-based methods, they compute the inter-subgraph similarities and then perform weighted federation. In contrast, as one of the local customization-based methods, FedProx~\cite{MLSYS2020_1f5fe839} customizes a personalized model for each client by adding a proximal term as a subproblem. Similarly, FedPer~\cite{Arivazhagan2019} only federates the weights of the backbone while training the personalized classification layer in each client. However, these methods only consider the inter-heterogeneity, while ignoring the intra-heterogeneity. Therefore, in this paper, we propose to characterize both inter- and intra- heterogeneity.

\section{Methodology}
This section details our proposed FedIIH. Specifically, we describe the modeling process of the intra-heterogeneity and inter-heterogeneity, respectively.
\vspace{-5pt}
\subsection{Modeling Intra-heterogeneity}

Since there are diverse connecting relations among nodes in a real-world subgraph~\cite{pmlrv97ma19a, NEURIPS2021_b6cda17a, guo2022learning}, there is inevitably heterogeneity within the subgraph. To deal with the intra-heterogeneity, we aim to disentangle the subgraph into $K$ latent factors, which are utilized to represent different relations within the subgraph. Note that most of the existing disentangled graph convolutional networks~\cite{pmlrv97ma19a, NEURIPS2021_b6cda17a, guo2022learning} are applicable to learning disentangled representations in GFL. In this paper, the well-known Disentangled Graph Convolutional Network (DisenGCN)~\cite{pmlrv97ma19a} is considered for its simplicity. Here we use the node $u$ as an example to describe the disentangling process of DisenGCN. Given one node $i \in \{u\} \cup \{v|(u, v) \in \mathcal{G}_m\}$ in the subgraph $\mathcal{G}_m$, let $\mathbf{x}^i \in \mathbb{R}^d$ denote its node feature vector. This node feature $\mathbf{x}^i$ is first projected to $K$ subspaces, and its node representation in the $k$-th subspace can be represented by
\begin{equation}\label{eq1}
\mathbf{z}^{i, k}=\frac{\sigma[({{\mathbf{W}^k})^{\top}}\mathbf{x}^i+\mathbf{b}^k]}{||\sigma[({{\mathbf{W}^k})^{\top}}\mathbf{x}^i+\mathbf{b}^k]||_2},
\end{equation} 
where $\mathbf{W}^k \in \mathbb{R}^{d \times \frac{d_\mathrm{out}}{K}}$ and $\mathbf{b}^k \in \mathbb{R}^{\frac{d_\mathrm{out}}{K}}$ are learnable parameters. Here $\sigma[\cdot]$ denotes the nonlinear activation function, and $d_\mathrm{out}$ denotes the output dimension of node representations. After the projection operation via Eq.~(\eqref{eq1}), $\mathbf{z}^{i, k} \in \mathbb{R}^{\frac{d_\mathrm{out}}{K}}$ represents the aspect of node $i$ that are related with the $k$-th latent factor. Second, DisenGCN proposes a neighborhood routing mechanism to identify the latent factor that causes the connection between node $u$ and its neighbor node $v$, and accordingly extract features of $v$ that are specific to that factor. Due to space limitations, details of the neighborhood routing mechanism are provided in the Appendix~A. With this mechanism, the node representation in each subspace is aggregated independently, such that we can obtain the disentangled latent factors $\{\mathbf{c}^{u, 1}, \mathbf{c}^{u, 2}, \cdots, \mathbf{c}^{u, K}\}$, where $\mathbf{c}^{u, k} \in \mathbb{R}^\frac{d_\mathrm{out}}{K}$ represents the $k$-th aspect of node $u$. Note that there are no learnable parameters in the neighborhood routing mechanism. Finally, the disentangled node representation of $u$ can be obtained by
\begin{equation}\label{eq2}
    \mathbf{h}^{u}=\mathrm{Con}(\mathbf{c}^{u, 1}, \mathbf{c}^{u, 2}, \cdots, \mathbf{c}^{u, K}),
\end{equation}
where $\mathbf{h}^{u} \in \mathbb{R}^{d_\mathrm{out}}$, and `$\mathrm{Con}$' denotes the concatenation operation performed along the column. Similarly, we can obtain all of the disentangled node representations in $\mathcal{G}_m$ based on Eq.~(\eqref{eq2}). After that, we use the matrix $\mathbf{H}_m \in \mathbb{R}^{n_m \times d_\mathrm{out}}$ to denote the disentangled node representations in $\mathcal{G}_m$. According to Eq.~(\eqref{eq1}), we can find that $\mathbf{W}^k$ and $\mathbf{b}^k$ correspond exactly to the $k$-th latent factor. Consequently, we propose to perform the separate federation with parameters (\textit{i.e.}, $\mathbf{W}^k$ and $\mathbf{b}^k$) specific to the $k$-th latent factor, which is described in the following section.

\subsection{Modeling Inter-heterogeneity}
Since the local views of subgraphs may not be sufficient to estimate the inter-heterogeneity, we aim to model the inter-heterogeneity from a multi-level global perspective. Moreover, since the similarities between clients are essentially determined by the similarities of local subgraph data distributions, we seek to infer the data distribution of subgraphs and thereby compute the similarities between clients through divergences. Specifically, we propose the HVGAE to infer the subgraph data distribution of the $m$-th client based on the disentangled node representations (\textit{i.e.}, $\mathbf{H}_m$). For ease of expression, we rewrite $\mathbf{H}_m$ as $\mathbf{H}_m^1, \mathbf{H}_m^2, \cdots, \mathbf{H}_m^K$, where $\mathbf{H}_m^k \in \mathbb{R}^{n_m \times \frac{d_\mathrm{out}}{K}}$ denotes the disentangled node representations with the $k$-th latent factor. As $\mathbf{H}_m^1, \mathbf{H}_m^2, \cdots, \mathbf{H}_m^K$ are deterministic results output by DisenGCN, we use $\tilde{\mathbf{H}}_m^1, \tilde{\mathbf{H}}_m^2, \cdots, \tilde{\mathbf{H}}_m^K$ to represent random latent variables.

\subsubsection{Hierarchical model}
\begin{figure}[t]
	\centering
	\includegraphics[width=3.1cm]{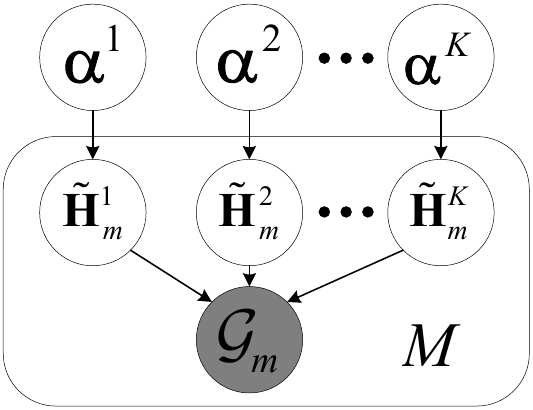}
	\caption{The graphical model of our proposed HVGAE, where $\tilde{\mathbf{H}}_m^1, \tilde{\mathbf{H}}_m^2, \cdots, \tilde{\mathbf{H}}_m^K$ denote the local latent factors on the $m$-th client, and $\bm{\alpha}^1, \bm{\alpha}^2, \cdots, \bm{\alpha}^K$ denote the global latent factors on the server.}
	\label{fig1}
    \vspace{-10pt}
\end{figure}

To build the posterior dependencies between the latent factors of the local subgraphs and those of the global graph, we assume that the latent factors of the subgraphs are governed by those of the global graph based on the theory of hierarchical Bayesian model~\cite{gelman2006data, tran2017hierarchical}. As shown in Fig.~\ref{fig1}, $\tilde{\mathbf{H}}_m^1, \tilde{\mathbf{H}}_m^2, \cdots, \tilde{\mathbf{H}}_m^K$ on the $m$-th client are local latent factors, which are linked via the higher-level global latent factors $\bm{\alpha}^1, \bm{\alpha}^2, \cdots, \bm{\alpha}^K$ on the server, respectively. Note that $\bm{\alpha}^1, \bm{\alpha}^2, \cdots, \bm{\alpha}^K$ are shared across all clients, where $\bm{\alpha}^k \in \mathbb{R}^{\frac{d_\mathrm{out}}{K}}$.

Based on the graphical model in Fig.~\ref{fig1}, the joint probability can be formulated as
\begin{equation}\label{eq3}
\begin{aligned}
&\quad p(\mathcal{G}_{1:M}, \tilde{\mathbf{H}}_{1:M}^1, \tilde{\mathbf{H}}_{1:M}^2, \cdots, \tilde{\mathbf{H}}_{1:M}^K, \bm{\alpha}^{1:K})=\\
&\prod_{k=1}^{K} p(\bm{\alpha}^k) \cdot \prod_{m=1}^{M}\prod_{k^\prime=1}^{K}p(\mathcal{G}_{m}|\tilde{\mathbf{H}}_m^1, \tilde{\mathbf{H}}_m^2, \cdots, \tilde{\mathbf{H}}_m^K)p(\tilde{\mathbf{H}}_m^k|\bm{\alpha}^{k^\prime}),
\end{aligned}
\end{equation}
where $\mathcal{G}_{1:M}$ is the abbreviation for ($\mathcal{G}_1, \mathcal{G}_2, \cdots, \mathcal{G}_M$). ($\tilde{\mathbf{H}}_{1:M}^1, \tilde{\mathbf{H}}_{1:M}^2, \cdots, \tilde{\mathbf{H}}_{1:M}^K$) is the abbreviation for ($\tilde{\mathbf{H}}_1^1, \tilde{\mathbf{H}}_2^1, \cdots, \tilde{\mathbf{H}}_M^1, \tilde{\mathbf{H}}_1^2, \tilde{\mathbf{H}}_2^2, \cdots, \tilde{\mathbf{H}}_M^2, \tilde{\mathbf{H}}_1^K, \tilde{\mathbf{H}}_2^K, \cdots, \tilde{\mathbf{H}}_M^K$), $\bm{\alpha}^{1:K}$ is the abbreviation for ($\bm{\alpha}^1, \bm{\alpha}^2, \cdots, \bm{\alpha}^K$), and $p(\bm{\alpha}^k)$ denotes the prior distribution of $\bm{\alpha}^k$. In Eq.~(\eqref{eq3}), $p(\mathcal{G}_{m}|\tilde{\mathbf{H}}_m^1, \tilde{\mathbf{H}}_m^2, \cdots, \tilde{\mathbf{H}}_m^K)$ denotes the conditional distribution, and $p(\tilde{\mathbf{H}}_m^k|\bm{\alpha}^{k^\prime})$ denotes the prior distribution of $\tilde{\mathbf{H}}_m^k$. Moreover, the true posterior distribution of both local and global latent factors can be formulated as
$p(\tilde{\mathbf{H}}_{1:M}^1, \tilde{\mathbf{H}}_{1:M}^2, \cdots, \tilde{\mathbf{H}}_{1:M}^K, \bm{\alpha}^{1:K}|\mathcal{G}_{1:M})$. However, this true posterior distribution is computationally intractable. Therefore, we attempt to approximate it with a tractable approximate posterior distribution $q(\tilde{\mathbf{H}}_{1:M}^1, \tilde{\mathbf{H}}_{1:M}^2, \cdots, \tilde{\mathbf{H}}_{1:M}^K, \bm{\alpha}^{1:K}|\mathcal{G}_{1:M})$, which can be formulated as
\begin{equation}\label{eq4}
\begin{aligned}
&q(\tilde{\mathbf{H}}_{1:M}^1, \tilde{\mathbf{H}}_{1:M}^2, \cdots, \tilde{\mathbf{H}}_{1:M}^K, \bm{\alpha}^{1:K}|\mathcal{G}_{1:M})\\
&=\prod_{k=1}^{K} q(\bm{\alpha}^k) \cdot \prod_{m=1}^{M}\prod_{k^\prime=1}^{K}q(\tilde{\mathbf{H}}_m^{k^\prime}|\mathcal{G}_m),
\end{aligned}
\end{equation}
where $q(\bm{\alpha}^k)$ denotes the marginal distribution, and $q(\tilde{\mathbf{H}}_m^{k^\prime}|\mathcal{G}_m)$ represents the approximate posterior distribution of $\tilde{\mathbf{H}}_m^{k^\prime}$. For ease of derivation, we can flexibly instantiate the above-mentioned distributions. The detailed instantiations are presented in the Appendix~B.

\subsubsection{Variational inference}
Since the approximate posterior distributions in Eq.~(\eqref{eq4}) are still intractable, we propose to use the variational inference~\cite{kingma2013auto, 8588399} to infer them. According to the graphical model in Fig.~\eqref{fig1}, the Evidence Lower BOund (ELBO) can be derived as follows:
\begin{equation}\label{eq6}
\small
\refstepcounter{equation}
\begin{aligned}
    \mathcal{L}_\mathrm{ELBO}= &\sum_{k=1}^{K} \log p(\tilde{\bm{\alpha}}^k) + \sum_{m=1}^{M} \Big\{ \mathbb{E}_{q(\tilde{\mathbf{H}}_m|\mathcal{G}_m)} \big[\log p(\mathcal{G}_m|\tilde{\mathbf{H}}_m)\big] \\
    &- \sum_{k^\prime=1}^{K} D_{\mathrm{KL}}\big(q(\tilde{\mathbf{H}}_m^{k^\prime}|\mathcal{G}_m) || p(\tilde{\mathbf{H}}_m^{k^\prime} | \tilde{\bm{\alpha}}^{k^\prime})\big)\Big\},
\end{aligned}
\tag{\theequation}
\normalsize
\end{equation}
where $\log p(\tilde{\bm{\alpha}}^k)$ denotes the log prior probability of $\tilde{\bm{\alpha}}^k$. Here $\tilde{\bm{\alpha}}^k$ denotes the posterior mean of $\bm{\alpha}^k$ for the $k$-th global latent factor, and $D_{\mathrm{KL}}$ denotes the Kullback-Leibler (KL) divergence~\cite{NIPS2003_0abdc563}. The detailed derivation of ELBO and $\tilde{\bm{\alpha}}^k$ are provided in the Appendix~C and~D, respectively. Eq.~(\eqref{eq6}) shows that the derived ELBO is the sum of two terms, where the first term is related to the global latent factors on the server, and the second term can be parallelly optimized on $M$ clients.

\subsubsection{Adaptation to FL}
However, since the server does not have access to the private data of the clients, the optimization process on the server (\textit{i.e.}, $\sum_{k=1}^{K}\log p(\tilde{\bm{\alpha}}^k)$) is challenging. To adapt Eq.~(\eqref{eq6}) to the federated learning scenario, we make some modifications to $\sum_{k=1}^{K}\log p(\tilde{\bm{\alpha}}^k)$. On one hand, we substitute $\sum_{k=1}^{K}\log p(\tilde{\bm{\alpha}}^k)$ by $\sum_{m=1}^{M} \sum_{k=1}^{K} \log p(\hat{\bm{\alpha}}^k_m)$, where $\hat{\bm{\alpha}}^k_m$ denotes the learnable parameter deployed on the $m$-th client. On the other hand, $\hat{\bm{\alpha}}^k_m$ on each client is constrained by the KL divergence between $p(\hat{\bm{\alpha}}^k_m)$ and $p(\tilde{\bm{\alpha}}^k)$. Finally, we have that
\begin{equation}\label{eq8}
    \begin{aligned}
        \hat{\mathcal{L}}_\mathrm{ELBO}=
        &\sum_{m=1}^{M} \sum_{k=1}^{K} \Big\{\log p(\hat{\bm{\alpha}}^k_m) - D_{\mathrm{KL}}\big(p(\hat{\bm{\alpha}}^k_m) || p(\tilde{\bm{\alpha}}^k)\big)\Big\} + \\
        &\sum_{{m^\prime}=1}^{M} \Big\{ \mathbb{E}_{q(\tilde{\mathbf{H}}_{m^\prime}|\mathcal{G}_{m^\prime})} \big[\log p(\mathcal{G}_{m^\prime}|\tilde{\mathbf{H}}_{m^\prime})\big] -\\
        &\sum_{{k^\prime}=1}^{K} D_{\mathrm{KL}}\big(q(\tilde{\mathbf{H}}_{m^\prime}^{k^\prime}|\mathcal{G}_{m^\prime}) || p(\tilde{\mathbf{H}}_{m^\prime}^{k^\prime} | \tilde{\bm{\alpha}}^{k^\prime})\big)\Big\}.
    \end{aligned}
    \end{equation}
Eq.~(\eqref{eq8}) implies that the optimizations of ELBO can be distributedly performed on $M$ clients in a parallel way.

\subsubsection{Variational graph autoencoder}
To implement the approximate posterior distribution $q(\tilde{\mathbf{H}}_m^k|\mathcal{G}_m)$ and the mathematical expectation $\mathbb{E}_{q(\tilde{\mathbf{H}}_m|\mathcal{G}_m)}\big[\log p(\mathcal{G}_m|\tilde{\mathbf{H}}_m)\big]$ in Eq.~(\eqref{eq8}), we introduce a simple yet effective framework (\textit{i.e.}, Variational Graph AutoEncoder (VGAE)~\cite{kipf2016variational}) on each client, which includes an inference network (\textit{a.k.a.} encoder) and a generative network (\textit{a.k.a.} decoder). First, we take an inference model parameterized by two DisenGCNs (\textit{i.e.}, $\mathrm{DisenGCN}_{\bm{\mu}_m}(\mathcal{G}_m)$ and $\mathrm{DisenGCN}_{\bm{\sigma}_m}(\mathcal{G}_m)$). Note that $\mathrm{DisenGCN}_{\bm{\mu}_m}(\mathcal{G}_m)$ and $\mathrm{DisenGCN}_{\bm{\sigma}_m}(\mathcal{G}_m)$ are used to infer the means and standard deviations of $\mathcal{G}_m$ for $K$ latent factors (\textit{i.e.}, $\mathbf{H}_{m, \bm{\mu}}$ and $\mathbf{H}_{m, \bm{\sigma}}$), respectively. For ease of expression, we rewrite them as $\mathbf{H}_{m, \bm{\mu}}^{1}, \mathbf{H}_{m, \bm{\mu}}^{2}, \cdots, \mathbf{H}_{m, \bm{\mu}}^{K}$ and $\mathbf{H}_{m, \bm{\sigma}}^{1}, \mathbf{H}_{m, \bm{\sigma}}^{2}, \cdots, \mathbf{H}_{m, \bm{\sigma}}^{K}$, respectively. By using the reparameterization trick~\cite{kingma2013auto}, we can have
\begin{equation}\label{eq9}
\tilde{\mathbf{H}}_m^k=\mathbf{H}_{m, \bm{\mu}}^{k} + \mathbf{H}_{m, \bm{\sigma}}^{k} \odot \bm{\epsilon},
\end{equation}
where $\bm{\epsilon}\sim \mathcal{N}(\mathbf{0}, \mathbf{I})$, and $\odot$ denotes the element-wise product. Second, our generative network is constructed by an inner product between latent variables. Since $\mathbb{E}_{q(\tilde{\mathbf{H}}_m|\mathcal{G}_m)}\big[\log p(\mathcal{G}_m|\tilde{\mathbf{H}}_m)\big]$ in Eq.~(\eqref{eq8}) can be regarded as a reconstruction loss, we implement it by
\begin{equation}\label{eq10}
\begin{aligned}
\quad \mathbb{E}_{q(\tilde{\mathbf{H}}_m|\mathcal{G}_m)}\big[\log p(\mathcal{G}_m|\tilde{\mathbf{H}}_m)\big]
&=p(\mathbf{A}_m|\tilde{\mathbf{H}}_m)\\
&=\prod_{i=1}^{n_m}\prod_{j=1}^{n_m}p(\mathbf{A}_{m}^{ij}|\mathbf{r}_i, \mathbf{r}_j),
\end{aligned}
\end{equation}
where $p(\mathbf{A}_{m}^{ij}=1|\mathbf{r}_i, \mathbf{r}_j)=\sigma(\mathbf{r}_i^{\top }\mathbf{r}_j)$, and $\mathbf{A}_{m}^{ij}$ denotes the element of $\mathbf{A}_m$. In Eq.~(\eqref{eq10}), $\mathbf{r}_i$ and $\mathbf{r}_j$ are the $i$-th and $j$-th rows of the matrix $\tilde{\mathbf{H}}_m$, respectively. The detailed implementation of HVGAE can be found in the Appendix~E.

\subsubsection{Similarity calculation}
Since the similarities between clients are essentially determined by the similarities of the local data distributions, we seek to calculate the similarities between clients based on the divergences of the inferred subgraph data distributions. Moreover, since HVGAE has $K$ disentangled latent factors, we can compute the similarities corresponding to each latent factor separately. Specifically, the similarity with respect to the $k$-th latent factor between clients $m$ and $j$ can be computed as
\begin{equation}\label{eq12}
S(m,j)^{k}=1-D_{\mathrm{JS}}\big(q(\tilde{\mathbf{H}}_m^k|\mathcal{G}_m) || q(\tilde{\mathbf{H}}_j^k|\mathcal{G}_j)\big),
\end{equation}
where $S(m,j)^{k}$ denotes the similarity, and $D_{\mathrm{JS}}$ denotes the Jensen-Shannon (JS) divergence~\cite{NEURIPS2020_43bb733c}. Note that Eq.~(\eqref{eq12}) uses only approximate posterior distributions sent to the server, which does not compromise data privacy.

\subsection{Federated Aggregation}
\label{federated_aggregation}
Before describing the federation process, let us analyze the learnable parameters that should be federated in our FedIIH. The first part of the federated parameters are $\mathbf{W}_m^1, \mathbf{W}_m^2, \cdots, \mathbf{W}_m^K$ and $\mathbf{b}_m^1, \mathbf{b}_m^2, \cdots, \mathbf{b}_m^K$ in Eq.~(\eqref{eq1}). The second part of the federated parameters are $\mathbf{W}_m^{\mathrm{cls}} \in \mathbb{R}^{c \times d_\mathrm{out}}$ and $\mathbf{b}_m^{\mathrm{cls}} \in \mathbb{R}^{c}$, which come from the node classifier. Here $c$ is the number of node classes. The detailed analysis can be found in the Appendix~G. For the federated aggregation of $\mathbf{W}_m^1, \mathbf{W}_m^2, \cdots, \mathbf{W}_m^K$, $\mathbf{b}_m^1, \mathbf{b}_m^2, \cdots, \mathbf{b}_m^K$, and $\mathbf{W}_m^{\mathrm{cls}}$, we propose a separate federation method. Specifically, with the computed similarity $S(i,j)^{k}$, we use the weighted averaging of parameters across different clients. Moreover, since HVGAE has $K$ disentangled latent factors, and parameters (\textit{e.g.}, $\mathbf{W}^k$ and $\mathbf{b}^k$ in Eq.~(\eqref{eq1})) correspond exactly to the $k$-th latent factor, we can therefore perform the separate federation according to each latent factor. For ease of expression, we rewrite $\mathbf{W}_m^{\mathrm{cls}}$ as $\mathbf{W}_{m}^{\mathrm{cls}, 1}, \mathbf{W}_{m}^{\mathrm{cls}, 2}, \cdots, \mathbf{W}_{m}^{\mathrm{cls}, K}$, where $\mathbf{W}_m^{\mathrm{cls}, k} \in \mathbb{R}^{c \times \frac{d_\mathrm{out}}{K}}$ denotes the parameters with respect to the $k$-th latent factor. Our proposed separate federation can be defined as
\begin{equation}\label{eq13}
\begin{aligned}
\overline {\mathbf{W}}_m^k\leftarrow \sum_{j=1}^{M} \beta_{mj}^{k} \cdot \mathbf{W}_j^k, \quad \overline {\mathbf{b}}_m^k\leftarrow \sum_{j=1}^{M} \beta_{mj}^{k} \cdot \mathbf{b}_j^k, \\
\quad \overline {\mathbf{W}}_m^{\mathrm{cls}, k}\leftarrow \sum_{j=1}^{M} \beta_{mj}^{k} \cdot \mathbf{W}_j^{\mathrm{cls}, k},
\end{aligned}
\end{equation}
where
\begin{equation}\label{eq14}
\beta_{mj}^k=\frac{\exp \big(\tau \cdot S(m,j)^k\big)}{\sum_{l=1}^{M} \exp \big(\tau \cdot S(m, l)^k\big)},
\end{equation}
and $\beta_{mj}^k$ denotes the normalized similarity with respect to the $k$-th latent factor between clients $m$ and $j$. Here `$\leftarrow$' denotes the assignment operation, and $\tau$ denotes a hyperparameter for scaling the similarity score. Our proposed separate federation not only allows different clients to obtain personalized parameters, which is beneficial for dealing with inter-heterogeneity, but also allows parameters in the same client to be federated separately according to each latent factor, which is beneficial for dealing with intra-heterogeneity. For the federation of the remaining parameter $\mathbf{b}_m^{\mathrm{cls}} \in \mathbb{R}^{c}$, we use the federation method as proposed in FedAvg~\cite{mcmahan2017communication}, which is simple but effective. The pseudocode of our FedIIH is listed in the Appendix~H.
\vspace{-5pt}
\section{Experiments}
To validate the effectiveness of our FedIIH, we perform extensive experiments on eleven widely used benchmark datasets. These datasets include both homophilic and heterophilic graphs, the statistical information and descriptions of which are included in the Appendix~I.2. We use both the non-overlapping and overlapping subgraph partitioning settings. To ensure a fair comparison, we compute the mean accuracy (or mean AUC), and the corresponding standard deviation over three independent runs according to~\cite{baek2023personalized}. See Appendix~I for more details.

\subsection{Main Results}
\subsubsection{Homophilic datasets} Tab.~\ref{table1} and Tab.~\ref{table2} show the node classification results on the homophilic datasets in two partitioning settings, respectively. We observe that our FedIIH achieves the best performance among all the methods, and the standard deviations are also relatively small as well, suggesting that FedIIH is more effective and stable than the compared methods. Moreover, the average accuracy of FedIIH for all six datasets in the non-overlapping scenario is 83.10\%, which is 1.51\% higher than the second-best method (\textit{i.e.}, FED-PUB). In the overlapping scenario, the average accuracy of FedIIH is 81.01\%, which is 1.48\% higher than the second-best method (\textit{i.e.}, FED-PUB). Generally, the non-overlapping scenario is more challenging than the overlapping scenario~\cite{baek2023personalized, zhang2024fedgt} due to the increasing heterogeneity. However, our FedIIH can still outperform the second-best method in both the non-overlapping and overlapping scenarios, and this validates the effectiveness of modeling inter-intra heterogeneity. For example, as shown in Tab.~\ref{table1}, the classification accuracy of FedIIH increases from 93.42\% to 93.55\% when the number of clients increases from 5 to 20 on the \textit{Amazon-Photo} dataset. This is non-trivial because the data heterogeneity increases from 0.664 to 0.759~\cite{zhang2024fedgt} during this change, indicating that FedIIH is robust to the heterogeneity.

\subsubsection{Heterophilic datasets} Tab.~\ref{table3} and Tab.~\ref{table4} show the node classification results on the heterophilic datasets in two partitioning settings, respectively. We can find that our proposed FedIIH not only achieves the best average performance among all baseline methods, but also outperforms the second-best method (\textit{i.e.}, FedSage+) by 5.79\% and 4.53\% in the non-overlapping and overlapping scenarios, respectively. This is because our FedIIH not only considers the inter-heterogeneity as these methods do, but also successfully deals with the intra-heterogeneity that they ignore. Moreover, although the intra-heterogeneity on the heterophilic datasets is stronger than that on the homophilic datasets~\cite{platonov2023a}, FedIIH achieves a larger performance improvement than on the homophilic datasets, which further validates the effectiveness of our FedIIH.
\vspace{-5pt}

\begin{table*}[]
    \centering
    \scriptsize
    \caption{Node classification results of different methods on the \textbf{homophilic} graph datasets in the \textbf{non-overlapping} subgraph partitioning setting. The best results are shown in \textbf{bold}.}
      \label{table1}
      \renewcommand{\arraystretch}{0.8} 
         \scalebox{0.82}{
    \begin{tabular}{lcccccccccc}
    \hline
    \textbf{}     & \multicolumn{3}{c}{Cora}                                                    & \multicolumn{3}{c}{CiteSeer}                                                & \multicolumn{3}{c}{PubMed}                                                  & -              \\ \cline{2-11} 
    Methods       & 5 Clients               & 10 Clients              & 20 Clients              & 5 Clients               & 10 Clients              & 20 Clients              & 5 Clients               & 10 Clients              & 20 Clients              & -              \\ \hline
    Local         & 81.30$\pm$0.21          & 79.94$\pm$0.24          & 80.30$\pm$0.25          & 69.02$\pm$0.05          & 67.82$\pm$0.13          & 65.98$\pm$0.17          & 84.04$\pm$0.18          & 82.81$\pm$0.39          & 82.65$\pm$0.03          & -              \\ \hline
    FedAvg~\cite{mcmahan2017communication}        & 74.45$\pm$5.64          & 69.19$\pm$0.67          & 69.50$\pm$3.58          & 71.06$\pm$0.60          & 63.61$\pm$3.59          & 64.68$\pm$1.83          & 79.40$\pm$0.11          & 82.71$\pm$0.29          & 80.97$\pm$0.26          & -              \\
    FedProx~\cite{MLSYS2020_1f5fe839}       & 72.03$\pm$4.56          & 60.18$\pm$7.04          & 48.22$\pm$6.18          & 71.73$\pm$1.11          & 63.33$\pm$3.25          & 64.85$\pm$1.35          & 79.45$\pm$0.25          & 82.55$\pm$0.24          & 80.50$\pm$0.25          & -              \\
    FedPer~\cite{Arivazhagan2019}        & 81.68$\pm$0.40          & 79.35$\pm$0.04          & 78.01$\pm$0.32          & 70.41$\pm$0.32          & 70.53$\pm$0.28          & 66.64$\pm$0.27          & 85.80$\pm$0.21          & 84.20$\pm$0.28          & 84.72$\pm$0.31          & -              \\
    GCFL~\cite{NEURIPS2021_9c6947bd}          & 81.47$\pm$0.65          & 78.66$\pm$0.27          & 79.21$\pm$0.70          & 70.34$\pm$0.57          & 69.01$\pm$0.12          & 66.33$\pm$0.05          & 85.14$\pm$0.33          & 84.18$\pm$0.19          & 83.94$\pm$0.36          & -              \\
    FedGNN~\cite{wu2021fedgnn}        & 81.51$\pm$0.68          & 70.12$\pm$0.99          & 70.10$\pm$3.52          & 69.06$\pm$0.92          & 55.52$\pm$3.17          & 52.23$\pm$6.00          & 79.52$\pm$0.23          & 83.25$\pm$0.45          & 81.61$\pm$0.59          & -              \\
    FedSage+\cite{NEURIPS2021_34adeb8e}      & 72.97$\pm$5.94          & 69.05$\pm$1.59          & 57.97$\pm$12.6          & 70.74$\pm$0.69          & 65.63$\pm$3.10          & 65.46$\pm$0.74          & 79.57$\pm$0.24          & 82.62$\pm$0.31          & 80.82$\pm$0.25          & -              \\
    FED-PUB~\cite{baek2023personalized}       & 83.70$\pm$0.19          & 81.54$\pm$0.12          & 81.75$\pm$0.56          & 72.68$\pm$0.44          & 72.35$\pm$0.53          & 67.62$\pm$0.12          & 86.79$\pm$0.09          & 86.28$\pm$0.18          & 85.53$\pm$0.30          & -              \\
    FedGTA~\cite{li2023fedgta}        & 80.06$\pm$0.63          & 80.59$\pm$0.38          & 79.01$\pm$0.31          & 70.12$\pm$0.10          & 71.57$\pm$0.34          & 69.94$\pm$0.14          & 87.75$\pm$0.01          & 86.80$\pm$0.01          & 87.12$\pm$0.05          & -              \\
    AdaFGL~\cite{li2024adafgl}        & 82.01$\pm$0.51          & 80.09$\pm$0.00          & 79.74$\pm$0.05          & 71.44$\pm$0.27          & 72.34$\pm$0.00          & 70.95$\pm$0.45          & 86.91$\pm$0.28          & 86.97$\pm$0.10          & 86.59$\pm$0.21          & -              \\ \hline
    FedIIH (Ours) & \textbf{84.11$\pm$0.17} & \textbf{81.85$\pm$0.09} & \textbf{83.01$\pm$0.15} & \textbf{72.86$\pm$0.25} & \textbf{76.50$\pm$0.06} & \textbf{73.36$\pm$0.41} & \textbf{87.80$\pm$0.18} & \textbf{87.65$\pm$0.18} & \textbf{87.19$\pm$0.25} & -              \\ \hline
                  & \multicolumn{3}{c}{Amazon-Computer}                                         & \multicolumn{3}{c}{Amazon-Photo}                                            & \multicolumn{3}{c}{ogbn-arxiv}                                              & Avg.            \\ \cline{2-11} 
    Methods       & 5 Clients               & 10 Clients              & 20 Clients              & 5 Clients               & 10 Clients              & 20 Clients              & 5 Clients               & 10 Clients              & 20 Clients              & All           \\ \hline
    Local         & 89.22$\pm$0.13          & 88.91$\pm$0.17          & 89.52$\pm$0.20          & 91.67$\pm$0.09          & 91.80$\pm$0.02          & 90.47$\pm$0.15          & 66.76$\pm$0.07          & 64.92$\pm$0.09          & 65.06$\pm$0.05          & 79.57          \\ \hline
    FedAvg~\cite{mcmahan2017communication}        & 84.88$\pm$1.96          & 79.54$\pm$0.23          & 74.79$\pm$0.24          & 89.89$\pm$0.83          & 83.15$\pm$3.71          & 81.35$\pm$1.04          & 65.54$\pm$0.07          & 64.44$\pm$0.10          & 63.24$\pm$0.13          & 74.58          \\
    FedProx~\cite{MLSYS2020_1f5fe839}       & 85.25$\pm$1.27          & 83.81$\pm$1.09          & 73.05$\pm$1.30          & 90.38$\pm$0.48          & 80.92$\pm$4.64          & 82.32$\pm$0.29          & 65.21$\pm$0.20          & 64.37$\pm$0.18          & 63.03$\pm$0.04          & 72.84          \\
    FedPer~\cite{Arivazhagan2019}        & 89.67$\pm$0.34          & 89.73$\pm$0.04          & 87.86$\pm$0.43          & 91.44$\pm$0.37          & 91.76$\pm$0.23          & 90.59$\pm$0.06          & 66.87$\pm$0.05          & 64.99$\pm$0.18          & 64.66$\pm$0.11          & 79.94          \\
    GCFL~\cite{NEURIPS2021_9c6947bd}          & 89.07$\pm$0.91          & 90.03$\pm$0.16          & 89.08$\pm$0.25          & 91.99$\pm$0.29          & 92.06$\pm$0.25          & 90.79$\pm$0.17          & 66.80$\pm$0.12          & 65.09$\pm$0.08          & 65.08$\pm$0.04          & 79.90          \\
    FedGNN~\cite{wu2021fedgnn}        & 88.08$\pm$0.15          & 88.18$\pm$0.41          & 83.16$\pm$0.13          & 90.25$\pm$0.70          & 87.12$\pm$2.01          & 81.00$\pm$4.48          & 65.47$\pm$0.22          & 64.21$\pm$0.32          & 63.80$\pm$0.05          & 75.23          \\
    FedSage+\cite{NEURIPS2021_34adeb8e}      & 85.04$\pm$0.61          & 80.50$\pm$1.13          & 70.42$\pm$0.85          & 90.77$\pm$0.44          & 76.81$\pm$8.24          & 80.58$\pm$1.15          & 65.69$\pm$0.09          & 64.52$\pm$0.14          & 63.31$\pm$0.20          & 73.47          \\
    FED-PUB~\cite{baek2023personalized}       & \textbf{90.74$\pm$0.05} & 90.55$\pm$0.13          & 90.12$\pm$0.09          & 93.29$\pm$0.19          & 92.73$\pm$0.18          & 91.92$\pm$0.12          & 67.77$\pm$0.09          & 66.58$\pm$0.08          & 66.64$\pm$0.12          & 81.59          \\
    FedGTA~\cite{li2023fedgta}        & 86.69$\pm$0.18          & 86.66$\pm$0.23          & 85.01$\pm$0.87          & 93.33$\pm$0.12          & 93.50$\pm$0.21          & 92.61$\pm$0.15          & 60.32$\pm$0.04                   & 60.22$\pm$0.09                   & 58.74$\pm$0.14                         & 79.45          \\ 
    AdaFGL~\cite{li2024adafgl}        & 80.20$\pm$0.05          & 83.62$\pm$0.26          & 84.53$\pm$0.23          & 86.69$\pm$0.19          & 89.85$\pm$0.83          & 88.11$\pm$0.05          & 52.73$\pm$0.19          & 51.77$\pm$0.36          & 50.94$\pm$0.08          & 76.97          \\ \hline
    FedIIH (Ours) & \textbf{90.74$\pm$0.13}          & \textbf{90.86$\pm$0.23} & \textbf{90.44$\pm$0.05} & \textbf{93.42$\pm$0.02} & \textbf{94.22$\pm$0.08} & \textbf{93.55$\pm$0.09} & \textbf{70.30$\pm$0.06} & \textbf{69.34$\pm$0.02} & \textbf{68.65$\pm$0.04} & \textbf{83.10} \\ \hline
    \end{tabular}
    }
    \end{table*}

    \begin{table*}[]
        \centering
            \scriptsize
            \caption{Node classification results of different methods on the \textbf{homophilic} graph datasets in the \textbf{overlapping} subgraph partitioning setting. The best results are shown in \textbf{bold}.}
            \label{table2}
      \renewcommand{\arraystretch}{0.8} 
             \scalebox{0.82}{
        \begin{tabular}{lcccccccccc}
        \hline
        \multicolumn{1}{c}{} & \multicolumn{3}{c}{Cora}                                                    & \multicolumn{3}{c}{CiteSeer}                                                      & \multicolumn{3}{c}{PubMed}                                                  & -              \\ \cline{2-11} 
        Methods              & 10 Clients              & 30 Clients              & 50 Clients              & 10 Clients                    & 30 Clients              & 50 Clients              & 10 Clients              & 30 Clients              & 50 Clients              & -              \\ \hline
        Local                & 73.98$\pm$0.25          & 71.65$\pm$0.12          & 76.63$\pm$0.10          & 65.12$\pm$0.08                & 64.54$\pm$0.42          & 66.68$\pm$0.44          & 82.32$\pm$0.07          & 80.72$\pm$0.16          & 80.54$\pm$0.11          & -              \\ \hline
        FedAvg~\cite{mcmahan2017communication}               & 76.48$\pm$0.36          & 53.99$\pm$0.98          & 53.99$\pm$4.53          & 69.48$\pm$0.15                & 66.15$\pm$0.64          & 66.51$\pm$1.00          & 82.67$\pm$0.11          & 82.05$\pm$0.12          & 80.24$\pm$0.35          & -              \\
        FedProx~\cite{MLSYS2020_1f5fe839}              & 77.85$\pm$0.50          & 51.38$\pm$1.74          & 56.27$\pm$9.04          & 69.39$\pm$0.35                & 66.11$\pm$0.75          & 66.53$\pm$0.43          & 82.63$\pm$0.17          & 82.13$\pm$0.13          & 80.50$\pm$0.46          & -              \\
        FedPer~\cite{Arivazhagan2019}               & 78.73$\pm$0.31          & 74.18$\pm$0.24          & 74.42$\pm$0.37          & 69.81$\pm$0.28                & 65.19$\pm$0.81          & 67.64$\pm$0.44          & 85.31$\pm$0.06          & 84.35$\pm$0.38          & 83.94$\pm$0.10          & -              \\
        GCFL~\cite{NEURIPS2021_9c6947bd}                 & 78.84$\pm$0.26          & 73.41$\pm$0.27          & 76.63$\pm$0.16          & 69.48$\pm$0.39                & 64.92$\pm$0.18          & 65.98$\pm$0.30          & 83.59$\pm$0.25          & 80.77$\pm$0.12          & 81.36$\pm$0.11          & -              \\
        FedGNN~\cite{wu2021fedgnn}               & 70.63$\pm$0.83          & 61.38$\pm$2.33          & 56.91$\pm$0.82          & 68.72$\pm$0.39                & 59.98$\pm$1.52          & 58.98$\pm$0.98          & 84.25$\pm$0.07          & 82.02$\pm$0.22          & 81.85$\pm$0.10          & -              \\
        FedSage+\cite{NEURIPS2021_34adeb8e}             & 77.52$\pm$0.46          & 51.99$\pm$0.42          & 55.48$\pm$11.5          & 68.75$\pm$0.48                & 65.97$\pm$0.02          & 65.93$\pm$0.30          & 82.77$\pm$0.08          & 82.14$\pm$0.11          & 80.31$\pm$0.68          & -              \\
        FED-PUB~\cite{baek2023personalized}              & 79.60$\pm$0.12          & 75.40$\pm$0.54          & 77.84$\pm$0.23          & 70.58$\pm$0.20                & 68.33$\pm$0.45          & 69.21$\pm$0.30          & 85.70$\pm$0.08          & 85.16$\pm$0.10          & 84.84$\pm$0.12          & -              \\
        FedGTA~\cite{li2023fedgta}               & 76.42$\pm$0.62          & 75.63$\pm$0.33          & 77.69$\pm$0.14          & 70.43$\pm$0.08 & 71.71$\pm$0.33          & 69.19$\pm$0.32          & 85.34$\pm$0.42          & 84.99$\pm$0.05          & 84.47$\pm$0.06          & -              \\
        AdaFGL~\cite{li2024adafgl}               & 78.50$\pm$0.19          & 75.80$\pm$0.23          & 74.41$\pm$0.00          & 72.63$\pm$0.15 & 68.18$\pm$0.31          & 62.90$\pm$0.75          & 85.58$\pm$0.23          & 85.85$\pm$0.41          & 84.45$\pm$0.07          & -              \\ \hline
        FedIIH (Ours)        & \textbf{80.57$\pm$0.23} & \textbf{76.82$\pm$0.24} & \textbf{78.58$\pm$0.25} & \textbf{73.16$\pm$0.18}       & \textbf{72.27$\pm$0.21} & \textbf{69.56$\pm$0.11} & \textbf{85.87$\pm$0.03} & \textbf{86.65$\pm$0.11} & \textbf{85.65$\pm$0.12} & -              \\ \hline
                             & \multicolumn{3}{c}{Amazon-Computer}                                         & \multicolumn{3}{c}{Amazon-Photo}                                                  & \multicolumn{3}{c}{ogbn-arxiv}                                              & Avg.            \\ \cline{2-11} 
        Methods              & 10 Clients              & 30 Clients              & 50 Clients              & 10 Clients                    & 30 Clients              & 50 Clients              & 10 Clients              & 30 Clients              & 50 Clients              & All              \\ \hline
        Local                & 88.50$\pm$0.20          & 86.66$\pm$0.00          & 87.04$\pm$0.02          & 92.17$\pm$0.12                & 90.16$\pm$0.12          & 90.42$\pm$0.15          & 62.52$\pm$0.07          & 61.32$\pm$0.04          & 60.04$\pm$0.04          & 76.72          \\ \hline
        FedAvg~\cite{mcmahan2017communication}               & 88.99$\pm$0.19          & 83.37$\pm$0.47          & 76.34$\pm$0.12          & 92.91$\pm$0.07                & 89.30$\pm$0.22          & 74.19$\pm$0.57          & 63.56$\pm$0.02          & 59.72$\pm$0.06          & 60.94$\pm$0.24          & 73.38          \\
        FedProx~\cite{MLSYS2020_1f5fe839}              & 88.84$\pm$0.20          & 83.84$\pm$0.89          & 76.60$\pm$0.47          & 92.67$\pm$0.19                & 89.17$\pm$0.40          & 72.36$\pm$2.06          & 63.52$\pm$0.11          & 59.86$\pm$0.16          & 61.12$\pm$0.04          & 73.38          \\
        FedPer~\cite{Arivazhagan2019}               & 89.30$\pm$0.04          & 87.99$\pm$0.23          & 88.22$\pm$0.27          & 92.88$\pm$0.24                & 91.23$\pm$0.16          & 90.92$\pm$0.38          & 63.97$\pm$0.08          & 62.29$\pm$0.04          & 61.24$\pm$0.11          & 78.42          \\
        GCFL~\cite{NEURIPS2021_9c6947bd}                 & 89.01$\pm$0.22          & 87.24$\pm$0.09          & 87.02$\pm$0.22          & 92.45$\pm$0.10                & 90.58$\pm$0.11          & 90.54$\pm$0.08          & 63.24$\pm$0.02          & 61.66$\pm$0.10          & 60.32$\pm$0.01          & 77.61          \\
        FedGNN~\cite{wu2021fedgnn}               & 88.15$\pm$0.09          & 87.00$\pm$0.10          & 83.96$\pm$0.88          & 91.47$\pm$0.11                & 87.91$\pm$1.34          & 78.90$\pm$6.46          & 63.08$\pm$0.19          & 60.09$\pm$0.04          & 60.51$\pm$0.11          & 73.66          \\
        FedSage+\cite{NEURIPS2021_34adeb8e}             & 89.24$\pm$0.15          & 81.33$\pm$1.20          & 76.72$\pm$0.39          & 92.76$\pm$0.05                & 88.69$\pm$0.99          & 72.41$\pm$1.36          & 63.24$\pm$0.02          & 59.90$\pm$0.12          & 60.95$\pm$0.09          & 73.12          \\
        FED-PUB~\cite{baek2023personalized}              & 89.98$\pm$0.08          & 89.15$\pm$0.06          & 88.76$\pm$0.14          & 93.22$\pm$0.07                & 92.01$\pm$0.07          & 91.71$\pm$0.11          & 64.18$\pm$0.04          & 63.34$\pm$0.12          & 62.55$\pm$0.12          & 79.53          \\
        FedGTA~\cite{li2023fedgta}               & 90.10$\pm$0.18          & 88.79$\pm$0.27          & 88.15$\pm$0.21          & 93.13$\pm$0.14                & 92.49$\pm$0.06          & 91.77$\pm$0.06          & 55.98$\pm$0.09          & 56.76$\pm$0.07          & 57.89$\pm$0.09          & 74.40          \\
        AdaFGL~\cite{li2024adafgl}        & 80.49$\pm$0.00          & 80.42$\pm$0.00          & 82.12$\pm$0.00          & 89.24$\pm$0.00          & 88.34$\pm$0.00          & 87.68$\pm$0.00          & 56.81$\pm$0.06                   & 55.17$\pm$0.00                   & 54.82$\pm$0.00                         & 75.74          \\ \hline
        FedIIH (Ours)        & \textbf{90.15$\pm$0.04} & \textbf{89.56$\pm$0.19} & \textbf{89.99$\pm$0.00} & \textbf{93.38$\pm$0.00}       & \textbf{94.17$\pm$0.04} & \textbf{93.25$\pm$0.16} & \textbf{66.69$\pm$0.09} & \textbf{66.10$\pm$0.03} & \textbf{65.67$\pm$0.06} & \textbf{81.01} \\ \hline
        \end{tabular}
        }
    \end{table*}

\begin{table*}[]
\centering
\scriptsize
\caption{Node classification results of different methods on the \textbf{heterophilic} graph datasets in the \textbf{non-overlapping} subgraph partitioning setting. Accuracy (\%) is reported for \textit{Roman-empire} and \textit{Amazon-ratings}, and AUC (\%) is reported for \textit{Minesweeper}, \textit{Tolokers}, and \textit{Questions}. The best and second-best results are shown in \textbf{bold} and \underline{underlined}, respectively.}
\label{table3}
\renewcommand{\arraystretch}{0.8} 
    \scalebox{0.82}{
    \begin{tabular}{lcccccccccc}
    \hline
    \textbf{}     & \multicolumn{3}{c}{Roman-empire}                                            & \multicolumn{3}{c}{Amazon-ratings}                                          & \multicolumn{3}{c}{Minesweeper}                                             & -              \\ \cline{2-11} 
    Methods       & 5 Clients               & 10 Clients              & 20 Clients              & 5 Clients               & 10 Clients              & 20 Clients              & 5 Clients               & 10 Clients              & 20 Clients              & -              \\ \hline
    Local         & 33.65$\pm$0.13          & 28.42$\pm$0.26          & 23.89$\pm$0.32          & \textbf{45.03$\pm$0.31} & \textbf{45.89$\pm$0.19} & \textbf{46.02$\pm$0.02} & 71.35$\pm$0.17          & 69.96$\pm$0.16          & 69.31$\pm$0.09          & -              \\ \hline
    FedAvg~\cite{mcmahan2017communication}        & 38.93$\pm$0.32          & 35.43$\pm$0.32          & 32.00$\pm$0.39          & 41.26$\pm$0.53          & 41.66$\pm$0.14          & 42.20$\pm$0.21          & 72.60$\pm$0.08          & 71.84$\pm$0.02          & 71.36$\pm$0.16          & -              \\
    FedProx~\cite{MLSYS2020_1f5fe839}       & 27.95$\pm$0.59          & 26.43$\pm$1.41          & 23.12$\pm$0.49          & 36.92$\pm$0.00          & 36.86$\pm$0.14          & 36.96$\pm$0.00          & 71.91$\pm$0.27          & 70.66$\pm$0.20          & 71.50$\pm$0.37          & -              \\
    FedPer~\cite{Arivazhagan2019}        & 20.75$\pm$1.75          & 15.51$\pm$1.13          & 15.45$\pm$2.76          & 36.62$\pm$0.30          & 32.34$\pm$1.01          & 36.96$\pm$0.00          & 58.73$\pm$10.45         & 65.35$\pm$7.02          & 53.80$\pm$11.40         & -              \\
    GCFL~\cite{NEURIPS2021_9c6947bd}          & 30.40$\pm$0.16          & 29.44$\pm$0.49          & 26.73$\pm$0.19          & 36.92$\pm$0.00          & 36.86$\pm$0.14          & 36.96$\pm$0.00          & 72.04$\pm$0.13          & 71.14$\pm$0.09          & 47.77$\pm$0.14          & -              \\
    FedGNN~\cite{wu2021fedgnn}        & 30.26$\pm$0.11          & 29.09$\pm$0.01          & 26.60$\pm$0.02          & 36.92$\pm$0.00          & 36.72$\pm$0.00          & 36.96$\pm$0.00          & 72.03$\pm$0.13          & 71.12$\pm$0.09          & \underline{ 71.71$\pm$0.27}    & -              \\
    FedSage+\cite{NEURIPS2021_34adeb8e}      & 57.26$\pm$0.00          & 49.07$\pm$0.00          & 38.36$\pm$0.00          & 36.82$\pm$0.00          & 36.71$\pm$0.00          & 37.03$\pm$0.00          & \textbf{77.74$\pm$0.00} & \underline{ 72.80$\pm$0.00}    & 69.70$\pm$0.00          & -              \\
    FED-PUB~\cite{baek2023personalized}       & 40.80$\pm$0.26          & 36.77$\pm$0.30          & 32.67$\pm$0.39          & \underline{ 44.41$\pm$0.41}    & \underline{ 44.85$\pm$0.17}    & \underline{ 45.39$\pm$0.50}    & 72.18$\pm$0.02          & 71.56$\pm$0.05          & 70.72$\pm$0.40          & -              \\
    FedGTA~\cite{li2023fedgta}        & 61.56$\pm$0.27    & 60.94$\pm$0.19    & 59.65$\pm$0.28    & 41.22$\pm$0.66          & 39.40$\pm$0.44          & 39.24$\pm$0.12          & 45.60$\pm$1.41          & 64.97$\pm$0.35          & 49.63$\pm$8.64          & -              \\
    AdaFGL~\cite{li2024adafgl}        & \underline{67.64$\pm$0.18}          & \underline{64.55$\pm$0.00}          & \underline{62.42$\pm$0.26}          & 41.70$\pm$0.06          & 42.30$\pm$0.00          & 42.59$\pm$0.14          & 47.45$\pm$2.10                  & 65.59$\pm$0.56                   & 51.48$\pm$7.14                         & -          \\ \hline
    FedIIH (Ours) & \textbf{68.32$\pm$0.05} & \textbf{66.44$\pm$0.28} & \textbf{64.61$\pm$0.13} & 44.26$\pm$0.24          & 44.24$\pm$0.10          & 45.19$\pm$0.04          & \underline{ 74.29$\pm$0.02}    & \textbf{73.23$\pm$0.04} & \textbf{72.81$\pm$0.02} &                \\ \hline
                  & \multicolumn{3}{c}{Tolokers}                                                & \multicolumn{3}{c}{Questions}                                               & \multicolumn{4}{c}{Avg.}                                                                     \\ \cline{2-11} 
    Methods       & 5 Clients               & 10 Clients              & 20 Clients              & 5 Clients               & 10 Clients              & 20 Clients              & 5 Clients               & 10 Clients              & 20 Clients              & All            \\ \hline
    Local         & 67.81$\pm$0.17          & 70.04$\pm$0.23          & 62.34$\pm$0.67          & 66.73$\pm$0.57          & 57.96$\pm$0.10          & 60.00$\pm$0.21          & 56.91                   & 54.45                   & 52.31                   & 54.56          \\ \hline
    FedAvg~\cite{mcmahan2017communication}        & 60.74$\pm$0.31          & 54.73$\pm$0.50          & 56.36$\pm$0.39          & 65.68$\pm$0.23          & 58.91$\pm$0.22          & 60.33$\pm$0.15          & 55.84                   & 52.51                   & 52.45                   & 53.60          \\
    FedProx~\cite{MLSYS2020_1f5fe839}       & 42.90$\pm$0.24          & 41.15$\pm$0.22          & 40.42$\pm$0.62          & 47.36$\pm$0.38          & 45.46$\pm$0.34          & 46.83$\pm$0.11          & 45.41                   & 44.11                   & 43.77                   & 44.43          \\
    FedPer~\cite{Arivazhagan2019}        & 46.61$\pm$9.88          & 54.97$\pm$13.23         & 44.82$\pm$11.61         & 58.38$\pm$9.39          & 59.40$\pm$9.71          & 62.32$\pm$1.56          & 44.22                   & 45.51                   & 42.67                   & 44.13          \\
    GCFL~\cite{NEURIPS2021_9c6947bd}          & 27.61$\pm$2.55          & 19.81$\pm$0.57          & 17.53$\pm$0.04          & 47.94$\pm$0.41          & 45.71$\pm$0.25          & 47.47$\pm$0.21          & 42.98                   & 40.59                   & 35.29                   & 39.62          \\
    FedGNN~\cite{wu2021fedgnn}        & 43.10$\pm$0.27          & 41.57$\pm$0.07          & 40.70$\pm$0.74          & 47.55$\pm$0.02          & 45.73$\pm$0.26          & 47.46$\pm$0.25          & 45.97                   & 44.85                   & 44.69                   & 45.17          \\
    FedSage+\cite{NEURIPS2021_34adeb8e}      & \textbf{75.06$\pm$0.00} & 71.31$\pm$0.00          & \underline{ 69.73$\pm$0.00}    & 64.95$\pm$0.00          & \underline{ 65.06$\pm$0.00}    & 59.33$\pm$0.00          & \underline{ 62.37}             & \underline{ 58.99}             & 54.83                   & \underline{ 58.73}    \\
    FED-PUB~\cite{baek2023personalized}       & 70.88$\pm$0.58          & \textbf{72.46$\pm$0.68} & 65.26$\pm$0.59          & \underline{ 67.71$\pm$3.99}    & 54.91$\pm$0.42          & 62.48$\pm$2.92    & 59.20                   & 56.11                   & \underline{ 55.30}             & 56.87          \\
    FedGTA~\cite{li2023fedgta}        & 33.33$\pm$0.51          & 49.97$\pm$2.68          & 50.68$\pm$3.94          & 53.61$\pm$0.36          & 53.79$\pm$0.41          & 61.70$\pm$0.35          & 47.06                   & 53.81                   & 52.18                   & 51.02          \\
    AdaFGL~\cite{li2024adafgl}        & 34.41$\pm$0.63          & 49.82$\pm$2.17          & 50.62$\pm$4.19          & 54.18$\pm$0.45          & 54.87$\pm$0.52          & \underline{62.84$\pm$0.49}          & 49.08                   & 55.43                   & 53.99                         & 52.83          \\ \hline
    FedIIH (Ours) & \underline{ 71.09$\pm$0.26}    & \underline{ 71.32$\pm$0.09}    & \textbf{70.30$\pm$0.10} & \textbf{68.32$\pm$0.03} & \textbf{67.99$\pm$0.09} & \textbf{65.40$\pm$0.07} & \textbf{65.26}          & \textbf{64.64}          & \textbf{63.66}          & \textbf{64.52} \\ \hline
        \end{tabular}
         }
        \end{table*}

    \begin{table*}[]
        \centering
        \scriptsize
        \caption{Node classification results of different methods on the \textbf{heterophilic} graph datasets in the \textbf{overlapping} subgraph partitioning setting. Accuracy (\%) is reported for \textit{Roman-empire} and \textit{Amazon-ratings}, and AUC (\%) is reported for \textit{Minesweeper}, \textit{Tolokers}, and \textit{Questions}. The best and second-best results are shown in \textbf{bold} and \underline{underlined}, respectively.}
        \label{table4}
      \renewcommand{\arraystretch}{0.82} 
         \scalebox{0.83}{
    \begin{tabular}{lcccccccccc}
    \hline
    \textbf{}     & \multicolumn{3}{c}{Roman-empire}                                            & \multicolumn{3}{c}{Amazon-ratings}                                          & \multicolumn{3}{c}{Minesweeper}                                                   & -              \\ \cline{2-11} 
    Methods       & 10 Clients              & 30 Clients              & 50 Clients              & 10 Clients              & 30 Clients              & 50 Clients              & 10 Clients              & 30 Clients              & 50 Clients                    & -              \\ \hline
    Local         & 39.47$\pm$0.03          & 34.43$\pm$0.14          & 31.28$\pm$0.18          & 41.43$\pm$0.04          & 41.81$\pm$0.14          & 42.57$\pm$0.12          & 67.98$\pm$0.07          & 64.39$\pm$0.10          & 62.73$\pm$0.23                & -              \\ \hline
    FedAvg~\cite{mcmahan2017communication}        & 40.89$\pm$0.25          & 38.66$\pm$0.08          & 36.71$\pm$0.20          & 39.86$\pm$0.06          & 41.40$\pm$0.02          & 41.02$\pm$0.16          & 69.06$\pm$0.07          & 67.95$\pm$0.04          & 66.89$\pm$0.08                & -              \\
    FedProx~\cite{MLSYS2020_1f5fe839}       & 36.63$\pm$0.14          & 35.31$\pm$0.17          & 33.61$\pm$0.59          & 37.00$\pm$0.00          & 36.60$\pm$0.00          & 36.89$\pm$0.00          & 68.27$\pm$0.05          & 66.75$\pm$0.19          & 66.03$\pm$0.16                & -              \\
    FedPer~\cite{Arivazhagan2019}        & 23.66$\pm$3.27          & 23.27$\pm$3.09          & 22.23$\pm$3.58          & 32.33$\pm$4.23          & 31.58$\pm$0.54          & 34.48$\pm$2.25          & 61.85$\pm$1.02          & 60.13$\pm$1.38          & 60.06$\pm$3.61                & -              \\
    GCFL~\cite{NEURIPS2021_9c6947bd}          & 37.65$\pm$0.27          & 36.32$\pm$0.19          & 34.80$\pm$0.09          & 37.00$\pm$0.00          & 36.60$\pm$0.00          & 36.89$\pm$0.00          & 68.47$\pm$0.06          & 67.13$\pm$0.10          & 57.41$\pm$12.56               & -              \\
    FedGNN~\cite{wu2021fedgnn}        & 37.46$\pm$0.12          & 36.30$\pm$0.16          & 34.84$\pm$0.13          & 37.00$\pm$0.00          & 36.60$\pm$0.00          & 36.89$\pm$0.00          & 68.47$\pm$0.06          & 67.12$\pm$0.11          & 66.41$\pm$0.23                & -              \\
    FedSage+\cite{NEURIPS2021_34adeb8e}      & 57.48$\pm$0.00          & 42.55$\pm$0.00          & 33.99$\pm$0.00          & 36.86$\pm$0.00          & 36.71$\pm$0.00          & 37.03$\pm$0.00          & \textbf{76.64$\pm$0.00} & \textbf{70.56$\pm$0.00} &  \textbf{70.34$\pm$0.00} & -              \\
    FED-PUB~\cite{baek2023personalized}       & 43.80$\pm$0.25          & 40.46$\pm$0.16          & 37.73$\pm$0.09          & \underline{ 42.25$\pm$0.25}    & \underline{ 42.25$\pm$0.06}    & \textbf{42.88$\pm$0.34} & 68.80$\pm$0.09          & 67.43$\pm$0.25          & 65.98$\pm$0.15                & -              \\
    FedGTA~\cite{li2023fedgta}        & 59.86$\pm$0.04    & 58.32$\pm$0.09    & 57.57$\pm$0.21    & 40.81$\pm$0.24          & 39.44$\pm$0.06          & 39.37$\pm$0.04          & 54.35$\pm$0.73          & 48.20$\pm$1.28          & 52.94$\pm$1.77                & -              \\
    AdaFGL~\cite{li2024adafgl}        & \underline{64.44$\pm$0.03}          & \underline{61.77$\pm$0.02}          & \underline{59.55$\pm$0.01}          & 39.39$\pm$0.05          & 41.19$\pm$0.15          & 40.71$\pm$0.25          & 55.15$\pm$0.84                   & 50.15$\pm$1.63                   & 54.18$\pm$2.15                         & -          \\ \hline
    FedIIH (Ours) & \textbf{65.48$\pm$0.12} & \textbf{63.32$\pm$0.06} & \textbf{62.42$\pm$0.10} & \textbf{42.63$\pm$0.02} & \textbf{42.40$\pm$0.05} & \underline{ 42.65$\pm$0.21}    & \underline{ 69.35$\pm$0.25}    & \underline{ 68.09$\pm$0.26}    & \underline{ 67.37$\pm$0.14} & -              \\ \hline
                  & \multicolumn{3}{c}{Tolokers}                                                & \multicolumn{3}{c}{Questions}                                               & \multicolumn{4}{c}{Avg.}                                                                           \\ \cline{2-11} 
    Methods       & 10 Clients              & 30 Clients              & 50 Clients              & 10 Clients              & 30 Clients              & 50 Clients              & 10 Clients              & 30 Clients              & 50 Clients                    & All            \\ \hline
    Local         & 73.83$\pm$0.03          & 69.01$\pm$0.31          & 66.63$\pm$0.20          & 63.17$\pm$0.02          & 57.17$\pm$0.08          & 56.13$\pm$0.02          & 57.18                   & 53.36                   & 51.87                         & 54.14          \\ \hline
    FedAvg~\cite{mcmahan2017communication}        & 72.99$\pm$0.40          & 58.51$\pm$0.27          & 55.47$\pm$0.42          & 62.80$\pm$0.63          & 58.88$\pm$0.18          & 60.78$\pm$0.27          & 57.12                   & 53.08                   & 52.17                         & 54.12          \\
    FedProx~\cite{MLSYS2020_1f5fe839}       & 54.49$\pm$1.69          & 45.59$\pm$0.41          & 41.49$\pm$0.45          & 52.53$\pm$0.34          & 51.54$\pm$0.41          & 50.72$\pm$0.40          & 49.78                   & 47.16                   & 45.75                         & 47.56          \\
    FedPer~\cite{Arivazhagan2019}        & 39.60$\pm$0.11          & 59.44$\pm$0.79          & 41.92$\pm$0.06          & 61.31$\pm$0.29          & 53.41$\pm$1.53          & 50.29$\pm$0.10          & 43.75                   & 45.57                   & 41.80                         & 43.70          \\
    GCFL~\cite{NEURIPS2021_9c6947bd}          & 55.91$\pm$1.13          & 23.26$\pm$0.70          & 18.40$\pm$0.25          & 53.04$\pm$0.47          & 51.84$\pm$0.38          & 51.10$\pm$0.38          & 50.41                   & 43.03                   & 39.72                         & 44.39          \\
    FedGNN~\cite{wu2021fedgnn}        & 56.21$\pm$1.20          & 46.85$\pm$0.31          & 42.18$\pm$0.45          & 53.04$\pm$0.47          & 51.86$\pm$0.36          & 51.11$\pm$0.38          & 50.44                   & 47.75                   & 46.29                         & 48.16          \\
    FedSage+\cite{NEURIPS2021_34adeb8e}      & \textbf{74.54$\pm$0.00} & \underline{ 70.88$\pm$0.00}    & \underline{ 69.61$\pm$0.00}    & 64.22$\pm$0.00          & \underline{ 65.34$\pm$0.00}    & \underline{ 62.76$\pm$0.00}    & \underline{ 61.95}             & \underline{ 57.21}             & \underline{ 54.75}                   & \underline{57.97}    \\
    FED-PUB~\cite{baek2023personalized}       & \underline{ 74.17$\pm$0.29}    & 70.35$\pm$0.54          & 66.80$\pm$0.85          & \underline{ 65.39$\pm$2.44}    & 58.38$\pm$1.19          & 58.76$\pm$0.16          & 58.88                   & 55.77                   & 54.43                         & 56.36          \\
    FedGTA~\cite{li2023fedgta}        & 40.02$\pm$1.70          & 47.34$\pm$0.75          & 45.81$\pm$1.96          & 35.56$\pm$5.46          & 50.43$\pm$1.05          & 53.33$\pm$0.40          & 46.12                   & 48.75                   & 49.80                         & 48.22          \\
    AdaFGL~\cite{li2024adafgl}        & 45.15$\pm$2.15          & 49.18$\pm$0.84          & 47.54$\pm$2.48          & 41.05$\pm$6.49          & 52.18$\pm$2.16          & 56.46$\pm$0.92          & 49.04                   & 50.89                   & 51.69                         & 50.54          \\ \hline
    FedIIH (Ours) & 71.67$\pm$0.02          & \textbf{71.69$\pm$0.12} & \textbf{69.99$\pm$0.03} & \textbf{68.79$\pm$0.09} & \textbf{66.98$\pm$0.04} & \textbf{64.73$\pm$0.35} & \textbf{63.58}          & \textbf{62.50}          & \textbf{61.43}                & \textbf{62.50} \\ \hline
        \end{tabular}
         }
\end{table*}

\subsection{Ablation Study}
Our FedIIH employs the hierarchical model and variational inference to compute the inter-subgraph similarities. In addition, we use the disentanglement to learn disentangled representations. To shed light on the contributions of these components, we report the experimental results of FedIIH when each of these components is removed on the \textit{Cora} and \textit{Amazon-ratings} datasets in Tab.~\ref{table5}. For simplicity, `w/o HM', `w/o VI', and `w/o Dis' denote the reduced models by removing the hierarchical model, variational inference, and disentanglement, respectively. We can clearly find that the performance decreases when any component is removed, showing that each component contributes a lot to the final performance. For example, the accuracies on the \textit{Cora} dataset are obviously reduced when the disentanglement component is removed. The ablation studies on other datasets are shown in the Appendix~J.1.

\begin{table}[]
    \centering
    \scriptsize
    \caption{Ablation studies in both non-overlapping and overlapping partitioning settings on two datasets with 10 clients.}
    \label{table5}
    \renewcommand{\arraystretch}{0.8} 
     \scalebox{0.75}{
        \begin{tabular}{lcccc}
    \hline
    \multicolumn{1}{c}{} & \multicolumn{2}{c}{Cora}                          & \multicolumn{2}{c}{Amazon-ratings}                \\ \hline
    Methods              & non-overlapping         & overlapping             & non-overlapping         & overlapping             \\ \hline
    w/o HM      & 78.67$\pm$1.17 ($\downarrow$3.18)          & 78.58$\pm$0.03 ($\downarrow$1.99)         & 41.69$\pm$0.09 ($\downarrow$2.55)          & 38.04$\pm$0.09 ($\downarrow$4.59)          \\ \hline
    w/o VI      & 73.51$\pm$0.52 ($\downarrow$8.34)          & 78.40$\pm$0.25 ($\downarrow$2.17)         & 41.68$\pm$0.05 ($\downarrow$2.56)          & 40.70$\pm$0.42 ($\downarrow$1.93)          \\ \hline
    w/o Dis     & 78.78$\pm$0.63 ($\downarrow$3.07)          & 77.18$\pm$0.23 ($\downarrow$3.39)         & 41.20$\pm$0.14 ($\downarrow$3.04)          & 39.98$\pm$0.06 ($\downarrow$2.65)          \\ \hline
    FedIIH               & \textbf{81.85$\pm$0.09} & \textbf{80.57$\pm$0.23} & \textbf{44.24$\pm$0.10} & \textbf{42.63$\pm$0.02} \\ \hline
    \end{tabular}
     }
    \end{table}

\subsection{Effectiveness of Inter-Subgraph Similarity and $K$}

\begin{figure}
    \centering
    \begin{subfigure}[t]{0.08\textwidth}
        \includegraphics[width=\linewidth]{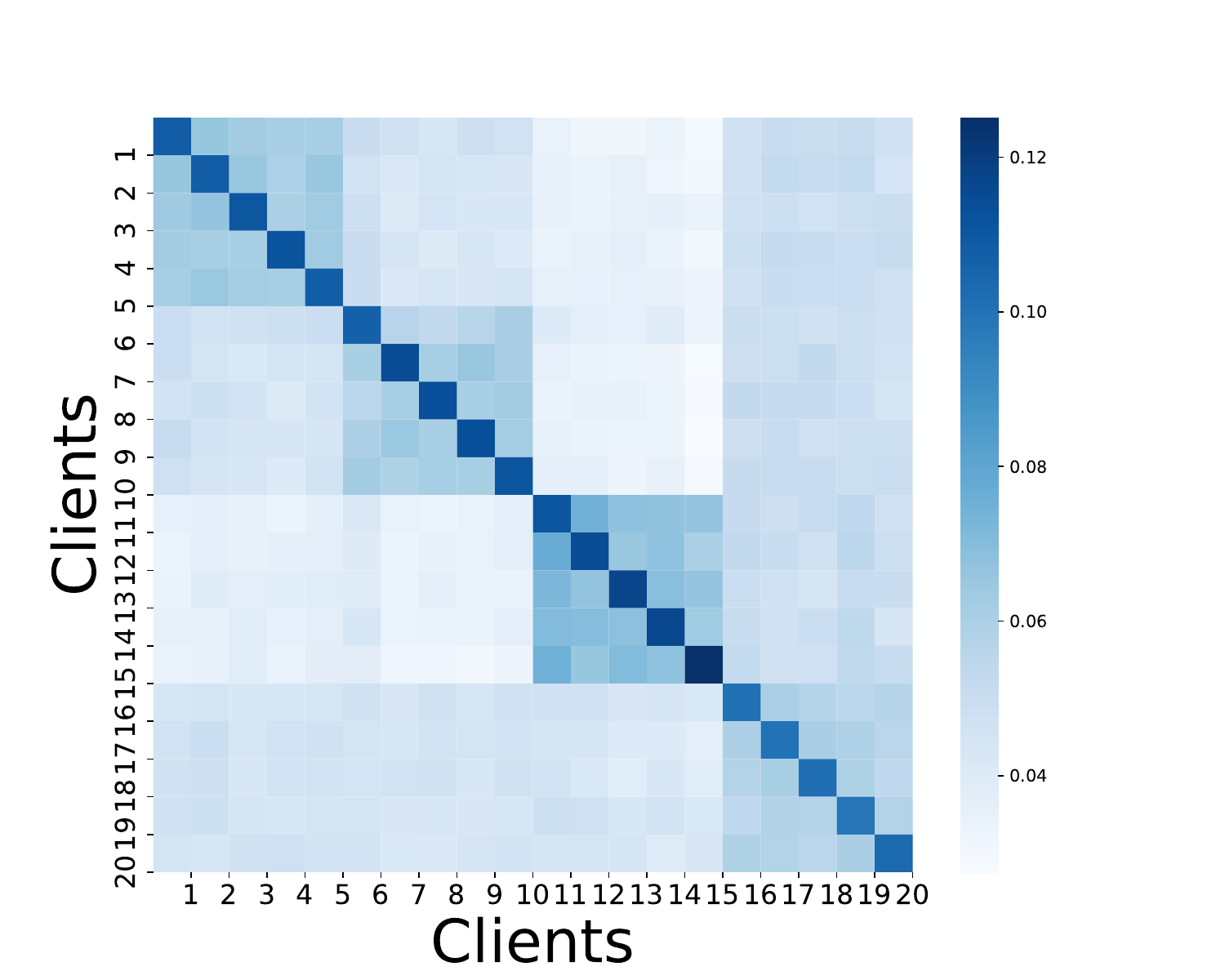}
        \captionsetup{font=tiny}
        \caption{Distr. Sim.} 
        \label{fig_s2_1}
    \end{subfigure}%
    \hfill
    \begin{subfigure}[t]{0.08\textwidth}
        \includegraphics[width=\linewidth]{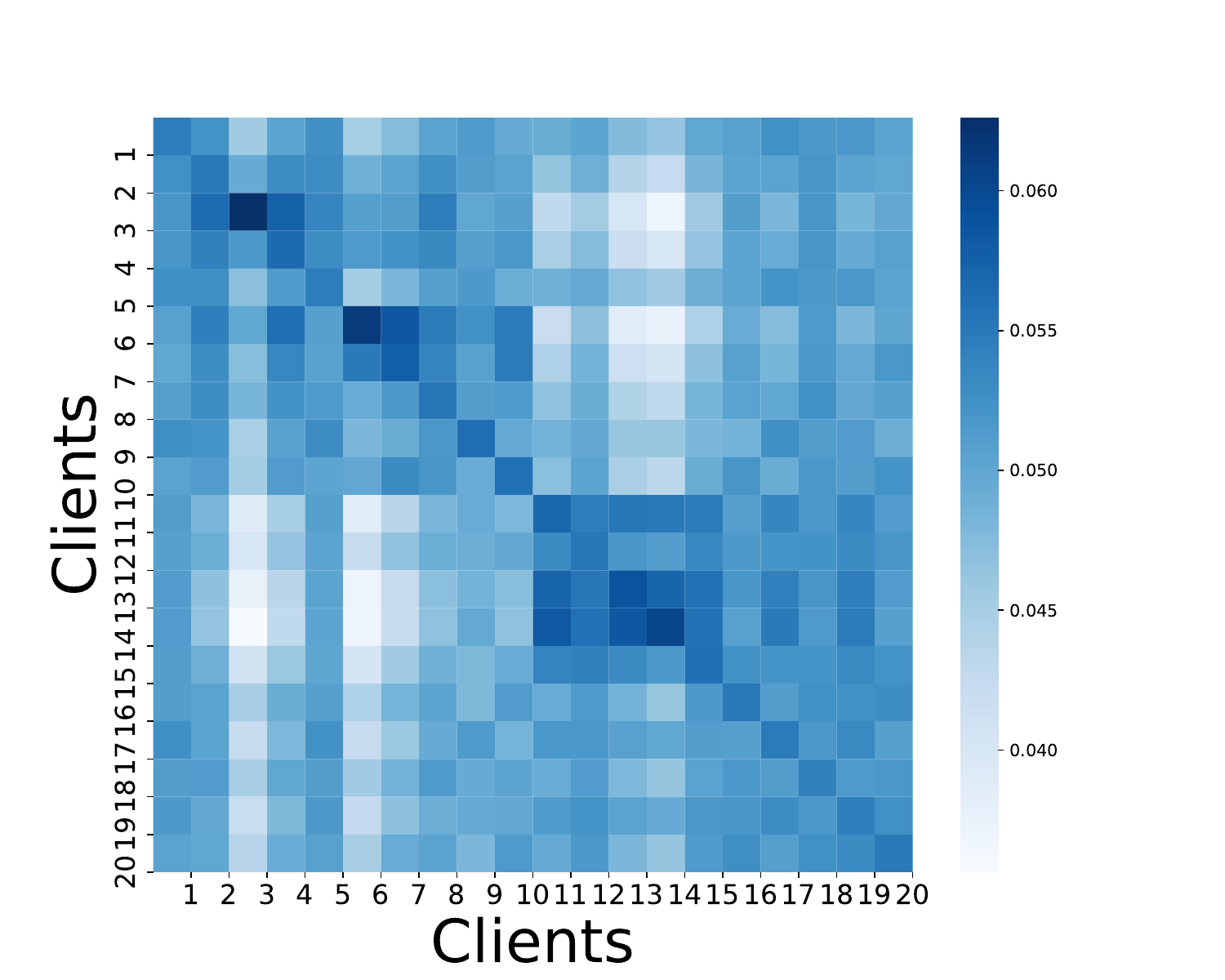}
        \captionsetup{font=tiny}
        \caption{FED-PUB} 
        \label{fig_s2_2}
    \end{subfigure}%
    \hfill
    \begin{subfigure}[t]{0.08\textwidth}
        \includegraphics[width=\linewidth]{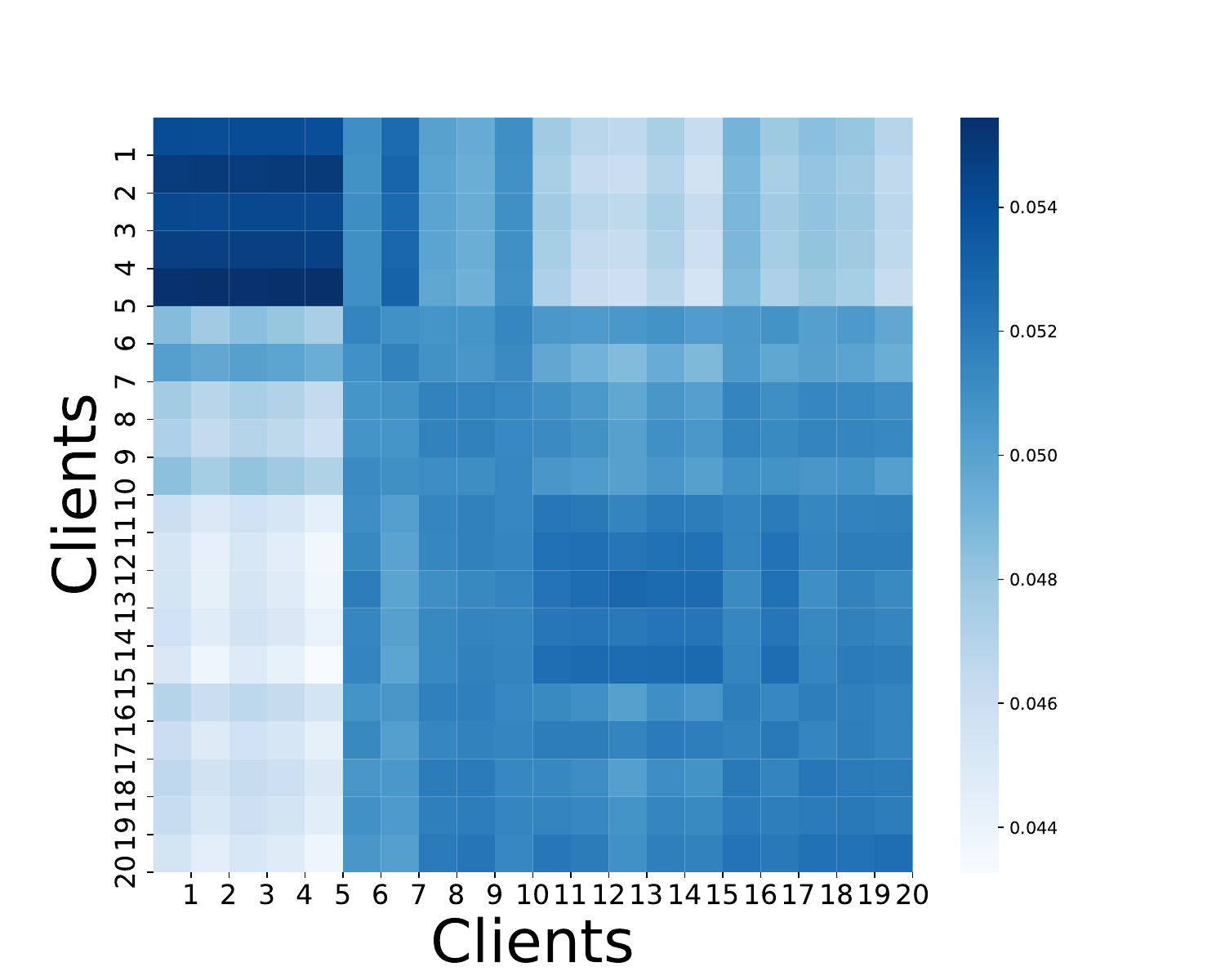}
        \captionsetup{font=tiny}
        \caption{FedGTA} 
        \label{fig_s2_3}
    \end{subfigure}%
    \hfill
    \begin{subfigure}[t]{0.08\textwidth}
        \includegraphics[width=\linewidth]{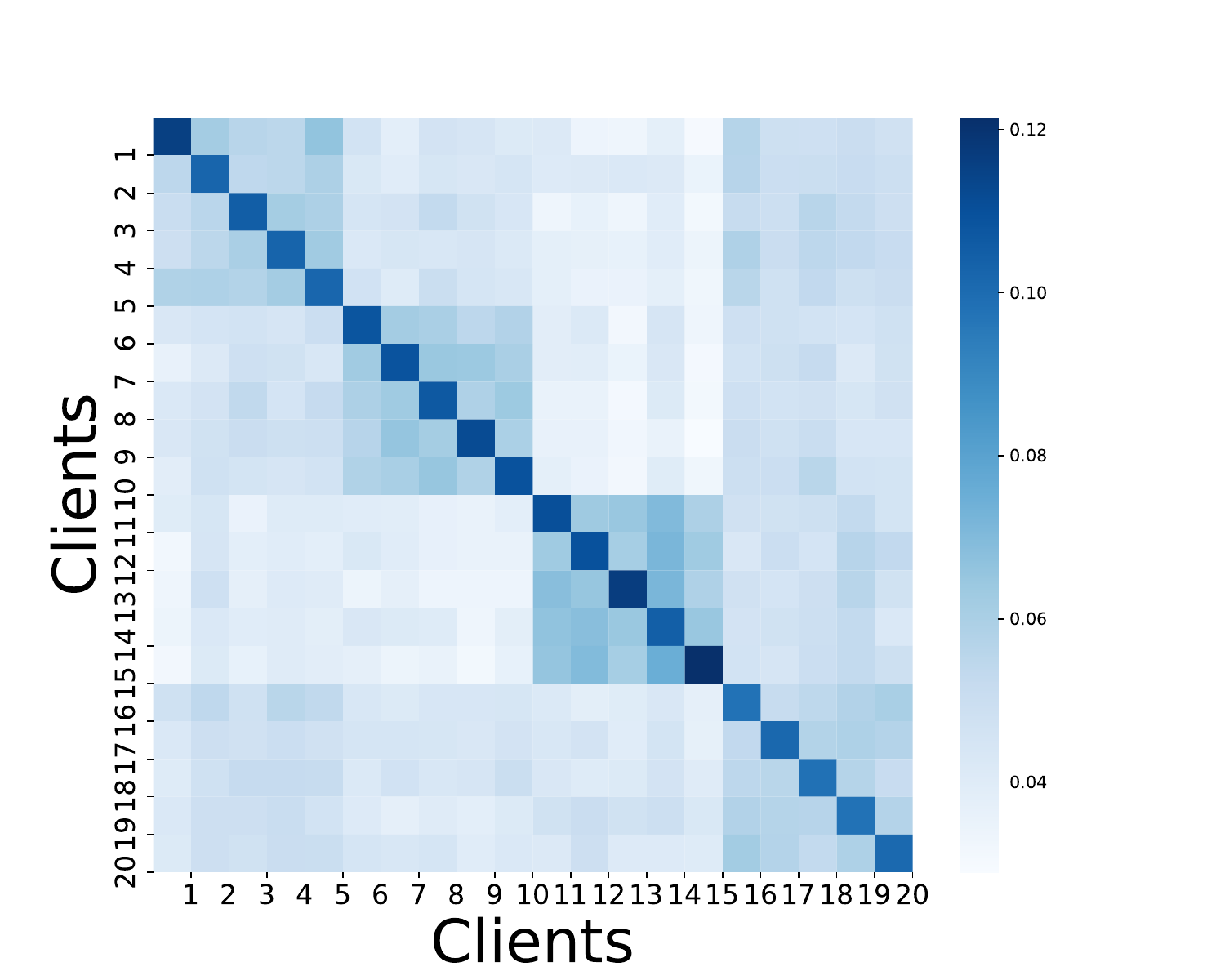}
        \captionsetup{font=tiny}
        \caption{FedIIH of the 1st latent factor ($K=2$)} 
        \label{fig_s2_4}
    \end{subfigure}
    \hfill
    \begin{subfigure}[t]{0.08\textwidth}
        \includegraphics[width=\linewidth]{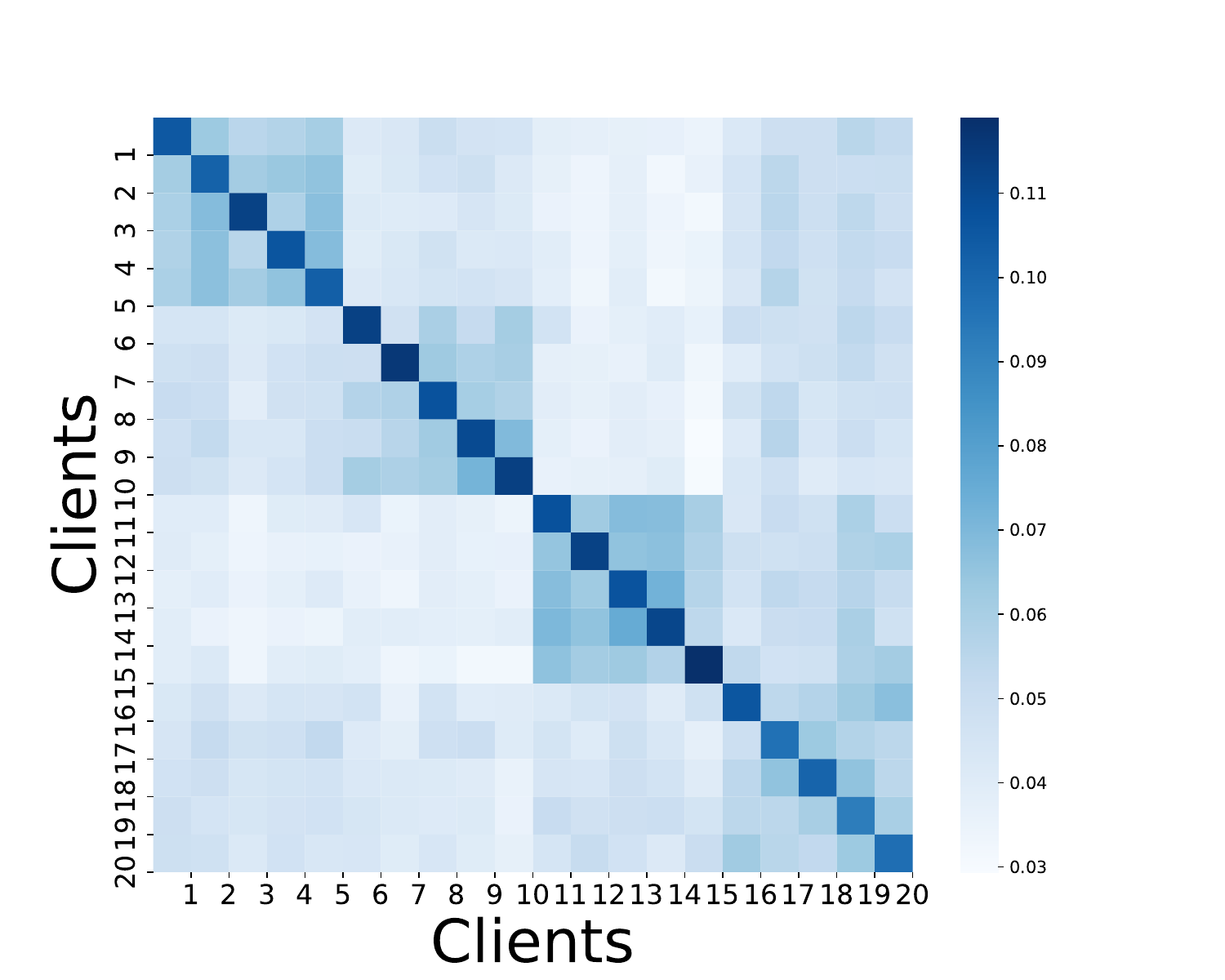}
        \captionsetup{font=tiny}
        \caption{FedIIH of the 2nd latent factor ($K=2$)}
        \label{fig_s2_5}
    \end{subfigure}
    \captionsetup{font=small}
    \caption{\small Similarity heatmaps on the \textit{Amazon-ratings} dataset in the overlapping setting with 20 clients.}
    \label{fig_s2}
    \vspace{-15pt}
\end{figure}

\begin{figure}[t]
	\centering
	\includegraphics[width=5cm]{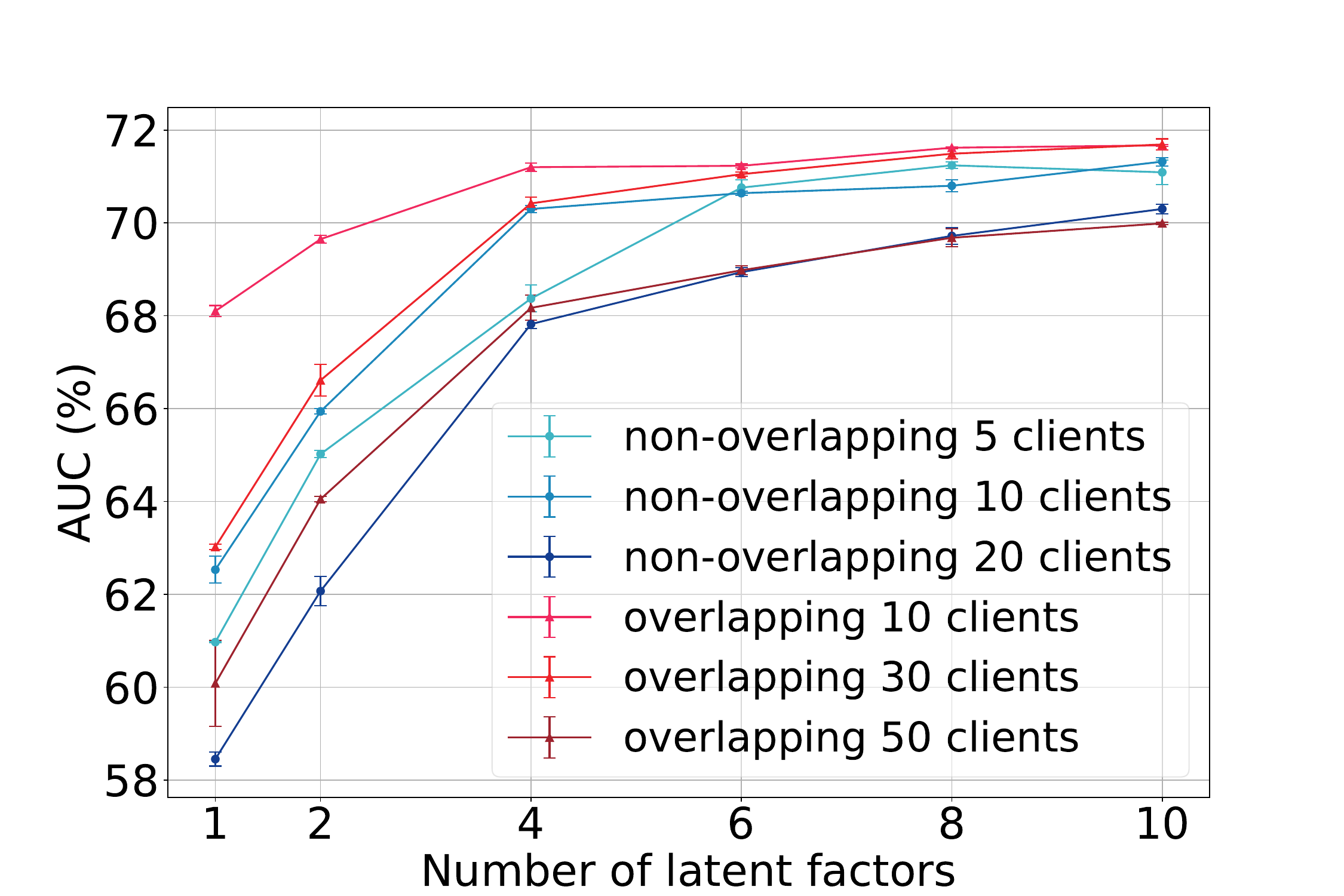}
	\caption{Since the intra-heterogeneity of the \textit{Tolokers} dataset is high, the performances increase consistently as the value of $K$ increases.}
	\label{fig3}
    \vspace{-10pt}
\end{figure}

First, we validate the effectiveness of the inter-subgraph similarities calculated by our method. In Fig.~\eqref{fig_s2}, we show the similarity heatmaps on the \textit{Amazon-ratings} dataset in the overlapping setting with 20 clients. The similarity heatmaps on other datasets are shown in the Appendix~K.2. Baek~\textit{et al.}~\cite{baek2023personalized} regard two clients have similar data distributions if they have similar label distributions. We argue that the label similarity cannot fully reflect the similarity of the subgraph data distribution, because both the node features and the structure are crucial properties of subgraphs. Therefore, we compute the subgraph distribution similarity by using the JS divergence, taking into account both semantic and structural information, and we treat such distribution similarity (Fig.~\ref{fig_s2_1}) as the ground truth. Both Fig.~\ref{fig_s2_4} and Fig.~\ref{fig_s2_5} are close to the ground truth, verifying the effectiveness of our similarity calculation scheme based on the inferred subgraph data distributions. In contrast, the similarity heatmaps of FED-PUB (Fig.~\ref{fig_s2_2}) and FedGTA (Fig.~\ref{fig_s2_3}) are both different from the ground truth, which implies that their calculated similarities are not good enough. Moreover, as shown in the Appendix~K.1, our calculated similarities are much more stable than the similarities calculated by FED-PUB. Second, we investigate the effectiveness of $K$. As shown in Fig.~\ref{fig3}, the performances on the \textit{Tolokers} dataset increase consistently as the value of $K$ increases. More results are provided in the Appendix~J.2. We can find that for most datasets the performances are usually bad when $K=1$, and it means that disentangling the subgraph into several latent factors is beneficial to improve the performances. Last but not least, the efficiency and robustness analysis in the Appendix~L and Appendix~M clearly shows that our FedIIH is more efficient and robust than the baseline methods.

\vspace{-5pt}

\section{Conclusion}
In this paper, we proposed a novel method FedIIH, which naturally integrates the inter- and intra- heterogeneity in GFL. On one hand, our new method characterizes the inter-heterogeneity from a multi-level global perspective, and thus it can properly compute the inter-subgraph similarities based on the whole distribution. On the other hand, it disentangles the subgraph into several latent factors, so that we can further consider the critical intra-heterogeneity. To the best of our knowledge, this is the first time in GFL that combines both inter- and intra- heterogeneity into a unified framework. Due to the adequate consideration of inter-intra heterogeneity, our method achieves satisfactory results on eleven datasets and outperforms the second-best method by a large margin of 5.79\% on the heterophilic graph data.

\section{Acknowledgments}
This research is supported by NSF of China (Nos: 62336003, 12371510), and NSF for Distinguished Young Scholar of Jiangsu Province (No: BK20220080).

\bibliography{aaai25}

\appendix
\onecolumn  

\section{A. Details of Neighborhood Routing Mechanism}
\label{Neighborhood_Routing_Mechanism}
Recall that we use the node $u$ as an example to describe the disentangling process of DisenGCN. By using Eq.~\eqref{eq1}, the obtained $\mathbf{z}^{i, k}$ describes the aspect of node $i$ that are related with the $k$-th latent factor. However, $\mathbf{z}^{i, k}$ only denotes the information of the node $i$ itself. Therefore, for the node $u$, we have to mine information from the neighborhoods, which are connected with node $u$ due to the $k$-th latent factor. Specifically, we aim to identify the latent factor that causes the connection between node $u$ and its neighbor node $v$, and accordingly extract features of $v$ that are specific to that factor. This is exactly the purpose of the neighborhood routing mechanism.

First, we define $p^{v, k}$, which represents the probability that the latent factor $k$ is the reason why the node $u$ reaches its neighbor $v$. Then, we can have $p^{v, k} \geq 0$, and $\sum_{k^{\prime}=1}^{K}p^{v, k^{\prime}}=1$. Moreover, $p^{v, k}$ is also the probability that we should use the neighbor $v$ to construct $\mathbf{c}^{u,k}$. The neighborhood routing mechanism will infer $p^{v, k}$ and construct $\mathbf{c}^{u,k}$ in an iterative manner. To initialize this process, we make $({p^{v,k}})^{(1)} \propto \exp[({\mathbf{z}^{v, k}})^{\top }\mathbf{z}^{u, k}/\tau_p]$. Then, it subsequently iterates to identify the largest cluster within each subspace, ensuring that each neighbor is primarily associated with a single subspace cluster:
\begin{equation}\label{eq_A_1}
({\mathbf{c}^{u,k}})^{(t)} = \frac{\mathbf{z}^{u, k}+\sum_{v:(u,v) \in \mathcal{G}_m}({p^{v,k}})^{(t-1)}\mathbf{z}^{v, k}}{||\mathbf{z}^{u, k}+\sum_{v:(u,v) \in \mathcal{G}_m}({p^{v,k}})^{(t-1)}\mathbf{z}^{v, k}||_2},
\end{equation}

\begin{equation}\label{eq_A_2}
({p^{v,k}})^{(t)} = \frac{\exp[({\mathbf{z}^{v, k}})^{\top}({\mathbf{c}^{u,k}})^{(t)}/\tau_p]}{\sum_{k^{\prime}=1}^{K} \exp[({\mathbf{z}^{v, k^{\prime}}})^{\top}({\mathbf{c}^{u,k^{\prime}}})^{(t)}/\tau_p]},
\end{equation}
where $({\mathbf{c}^{u,k}})^{(t)}$ denotes the center of each subspace cluster at the $t$-th iteration, $t=2, 3, \cdots, T$, and $\tau_p$ is a hyperparameter that controls the hardness of the assignment. Here $\tau_p$ is set to 1 according to~\cite{pmlrv97ma19a}. After $T$ times of iterations, the final output is $\mathbf{c}^{u,k}=({\mathbf{c}^{u,k}})^{(T)}$.

\section{B. Instantiations of Distributions}
\label{instantiations_distributions}
We summarize the family of distributions instantiated by our proposed FedIIH in Tab.~\ref{tableA1}. Specifically, the priors $p(\bm{\alpha}^k)$ and $p(\tilde{\mathbf{H}}_{m}^k)$ are both centered isotropic multivariate Gaussian distributions. Besides, the prior over the local latent factor $\tilde{\mathbf{H}}_{m}^k$ conditioned on $\bm{\alpha}^k$ (\textit{i.e.}, $p(\tilde{\mathbf{H}}_{m}^k|\bm{\alpha}^k)$) is an isotropic multivariate Gaussian distribution centered at $\bm{\alpha}^k$. Similarly, the marginal distribution of $\bm{\alpha}^k$ (\textit{i.e.}, $q(\bm{\alpha}^k)$) is a multivariate diagonal Gaussian distribution. As shown in Tab.~\ref{tableA1}, $\tilde{\bm{\alpha}}^k$ and $\sigma_{\tilde{\bm{\alpha}}^k}^2$ denote the posterior mean and variance of $\bm{\alpha}^k$, respectively. Moreover, $\hat{\bm{\mu}}_{\tilde{\mathbf{H}}_{m}^k}$ and $\hat{\bm{\sigma}}^{2}_{\tilde{\mathbf{H}}_{m}^k}$ denote the variational mean and variance evaluated at $\mathcal{G}_m$, respectively.

\begin{table}[]
    \centering
    \caption{Family of distributions instantiated by our proposed FedIIH.}
	\label{tableA1}
    \begin{tabular}{l|l}
    \hline
    \hline
    $p(\bm{\alpha}^k)$ & $\mathcal{N}(0, \sigma_{\bm{\alpha}^k}^2\mathbf{I})$ \\
    $p(\tilde{\mathbf{H}}_{m}^k)$ & $\mathcal{N}(0, \sigma_{\tilde{\mathbf{H}}_{m}^k}^2\mathbf{I})$ \\
    $p(\tilde{\mathbf{H}}_{m}^k|\bm{\alpha}^k)$ & $\mathcal{N}(\bm{\alpha}^k, \sigma_{\tilde{\mathbf{H}}_{m}^k}^2\mathbf{I})$ \\
    \hline
    \hline
    $q(\bm{\alpha}^k)$ & $\mathcal{N}(\tilde{\bm{\alpha}}^k, \sigma_{\tilde{\bm{\alpha}}^k}^2\mathbf{I})$ \\
    $q(\tilde{\mathbf{H}}_{m}^k|\mathcal{G}_m)$ & $\mathcal{N}(\hat{\bm{\mu}}_{\tilde{\mathbf{H}}_{m}^k}, \hat{\bm{\sigma}}^{2}_{\tilde{\mathbf{H}}_{m}^k})$ \\
    \hline
    \hline
    \end{tabular}
    \end{table}

\section{C. Derivation of the ELBO}
\label{derivation_ELBO}
First, the ELBO for the marginal likelihood of $\mathcal{G}_{1:M}$ (\textit{i.e.}, $\log p(\mathcal{G}_{1:M})$) can be obtained by using the Jensen inequality:
\begin{equation}\label{eq_elbo_1}
\tiny
\refstepcounter{equation}
\begin{aligned}
&\log p(\mathcal{G}_{1:M})\\
&= \log \sum_{\tilde{\mathbf{H}}_{1:M}^1, \tilde{\mathbf{H}}_{1:M}^2, \cdots, \tilde{\mathbf{H}}_{1:M}^K, \bm{\alpha}^{1:K}|\mathcal{G}_{1:M}} q(\tilde{\mathbf{H}}_{1:M}^1, \tilde{\mathbf{H}}_{1:M}^2, \cdots, \tilde{\mathbf{H}}_{1:M}^K, \bm{\alpha}^{1:K}|\mathcal{G}_{1:M}) \frac{p(\mathcal{G}_{1:M}, \tilde{\mathbf{H}}_{1:M}^1, \tilde{\mathbf{H}}_{1:M}^2, \cdots, \tilde{\mathbf{H}}_{1:M}^K, \bm{\alpha}^{1:K})}{q(\tilde{\mathbf{H}}_{1:M}^1, \tilde{\mathbf{H}}_{1:M}^2, \cdots, \tilde{\mathbf{H}}_{1:M}^K, \bm{\alpha}^{1:K}|\mathcal{G}_{1:M})},\\ 
&\geqq \sum_{\tilde{\mathbf{H}}_{1:M}^1, \tilde{\mathbf{H}}_{1:M}^2, \cdots, \tilde{\mathbf{H}}_{1:M}^K, \bm{\alpha}^{1:K}|\mathcal{G}_{1:M}} q(\tilde{\mathbf{H}}_{1:M}^1, \tilde{\mathbf{H}}_{1:M}^2, \cdots, \tilde{\mathbf{H}}_{1:M}^K, \bm{\alpha}^{1:K}|\mathcal{G}_{1:M}) \log \frac{p(\mathcal{G}_{1:M}, \tilde{\mathbf{H}}_{1:M}^1, \tilde{\mathbf{H}}_{1:M}^2, \cdots, \tilde{\mathbf{H}}_{1:M}^K, \bm{\alpha}^{1:K})}{q(\tilde{\mathbf{H}}_{1:M}^1, \tilde{\mathbf{H}}_{1:M}^2, \cdots, \tilde{\mathbf{H}}_{1:M}^K, \bm{\alpha}^{1:K}|\mathcal{G}_{1:M})},\\ 
&\triangleq \mathrm{ELBO}\Big(q(\tilde{\mathbf{H}}_{1:M}^1, \tilde{\mathbf{H}}_{1:M}^2, \cdots, \tilde{\mathbf{H}}_{1:M}^K, \bm{\alpha}^{1:K}|\mathcal{G}_{1:M}), \mathcal{G}_{1:M}\Big),\\ 
&= \mathbb{E}_{q(\tilde{\mathbf{H}}_{1:M}^1, \tilde{\mathbf{H}}_{1:M}^2, \cdots, \tilde{\mathbf{H}}_{1:M}^K, \bm{\alpha}^{1:K}|\mathcal{G}_{1:M})}\Big[\log p(\mathcal{G}_{1:M}, \tilde{\mathbf{H}}_{1:M}^1, \tilde{\mathbf{H}}_{1:M}^2, \cdots, \tilde{\mathbf{H}}_{1:M}^K, \bm{\alpha}^{1:K})\\
&\quad -\log q(\tilde{\mathbf{H}}_{1:M}^1, \tilde{\mathbf{H}}_{1:M}^2, \cdots, \tilde{\mathbf{H}}_{1:M}^K, \bm{\alpha}^{1:K}|\mathcal{G}_{1:M})\Big].
\end{aligned}
\tag{\theequation}
\normalsize
\end{equation}

Second, inspired by~\cite{Hsu_Zhang_Glass_2017}, the ELBO can be derived as follows:
\begin{equation}\label{eq_elbo_2}
    \begin{aligned}
    &\quad \mathbb{E}_{q(\tilde{\mathbf{H}}_{1:M}^1, \tilde{\mathbf{H}}_{1:M}^2, \cdots, \tilde{\mathbf{H}}_{1:M}^K, \bm{\alpha}^{1:K}|\mathcal{G}_{1:M})}\big[\log p(\mathcal{G}_{1:M}, \tilde{\mathbf{H}}_{1:M}^1, \tilde{\mathbf{H}}_{1:M}^2, \cdots, \tilde{\mathbf{H}}_{1:M}^K, \bm{\alpha}^{1:K}) \\
&\quad - \log q(\tilde{\mathbf{H}}_{1:M}^1, \tilde{\mathbf{H}}_{1:M}^2, \cdots, \tilde{\mathbf{H}}_{1:M}^K, \bm{\alpha}^{1:K}|\mathcal{G}_{1:M})\big]\\
    &=\sum_{m=1}^{M} \mathbb{E}_{q(\tilde{\mathbf{H}}_m^1, \tilde{\mathbf{H}}_m^2, \cdots, \tilde{\mathbf{H}}_m^K|\mathcal{G}_m)} \big[\log p(\mathcal{G}_m|\tilde{\mathbf{H}}_m^1, \tilde{\mathbf{H}}_m^2, \cdots, \tilde{\mathbf{H}}_m^K)\big]\\
    &\quad - \sum_{m=1}^{M} \sum_{k=1}^{K} \mathbb{E}_{q(\bm{\alpha}^k)}\big[D_{\mathrm{KL}}\big(q(\tilde{\mathbf{H}}_m^k|\mathcal{G}_m) || p(\tilde{\mathbf{H}}_m^k | \bm{\alpha}^k)\big)\big]\\
    &\quad -\sum_{k=1}^{K} D_{\mathrm{KL}}\big(q(\bm{\alpha}^k) || p(\bm{\alpha}^k)\big).
    \end{aligned}
    \end{equation}

Third, we compute the expected KL divergence of two Gaussian distributions (\textit{i.e.}, $q(\tilde{\mathbf{H}}_m^k|\mathcal{G}_m)$ and $p(\tilde{\mathbf{H}}_m^k | \bm{\alpha}^k)$) over a Gaussian distribution (\textit{i.e.}, $q(\bm{\alpha}^k)$) analytically. According to Tab.~\ref{tableA1}, we can have
\begin{equation}\label{eq_elbo_3}
    \begin{aligned}
    & \quad \mathbb{E}_{q(\bm{\alpha}^k)}\big[D_{\mathrm{KL}}\big(q(\tilde{\mathbf{H}}_m^k|\mathcal{G}_m) || p(\tilde{\mathbf{H}}_m^k | \bm{\alpha}^k)\big)\big] \\
&= \mathbb{E}_{q(\bm{\alpha}^k)}\big[D_{\mathrm{KL}}\big(\mathcal{N}(\hat{\bm{\mu}}_{\tilde{\mathbf{H}}_{m}^k}, \hat{\bm{\sigma}}^{2}_{\tilde{\mathbf{H}}_{m}^k}) || \mathcal{N}(\bm{\alpha}^k, \sigma_{\tilde{\mathbf{H}}_{m}^k}^2\mathbf{I})\big)\big]\\
&=\mathbb{E}_{q(\bm{\alpha}^k)}\big[-\frac{1}{2}\sum_{j=1}^J\big(1+\log\frac{\hat{\sigma}^{2}_{\tilde{\mathbf{H}}_{m}^k,j}}{\sigma_{\tilde{\mathbf{H}}_{m}^k}^2}-\frac{(\hat{\mu}_{\tilde{\mathbf{H}}_{m}^k,j}-\alpha^{k,j})^2+\hat{\sigma}^{2}_{\tilde{\mathbf{H}}_{m}^k,j}}{\sigma_{\tilde{\mathbf{H}}_{m}^k}^2}\big)\big]\\
&=-\frac{1}{2}\sum_{j=1}^{J}\big(1+\log \frac{\hat{\sigma}^{2}_{\tilde{\mathbf{H}}_{m}^k,j}}{\sigma_{\tilde{\mathbf{H}}_{m}^k}^2}-\frac{\hat{\sigma}^{2}_{\tilde{\mathbf{H}}_{m}^k,j}}{\sigma_{\tilde{\mathbf{H}}_{m}^k}^2}-\mathbb{E}_{q(\bm{\alpha}^k)}[\frac{(\hat{\mu}_{\tilde{\mathbf{H}}_{m}^k,j}-\alpha^{k,j})^2}{\sigma_{\tilde{\mathbf{H}}_{m}^k}^2}]\big)\\
&=D_{\mathrm{KL}}\big(\mathcal{N}(\hat{\bm{\mu}}_{\tilde{\mathbf{H}}_{m}^k}, \hat{\bm{\sigma}}^{2}_{\tilde{\mathbf{H}}_{m}^k}) || \mathcal{N}(\tilde{\bm{\alpha}}^k, \sigma_{\tilde{\mathbf{H}}_{m}^k}^2)\big) + \frac{J}{2} \frac{\sigma_{\tilde{\bm{\alpha}}^k}^2}{\sigma_{\tilde{\mathbf{H}}_{m}^k}^2}\\
&=D_{\mathrm{KL}}\big(q(\tilde{\mathbf{H}}_m^k|\mathcal{G}_m) || p(\tilde{\mathbf{H}}_m^k | \tilde{\bm{\alpha}}^k)\big) + \frac{J}{2} \frac{\sigma_{\tilde{\bm{\alpha}}^k}^2}{\sigma_{\tilde{\mathbf{H}}_{m}^k}^2},
\end{aligned}
\end{equation}
where $J$ denotes the dimension of $\tilde{\mathbf{H}}_m^k$. Moreover, $\alpha^{k,j}$, $\hat{\mu}_{\tilde{\mathbf{H}}_{m}^k,j}$, and $\hat{\sigma}^{2}_{\tilde{\mathbf{H}}_{m}^k,j}$ denote the $j$-th element of $\bm{\alpha}^{k}$, $\hat{\bm{\mu}}_{\tilde{\mathbf{H}}_{m}^k}$, and $\hat{\bm{\sigma}}^{2}_{\tilde{\mathbf{H}}_{m}^k}$, respectively.

Fourth, the KL divergence between $q(\bm{\alpha}^k)$ and $p(\bm{\alpha}^k)$ can be computed analytically as
\begin{equation}\label{eq_elbo_4}
    \begin{aligned}
    &\quad D_{\mathrm{KL}}\big(q(\bm{\alpha}^k) || p(\bm{\alpha}^k)\big) \\
    &=D_{\mathrm{KL}}\big(\mathcal{N}(\tilde{\bm{\alpha}}^k, \sigma_{\tilde{\bm{\alpha}}^k}^2\mathbf{I})||\mathcal{N}(0, \sigma_{\bm{\alpha}^k}^2\mathbf{I})\big)\\
    &=-\frac{1}{2} \sum_{j=1}^{J} \big(1+\log \frac{\sigma_{\tilde{\bm{\alpha}}^k}^2}{\sigma_{\bm{\alpha}^k}^2}-\frac{(\tilde{\alpha}^{k, j}-0)^2+\sigma_{\tilde{\bm{\alpha}}^k}^2}{\sigma_{\bm{\alpha}^k}^2}\big)\\
    &=-\frac{1}{2} \sum_{j=1}^{J}(1+\log \sigma_{\tilde{\bm{\alpha}}^k}^2)-\frac{1}{2}\log 2\pi - \log p(\tilde{\bm{\alpha}}^k),
\end{aligned}
\end{equation}
where $\tilde{\alpha}^{k, j}$ denote the $j$-th element of $\tilde{\bm{\alpha}}^k$.

Fifth, we substitute the result of Eq.~\eqref{eq_elbo_3} and Eq.~\eqref{eq_elbo_4} into Eq.~\eqref{eq_elbo_2}, respectively. Then, we can have
\begin{equation}\label{eq_elbo_5}
    \begin{aligned}
    &\quad \mathbb{E}_{q(\tilde{\mathbf{H}}_{1:M}^1, \tilde{\mathbf{H}}_{1:M}^2, \cdots, \tilde{\mathbf{H}}_{1:M}^K, \bm{\alpha}^{1:K}|\mathcal{G}_{1:M})}\big[\log p(\mathcal{G}_{1:M}, \tilde{\mathbf{H}}_{1:M}^1, \tilde{\mathbf{H}}_{1:M}^2, \cdots, \tilde{\mathbf{H}}_{1:M}^K, \bm{\alpha}^{1:K}) \\
&\quad - \log q(\tilde{\mathbf{H}}_{1:M}^1, \tilde{\mathbf{H}}_{1:M}^2, \cdots, \tilde{\mathbf{H}}_{1:M}^K, \bm{\alpha}^{1:K}|\mathcal{G}_{1:M})\big]\\
    &=\sum_{m=1}^{M} \mathbb{E}_{q(\tilde{\mathbf{H}}_m^1, \tilde{\mathbf{H}}_m^2, \cdots, \tilde{\mathbf{H}}_m^K|\mathcal{G}_m)} \big[\log p(\mathcal{G}_m|\tilde{\mathbf{H}}_m^1, \tilde{\mathbf{H}}_m^2, \cdots, \tilde{\mathbf{H}}_m^K)\big]\\
    &\quad - \sum_{m=1}^{M} \sum_{k=1}^{K} \mathbb{E}_{q(\bm{\alpha}^k)}\big[D_{\mathrm{KL}}\big(q(\tilde{\mathbf{H}}_m^k|\mathcal{G}_m) || p(\tilde{\mathbf{H}}_m^k | \bm{\alpha}^k)\big)\big]\\
\end{aligned}
\end{equation}

\begin{equation}\label{eq_elbo_5_2}
    \begin{aligned}
    &\hspace{-7em} -\sum_{k=1}^{K}D_{\mathrm{KL}}\big(q(\bm{\alpha}^k) || p(\bm{\alpha}^k)\big)\\
    &\hspace{-7em}=\sum_{m=1}^{M} \mathbb{E}_{q(\tilde{\mathbf{H}}_m^1, \tilde{\mathbf{H}}_m^2, \cdots, \tilde{\mathbf{H}}_m^K|\mathcal{G}_m)} \big[\log p(\mathcal{G}_m|\tilde{\mathbf{H}}_m^1, \tilde{\mathbf{H}}_m^2, \cdots, \tilde{\mathbf{H}}_m^K)\big]\\
    &\hspace{-7em} -\sum_{m=1}^{M} \sum_{k=1}^{K}D_{\mathrm{KL}}\big(q(\tilde{\mathbf{H}}_m^k|\mathcal{G}_m) || p(\tilde{\mathbf{H}}_m^k | \tilde{\bm{\alpha}}^k)\big) - \sum_{m=1}^{M} \sum_{k=1}^{K}\frac{J}{2} \frac{\sigma_{\tilde{\bm{\alpha}}^k}^2}{\sigma_{\tilde{\mathbf{H}}_{m}^k}^2}\\
    &\hspace{-7em} + \sum_{k=1}^{K} \big[\frac{1}{2} \sum_{j=1}^{J}(1+\log \sigma_{\tilde{\bm{\alpha}}^k}^2)+\frac{1}{2}\log 2\pi + \log p(\tilde{\bm{\alpha}}^k)\big]\\
    &\hspace{-7em}=\sum_{m=1}^{M} \Big\{\mathbb{E}_{q(\tilde{\mathbf{H}}_m^1, \tilde{\mathbf{H}}_m^2, \cdots, \tilde{\mathbf{H}}_m^K|\mathcal{G}_m)} \big[\log p(\mathcal{G}_m|\tilde{\mathbf{H}}_m^1, \tilde{\mathbf{H}}_m^2, \cdots, \tilde{\mathbf{H}}_m^K)\big]\\
    &\hspace{-7em} -\sum_{k=1}^{K}D_{\mathrm{KL}}\big(q(\tilde{\mathbf{H}}_m^k|\mathcal{G}_m) || p(\tilde{\mathbf{H}}_m^k | \tilde{\bm{\alpha}}^k)\big)\\
    &\hspace{-7em} - \underbrace{\sum_{k=1}^{K}\frac{J}{2} \frac{\sigma_{\tilde{\bm{\alpha}}^k}^2}{\sigma_{\tilde{\mathbf{H}}_{m}^k}^2}}_{\mathrm{constant}}\Big\}\\
    &\hspace{-7em} + \sum_{k=1}^{K}\big[\log p(\tilde{\bm{\alpha}}^k) + \underbrace{\frac{1}{2} \sum_{j=1}^{J}(1+\log \sigma_{\tilde{\bm{\alpha}}^k}^2)+\frac{1}{2}\log 2\pi}_{\mathrm{constant}}\big].
\end{aligned}
\end{equation}

Finally, since two constants in the above Eq.~\eqref{eq_elbo_5} can be omitted, we can have
\begin{equation}\label{eq_elbo_5_B}
    \begin{aligned}
    &\quad \mathbb{E}_{q(\tilde{\mathbf{H}}_{1:M}^1, \tilde{\mathbf{H}}_{1:M}^2, \cdots, \tilde{\mathbf{H}}_{1:M}^K, \bm{\alpha}^{1:K}|\mathcal{G}_{1:M})}\big[\log p(\mathcal{G}_{1:M}, \tilde{\mathbf{H}}_{1:M}^1, \tilde{\mathbf{H}}_{1:M}^2, \cdots, \tilde{\mathbf{H}}_{1:M}^K, \bm{\alpha}^{1:K}) \\
    &\quad - \log q(\tilde{\mathbf{H}}_{1:M}^1, \tilde{\mathbf{H}}_{1:M}^2, \cdots, \tilde{\mathbf{H}}_{1:M}^K, \bm{\alpha}^{1:K}|\mathcal{G}_{1:M})\big]\\
    &\approx \sum_{m=1}^{M} \Big\{\mathbb{E}_{q(\tilde{\mathbf{H}}_m^1, \tilde{\mathbf{H}}_m^2, \cdots, \tilde{\mathbf{H}}_m^K|\mathcal{G}_m)} \big[\log p(\mathcal{G}_m|\tilde{\mathbf{H}}_m^1, \tilde{\mathbf{H}}_m^2, \cdots, \tilde{\mathbf{H}}_m^K)\big]\\
    &\quad -\sum_{k=1}^{K}D_{\mathrm{KL}}\big(q(\tilde{\mathbf{H}}_m^k|\mathcal{G}_m) || p(\tilde{\mathbf{H}}_m^k | \tilde{\bm{\alpha}}^k)\big)\Big\}\\
    &\quad + \sum_{k=1}^{K} \log p(\tilde{\bm{\alpha}}^k)\\
    &= \sum_{m=1}^{M} \Big\{\mathbb{E}_{q(\tilde{\mathbf{H}}_m|\mathcal{G}_m)} \big[\log p(\mathcal{G}_m|\tilde{\mathbf{H}}_m)\big]\\
    &\quad -\sum_{k=1}^{K}D_{\mathrm{KL}}\big(q(\tilde{\mathbf{H}}_m^k|\mathcal{G}_m) || p(\tilde{\mathbf{H}}_m^k | \tilde{\bm{\alpha}}^k)\big)\Big\}\\
    &\quad + \sum_{k=1}^{K} \log p(\tilde{\bm{\alpha}}^k).
\end{aligned}
\end{equation}

\section{D. Derivation of $\tilde{\bm{\alpha}}^k$}
\label{derivation_alpha}
Since estimating the exact Maximum A Posterior (MAP) of $\bm{\alpha}^k$ is intractable, inspired by~\cite{Hsu_Zhang_Glass_2017}, we approximate $\bm{\alpha}^k$ with the help of ELBO as follows:
\begin{equation}\label{eq_elbo_6}
    \begin{aligned}
    \tilde{\bm{\alpha}}^k &= \operatorname*{argmax}_{\bm{\alpha}^k} \log p(\bm{\alpha}^k|\mathcal{G}_{1:M})\\
    &= \operatorname*{argmax}_{\bm{\alpha}^k} \log \frac{p(\mathcal{G}_{1:M}, \bm{\alpha}^k)}{p(\mathcal{G}_{1:M})}\\
    &= \operatorname*{argmax}_{\bm{\alpha}^k} \log p(\mathcal{G}_{1:M}, \bm{\alpha}^k)\\
    &= \operatorname*{argmax}_{\bm{\alpha}^k}\big(\sum_{m=1}^{M}\log p(\mathcal{G}_m|\bm{\alpha}^k)\big)+\log p(\bm{\alpha}^k)\\
    &\approx \operatorname*{argmax}_{\bm{\alpha}^k} \sum_{m=1}^{M} \Big\{\mathbb{E}_{q(\tilde{\mathbf{H}}_m|\mathcal{G}_m)} \big[\log p(\mathcal{G}_m|\tilde{\mathbf{H}}_m)\big] -\sum_{k=1}^{K}D_{\mathrm{KL}}\big(q(\tilde{\mathbf{H}}_m^k|\mathcal{G}_m) || p(\tilde{\mathbf{H}}_m^k | \tilde{\bm{\alpha}}^k)\big)\Big\} + \log p(\bm{\alpha}^k)\\
    & = \operatorname*{argmax}_{\bm{\alpha}^k} \sum_{m=1}^{M} \sum_{k=1}^{K}-D_{\mathrm{KL}}\big(q(\tilde{\mathbf{H}}_m^k|\mathcal{G}_m) || p(\tilde{\mathbf{H}}_m^k | \bm{\alpha}^k)\big) + \log p(\bm{\alpha}^k)\\
    &= \operatorname*{argmax}_{\bm{\alpha}^k} \sum_{m=1}^{M} \sum_{k=1}^{K}\sum_{j=1}^{J} \frac{(\hat{\mu}_{\tilde{\mathbf{H}}_{m}^k,j}-\alpha^{k,j})^2}{\sigma_{\tilde{\mathbf{H}}_{m}^k}^2} - \sum_{j=1}^{J}\frac{(\alpha^{k,j}-0)^2}{\sigma_{\bm{\alpha}^k}^2}\\
    &= \operatorname*{argmax}_{\bm{\alpha}^k} f(\bm{\alpha}^k),
\end{aligned}
\end{equation}
where $f(\bm{\alpha}^k)$ is a concave quadratic function with only one maximum point, and $\alpha^{k, j}$ denotes the $j$-th element of $\bm{\alpha}^k$. The closed-form solution of $\tilde{\bm{\alpha}}^k$ can be derived by differentiating Eq.~\eqref{eq_elbo_6} with respect to $\bm{\alpha}^k$. Specifically, we let
\begin{equation}\label{eq_elbo_8}
\left. \frac{\partial f(\bm{\alpha}^k)}{\partial \bm{\alpha}^k} \right|_{\bm{\alpha}^k=\tilde{\bm{\alpha}}^k} = 0,
\end{equation}
and then we can have
\begin{equation}\label{eq_elbo_9}
    \tilde{\bm{\alpha}}^k = \frac{\sum_{m=1}^{M}\hat{\bm{\mu}}_{\tilde{\mathbf{H}}_{m}^k}}{M+\frac{\sigma_{\tilde{\mathbf{H}}_{m}^k}^2}{\sigma_{\bm{\alpha}^k}^2}}.
\end{equation}

\section{E. Details of HVGAE}
\label{detail_hvage}
In this section, we introduce the detailed neural network architectures for our proposed HVGAE. Since HVGAE is deployed in each client, we take the $m$-th client as an example. Recall that $\mathcal{G}_m$ is the subgraph on the $m$-th client, which contains the node feature matrix $\mathbf{X}_m$ and adjacency matrix $\mathbf{A}_m$. We use two DisenGCNs (\textit{i.e.}, $\mathrm{DisenGCN}_{\bm{\mu}_m}(\mathcal{G}_m)$ and $\mathrm{DisenGCN}_{\bm{\sigma}_m}(\mathcal{G}_m)$) as the encoder and an inner product as the decoder of HVGAE, respectively. Note that $\mathrm{DisenGCN}_{\bm{\mu}_m}(\mathcal{G}_m)$ and $\mathrm{DisenGCN}_{\bm{\sigma}_m}(\mathcal{G}_m)$ are used to infer the means and standard deviations of $\mathcal{G}_m$ for $K$ latent factors, respectively. Moreover, $\mathrm{DisenGCN}_{\bm{\mu}}(\mathcal{G}_m)$ and $\mathrm{DisenGCN}_{\bm{\sigma}}(\mathcal{G}_m)$ share the same node feature projection layer.

\begin{figure}[t]
	\centering
	\includegraphics[width=11cm]{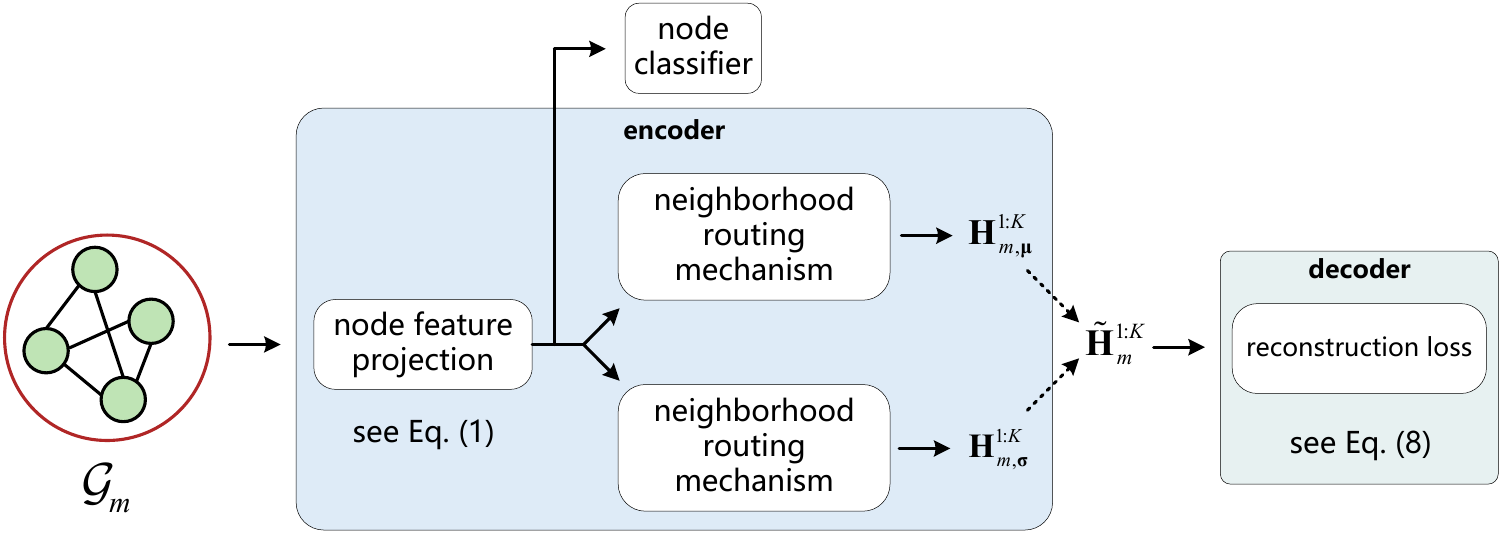}
	\caption{The architecture of our proposed HVGAE. The dashed lines show the process of sampling using the reparameterization trick~\cite{kingma2013auto}.}
	\label{fig_A_1}
\end{figure}

Fig.~\ref{fig_A_1} shows the architecture of our proposed HVGAE. First, we project the node feature to $K$ subspaces according to Eq.~\eqref{eq1}. Meanwhile, the node representations after the node feature projection layer are used to train a local node classifier. Second, we use the neighborhood routing mechanism to obtain $\mathbf{H}_{m, \bm{\mu}}^{1:K}$ and $\mathbf{H}_{m, \bm{\sigma}}^{1:K}$, respectively. Third, we use the reparameterization trick~\cite{kingma2013auto} to sample $\tilde{\mathbf{H}}_{m}^{1:K}$ from $\mathbf{H}_{m, \bm{\mu}}^{1:K}$ and $\mathbf{H}_{m, \bm{\sigma}}^{1:K}$ (see Eq.~\eqref{eq9}). Fourth, HVGAE decodes from $\tilde{\mathbf{H}}_{m}^{1:K}$ and then computes $\mathbb{E}_{q(\tilde{\mathbf{H}}_m|\mathcal{G}_m)}\big[\log p(\mathcal{G}_m|\tilde{\mathbf{H}}_m)\big]$ (see Eq.~\eqref{eq10}). The implementations of two prior distributions $p(\tilde{\bm{\alpha}}^k)$ and $p(\tilde{\mathbf{H}}_m^k | \tilde{\bm{\alpha}}^k)$ are shown in the Appendix I.6.

\section{F. Additional Related Work}
\label{additional_related_work}
Here we review the typical works related to the FL via Bayesian methods. Recently, some methods~\cite{shedivat2021federated, pmlr-v162-zhang22o, kim2023fedhb} have tried to model the FL problem by using the Bayesian theories. Specifically, they take the network weights as a whole entity and treat them as a single random variable shared by all clients. For example, FedPA~\cite{shedivat2021federated} infers the global posterior by averaging the local posteriors. However, FedPA targets a general FL setting, so it does not apply to the personalized FL, and its performance may be degraded. Different from FedPA, pFedBayes~\cite{pmlr-v162-zhang22o} assigns a personalized Bayesian neural network to each client. It infers the global posterior from individual posteriors under a regularizer based on the KL divergence. However, it cannot model the posterior of each client from a global perspective. To address this deficiency, Kim~\textit{et. al} propose a hierarchical Bayesian approach called FedHB~\cite{kim2023fedhb}, which mitigates the heterogeneity problem in a personalized way. Specifically, it introduces two types of latent random variables, one used as the network weights for each client's backbone, and the other used as a globally shared random variable to be associated with each client. Unlike FedHB, we model the local subgraph data rather than the network weight, so that we can infer the subgraph data distribution on each client.

\section{G. Analysis of Federated Parameters}
\label{analysis_parameter}
According to the architecture of our proposed HVGAE, there are several components: a node feature projection layer, two neighborhood routing mechanism layers, a node classifier, and a reconstruction layer. First, recall that there are no learnable parameters in the neighborhood routing mechanism. Second, the reconstruction layer is constructed by an inner product, so there are no learnable parameters. Third, although we use two DisenGCNs (\textit{i.e.}, $\mathrm{DisenGCN}_{\bm{\mu}_m}(\mathcal{G}_m)$ and $\mathrm{DisenGCN}_{\bm{\sigma}_m}(\mathcal{G}_m)$) as the encoder of our proposed HVGAE, they share the same node feature projection layer (see Appendix~\ref{detail_hvage}). Consequently, according to the node feature projection layer defined by Eq.~\eqref{eq1}, we can find that $\mathbf{W}_m^1, \mathbf{W}_m^2, \cdots, \mathbf{W}_m^K$ and $\mathbf{b}_m^1, \mathbf{b}_m^2, \cdots, \mathbf{b}_m^K$ are learnable parameters that should be federated. Last but not least, since we introduce a node classifier, which is actually a Multi-Layer Perceptron (MLP), we can find that $\mathbf{W}_m^{\mathrm{cls}} \in \mathbb{R}^{c \times d_\mathrm{out}}$ and $\mathbf{b}_m^{\mathrm{cls}} \in \mathbb{R}^{c}$ are learnable parameters that should be federated.

In Eq.~\eqref{eq8}, although we have defined the learnable parameter $\hat{\bm{\alpha}}^k_m$ on each client, $\hat{\bm{\alpha}}^k_m$ is only used to approximate the data distribution instead of a parameter in a neural network. Consequently, $\hat{\bm{\alpha}}^k_m$ does not participate in the federation.

\section{H. Pseudocode of FedIIH}
\label{pseudocode}
\begin{figure}[t]
	\centering
	\includegraphics[width=9cm]{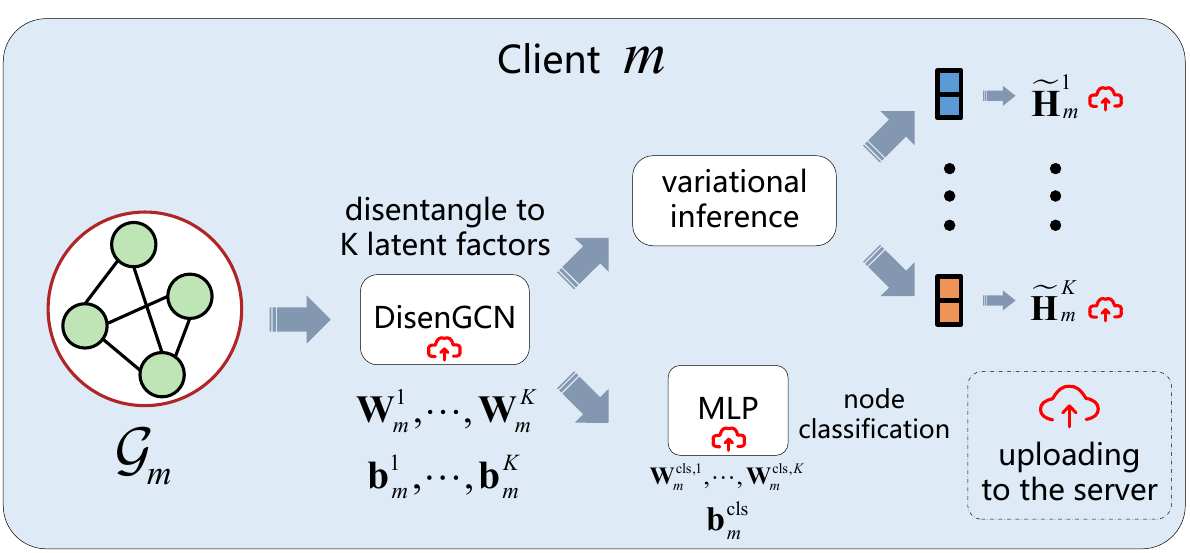}
	\caption{The framework of our proposed FedIIH from the perspective of the clients. Let us take the client $m$ as an example.}
	\label{fig_A_2_client}
\end{figure}

\begin{figure}[t]
	\centering
	\includegraphics[width=14cm]{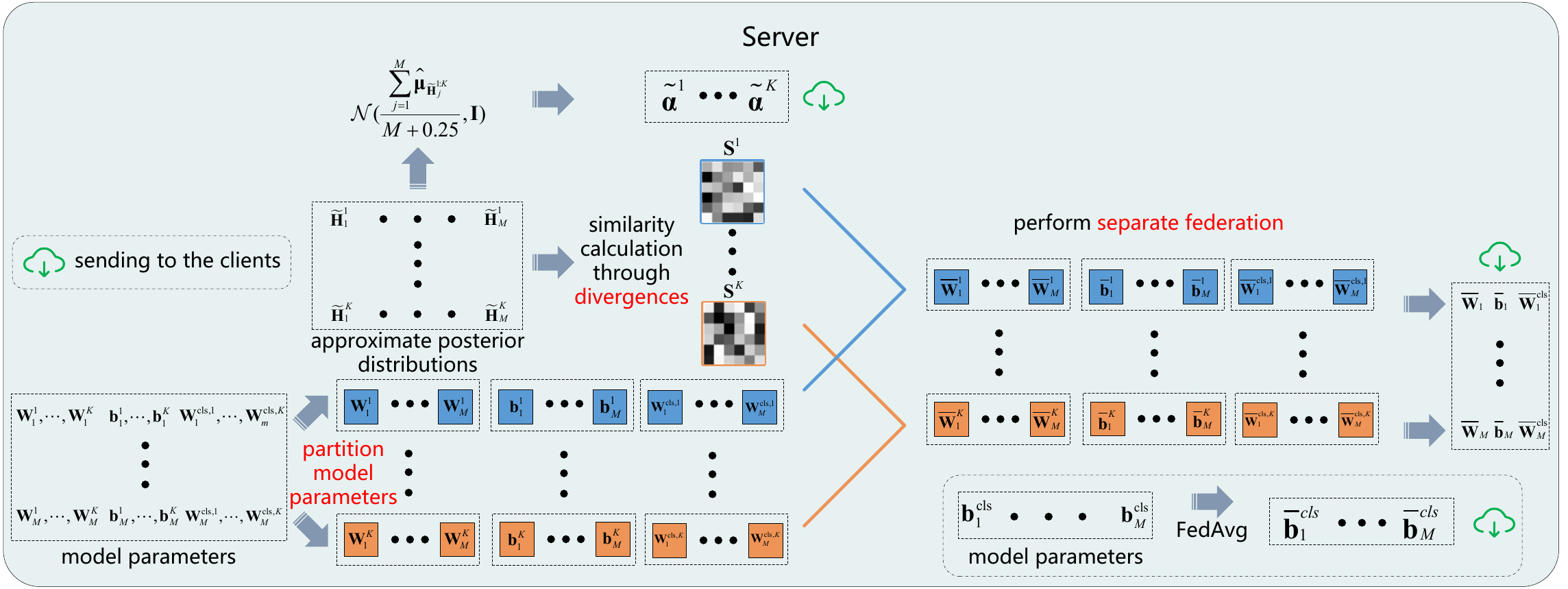}
	\caption{The framework of our proposed FedIIH from the perspective of the server, where $\mathbf{S}^k$ denotes the similarity matrix with respect to the $k$-th latent factor. Here we use blue and orange to represent the first and the $K$-th latent factors, respectively.}
	\label{fig_A_2_server}
\end{figure}

In this section, we show the pseudocode of our proposed FedIIH for the clients and server in Algorithm~\ref{algorithm1} and Algorithm~\ref{algorithm2}, respectively. Moreover, the framework of our proposed FedIIH from the perspective of the clients and the server is shown in Fig.~\ref{fig_A_2_client} and Fig.~\ref{fig_A_2_server}, respectively. Inspired by~\cite{liu2023bayesian}, we regard the posterior mean of $\bm{\alpha}^k$ for the $k$-th global latent factor (\textit{i.e.}, $\tilde{\bm{\alpha}}^k$) in the last round as the prior for the $k$-th local latent factor on each client in the current round.

\begin{algorithm}[tb]
	\caption{\textbf{FedIIH} Client Algorithm}
	\label{algorithm1}
	\textbf{Input}: Number of local epochs $E$; number of latent factors $K$; subgraph $\mathcal{G}_m$ on client $m$; node feature matrix $\mathbf{X}_m$ on client $m$; label matrix $\mathbf{Y}_m$ on client $m$; parameters of two DisenGCNs $\mathbf{W}_m^{1:K}$ and $\mathbf{b}_{m}^{1:K}$ on client $m$; parameters of the node classifier $\mathbf{W}_m^{\mathrm{cls}}$ and $\mathbf{b}_m^{\mathrm{cls}}$ on client $m$; federated parameters $\overline{\mathbf{W}}_m^{\mathrm{cls}}$, $\overline{\mathbf{b}}_m^{\mathrm{cls}}$, $\overline{\mathbf{W}}_{m}^{1:K}$, and $\overline{\mathbf{b}}_{m}^{1:K}$ from the server; global latent factors $\tilde{\bm{\alpha}}^{1:K}$ from the server.\\
	\textbf{Output}: Predicted label for each unlabeled graph node in subgraph $\mathcal{G}_m$.\\
	\begin{algorithmic}[1] 
    \STATE Download federated parameters $\overline{\mathbf{W}}_m^{\mathrm{cls}}$, $\overline{\mathbf{b}}_m^{\mathrm{cls}}$, $\overline{\mathbf{W}}_{m}^{1:K}$, and $\overline{\mathbf{b}}_{m}^{1:K}$ from the server;

    \STATE Download the global latent factors $\tilde{\bm{\alpha}}^{1:K}$ from the server;

    \STATE $\mathbf{W}_m^{\mathrm{cls}}\leftarrow \overline{\mathbf{W}}_m^{\mathrm{cls}}$, $\mathbf{b}_m^{\mathrm{cls}}\leftarrow \overline{\mathbf{b}}_m^{\mathrm{cls}}$, $\mathbf{W}_{m}^{1:K}\leftarrow \overline{\mathbf{W}}_{m}^{1:K}$, $\mathbf{b}_{m}^{1:K}\leftarrow \overline{\mathbf{b}}_{m}^{1:K}$;

	\FOR{\textnormal{each local epoch} $e$ \textnormal{from 1 to} $E$}
        \STATE Project the node feature into $K$ subspaces via Eq.~\eqref{eq1};

        \STATE Optimize $\mathbf{W}_m^{\mathrm{cls}}$ and $\mathbf{b}_m^{\mathrm{cls}}$ via the cross-entropy loss;
 
		\STATE Disentangle projected node representations into $K$ latent factors via the neighborhood routing mechanism, and then obtain $\mathbf{H}_{m, \bm{\mu}}^{1:K}$ and $\mathbf{H}_{m, \bm{\sigma}}^{1:K}$, respectively;

        \STATE Sample $\tilde{\mathbf{H}}_{m}^{1:K}$ from $\mathbf{H}_{m, \bm{\mu}}^{1:K}$ and $\mathbf{H}_{m, \bm{\sigma}}^{1:K}$ via Eq.~\eqref{eq9} so as to obtain $q(\tilde{\mathbf{H}}_m^{1:K}|\mathcal{G}_m)$;
		
        \STATE Compute $\mathbb{E}_{q(\tilde{\mathbf{H}}_m|\mathcal{G}_m)}\big[\log p(\mathcal{G}_m|\tilde{\mathbf{H}}_m)\big]$ via Eq.~\eqref{eq10};

        \STATE Approximate $\tilde{\bm{\alpha}}^{1:K}$ by using $\hat{\bm{\alpha}}^{1:K}_m$ so as to compute $\sum_{k=1}^{K} \Big\{\log p(\hat{\bm{\alpha}}^k_m) - D_{\mathrm{KL}}\big(p(\hat{\bm{\alpha}}^k_m) || p(\tilde{\bm{\alpha}}^k)\big)\Big\}$;
        
        \STATE Compute $\sum_{k=1}^{K} D_{\mathrm{KL}}\big(q(\tilde{\mathbf{H}}_m^k|\mathcal{G}_m) || p(\tilde{\mathbf{H}}_m^k | \tilde{\bm{\alpha}}^k)\big)$, where $p(\tilde{\mathbf{H}}_m^{1:K} | \tilde{\bm{\alpha}}^{1:K}) \sim \mathcal{N}(\tilde{\bm{\alpha}}^{1:K}, \sigma_{\tilde{\mathbf{H}}_{m}^{1:K}}^2\mathbf{I})$;

        \STATE Optimize $\mathbf{W}_{m}^{1:K}$ and $\mathbf{b}_{m}^{1:K}$ via Eq.~\eqref{eq8};
    \ENDFOR
	
	\STATE Upload $\mathbf{W}_m^{\mathrm{cls}}$, $\mathbf{b}_m^{\mathrm{cls}}$, $\mathbf{W}_{m}^{1:K}$, $\mathbf{b}_{m}^{1:K}$, and $\tilde{\mathbf{H}}_{m}^{1:K}$ to the server;
	
	\STATE Predict labels based on the trained node classifier.
\end{algorithmic}
\end{algorithm}

\begin{algorithm}[tb]
	\caption{\textbf{FedIIH} Server Algorithm}
	\label{algorithm2}
	\textbf{Input}: Number of rounds $R$; number of clients $M$; number of latent factors $K$; parameters of two DisenGCNs $\mathbf{W}_{m}^{1:K}$ and $\mathbf{b}_{m}^{1:K}$ from client $m$; parameters of the node classifier $\mathbf{W}_m^{\mathrm{cls}}$ and $\mathbf{b}_m^{\mathrm{cls}}$ from client $m$; approximate posterior distributions $\tilde{\mathbf{H}}_{m}^{1:K}$ from client $m$.\\
	\textbf{Output}: Federated model parameters for client $m$.\\
	\begin{algorithmic}[1] 
    \STATE Initialize parameters $({\overline{\mathbf{W}}^{\mathrm{cls}}})^{(1)}$, $({\overline{\mathbf{b}}^{\mathrm{cls}}})^{(1)}$, $({\overline{\mathbf{W}}^{1:K}})^{(1)}$, and $({\overline{\mathbf{b}}^{1:K}})^{(1)}$;

    \STATE Initialize $(\tilde{\bm{\alpha}}^{1:K})^{(1)}$;

	\FOR{\textnormal{each round} $r$ \textnormal{from 1 to} $R$}
        \FOR{\textnormal{client} $m \in \{1, 2, \cdots, M\}$ \textnormal{\textbf{in parallel}}}
            \IF {$r=1$}
                \STATE Send $({\overline{\mathbf{W}}^{\mathrm{cls}}})^{(r)}$, $({\overline{\mathbf{b}}^{\mathrm{cls}}})^{(r)}$, $({\overline{\mathbf{W}}^{1:K}})^{(r)}$, and $({\overline{\mathbf{b}}^{1:K}})^{(r)}$ to client $m$;

                \STATE Send $(\tilde{\bm{\alpha}}^{1:K})^{(r)}$ to client $m$;
            \ELSE
                \STATE Receive $\mathbf{W}_m^{\mathrm{cls}}$, $\mathbf{b}_m^{\mathrm{cls}}$, $\mathbf{W}_{m}^{1:K}$, $\mathbf{b}_{m}^{1:K}$, and $\tilde{\mathbf{H}}_{m}^{1:K}$ from client $m$;
            
                \STATE Compute $S(m,j)^{1:K}$ via Eq.~\eqref{eq12}, where $ j \in \{1, 2, \cdots, M\}$;

                \STATE Compute $\beta_{mj}^{1:K}$ via Eq.~\eqref{eq14}, where $ j \in \{1, 2, \cdots, M\}$;

                \STATE Perform the separate federation to obtain $({\overline{\mathbf{W}}^{\mathrm{cls}}_m})^{(r)}$, $({\overline{\mathbf{W}}_{m}^{1:K}})^{(r)}$, and $({\overline{\mathbf{b}}_{m}^{1:K}})^{(r)}$ via Eq.~\eqref{eq13};
                
                \STATE Perform the FedAvg~\cite{mcmahan2017communication} to obtain $({\overline{\mathbf{b}}^{\mathrm{cls}}_m})^{(r)}$;

                \STATE Sample $(\tilde{\bm{\alpha}}^{1:K})^{(r)}$ from $\mathcal{N}(\frac{\sum_{j=1}^{M}\hat{\bm{\mu}}_{\tilde{\mathbf{H}}_{j}^{1:K}}}{M+0.25}, \mathbf{I})$;

                \STATE Send $({\overline{\mathbf{W}}^{\mathrm{cls}}_m})^{(r)}$, $({\overline{\mathbf{b}}^{\mathrm{cls}}_m})^{(r)}$, $({\overline{\mathbf{W}}_{m}^{1:K}})^{(r)}$, and $({\overline{\mathbf{b}}_{m}^{1:K}})^{(r)}$ to client $m$;

                \STATE Send $(\tilde{\bm{\alpha}}^{1:K})^{(r)}$ to client $m$;

            \ENDIF
        	\STATE Perform Algorithm~\ref{algorithm1} on client $m$;
        \ENDFOR

    \ENDFOR
\end{algorithmic}
\end{algorithm}

\section{I. Implementation Details}
\label{implementation_details}
In this section, we provide the implementation details, including the experimental platform, the dataset descriptions, the details of subgraph partitioning, the information on baseline methods, training details, and the implementations of two prior distributions (\textit{i.e.}, $p(\tilde{\bm{\alpha}}^k)$ and $p(\tilde{\mathbf{H}}_m^k | \tilde{\bm{\alpha}}^k)$).

\subsection{I.1 Experimental Platform}
\label{experimental_platform}
All the experiments in this work are conducted on a Linux server with a 2.90 GHz Intel Xeon Gold 6326 CPU, 64 GB of RAM, and two NVIDIA GeForce RTX 4090 GPUs with 48GB of memory. Our proposed method is implemented via Python 3.8.8, PyTorch 1.12.0, and PyTorch Geometric (PyG) 2.3.0 with the Adam optimizer.

\subsection{I.2 Datasets}
\label{dataset_info}
To validate the effectiveness of our proposed FedIIH, we perform extensive experiments on eleven widely used benchmark datasets, including six homophilic and five heterophilic graph datasets. In the homophilic graph datasets, there are \textit{Cora}, \textit{CiteSeer}, \textit{PubMed}, and \textit{ogbn-arxiv} for the citation graphs; \textit{Amazon-Computer} and \textit{Amazon-Photo} for Amazon product graphs. In the heterophilic graph datasets~\cite{platonov2023a}, there are \textit{Roman-empire}, \textit{Amazon-ratings}, \textit{Minesweeper}, \textit{Tolokers}, and \textit{Questions}. The statistical information of the above benchmark datasets is described in Tab.~\ref{datasets_statistics}. Note that we use the Area Under the ROC curve (AUC) as the evaluation metric (higher values are better) for the \textit{Minesweeper}, \textit{Tolokers}, and \textit{Questions} datasets, and use the accuracy as the evaluation metric (higher values are better) for other datasets.

\begin{table}[t]
    \centering
    \caption{Statistical information of eleven used graph datasets.}
    \label{datasets_statistics}
    \begin{tabular}{cccccc}
    \hline
    Types                               & Datasets                      & \# Nodes & \# Edges  & \# Classes & \# Node Features \\ \hline
    \multirow{6}{*}{homophilic graph}   & \textit{Cora}                 & 2,708    & 5,429     & 7          & 1,433            \\
                                        & \textit{CiteSeer}             & 3,327    & 4,732     & 6          & 3,703            \\
                                        & \textit{PubMed}               & 19,717   & 44,324    & 3          & 500              \\
                                        & \textit{Amazon-Computer}      & 13,752   & 491,722   & 10         & 767              \\
                                        & \textit{Amazon-Photo}         & 7,650    & 238,162   & 8          & 745              \\
                                        & \textit{ogbn-arxiv}           & 169,343  & 2,315,598 & 40         & 128              \\ \hline
    \multirow{6}{*}{heterophilic graph} & \textit{Roman-empire}         & 22,662   & 32,927    & 18         & 300              \\
                                        & \textit{Amazon-ratings}       & 24,492   & 93,050    & 5          & 300              \\
                                        & \textit{Minesweeper}          & 10,000   & 39,402    & 2          & 7                \\
                                        & \textit{Tolokers}             & 11,758   & 519,000   & 2          & 10               \\
                                        & \textit{Questions}            & 48,921   & 153,540   & 2          & 301              \\ \hline
    \end{tabular}
    \end{table}

In order to facilitate the division of datasets, a random sample of 20\% of nodes is selected for training purposes, 40\% for the purpose of validation, and 40\% for testing, with the exception of the \textit{ogbn-arxiv} dataset. This is due to the fact that the \textit{ogbn-arxiv} dataset comprises a relatively large number of nodes in comparison to the other datasets, as reported in Tab.~\ref{datasets_statistics}. Consequently, for the \textit{ogbn-arxiv} dataset, a random sample of 5\% of the nodes is used for training, while the remaining half of the nodes are used for validation and testing, respectively.

\subsection{I.3 Subgraph Partitioning}
\label{subgraph_partitioning_detail}
Inspired by real-world requirements and following~\cite{baek2023personalized}, we consider two subgraph partitioning settings: non-overlapping and overlapping. In the non-overlapping setting, 
$\cup_{m=1}^{M} \mathcal{V}_m=\mathcal{V}$ and $\mathcal{V}_m \cap \mathcal{V}_n=\emptyset $ for $\forall m \neq n \in \{1, 2, \cdots, M\}$, where $\mathcal{V}$ represents the node set of the global graph. Partitioning without this property is called overlapping. Here we present the details of how to partition the original graph into multiple subgraphs. It should be noted that the number of subgraphs is equal to the number of clients. Both non-overlapping and overlapping subgraph partitioning settings are used in the experiments for all datasets.

\textbf{Non-overlapping partitioning} First, if there are $M$ clients, the number of non-overlapping subgraphs to be generated is specified as $M$. Second, the METIS graph partitioning algorithm, as described in~\cite{karypis1997metis}, is used to divide the original graph into $M$ subgraphs. In other words, the non-overlapping partitioning subgraph for each client is directly obtained by the output of the METIS algorithm.

\textbf{Overlapping partitioning} First, if there are $M$ clients, the number of overlapping subgraphs to be generated is specified as $M$. Second, the METIS~\cite{karypis1997metis} graph partitioning algorithm is used to divide the original graph into $\lfloor \frac{M}{5} \rfloor$ subgraphs. Third, in each subgraph generated by METIS, half of the nodes and their associated edges are randomly sampled. This procedure is performed five times to generate five different yet overlapped subgraphs. By doing so, the number of overlapping subgraphs is equal to the number of clients.

\subsection{I.4 Baseline Methods}
\label{baseline_methods_info}
We compare our proposed FedIIH with the following baseline methods, which can be categorized into two groups. The first group comprises general FL baseline methods, including FedAvg~\cite{mcmahan2017communication}, FedProx~\cite{MLSYS2020_1f5fe839}, and FedPer~\cite{Arivazhagan2019}. The second group consists of six GFL baseline methods, namely GCFL~\cite{NEURIPS2021_9c6947bd}, FedGNN~\cite{wu2021fedgnn}, FedSage+~\cite{NEURIPS2021_34adeb8e}, FED-PUB~\cite{baek2023personalized}, FedGTA~\cite{li2023fedgta}, and AdaFGL~\cite{li2024adafgl}. Moreover, we perform experiments with local training, that is, training each client without federated aggregation. The detailed descriptions of these baseline methods are provided below.

\textbf{FedAvg} This method~\cite{mcmahan2017communication} represents one of the fundamental baseline methods in the field of FL. First, each client independently trains a model, which is subsequently transmitted to a server. Then, the server aggregates the locally updated models by averaging and transmits the aggregated model back to the clients.

\textbf{FedProx} This method~\cite{MLSYS2020_1f5fe839} is one of the personalized FL baseline methods. It customizes a personalized model for each client by adding a proximal term as a subproblem that minimizes weight differences between local and global models.

\textbf{FedPer} This method~\cite{Arivazhagan2019} is one of the personalized FL baseline methods. It only federates the weights of the backbone while training the personalized classification layer in each client.

\textbf{GCFL} This method~\cite{NEURIPS2021_9c6947bd} is one of the basic GFL methods. Specifically, GCFL is designed for vertical GFL (\textit{e.g.}, GFL for molecular graphs). In particular, it uses the bi-partitioning scheme, which divides a set of clients into two disjoint groups of clients based on the similarity of their gradients. This is similar to the mechanism proposed for image classification in clustered-FL~\cite{9174890}. Then, after partitioning, the model weights are shared only among clustered clients with similar gradients.

\textbf{FedGNN} This method~\cite{wu2021fedgnn} is one of the GFL baseline methods. It extends local subgraphs by exchanging node embeddings from other clients. Specifically, if two nodes in two different clients have exactly the same neighbors, FedGNN transfers the nodes with the same neighbors from other clients and expands them.

\textbf{FedSage+} This method~\cite{NEURIPS2021_34adeb8e} is one of the GFL baseline methods. It generates the missing edges between subgraphs and the corresponding neighbor nodes by using the missing neighbor generator. To train this neighbor generator, each client first receives node representations from other clients, and then computes the gradient of the distances between the local node features and the node representations of the other clients. After that, the gradient is sent back to the other clients, and this gradient is then used to train the neighbor generator.

\textbf{FED-PUB} This method~\cite{baek2023personalized} is one of the personalized GFL baseline methods. It estimates the similarities between the subgraphs based on the outputs of the local models that are given the same test graph. Then, based on the similarities, it performs a weighted averaging of the local models for each client. In addition, it learns a personalized sparse mask at each client in order to select and update only the subgraph-relevant subset of the aggregated parameters.

\textbf{FedGTA} This method~\cite{li2023fedgta} is one of the personalized GFL baseline methods. In this method, each client first computes topology-aware local smoothing confidence and mixed moments of neighbor features. They are then used to compute the inter-subgraph similarities, which are uploaded to the server along with the model parameters. Finally, the server is able to perform a weighted federation for each client.

\textbf{AdaFGL} This method~\cite{li2024adafgl} is one of the personalized GFL baseline methods. Actually, it is a decoupled two-step personalized approach. First, it uses standard multi-client federated collaborative training to acquire the federated knowledge extractor by aggregating uploaded models in the final round at the server. Second, each client performs personalized training based on the local subgraph and the federated knowledge extractor.

\textbf{Local} This method is the non-FL baseline, where the model is trained only locally for each client, with no weight sharing.

\subsection{I.5 Training Details}
\label{training_details}
\textbf{Training rounds and epochs} For the \textit{Cora}, \textit{CiteSeer}, \textit{PubMed}, \textit{Roman-empire}, \textit{Amazon-ratings}, \textit{Minesweeper}, \textit{Tolokers}, and \textit{Questions} datasets, we set the number of local training epochs and total rounds to 1 and 100, respectively. For larger datasets, such as \textit{Amazon-Computer}, \textit{Amazon-Photo}, and \textit{ogbn-arxiv}, we set the number of total rounds to 200. Note that the number of local epochs is set to 2 for the \textit{Amazon-Photo} and \textit{ogbn-arxiv} datasets, and to 3 for the \textit{Amazon-Computer} dataset. Finally, we report the test performance of all models at the best validation epoch, and the performance is measured by averaging across all clients in terms of node classification accuracies (or AUCs).

\textbf{Hyperparameters}
We report the detailed hyperparameters used to train our proposed FedIIH in Tab.~\ref{hyperparameters_table} and Tab.~\ref{hyperparameters_table2}. These hyperparameters are determined by grid search. Note that we set the similarity scaling hyperparameter (\textit{i.e.}, $\tau$ in Eq.~\eqref{eq14}) to 10 in both non-overlapping and overlapping subgraph partitioning scenarios according to the recommendation of FED-PUB~\cite{baek2023personalized}. We provide the detailed reasons in the Appendix K.7. The search range of hyperparameters is shown in Tab.~\ref{hyperparameters_range}. The code of our proposed FedIIH will be released on GitHub upon acceptance of the paper. The detailed hyperparameter sensitivity analysis can be found in the Appendix M.2.

\textbf{Hyperparameter tuning guidelines}
There are four vital hyperparameters (\textit{i.e.}, number of latent factors, number of neighborhood routing layers, number of neighborhood routing iterations, and $\tau$) in our proposed FedIIH. First, the number of latent factors (\textit{i.e.}, $K$) is related to whether the graph tends to be homophilic or heterophilic. If the graph dataset tends to be homophilic, a small $K$ is recommended. If the graph dataset tends to be heterophilic, a large $K$ is recommended. Second, the variations in performance under different numbers of neighborhood routing layers, different numbers of neighborhood routing iterations, and different values of $\tau$ are all small. Therefore, these three hyperparameters can be tuned by grid search.

\textbf{Network architectures} For the experiments of all baseline methods, except FedSage+, FedGTA, and our proposed FedIIH, we use two layers of the Graph Convolutional Network (GCN)~\cite{kipf2017semisupervised} and a linear classifier layer as their network architectures. For the hyperparameter settings of the baseline methods, we use the default settings given in their original papers. Because of the inductive and scalability advantages of GraphSAGE~\cite{NIPS2017_5dd9db5e}, FedSage+ uses GraphSAGE as the encoder and then trains a missing neighbor generator to handle missing links across local subgraphs. For our proposed FedIIH, we use the node feature projection layer of DisenGCN~\cite{pmlrv97ma19a} to obtain the node representations (see Fig.~\ref{fig_A_1}) and a linear classifier layer (\textit{i.e.}, MLP) to perform node classifications. In contrast, FedGTA uses a Graph Attention Multi-Layer Perceptron (GAMLP)~\cite{35346783539121} as its backbone and a linear classifier layer to classify nodes. Note that GAMLP~\cite{35346783539121} is one of the scalable Graph Neural Network (GNN) models, which can capture the underlying correlations between different scales of graph knowledge.

\begin{table}[t]
    \centering
    \caption{Hyperparameters used in our proposed FedIIH on the \textbf{homophilic} graph datasets in both the non-overlapping and overlapping subgraph partitioning settings.}
    \scriptsize
    \label{hyperparameters_table}
    \begin{tabular}{c|cc|cccc}
    \hline
    \textbf{}                                                                    & \multicolumn{2}{c|}{Cora}                          & \multicolumn{2}{c|}{CiteSeer}                                            & \multicolumn{2}{c}{PubMed}                         \\ \hline
    Hyperparameters                                                              & \multicolumn{1}{c|}{non-overlapping} & overlapping & \multicolumn{1}{c|}{non-overlapping} & \multicolumn{1}{c|}{overlapping} & \multicolumn{1}{c|}{non-overlapping} & overlapping \\ \hline
    \# latent factors                                                            & \multicolumn{1}{c|}{2}               & 2           & \multicolumn{1}{c|}{2}               & \multicolumn{1}{c|}{2}           & \multicolumn{1}{c|}{4}               & 2           \\ \hline
    learning rate                                                                & \multicolumn{1}{c|}{0.02}            & 0.01        & \multicolumn{1}{c|}{0.01}            & \multicolumn{1}{c|}{0.01}        & \multicolumn{1}{c|}{0.01}            & 0.015       \\ \hline
    \# hidden dimensions                                                         & \multicolumn{1}{c|}{128}             & 256         & \multicolumn{1}{c|}{256}             & \multicolumn{1}{c|}{256}         & \multicolumn{1}{c|}{256}             & 256         \\ \hline
    dropout rate                                                                 & \multicolumn{1}{c|}{0.3}             & 0.35        & \multicolumn{1}{c|}{0.35}            & \multicolumn{1}{c|}{0.35}        & \multicolumn{1}{c|}{0.25}            & 0.4         \\ \hline
    weight decay                                                                 & \multicolumn{1}{c|}{0.005}           & 1e-6        & \multicolumn{1}{c|}{1e-6}            & \multicolumn{1}{c|}{1e-6}        & \multicolumn{1}{c|}{0.0045}          & 1e-6        \\ \hline
    \begin{tabular}[c]{@{}c@{}}\# neighborhood\\ routing layers\end{tabular}     & \multicolumn{1}{c|}{4}               & 5           & \multicolumn{1}{c|}{5}               & \multicolumn{1}{c|}{5}           & \multicolumn{1}{c|}{1}               & 1           \\ \hline
    \begin{tabular}[c]{@{}c@{}}\# neighborhood\\ routing iterations\end{tabular} & \multicolumn{1}{c|}{6}               & 6           & \multicolumn{1}{c|}{6}               & \multicolumn{1}{c|}{6}           & \multicolumn{1}{c|}{6}               & 6           \\ \hline
    \hline
                                                                                 & \multicolumn{2}{c|}{Amazon-Computer}               & \multicolumn{2}{c|}{Amazon-Photo}                                       & \multicolumn{2}{c}{ogbn-arxiv}                     \\ \hline
    Hyperparameters                                                              & \multicolumn{1}{c|}{non-overlapping} & overlapping & \multicolumn{1}{c|}{non-overlapping} & \multicolumn{1}{c|}{overlapping} & \multicolumn{1}{c|}{non-overlapping} & overlapping \\ \hline
    \# latent factors                                                            & \multicolumn{1}{c|}{6}               & 4           & \multicolumn{1}{c|}{6}               & \multicolumn{1}{c|}{10}          & \multicolumn{1}{c|}{2}               & 2           \\ \hline
    learning rate                                                                & \multicolumn{1}{c|}{0.015}           & 0.015       & \multicolumn{1}{c|}{0.015}           & \multicolumn{1}{c|}{0.01}        & \multicolumn{1}{c|}{0.01}            & 0.01        \\ \hline
    \# hidden dimensions                                                         & \multicolumn{1}{c|}{128}             & 128         & \multicolumn{1}{c|}{256}             & \multicolumn{1}{c|}{128}         & \multicolumn{1}{c|}{128}             & 128         \\ \hline
    dropout rate                                                                 & \multicolumn{1}{c|}{0.4}             & 0.35        & \multicolumn{1}{c|}{0.4}             & \multicolumn{1}{c|}{0.35}        & \multicolumn{1}{c|}{0.35}            & 0.35        \\ \hline
    weight decay                                                                 & \multicolumn{1}{c|}{1e-6}            & 1e-6        & \multicolumn{1}{c|}{1e-6}            & \multicolumn{1}{c|}{1e-6}        & \multicolumn{1}{c|}{1e-6}            & 1e-6        \\ \hline
    \begin{tabular}[c]{@{}c@{}}\# neighborhood\\ routing layers\end{tabular}     & \multicolumn{1}{c|}{1}               & 5           & \multicolumn{1}{c|}{1}               & \multicolumn{1}{c|}{1}           & \multicolumn{1}{c|}{5}               & 5           \\ \hline
    \begin{tabular}[c]{@{}c@{}}\# neighborhood\\ routing iterations\end{tabular} & \multicolumn{1}{c|}{6}               & 6           & \multicolumn{1}{c|}{6}               & \multicolumn{1}{c|}{5}           & \multicolumn{1}{c|}{6}               & 6           \\ \hline
    \end{tabular}
    \end{table}

    \begin{table}[]
        \centering
        \caption{Hyperparameters used in our proposed FedIIH on the \textbf{heterophilic} graph datasets in both the non-overlapping and overlapping subgraph partitioning settings.}
        \scriptsize
        \label{hyperparameters_table2}
        \begin{tabular}{c|cc|cccc}
        \hline
        \textbf{}                                                                    & \multicolumn{2}{c|}{Roman-empire}                  & \multicolumn{2}{c|}{Amazon-ratings}                                      & \multicolumn{2}{c}{Minesweeper}                    \\ \hline
        Hyperparameters                                                              & \multicolumn{1}{c|}{non-overlapping} & overlapping & \multicolumn{1}{c|}{non-overlapping} & \multicolumn{1}{c|}{overlapping} & \multicolumn{1}{c|}{non-overlapping} & overlapping \\ \hline
        \# latent factors                                                            & \multicolumn{1}{c|}{4}               & 4           & \multicolumn{1}{c|}{4}               & \multicolumn{1}{c|}{4}           & \multicolumn{1}{c|}{6}               & 4           \\ \hline
        learning rate                                                                & \multicolumn{1}{c|}{0.015}           & 0.015       & \multicolumn{1}{c|}{0.01}            & \multicolumn{1}{c|}{0.01}        & \multicolumn{1}{c|}{0.01}            & 0.01        \\ \hline
        \# hidden dimensions                                                         & \multicolumn{1}{c|}{128}             & 128         & \multicolumn{1}{c|}{128}             & \multicolumn{1}{c|}{256}         & \multicolumn{1}{c|}{256}             & 128         \\ \hline
        dropout rate                                                                 & \multicolumn{1}{c|}{0.35}            & 0.35        & \multicolumn{1}{c|}{0.35}            & \multicolumn{1}{c|}{0.35}        & \multicolumn{1}{c|}{0.35}            & 0.35        \\ \hline
        weight decay                                                                 & \multicolumn{1}{c|}{1e-6}            & 1e-6        & \multicolumn{1}{c|}{1e-6}            & \multicolumn{1}{c|}{1e-6}        & \multicolumn{1}{c|}{1e-6}            & 1e-6        \\ \hline
        \begin{tabular}[c]{@{}c@{}}\# neighborhood\\ routing layers\end{tabular}     & \multicolumn{1}{c|}{1}               & 1           & \multicolumn{1}{c|}{3}               & \multicolumn{1}{c|}{5}           & \multicolumn{1}{c|}{5}               & 6           \\ \hline
        \begin{tabular}[c]{@{}c@{}}\# neighborhood\\ routing iterations\end{tabular} & \multicolumn{1}{c|}{6}               & 6           & \multicolumn{1}{c|}{7}               & \multicolumn{1}{c|}{6}           & \multicolumn{1}{c|}{6}               & 7           \\ \hline
                                                                                     & \multicolumn{2}{c|}{Tolokers}                      & \multicolumn{2}{c}{Questions}                                           &                                      &             \\ \cline{1-5}
        Hyperparameters                                                              & \multicolumn{1}{c|}{non-overlapping} & overlapping & \multicolumn{1}{c|}{non-overlapping} & overlapping                      &                                      &             \\ \cline{1-5}
        \# latent factors                                                            & \multicolumn{1}{c|}{10}              & 10          & \multicolumn{1}{c|}{8}               & 2                                &                                      &             \\ \cline{1-5}
        learning rate                                                                & \multicolumn{1}{c|}{0.01}            & 0.01        & \multicolumn{1}{c|}{0.01}            & 0.01                             &                                      &             \\ \cline{1-5}
        \# hidden dimensions                                                         & \multicolumn{1}{c|}{128}             & 128         & \multicolumn{1}{c|}{256}             & 256                              &                                      &             \\ \cline{1-5}
        dropout rate                                                                 & \multicolumn{1}{c|}{0.35}            & 0.35        & \multicolumn{1}{c|}{0.35}            & 0.35                             &                                      &             \\ \cline{1-5}
        weight decay                                                                 & \multicolumn{1}{c|}{0.0045}          & 0.0045      & \multicolumn{1}{c|}{1e-6}            & 1e-6                             &                                      &             \\ \cline{1-5}
        \begin{tabular}[c]{@{}c@{}}\# neighborhood\\ routing layers\end{tabular}     & \multicolumn{1}{c|}{1}               & 1           & \multicolumn{1}{c|}{5}               & 5                                &                                      &             \\ \cline{1-5}
        \begin{tabular}[c]{@{}c@{}}\# neighborhood\\ routing iterations\end{tabular} & \multicolumn{1}{c|}{2}               & 2           & \multicolumn{1}{c|}{6}               & 6                                &                                      &             \\ \cline{1-5}
        \end{tabular}
        \end{table}

\begin{table}[t]
    \centering
    \caption{The search range of the hyperparameters used in our proposed FedIIH.}
    \footnotesize	
    \label{hyperparameters_range}
\begin{tabular}{c|cccc}
\hline
Hyperparameters & \# latent factors     & learning rate                                                            & \# hidden dimensions                                                         & dropout rate    \\ \hline
Range           & \{1, 2, 4, 6, 8, 10\} & \{0.01, 0.015, 0.02\}                                                    & \{128, 256\}                                                                 & {[}0.25, 0.4{]} \\ \hline
Hyperparameters & weight decay          & \begin{tabular}[c]{@{}c@{}}\# neighborhood\\ routing layers\end{tabular} & \begin{tabular}[c]{@{}c@{}}\# neighborhood\\ routing iterations\end{tabular} &                 \\ \cline{1-4}
Range           & {[}1e-6, 5e-3{]}      & \{1, 2, 3, 4, 5, 6\}                                                     & \{2, 3, 4, 5, 6, 7\}                                                         &                 \\ \cline{1-4}
\end{tabular}
\end{table}

\subsection{I.6 Implementations of Two Prior Distributions}
\label{implementations_two_priors}
Here we describe the detailed implementations of $p(\tilde{\bm{\alpha}}^k)$ and $p(\tilde{\mathbf{H}}_m^k | \tilde{\bm{\alpha}}^k)$, respectively. Since $\tilde{\bm{\alpha}}^k$ denotes the posterior mean of $\bm{\alpha}^k$ for the $k$-th global latent factor, we assume that $p(\tilde{\bm{\alpha}}^k)\sim \mathcal{N}(\tilde{\bm{\alpha}}^k, \sigma_{\bm{\alpha}^k}^2\mathbf{I})$, where $\tilde{\bm{\alpha}}^k$ is given in Eq.~\eqref{eq_elbo_9}. Moreover, $\sigma_{\bm{\alpha}^k}^2$ and $\sigma_{\tilde{\mathbf{H}}_{m}^k}^2$ is set to 1 and 0.25, respectively. In other words, $p(\tilde{\bm{\alpha}}^k)\sim \mathcal{N}(\frac{\sum_{m=1}^{M}\hat{\bm{\mu}}_{\tilde{\mathbf{H}}_{m}^k}}{M+0.25}, \mathbf{I})$. According to Tab.~\ref{tableA1}, $p(\tilde{\mathbf{H}}_{m}^k|\bm{\alpha}^k) \sim \mathcal{N}(\bm{\alpha}^k, \sigma_{\tilde{\mathbf{H}}_{m}^k}^2\mathbf{I})$. Consequently, we can have
\begin{equation}\label{eq_elbo_10}
\begin{aligned}
p(\tilde{\mathbf{H}}_m^k | \tilde{\bm{\alpha}}^k) &\sim \mathcal{N}(\tilde{\bm{\alpha}}^k, 0.25\mathbf{I})\\
&\sim \mathcal{N}(\frac{\sum_{m=1}^{M}\hat{\bm{\mu}}_{\tilde{\mathbf{H}}_{m}^k}}{M+0.25}, 0.25\mathbf{I}).
\end{aligned}
\end{equation}

\section{J. Additional Experiments}
In this section, we provide the additional experiments related to the ablation studies and hyperparameter analysis, respectively.

\subsection{J.1 Ablation Studies on Other Datasets}
\label{ablation_studies}
To further analyze the contribution of each component, we conduct ablation studies on the remaining datasets in both non-overlapping and overlapping partitioning settings with 10 clients. As shown in Tab.~\ref{table_other_ablation}, we can find that the performance of FedIIH is significantly better than the three variants. It validates that each component indeed contributes a lot to the final performance.

\begin{table}[]
    \centering
    \scriptsize		
    \caption{Ablation studies in both non-overlapping and overlapping partitioning settings on other datasets with 10 clients.}
      \label{table_other_ablation}
         \scalebox{0.8}{
\begin{tabular}{lcccccc}
\hline
\multicolumn{1}{c}{} & \multicolumn{2}{c}{CiteSeer}                                           & \multicolumn{2}{c}{PubMed}                                              & \multicolumn{2}{c}{Amazon-Computer}                                     \\ \hline
Methods              & non-overlapping                    & overlapping                       & non-overlapping                    & overlapping                        & non-overlapping                    & overlapping                        \\ \hline
FedIIH (w/o HM)      & 74.91$\pm$0.27 ($\downarrow$1.59)  & 72.29$\pm$0.16 ($\downarrow$0.87) & 85.19$\pm$0.05 ($\downarrow$2.46)  & 85.16$\pm$0.17 ($\downarrow$0.71)  & 85.75$\pm$0.90 ($\downarrow$5.11)  & 87.71$\pm$0.20 ($\downarrow$2.44)  \\ \hline
FedIIH (w/o VI)      & 72.27$\pm$2.16 ($\downarrow$4.23)  & 72.60$\pm$0.16 ($\downarrow$0.56) & 81.35$\pm$0.20 ($\downarrow$6.30)  & 84.56$\pm$0.04 ($\downarrow$1.31)  & 69.16$\pm$1.15 ($\downarrow$21.70) & 74.90$\pm$1.24 ($\downarrow$15.25) \\ \hline
FedIIH (w/o Dis)     & 75.71$\pm$0.38 ($\downarrow$0.79) & 71.54$\pm$0.12 ($\downarrow$1.62) & 85.71$\pm$0.12 ($\downarrow$1.94) & 84.30$\pm$0.03 ($\downarrow$1.57)  & 88.96$\pm$0.08 ($\downarrow$1.90) & 88.51$\pm$0.04 ($\downarrow$1.64) \\ \hline
FedIIH               & \textbf{76.50$\pm$0.06}            & \textbf{73.16$\pm$0.18}           & \textbf{87.65$\pm$0.18}            & \textbf{85.87$\pm$0.03}            & \textbf{90.86$\pm$0.23}            & \textbf{90.15$\pm$0.04}            \\ \hline
\multicolumn{1}{c}{} & \multicolumn{2}{c}{Amazon-Photo}                                       & \multicolumn{2}{c}{ogbn-arxiv}                                          & \multicolumn{2}{c}{Roman-empire}                                        \\ \hline
Methods              & non-overlapping                    & overlapping                       & non-overlapping                    & overlapping                        & non-overlapping                    & overlapping                        \\ \hline
FedIIH (w/o HM)      & 91.84$\pm$0.05 ($\downarrow$2.38)  & 91.88$\pm$0.27 ($\downarrow$1.50) & 65.18$\pm$0.41 ($\downarrow$4.16)  & 63.54$\pm$0.15 ($\downarrow$3.15)  & 56.28$\pm$0.20 ($\downarrow$10.16) & 60.90$\pm$0.21 ($\downarrow$4.58)  \\ \hline
FedIIH (w/o VI)      & 79.25$\pm$0.96 ($\downarrow$14.97) & 82.16$\pm$0.56($\downarrow$11.22) & 49.26$\pm$0.74 ($\downarrow$20.08) & 46.20$\pm$1.73 ($\downarrow$20.49) & 57.38$\pm$0.18 ($\downarrow$9.06)  & 61.69$\pm$0.09 ($\downarrow$3.79)  \\\hline
FedIIH (w/o Dis)     & 92.43$\pm$0.01 ($\downarrow$1.79) & 92.01$\pm$0.04($\downarrow$1.37) & 61.51$\pm$0.15 ($\downarrow$7.83)  & 60.64$\pm$0.16 ($\downarrow$6.05)  & 40.51$\pm$0.27 ($\downarrow$25.93) & 42.84$\pm$0.09 ($\downarrow$22.64) \\\hline
FedIIH               & \textbf{94.22$\pm$0.08}            & \textbf{93.38$\pm$0.00}           & \textbf{69.34$\pm$0.02}            & \textbf{66.69$\pm$0.09}            & \textbf{66.44$\pm$0.28}            & \textbf{65.48$\pm$0.12}            \\ \hline
\multicolumn{1}{c}{} & \multicolumn{2}{c}{Minesweeper}                                        & \multicolumn{2}{c}{Tolokers}                                            & \multicolumn{2}{c}{Questions}                                           \\ \hline
Methods              & non-overlapping                    & overlapping                       & non-overlapping                    & overlapping                        & non-overlapping                    & overlapping                        \\\hline
FedIIH (w/o HM)      & 70.88$\pm$0.01 ($\downarrow$2.35)  & 66.56$\pm$0.07 ($\downarrow$2.79) & 64.56$\pm$0.17 ($\downarrow$6.76)  & 69.21$\pm$0.22 ($\downarrow$2.46)  & 65.90$\pm$0.09 ($\downarrow$2.09)  & 67.77$\pm$0.10 ($\downarrow$1.02)  \\\hline
FedIIH (w/o VI)      & 71.34$\pm$0.09 ($\downarrow$1.89)  & 68.47$\pm$0.06 ($\downarrow$0.88) & 64.34$\pm$0.12 ($\downarrow$6.98)  & 68.03$\pm$0.34 ($\downarrow$3.64)  & 66.63$\pm$0.08 ($\downarrow$1.36)  & 68.24$\pm$0.06 ($\downarrow$0.55)  \\\hline
FedIIH (w/o Dis)     & 70.67$\pm$0.02 ($\downarrow$2.56)   & 68.75$\pm$0.19 ($\downarrow$0.60) & 62.53$\pm$0.29 ($\downarrow$8.79) & 68.10$\pm$0.12 ($\downarrow$3.57)  & 66.42$\pm$0.01 ($\downarrow$1.57)  & 67.52$\pm$0.17 ($\downarrow$1.27)  \\\hline
FedIIH               & \textbf{73.23$\pm$0.04}            & \textbf{69.35$\pm$0.25}           & \textbf{71.32$\pm$0.09}            & \textbf{71.67$\pm$0.02}            & \textbf{67.99$\pm$0.09}            & \textbf{68.79$\pm$0.09}            \\ \hline
\end{tabular}
}
\end{table}

\subsection{J.2 Hyperparameter Sensitivity Analysis of $K$}
\label{sensitivity_k}
\begin{figure}[t]
	\centering
	\includegraphics[width=8cm]{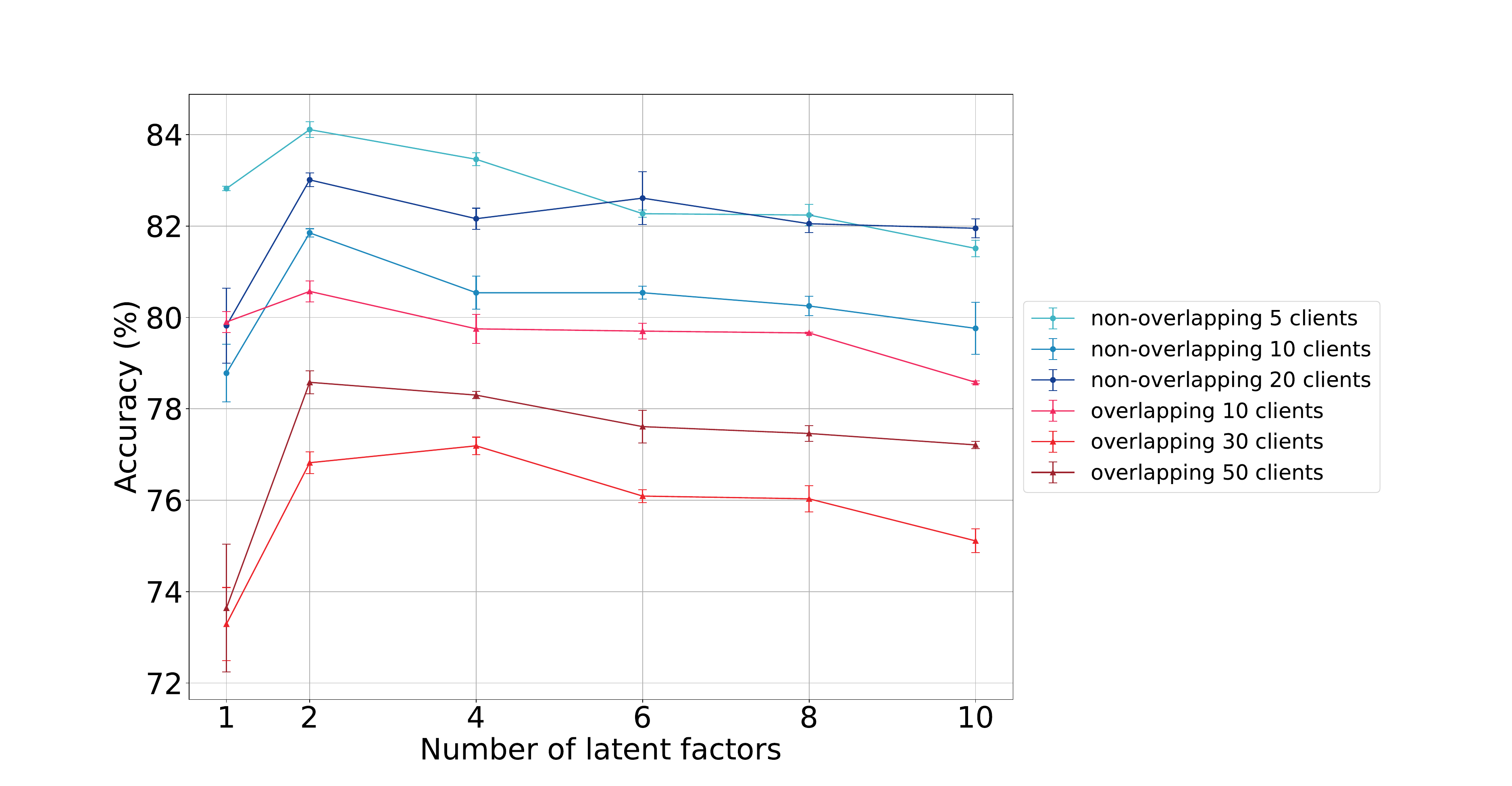}
	\caption{Sensitivity of the number of latent factors $K$ on the \textit{Cora} dataset.}
	\label{fig_K_10}
\end{figure}

\begin{figure}[t]
	\centering
	\includegraphics[width=8cm]{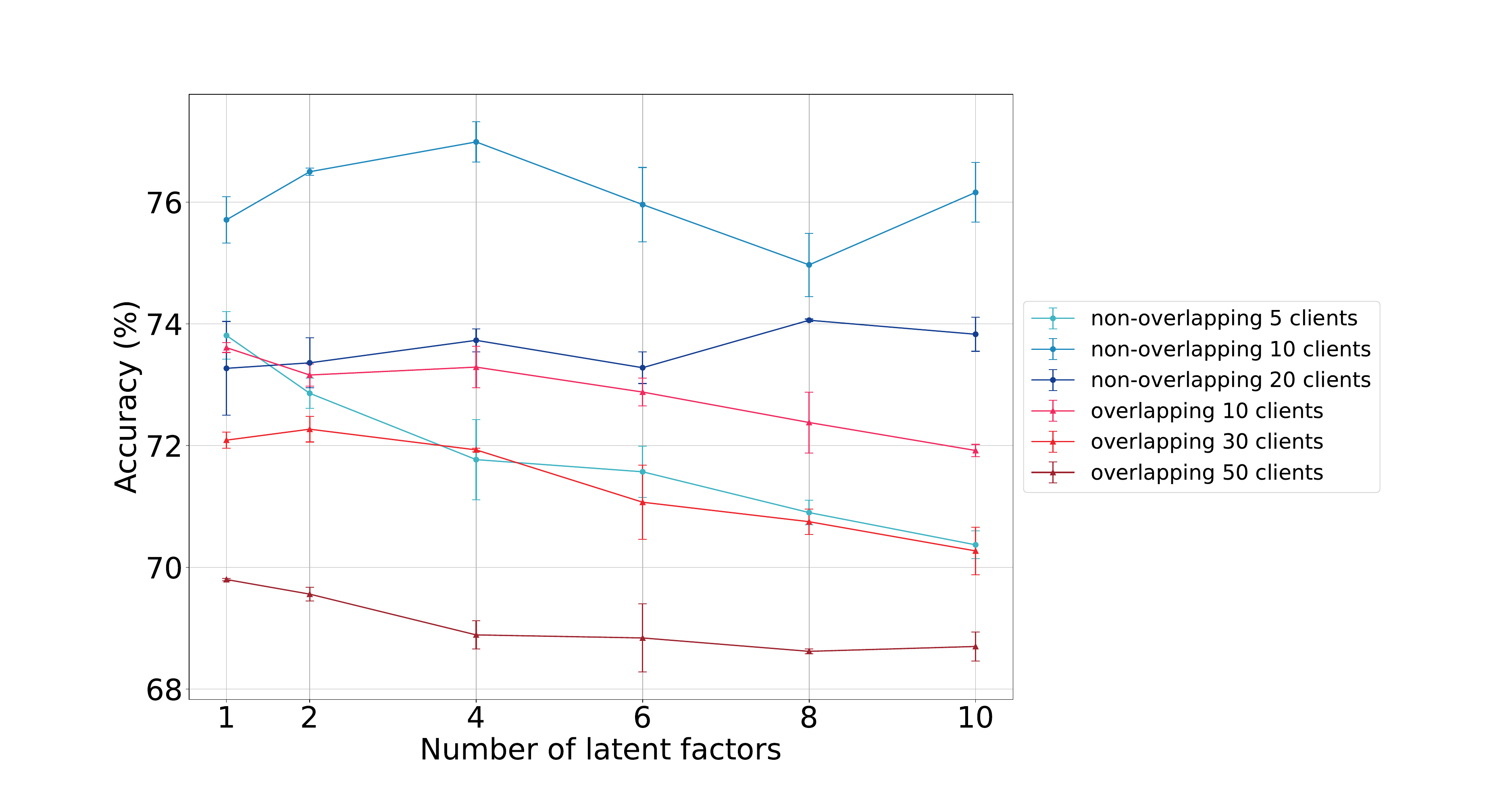}
	\caption{Sensitivity of the number of latent factors $K$ on the \textit{CiteSeer} dataset.}
	\label{fig_K_2}
\end{figure}

\begin{figure}[t]
	\centering
	\includegraphics[width=8cm]{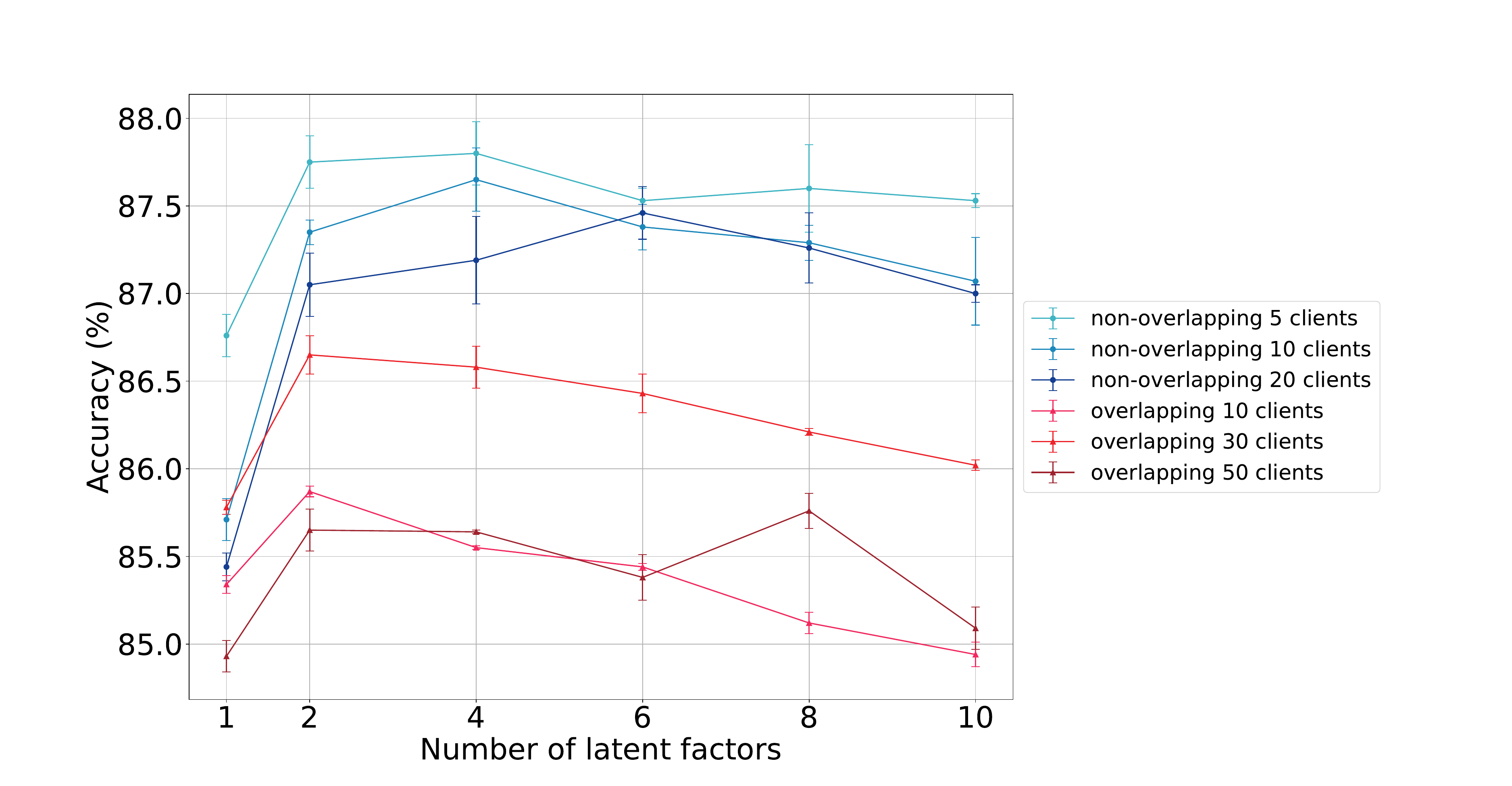}
	\caption{Sensitivity of the number of latent factors $K$ on the \textit{PubMed} dataset.}
	\label{fig_K_3}
\end{figure}

The hyperparameter sensitivity analysis of $K$ on the remaining datasets are shown in Fig.~\ref{fig_K_2}, Fig.~\ref{fig_K_3},~..., to Fig.~\ref{fig_K_11}. According to the experimental results, we have the following observations and insights: \textbf{1)} In general, the performance variation under different $K$ is small except for the \textit{Roman-empire} dataset (see Fig.~\ref{fig_K_7}). The \textit{Roman-empire} dataset has low homophily~\cite{platonov2023a}, which is based on the Roman Empire article from the English Wikipedia. Each node in the graph corresponds to a (non-unique) word in the text, and the node label is determined by the syntactic role of the word. Due to the syntactic dependencies within neighboring words, there exists strong heterogeneity within the \textit{Roman-empire} graph. Consequently, if we ignore this intra-heterogeneity (\textit{i.e.}, without disentangling), the performance will decrease significantly on the \textit{Roman-empire} dataset. This is consistent with the experimental results in Tab.~3 and Tab.~4, where our FedIIH outperforms other methods by a large margin. \textbf{2)} As the value of $K$ increases, the performance may increase or decrease depending on the datasets. For example, on the \textit{Roman-empire} (see Fig.~\ref{fig_K_7}) and \textit{Tolokers} (see Fig.~4) datasets, their performances increase consistently as the value of $K$ increases. On the contrary, on the \textit{Cora} dataset (see Fig.~\ref{fig_K_10}), the accuracy reaches its highest value when $K=2$, and then it decreases as the value of $K$ is further increased.

\begin{figure}[t]
	\centering
	\includegraphics[width=8cm]{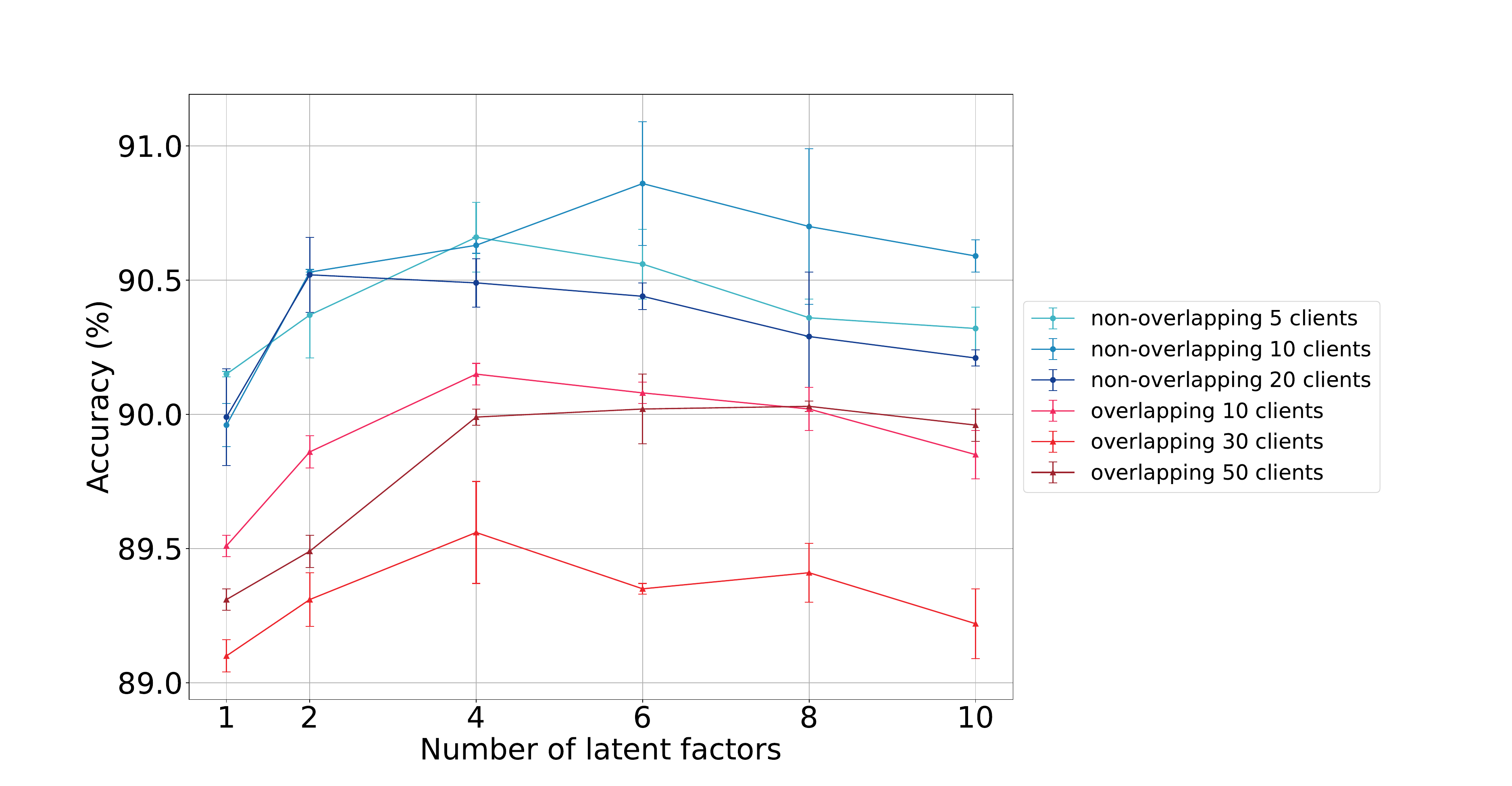}
	\caption{Sensitivity of the number of latent factors $K$ on the \textit{Amazon-Computer} dataset.}
	\label{fig_K_4}
\end{figure}

\begin{figure}[t]
	\centering
	\includegraphics[width=8cm]{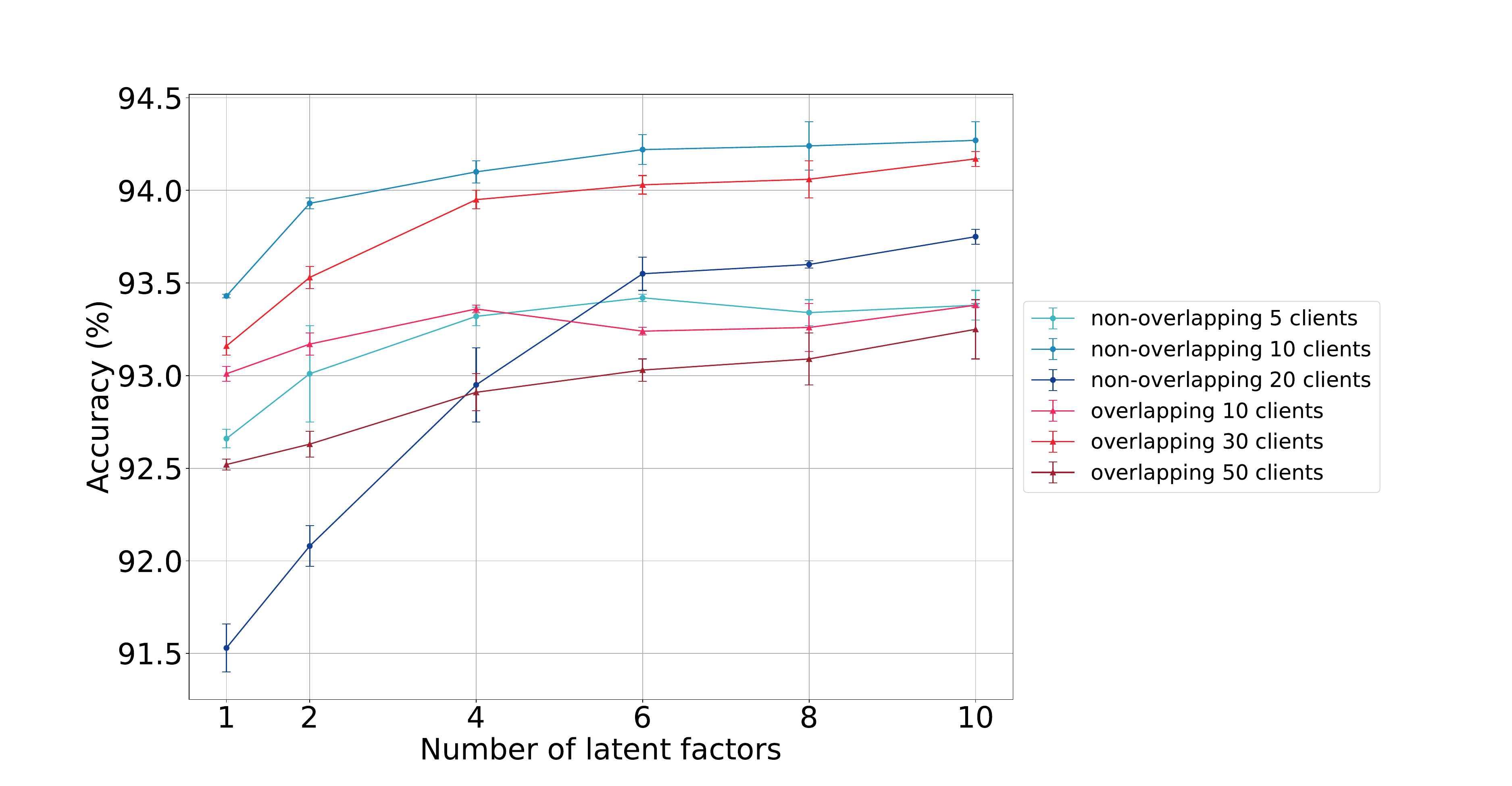}
	\caption{Sensitivity of the number of latent factors $K$ on the \textit{Amazon-Photo} dataset.}
	\label{fig_K_5}
\end{figure}

\begin{figure}[t]
	\centering
	\includegraphics[width=8cm]{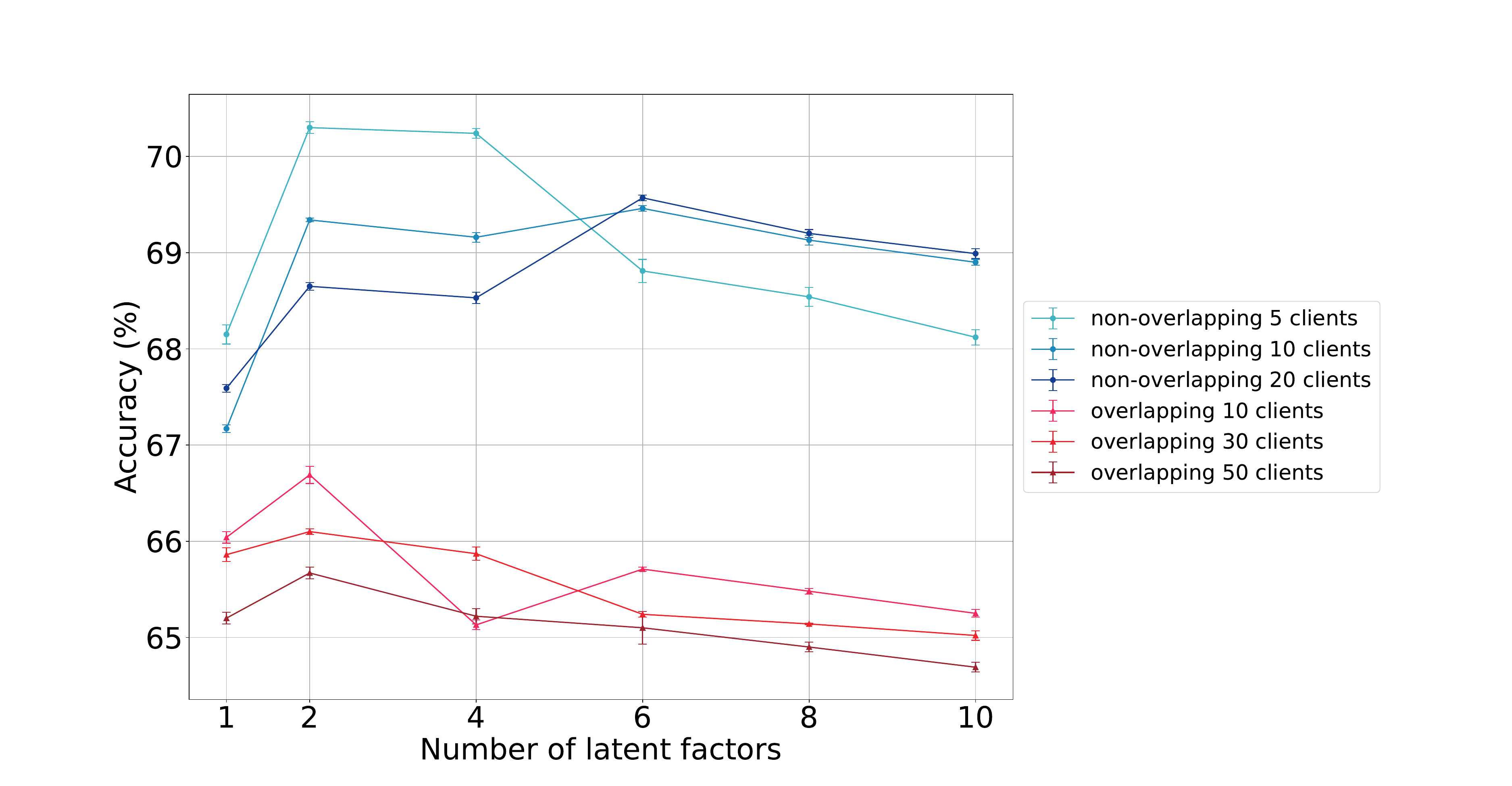}
	\caption{Sensitivity of the number of latent factors $K$ on the \textit{ogbn-arxiv} dataset.}
	\label{fig_K_6}
\end{figure}

\begin{figure}[t]
	\centering
	\includegraphics[width=8cm]{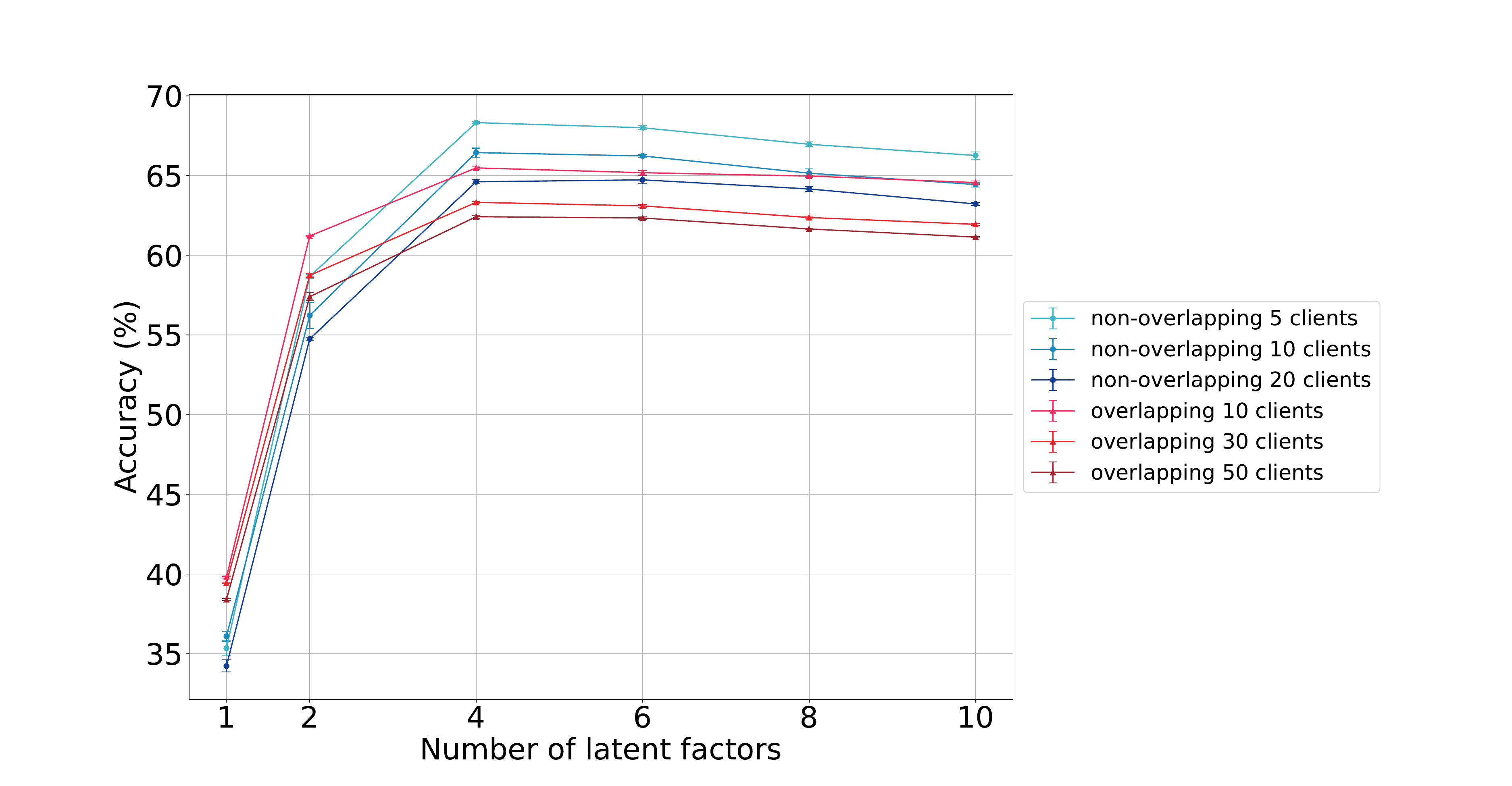}
	\caption{Sensitivity of the number of latent factors $K$ on the \textit{Roman-empire} dataset.}
	\label{fig_K_7}
\end{figure}

\begin{figure}[t]
	\centering
	\includegraphics[width=8cm]{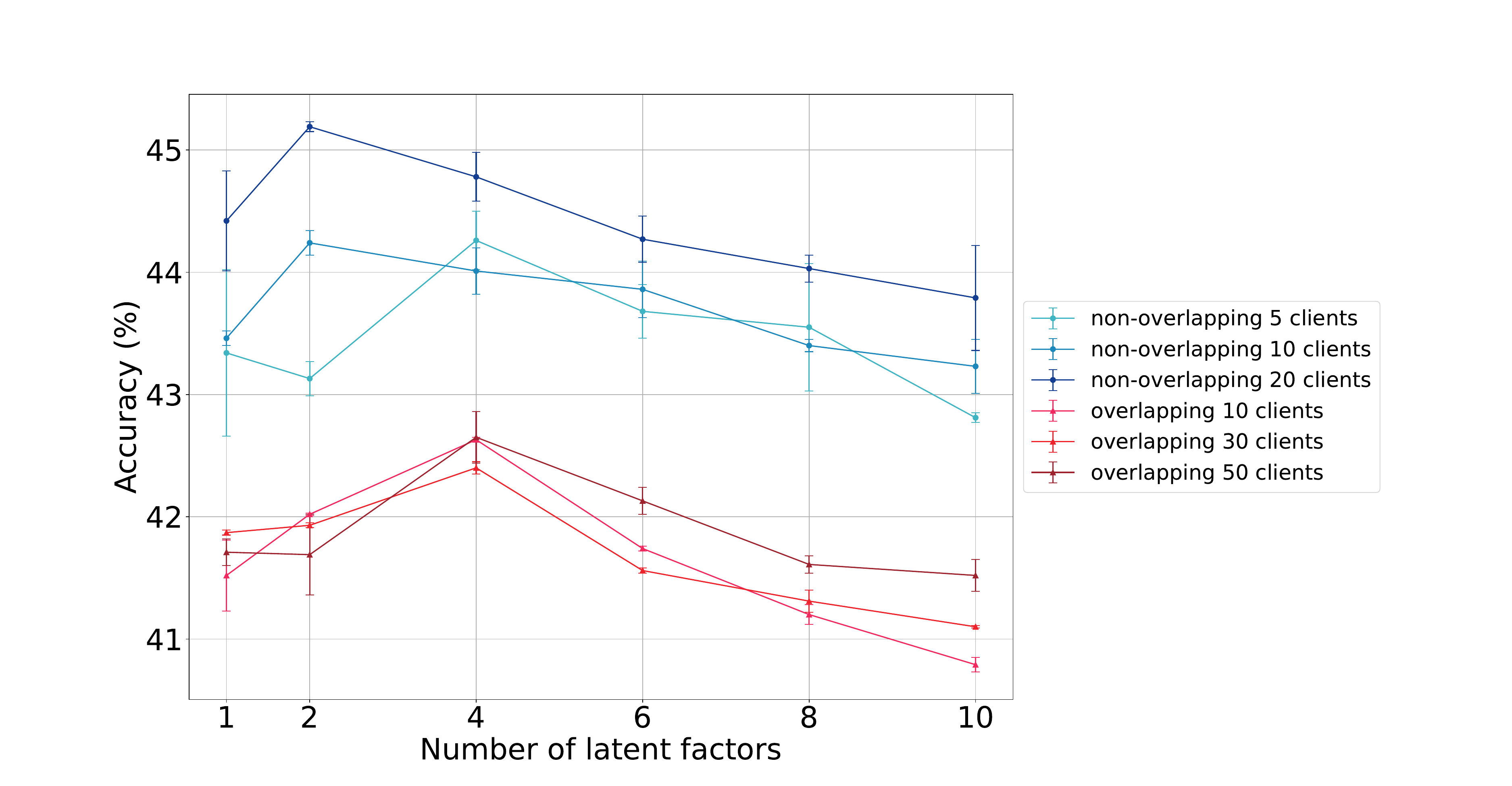}
	\caption{Sensitivity of the number of latent factors $K$ on the \textit{Amazon-ratings} dataset.}
	\label{fig_K_8}
\end{figure}

\begin{figure}[t]
	\centering
	\includegraphics[width=8cm]{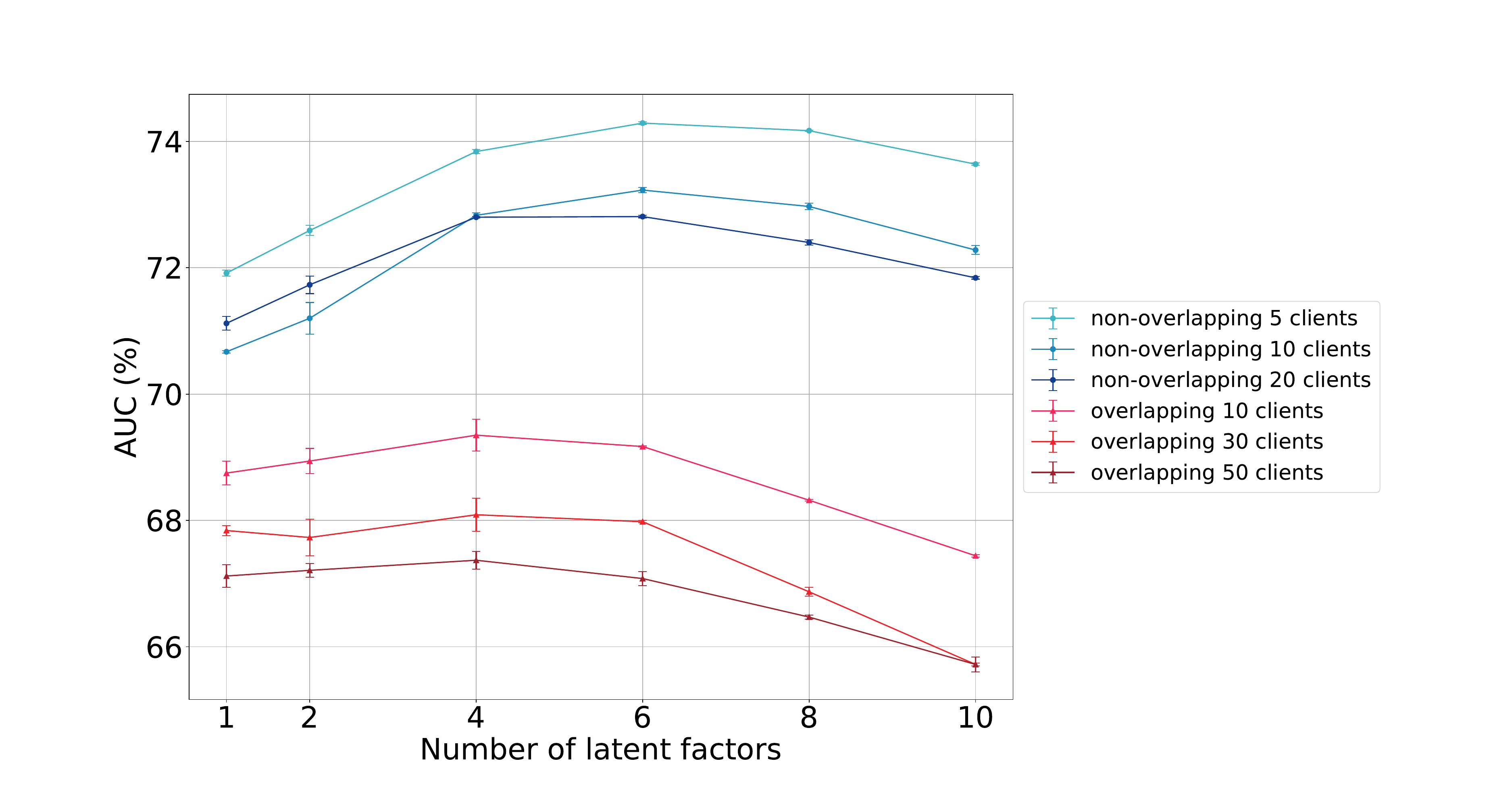}
	\caption{Sensitivity of the number of latent factors $K$ on the \textit{Minesweeper} dataset.}
	\label{fig_K_9}
\end{figure}

\begin{figure}[t]
	\centering
	\includegraphics[width=8cm]{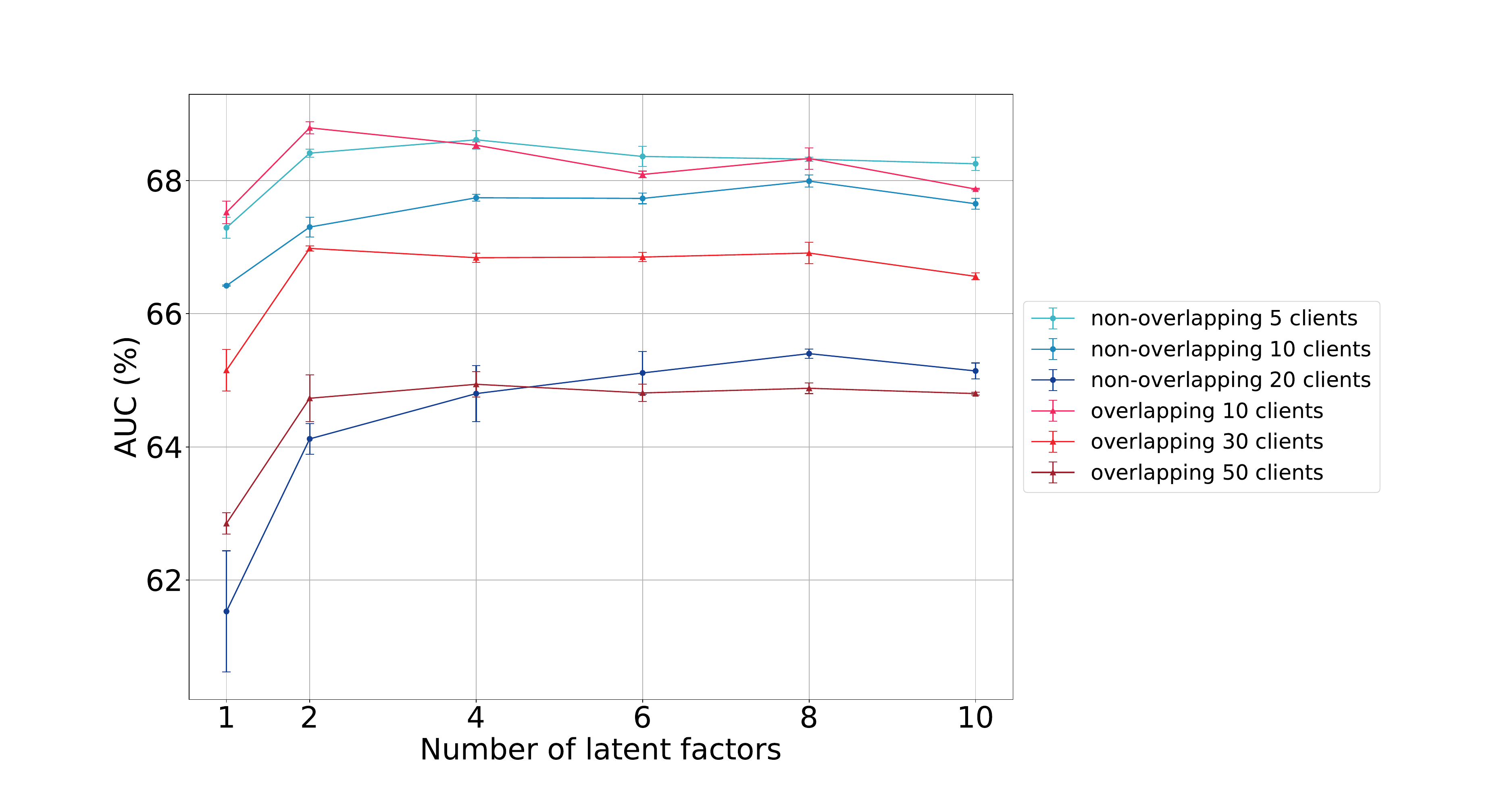}
	\caption{Sensitivity of the number of latent factors $K$ on the \textit{Questions} dataset.}
	\label{fig_K_11}
\end{figure}

\section{K. Discussions}
\subsection{K.1 Similarity Heatmaps of FED-PUB and FedIIH over Three Independent Runs}
\label{fediih_sim}
\begin{figure}[t]
    \centering
    \begin{subfigure}{0.23\textwidth}
        \includegraphics[width=\linewidth]{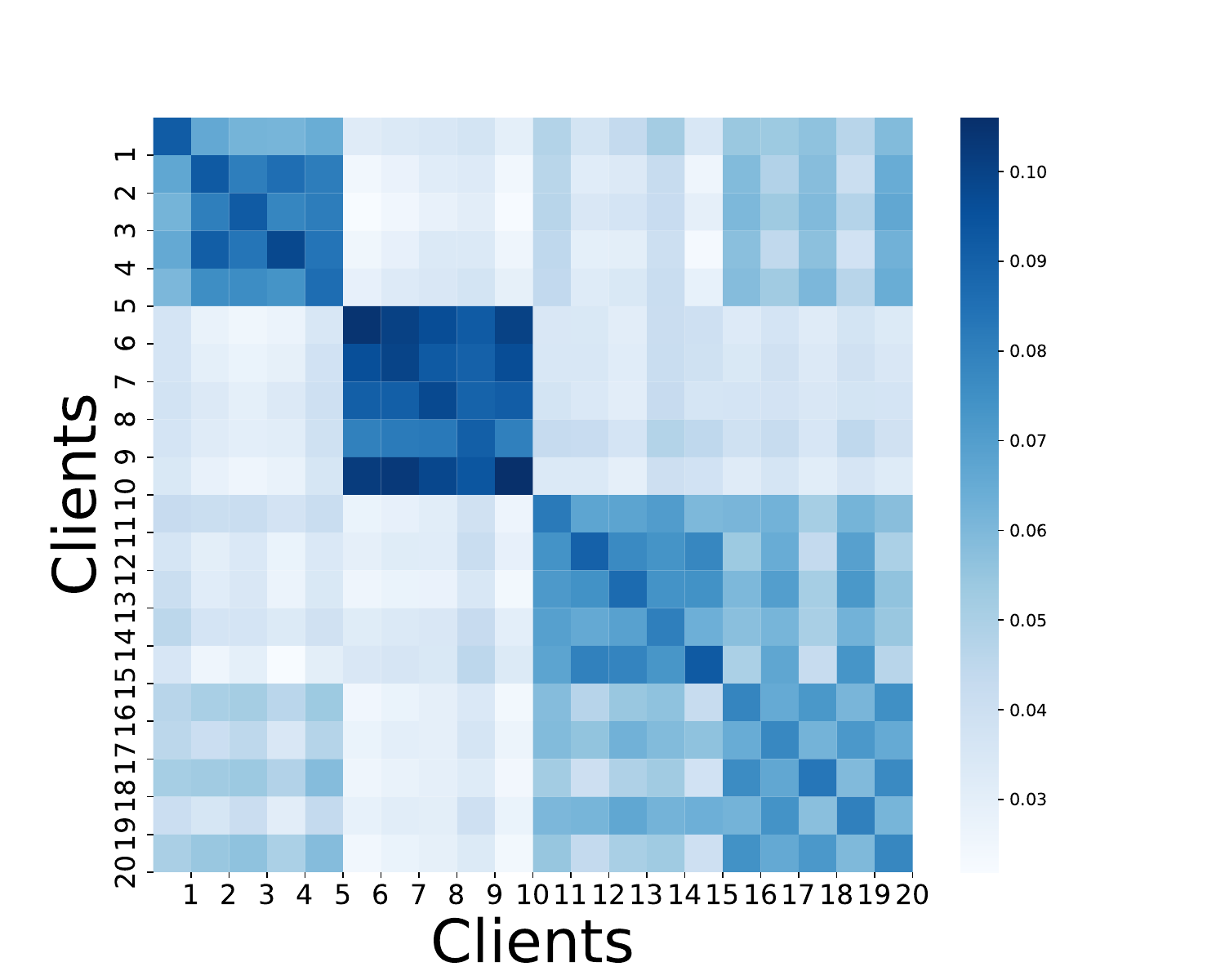}
        \caption{the first independent run}
        \label{fig_A_3_1}
    \end{subfigure}%
    \hfill
    \begin{subfigure}{0.23\textwidth}
        \includegraphics[width=\linewidth]{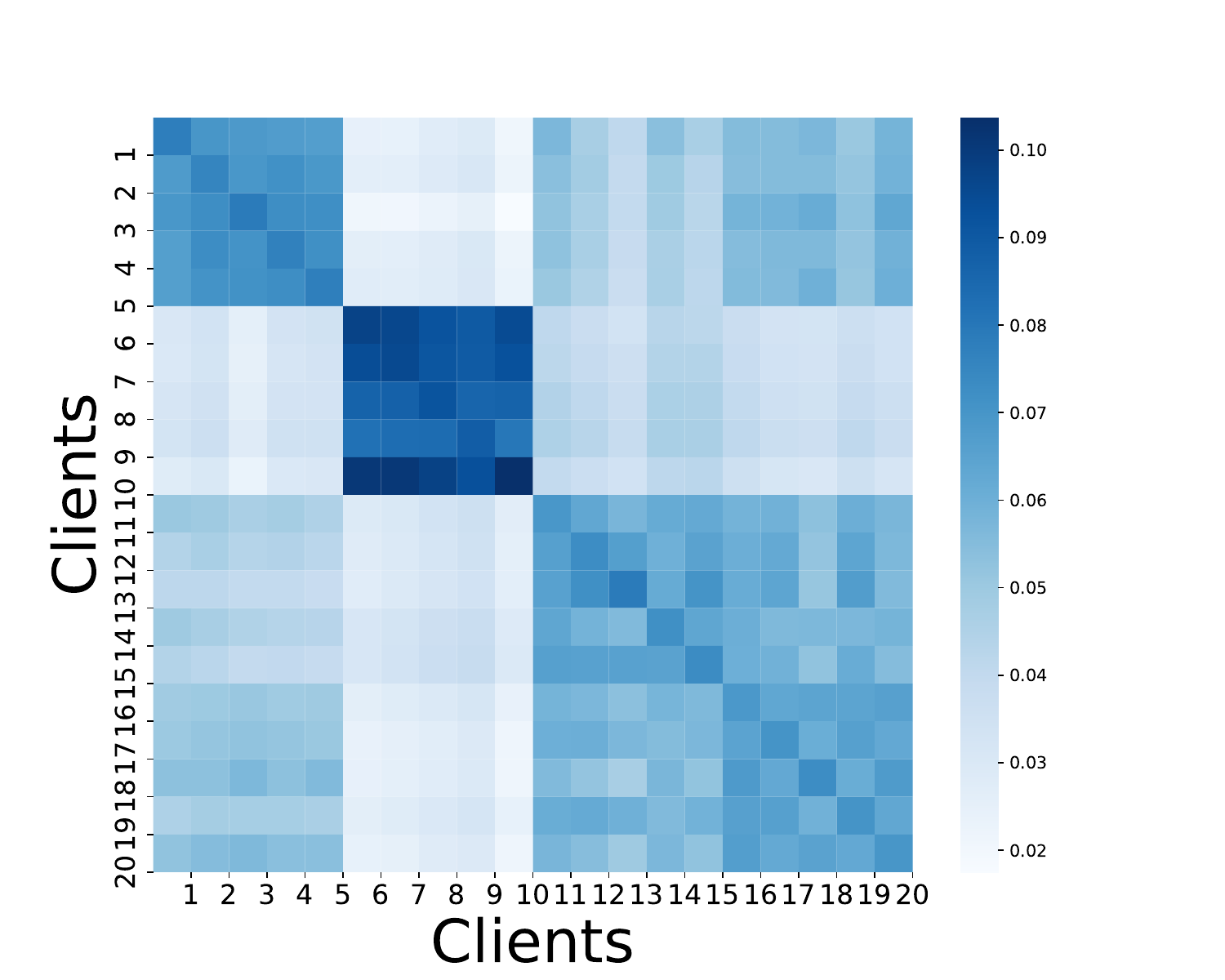}
        \caption{the second independent run}
        \label{fig_A_3_22}
    \end{subfigure}%
    \hfill
    \begin{subfigure}{0.23\textwidth}
        \includegraphics[width=\linewidth]{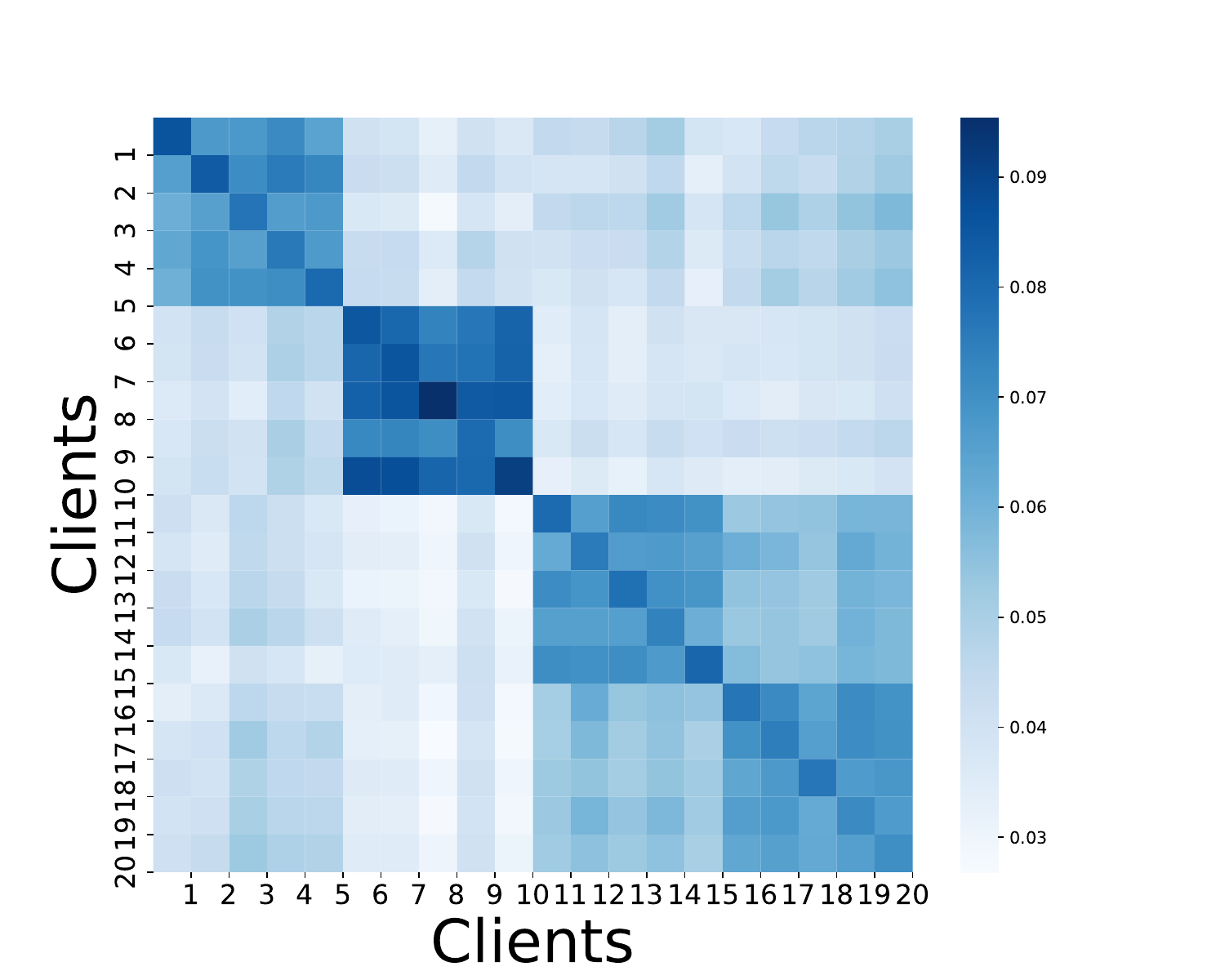}
        \caption{the third independent run}
        \label{fig_A_3_3}
    \end{subfigure}
    \caption{The similarity heatmaps of FED-PUB over three independent runs on the \textit{Cora} dataset in the overlapping setting with 20 clients.}
    \label{fig_A_3}
\end{figure}

\begin{figure}
    \centering
    \begin{subfigure}{0.23\textwidth}
        \includegraphics[width=\linewidth]{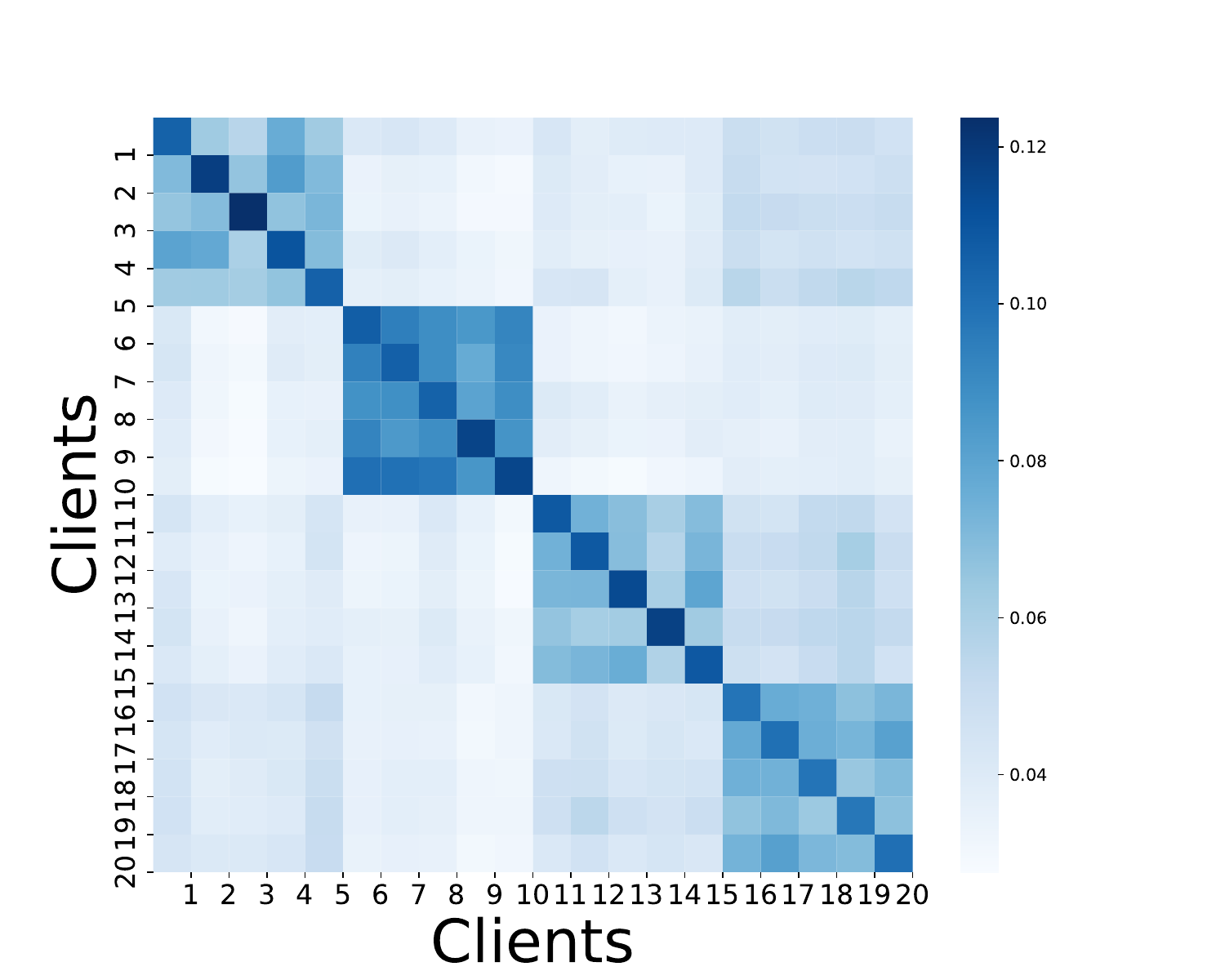}
        \caption{the first independent run of the 1st latent factor ($K=2$)}
        \label{fig_b_3_1}
    \end{subfigure}%
    \hfill
    \begin{subfigure}{0.23\textwidth}
        \includegraphics[width=\linewidth]{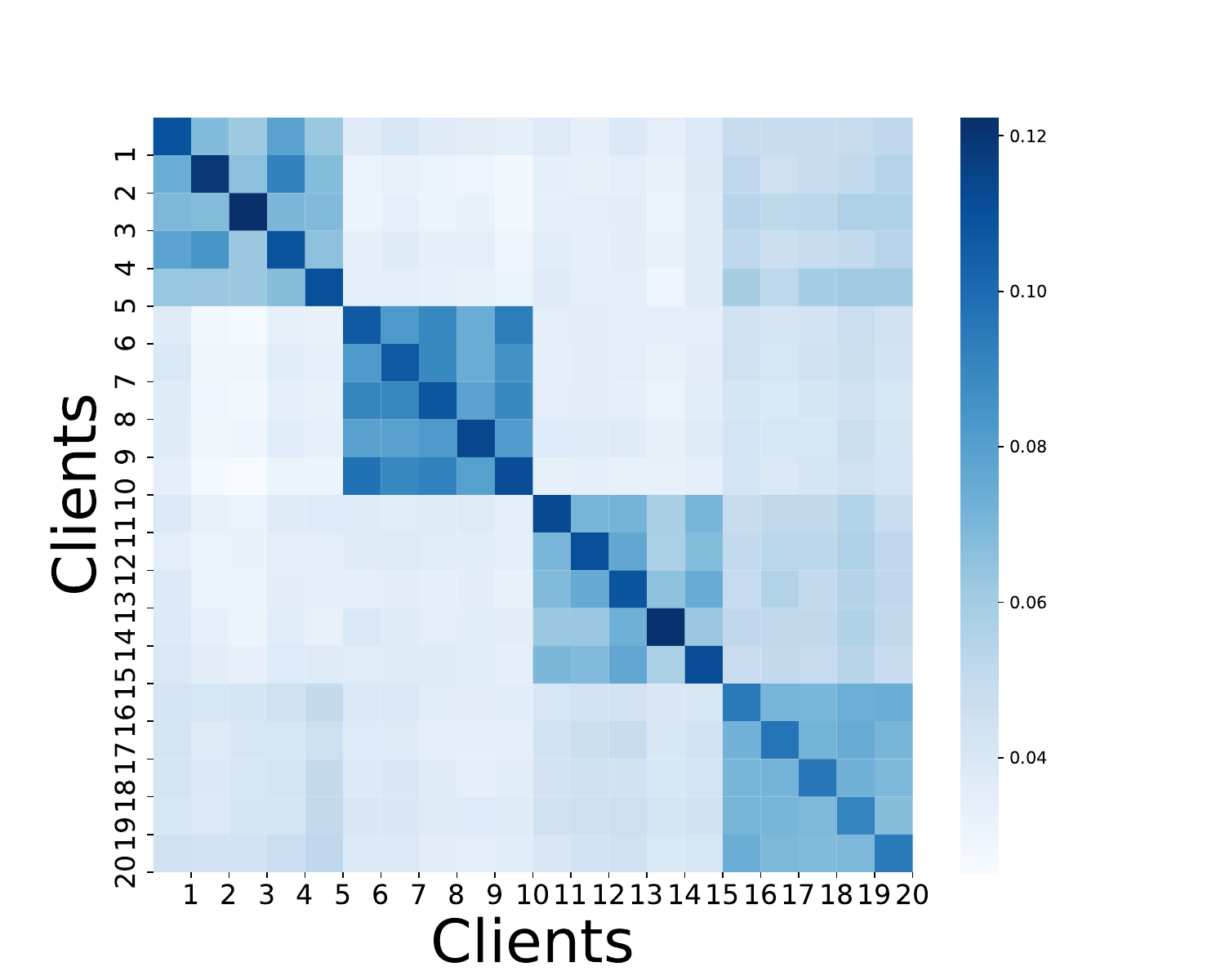}
        \caption{the second independent run of the 1st latent factor ($K=2$)}
        \label{fig_b_3_2}
    \end{subfigure}
    \hfill
    \begin{subfigure}{0.23\textwidth}
        \includegraphics[width=\linewidth]{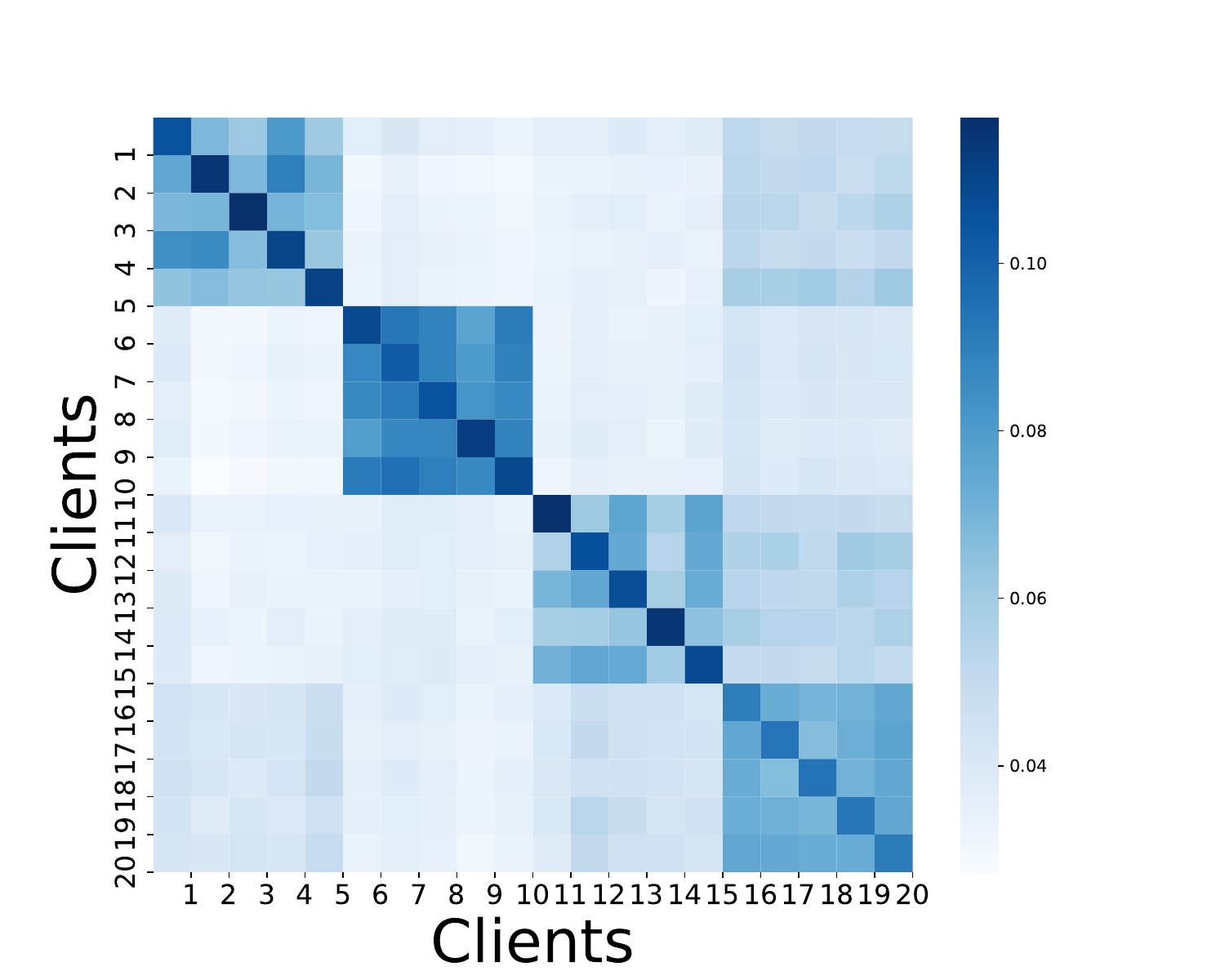}
        \caption{the third independent run of the 1st latent factor ($K=2$)}
        \label{fig_b_3_3}
    \end{subfigure}
    \hfill

    \begin{subfigure}{0.23\textwidth}
        \includegraphics[width=\linewidth]{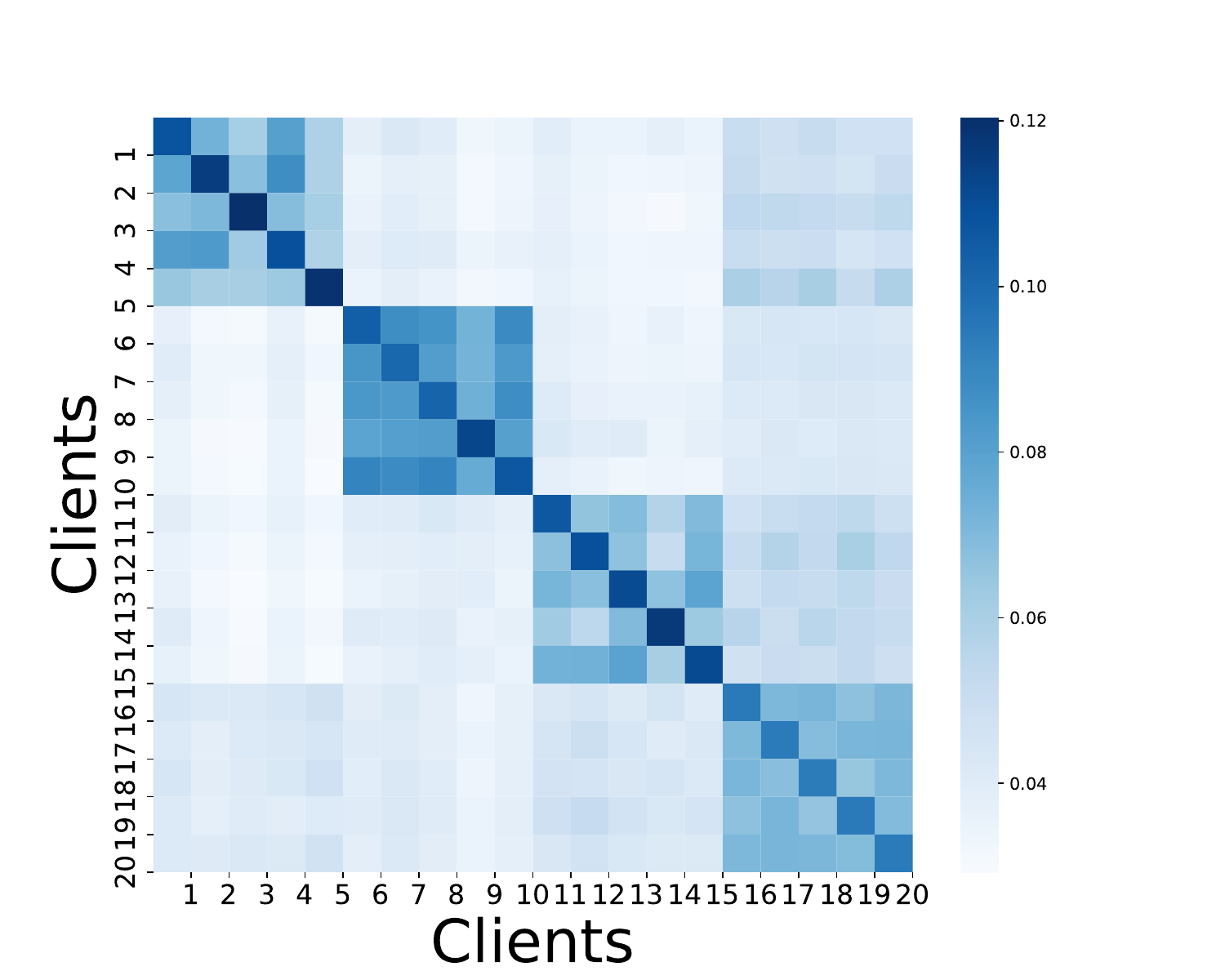}
        \caption{the first independent run of the 2nd latent factor ($K=2$)}
        \label{fig_b_3_4}
    \end{subfigure}%
    \hfill
    \begin{subfigure}{0.23\textwidth}
        \includegraphics[width=\linewidth]{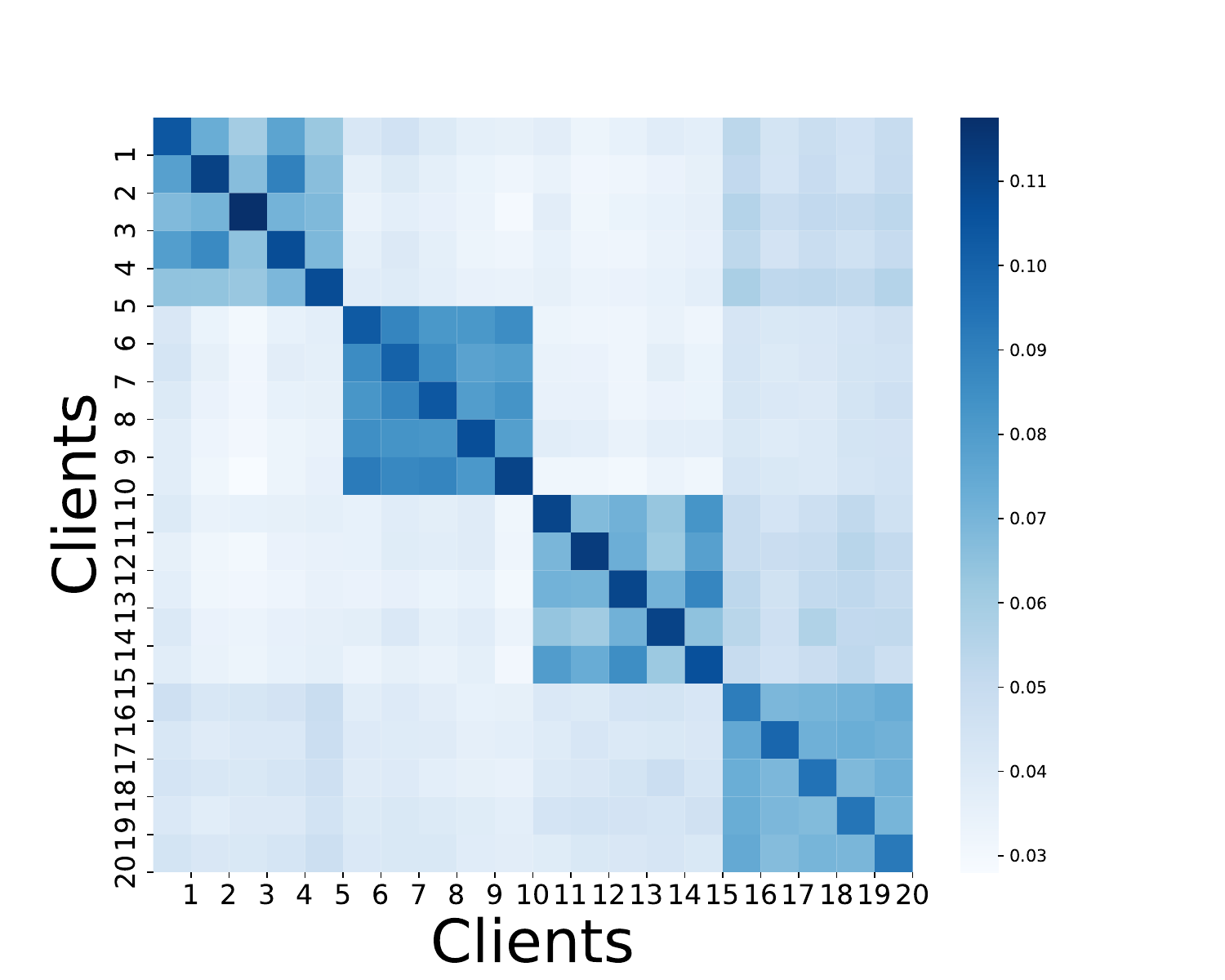}
        \caption{the second independent run of the 2nd latent factor ($K=2$)}
        \label{fig_b_3_5}
    \end{subfigure}
    \hfill
    \begin{subfigure}{0.23\textwidth}
        \includegraphics[width=\linewidth]{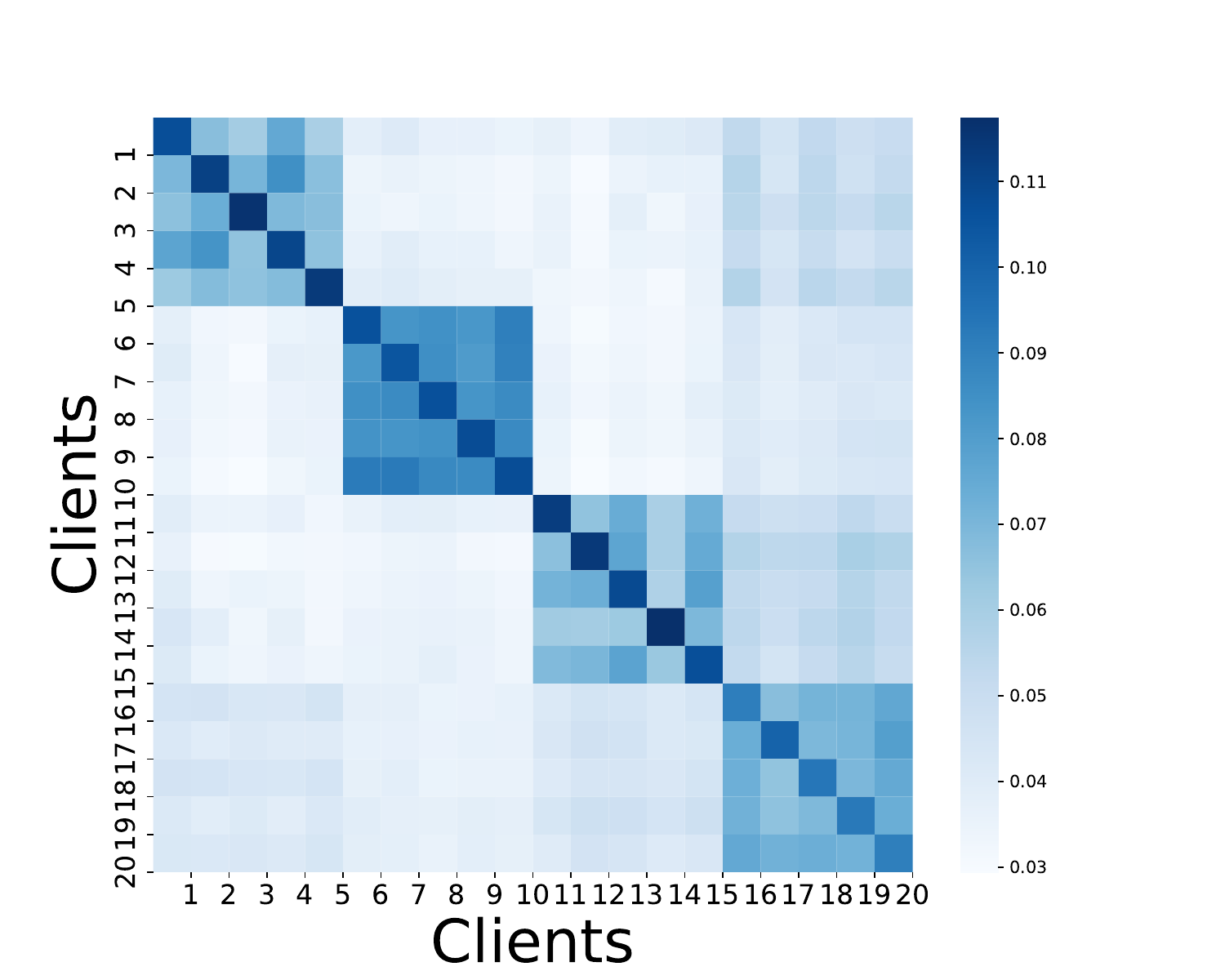}
        \caption{the third independent run of the 2nd latent factor ($K=2$)}
        \label{fig_b_3_6}
    \end{subfigure}
    \hfill
    \caption{The similarity heatmaps of FedIIH over three independent runs on the \textit{Cora} dataset in the overlapping setting with 20 clients.}
    \label{fig_b_3}
\end{figure}

\begin{figure}[t]
    \centering
    \begin{subfigure}{0.23\textwidth}
        \includegraphics[width=\linewidth]{fig/similarity_heatmaps/fedpub_1_amazon_ratings.pdf}
        \caption{the first independent run}
        \label{fig_A_3_2_1}
    \end{subfigure}%
    \hfill
    \begin{subfigure}{0.23\textwidth}
        \includegraphics[width=\linewidth]{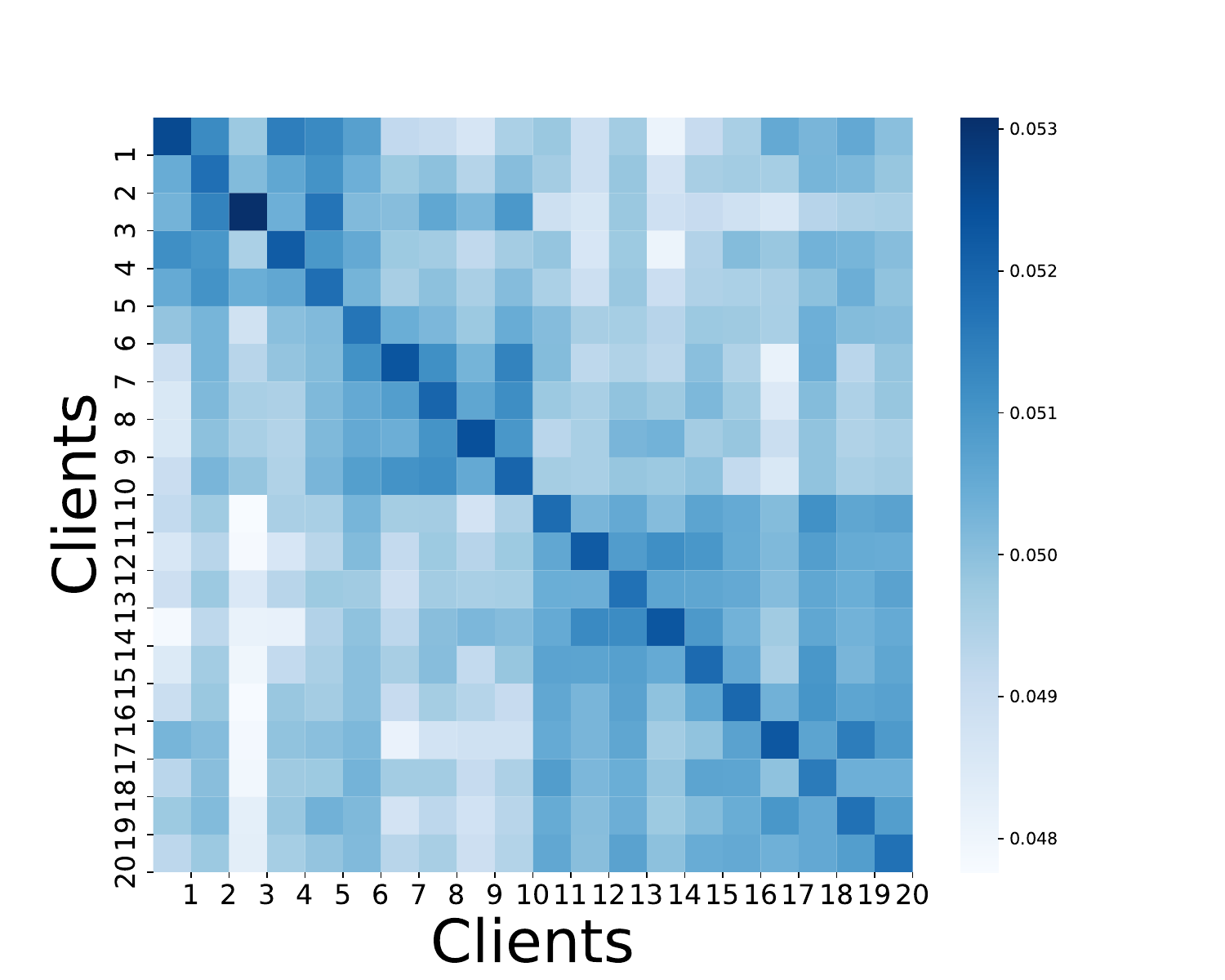}
        \caption{the second independent run}
        \label{fig_A_3_2_2}
    \end{subfigure}%
    \hfill
    \begin{subfigure}{0.23\textwidth}
        \includegraphics[width=\linewidth]{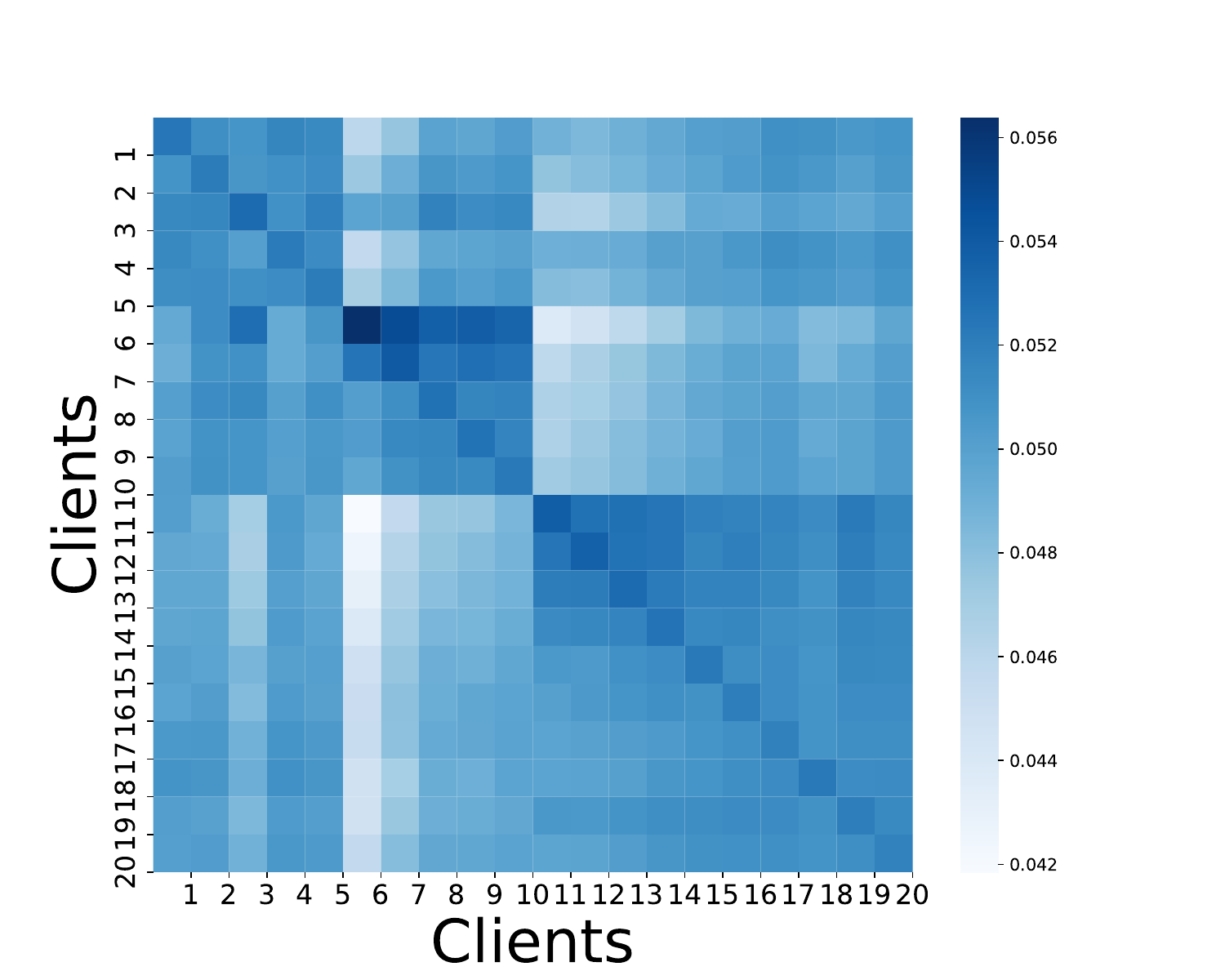}
        \caption{the third independent run}
        \label{fig_A_3_2_3}
    \end{subfigure}
    \caption{The similarity heatmaps of FED-PUB over three independent runs on the \textit{Amazon-ratings} dataset in the overlapping setting with 20 clients.}
    \label{fig_A_3_2}
\end{figure}

\begin{figure}
    \centering
    \begin{subfigure}{0.23\textwidth}
        \includegraphics[width=\linewidth]{fig/similarity_heatmaps/fediih_1_amazon_ratings_latent1.pdf}
        \caption{the first independent run of the 1st latent factor ($K=2$)}
        \label{fig_c_3_1}
    \end{subfigure}%
    \hfill
    \begin{subfigure}{0.23\textwidth}
        \includegraphics[width=\linewidth]{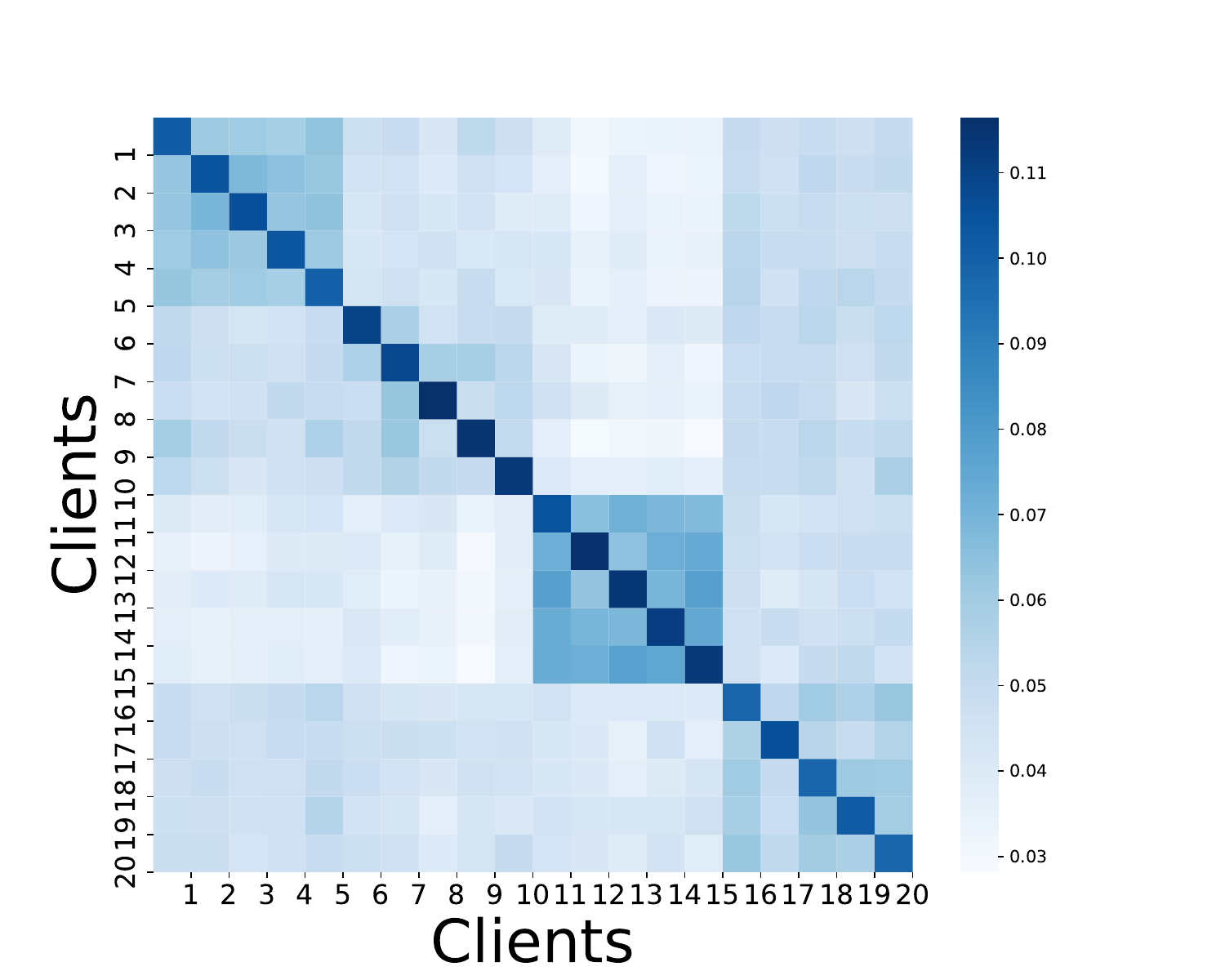}
        \caption{the second independent run of the 1st latent factor ($K=2$)}
        \label{fig_c_3_2}
    \end{subfigure}
    \hfill
    \begin{subfigure}{0.23\textwidth}
        \includegraphics[width=\linewidth]{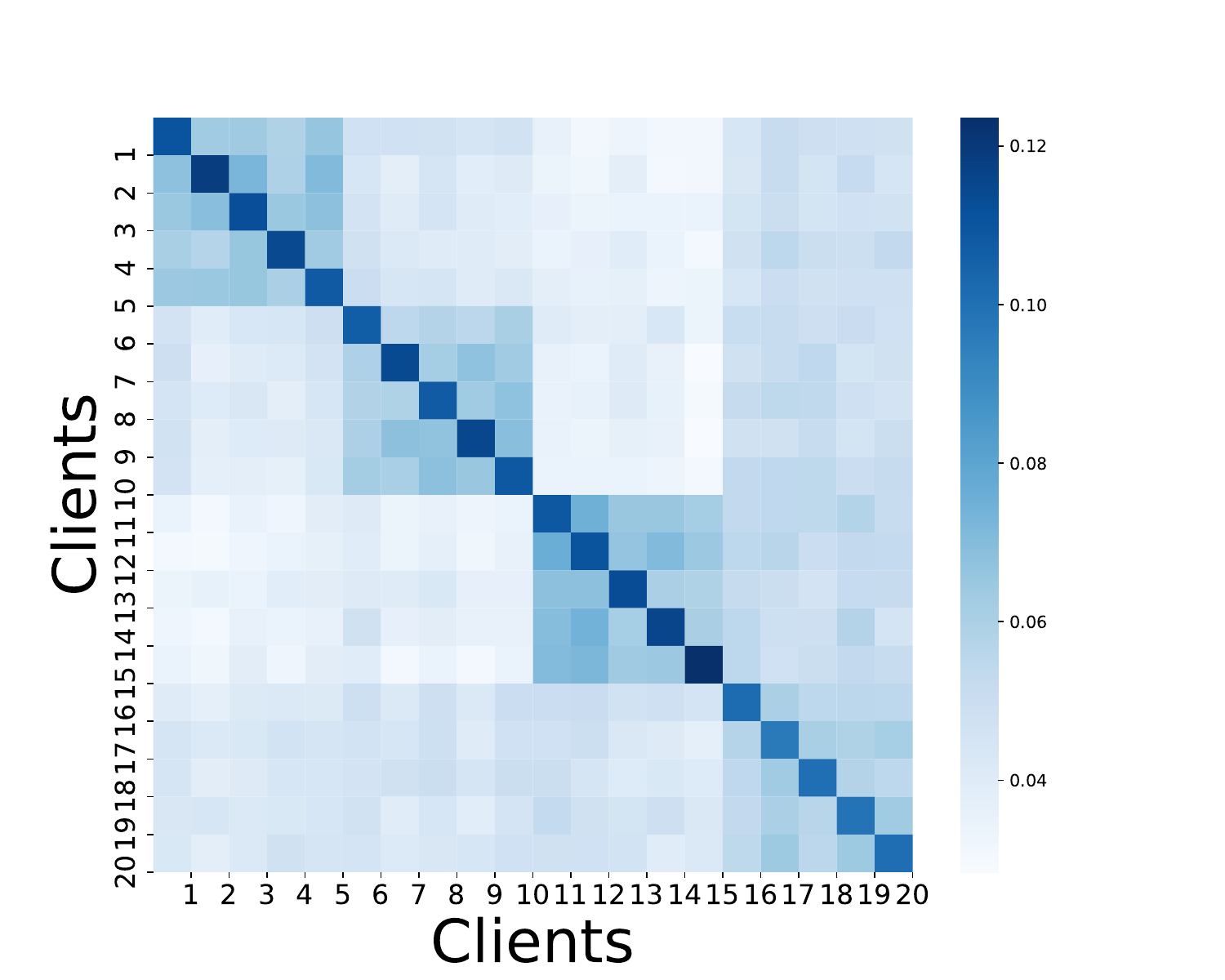}
        \caption{the third independent run of the 1st latent factor ($K=2$)}
        \label{fig_c_3_3}
    \end{subfigure}
    \hfill

    \begin{subfigure}{0.23\textwidth}
        \includegraphics[width=\linewidth]{fig/similarity_heatmaps/fediih_1_amazon_ratings_latent2.pdf}
        \caption{the first independent run of the 2nd latent factor ($K=2$)}
        \label{fig_c_3_4}
    \end{subfigure}%
    \hfill
    \begin{subfigure}{0.23\textwidth}
        \includegraphics[width=\linewidth]{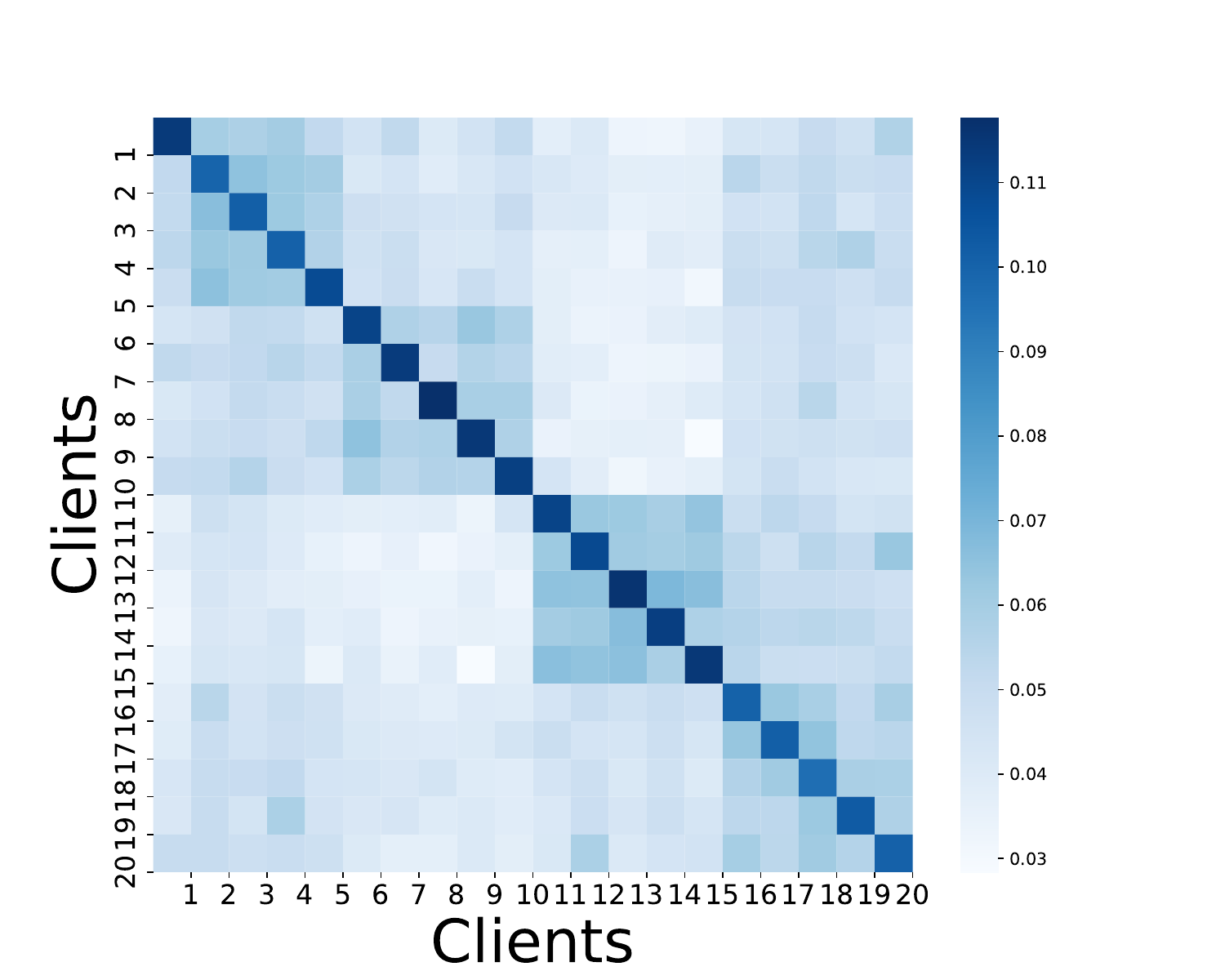}
        \caption{the second independent run of the 2nd latent factor ($K=2$)}
        \label{fig_c_3_5}
    \end{subfigure}
    \hfill
    \begin{subfigure}{0.23\textwidth}
        \includegraphics[width=\linewidth]{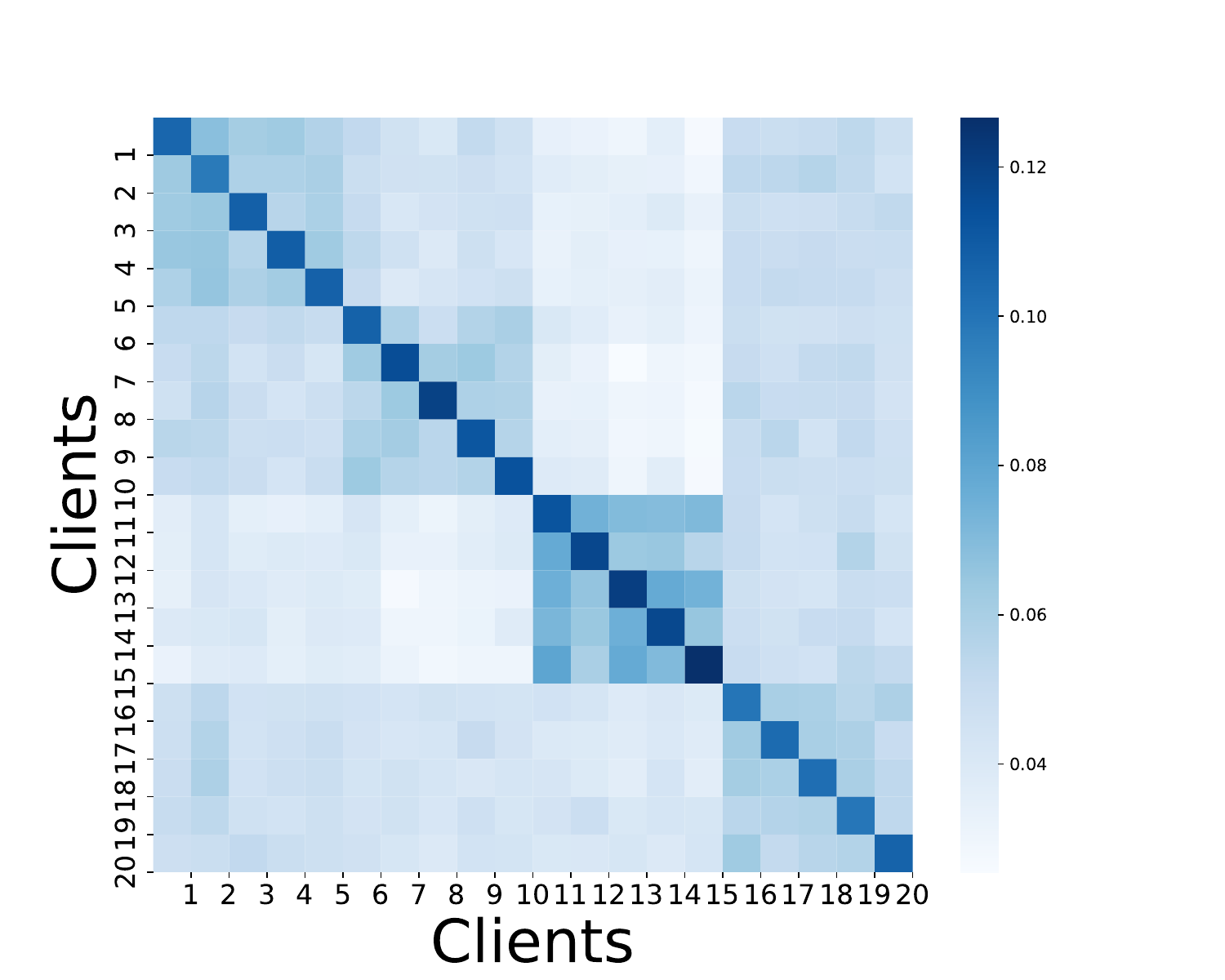}
        \caption{the third independent run of the 2nd latent factor ($K=2$)}
        \label{fig_c_3_6}
    \end{subfigure}
    \hfill
    \caption{The similarity heatmaps of FedIIH over three independent runs on the \textit{Amazon-ratings} dataset in the overlapping setting with 20 clients.}
    \label{fig_c_3}
\end{figure}

As shown in Fig.~\ref{fig_A_3}, Fig.~\ref{fig_b_3}, Fig.~\ref{fig_A_3_2}, and Fig.~\ref{fig_c_3}, we present the similarity heatmaps of FED-PUB and our FedIIH over three independent runs on the \textit{Cora} and \textit{Amazon-ratings} datasets in the overlapping setting with 20 clients, respectively. We can find that our calculated similarities are fairly much more stable than the similarities calculated by FED-PUB. This is because FED-PUB estimates the similarities between subgraphs based on the outputs of local models given the same random test graph. Since the random test graph varies over three independent runs, the outputs of the local models also change. In contrast, our FedIIH successfully infers the whole distribution of subgraph data in a multi-level global perspective, such that we can stably characterize the inter-subgraph similarities. Note that the similarity ground truth of the \textit{Cora} and \textit{Amazon-ratings} datasets in the overlapping setting with 20 clients are presented in Fig.~\ref{fig_s_1} and Fig.~3a, respectively.

\subsection{K.2 Similarity Heatmaps on Other Datasets}
\begin{figure}[t]
    \centering
    \begin{subfigure}[t]{0.15\textwidth}
        \includegraphics[width=\linewidth]{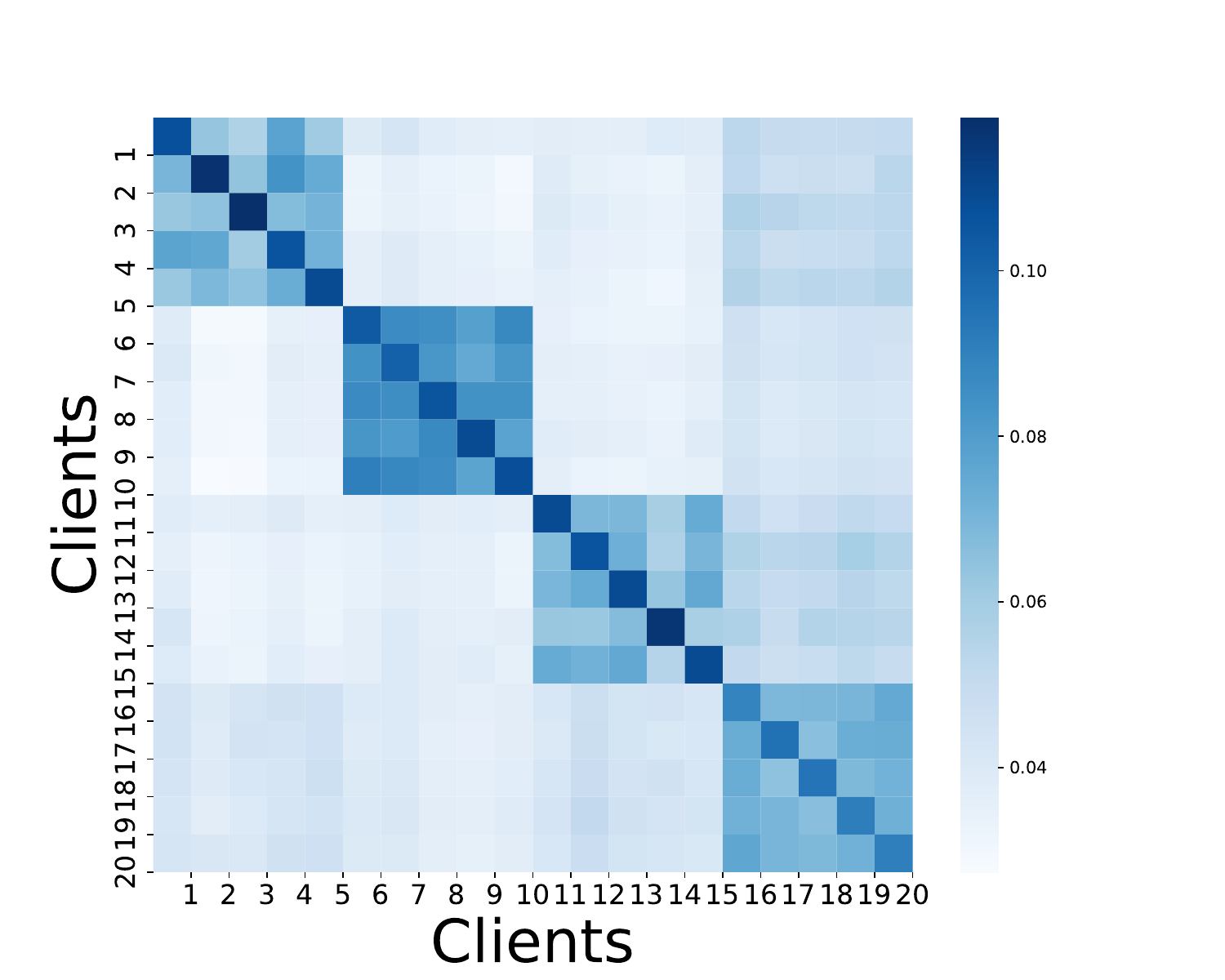}
        \caption{Distr. Sim.}
        \label{fig_s_1}
    \end{subfigure}%
    \hfill
    \begin{subfigure}[t]{0.15\textwidth}
        \includegraphics[width=\linewidth]{fig/similarity_heatmaps/fedpub_4.pdf}
        \caption{FED-PUB}
        \label{fig_s_2}
    \end{subfigure}%
    \hfill
    \begin{subfigure}[t]{0.15\textwidth}
        \includegraphics[width=\linewidth]{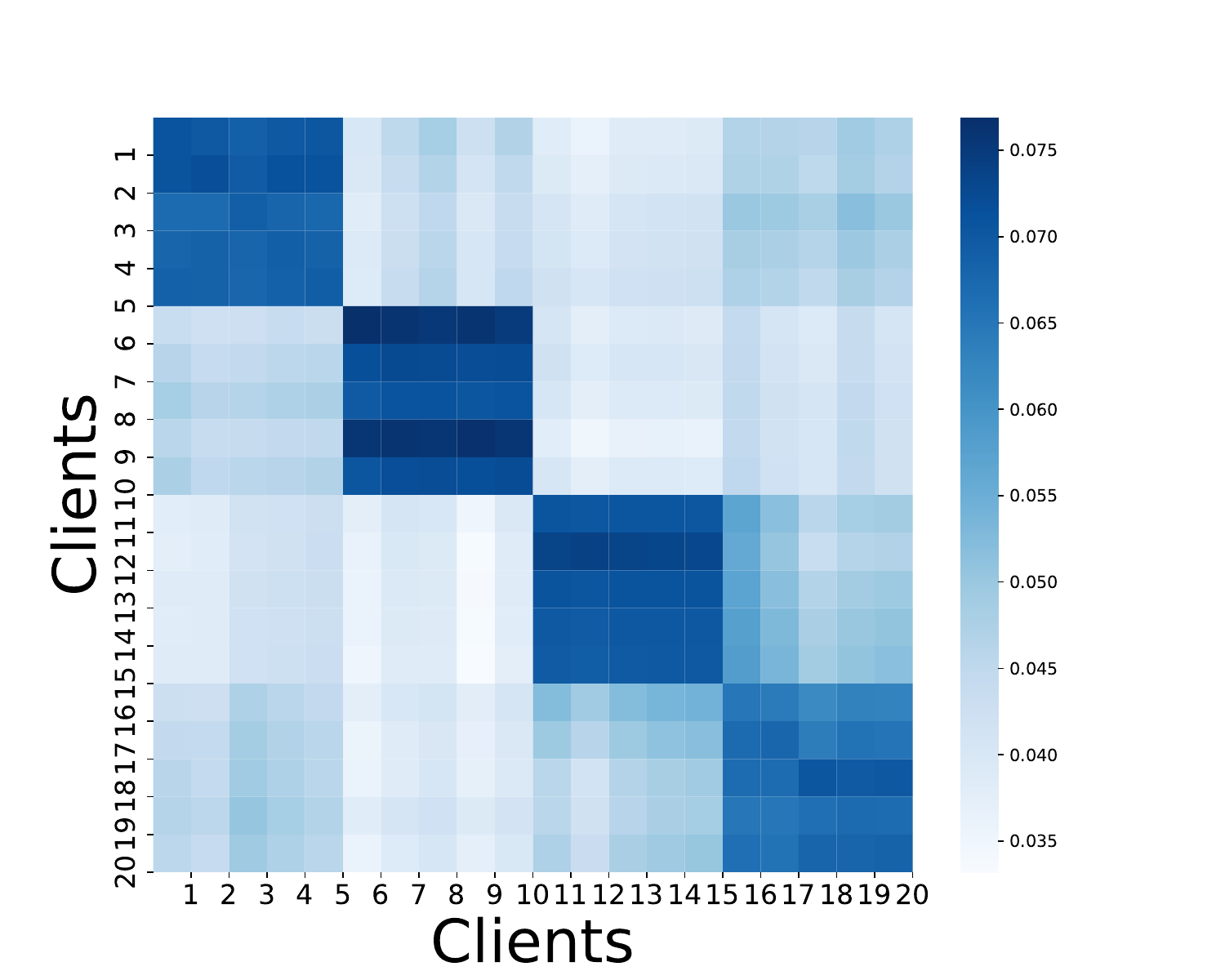}
        \caption{FedGTA}
        \label{fig_s_3}
    \end{subfigure}%
    \hfill
    \begin{subfigure}[t]{0.15\textwidth}
        \includegraphics[width=\linewidth]{fig/similarity_heatmaps/fediih_2_latent1.pdf}
        \caption{FedIIH of the 1st latent factor ($K=2$)}
        \label{fig_s_4}
    \end{subfigure}
    \hfill
    \begin{subfigure}[t]{0.15\textwidth}
        \includegraphics[width=\linewidth]{fig/similarity_heatmaps/fediih_2_latent2.pdf}
        \caption{FedIIH of the 2nd latent factor ($K=2$)}
        \label{fig_s_5}
    \end{subfigure}
    \caption{Similarity heatmaps on the \textit{Cora} dataset in the overlapping setting with 20 clients.}
    \label{fig_s}
\end{figure}

\begin{figure}
    \centering
    \begin{subfigure}[t]{0.15\textwidth}
        \includegraphics[width=\linewidth]{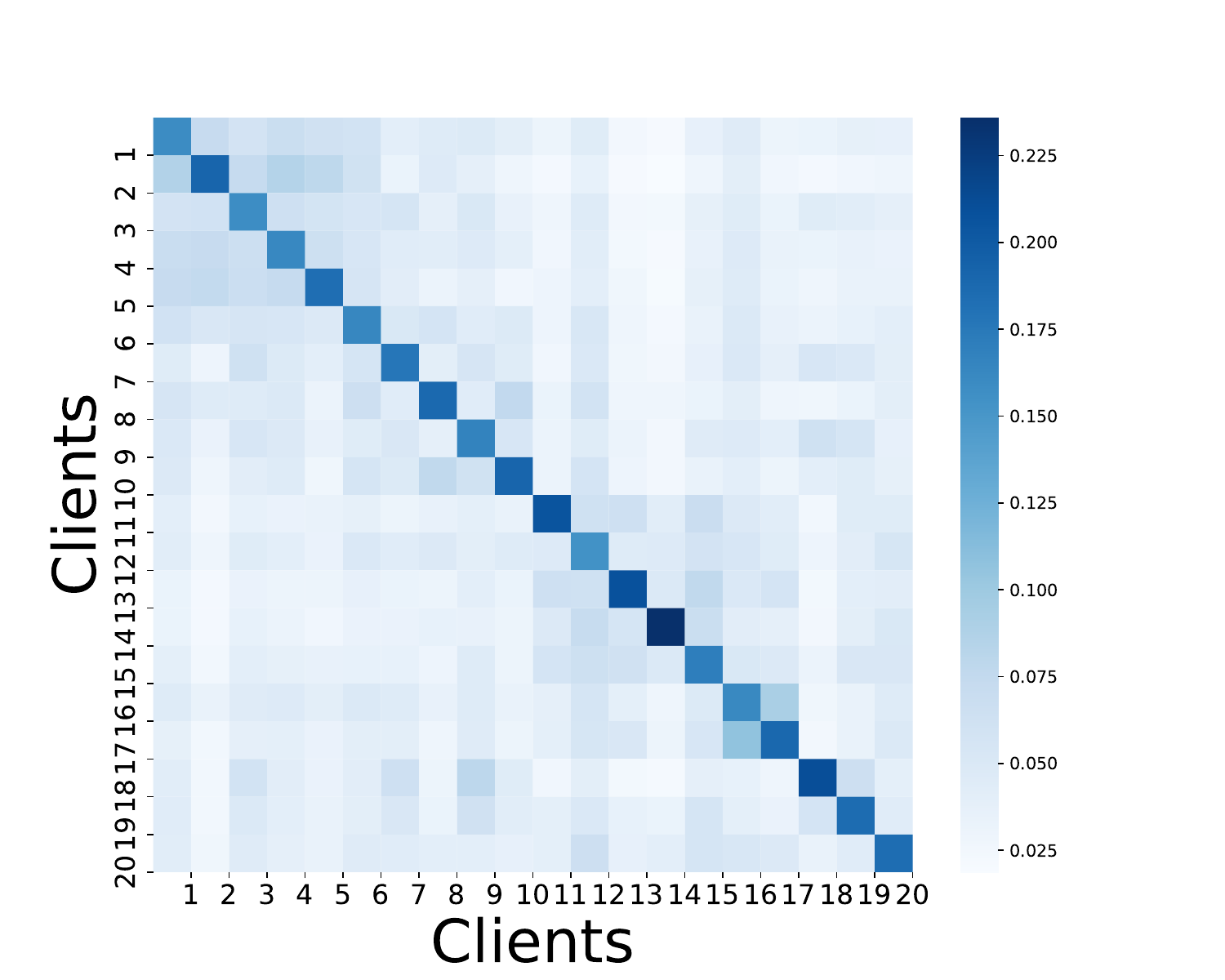}
        \caption{Distr. Sim.}
        \label{fig_CiteSeer_D1}
    \end{subfigure}%
    \hfill
    \begin{subfigure}[t]{0.15\textwidth}
        \includegraphics[width=\linewidth]{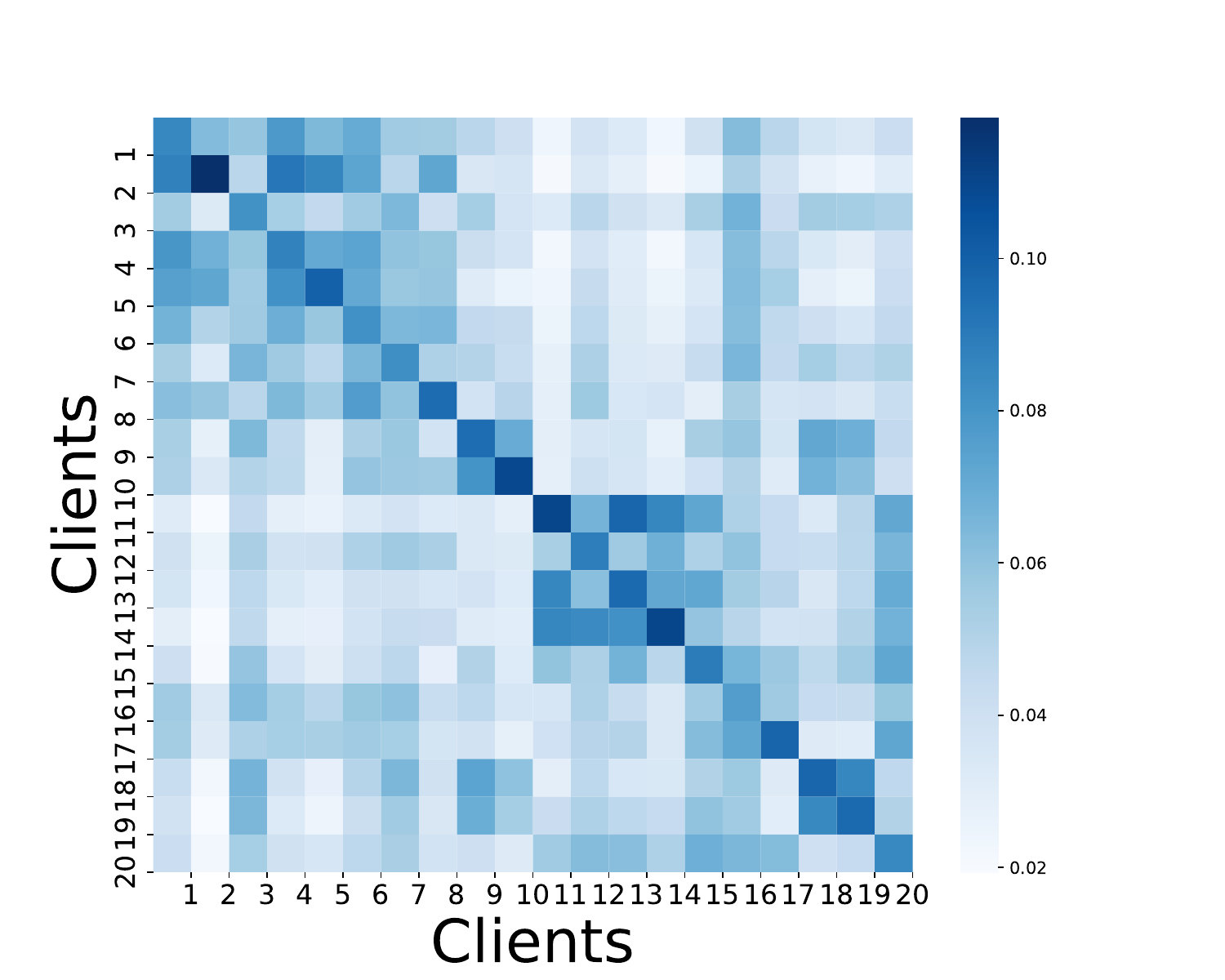}
        \caption{FED-PUB}
        \label{fig_CiteSeer_D2}
    \end{subfigure}%
    \hfill
    \begin{subfigure}[t]{0.15\textwidth}
        \includegraphics[width=\linewidth]{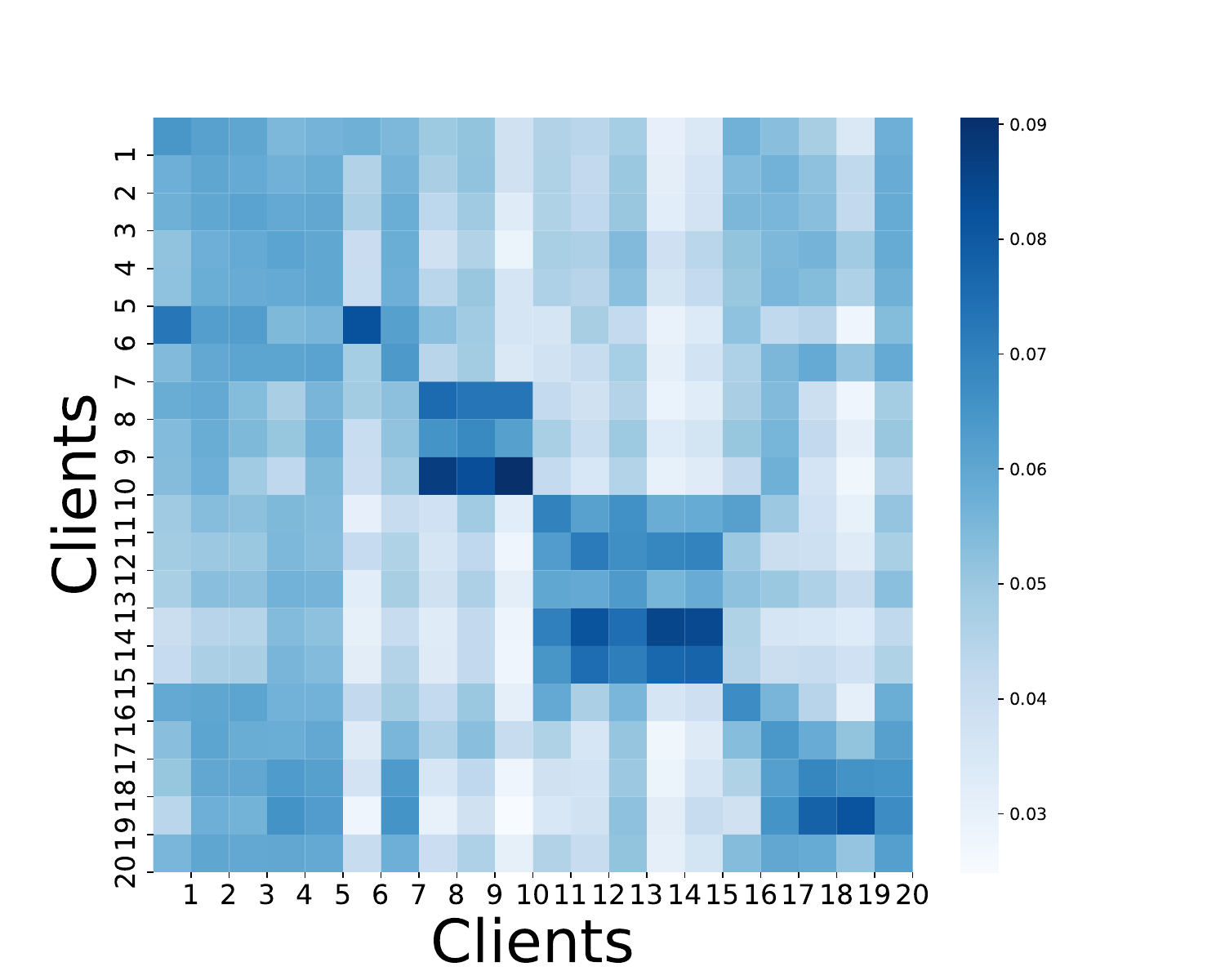}
        \caption{FedGTA}
        \label{fig_CiteSeer_D3}
    \end{subfigure}%
    \hfill
    \begin{subfigure}[t]{0.15\textwidth}
        \includegraphics[width=\linewidth]{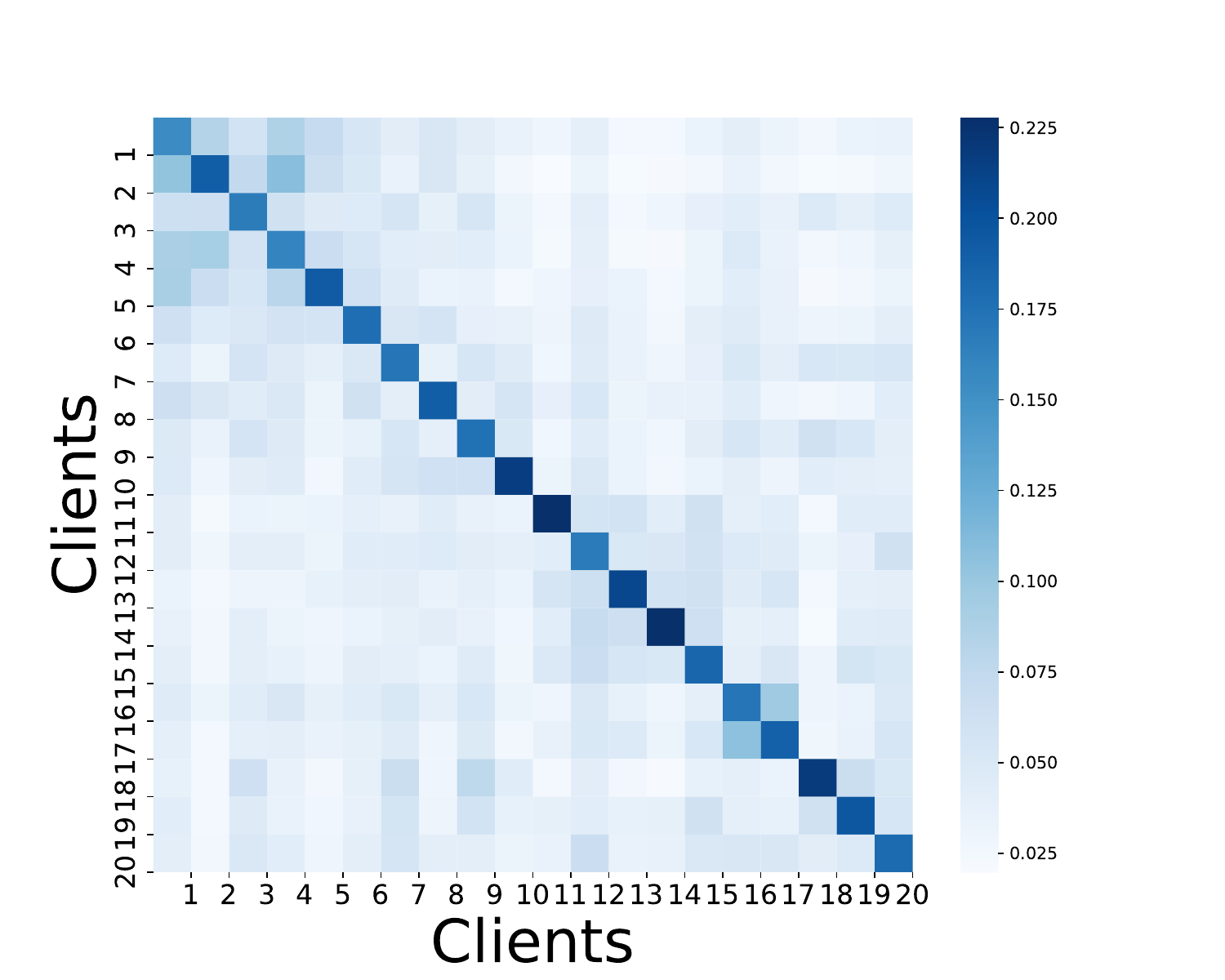}
        \caption{FedIIH of the 1st latent factor ($K=2$)}
        \label{fig_CiteSeer_D4}
    \end{subfigure}
    \hfill
    \begin{subfigure}[t]{0.15\textwidth}
        \includegraphics[width=\linewidth]{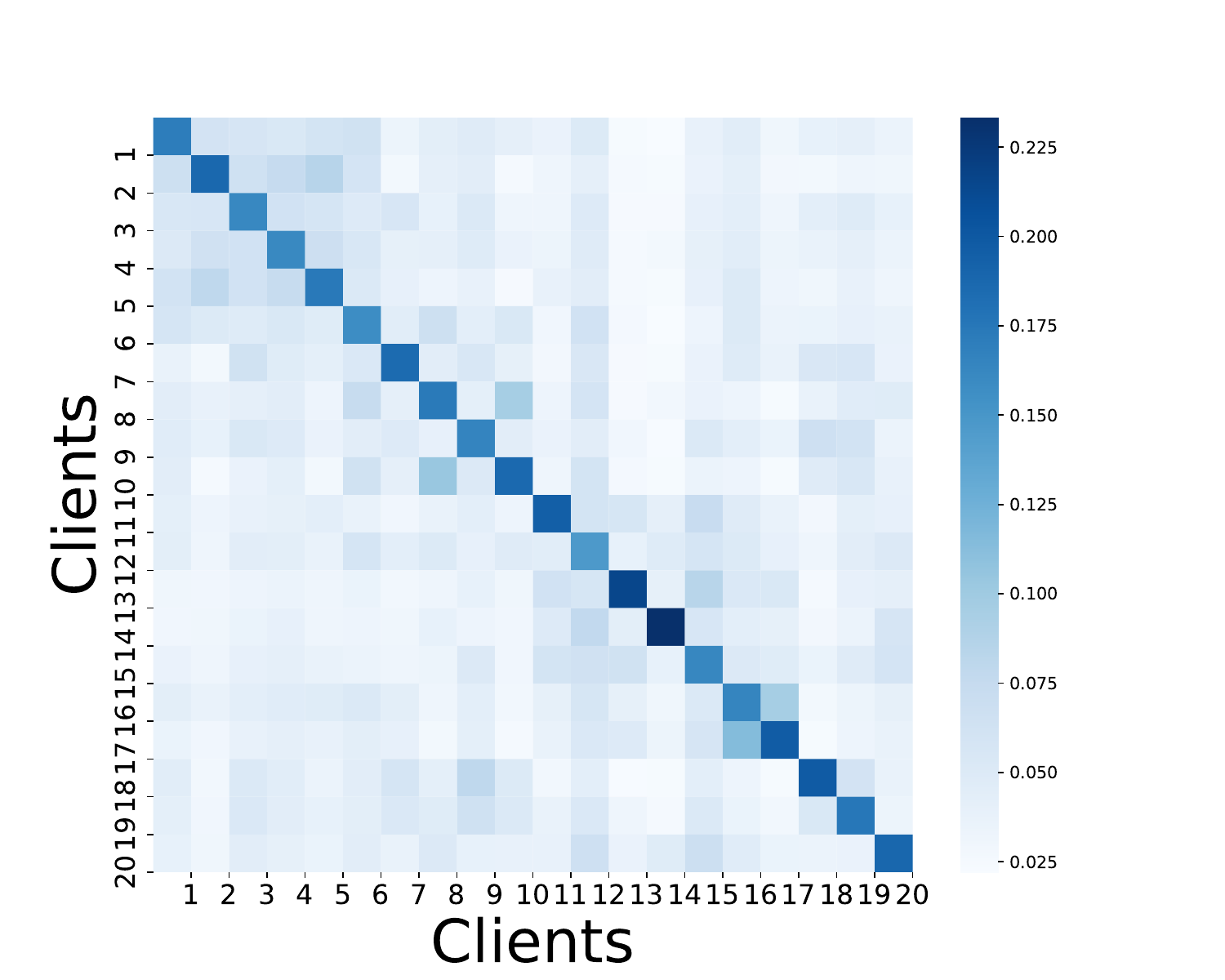}
        \caption{FedIIH of the 2nd latent factor ($K=2$)}
        \label{fig_CiteSeer_D5}
    \end{subfigure}
    \caption{Similarity heatmaps on the \textit{CiteSeer} dataset in the non-overlapping setting with 20 clients.}
    \label{fig_CiteSeer_D}
\end{figure}

The similarity heatmaps on other datasets are shown in Fig.~\ref{fig_s}, Fig.~\ref{fig_CiteSeer_D}, ..., to Fig.~\ref{fig_Questions_O}. We present the similarity heatmaps for each dataset (except the \textit{Cora} dataset) in two settings, namely, non-overlapping with 20 clients and overlapping with 30 clients. However, for the \textit{Cora} dataset, we present the similarity heatmaps in the overlapping setting with 20 clients. This is because in~\cite{baek2023personalized}, Baek~\textit{et al.} present the heatmaps on the \textit{Cora} dataset in the overlapping setting with 20 clients, and we specifically want to be consistent with that here. According to the experimental results on these datasets, we can find that the similarity heatmaps of our FedIIH are always much closer to the ground truth than FED-PUB and FedGTA, verifying the effectiveness of our similarity calculation scheme based on the inferred subgraph data distribution.

\label{more_similarity_heatmaps}

\begin{figure}[t]
    \centering
    \begin{subfigure}[t]{0.15\textwidth}
        \includegraphics[width=\linewidth]{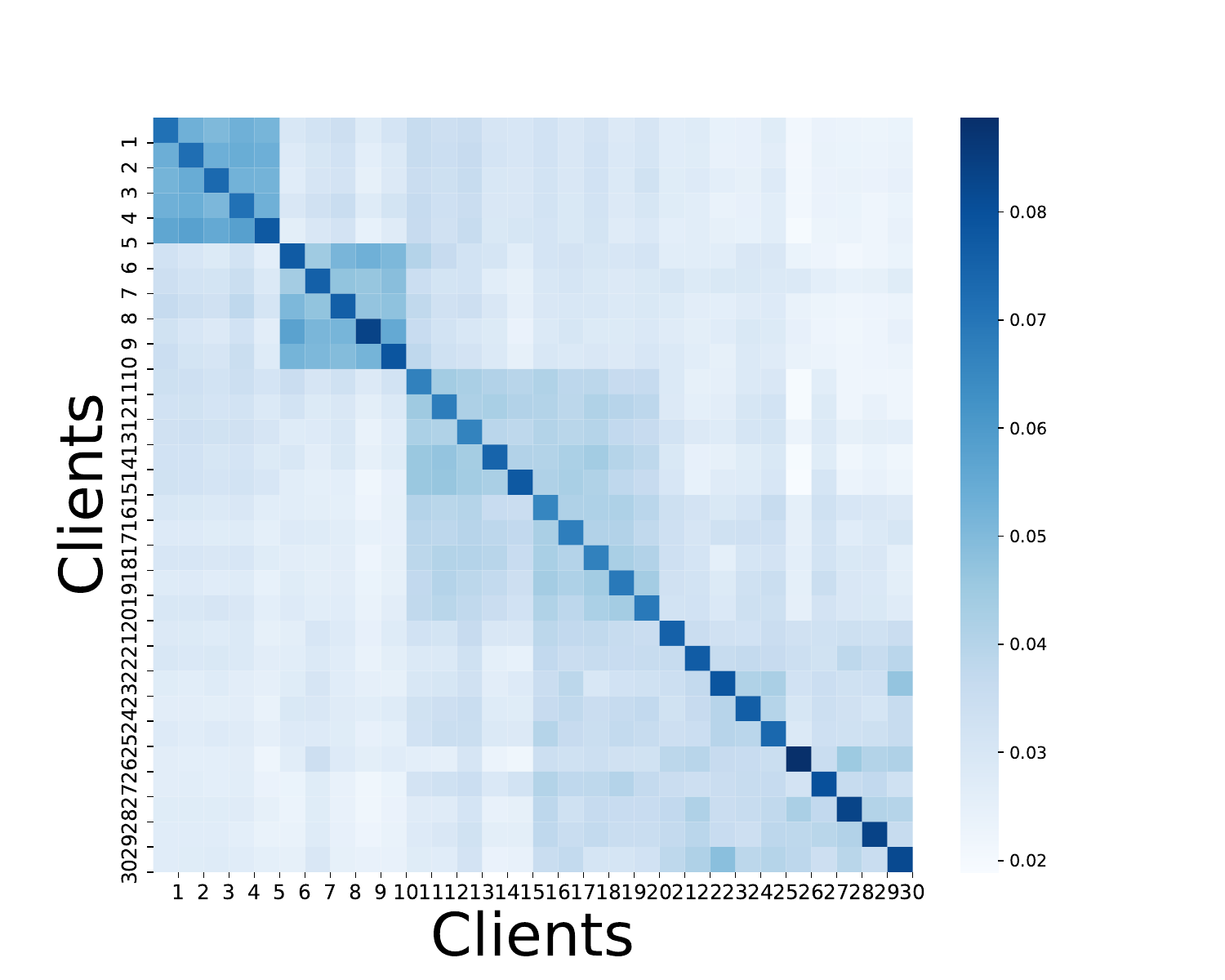}
        \caption{Distr. Sim.}
        \label{fig_CiteSeer_01}
    \end{subfigure}%
    \hfill
    \begin{subfigure}[t]{0.15\textwidth}
        \includegraphics[width=\linewidth]{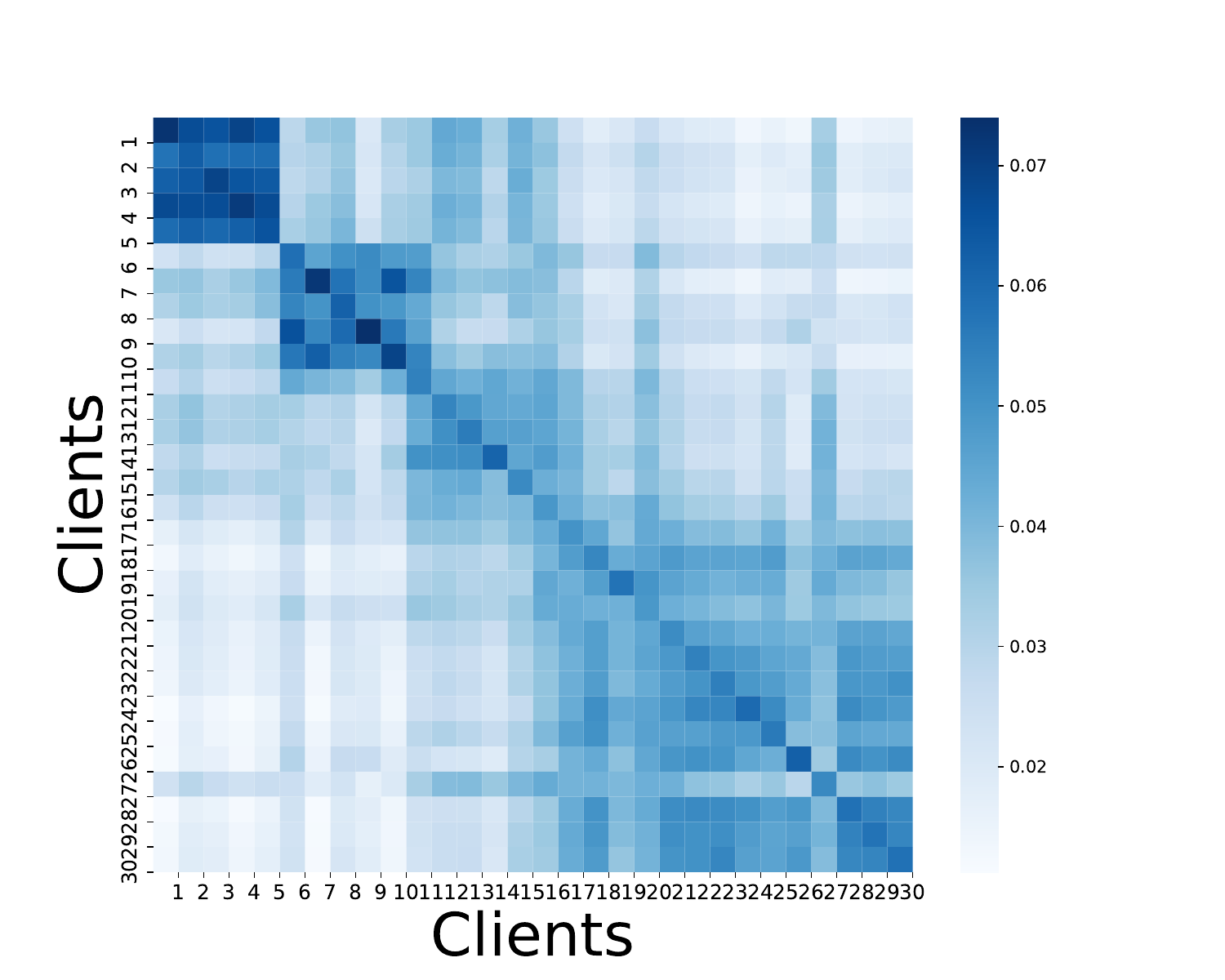}
        \caption{FED-PUB}
        \label{fig_CiteSeer_02}
    \end{subfigure}%
    \hfill
    \begin{subfigure}[t]{0.15\textwidth}
        \includegraphics[width=\linewidth]{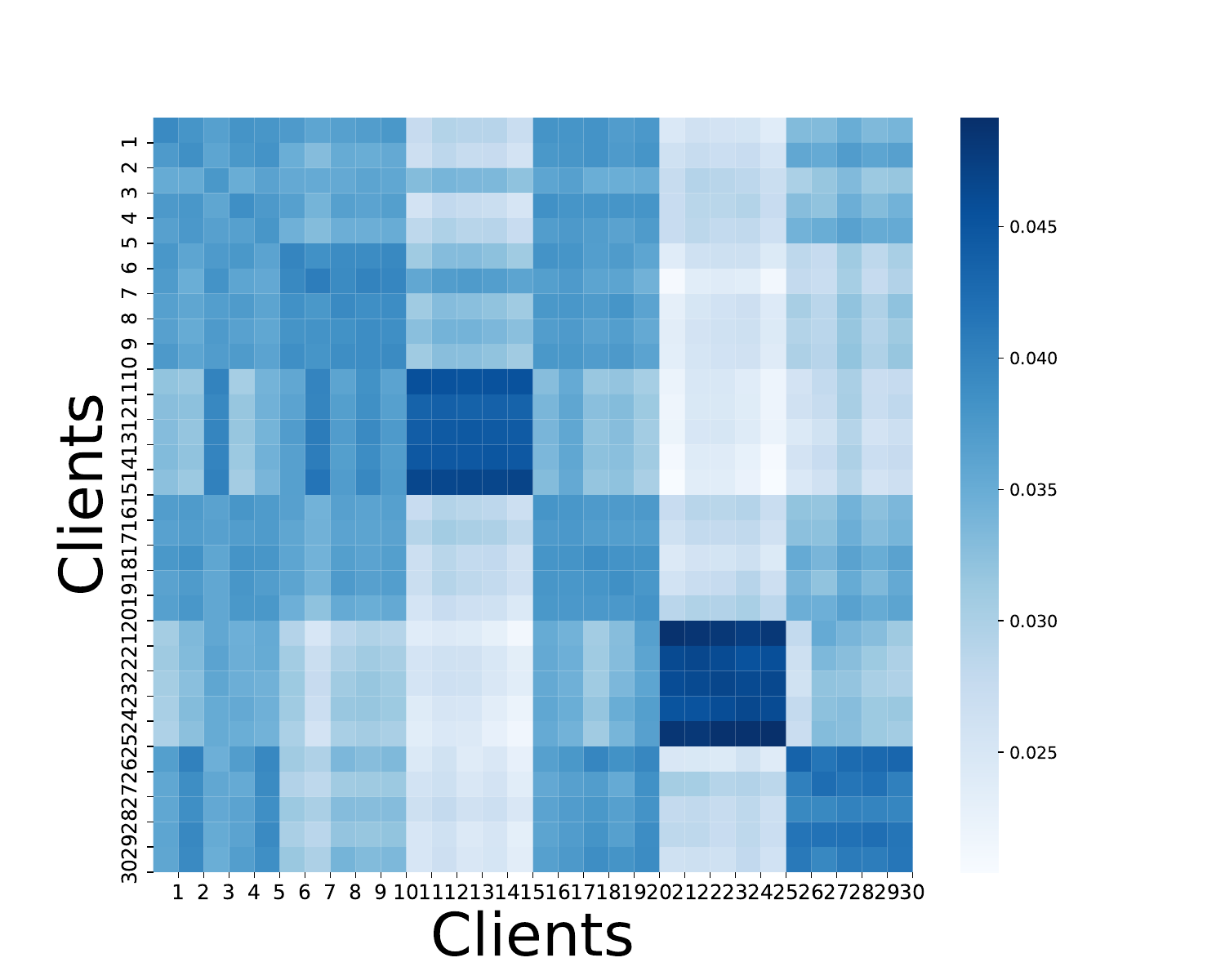}
        \caption{FedGTA}
        \label{fig_CiteSeer_03}
    \end{subfigure}%
    \hfill
    \begin{subfigure}[t]{0.15\textwidth}
        \includegraphics[width=\linewidth]{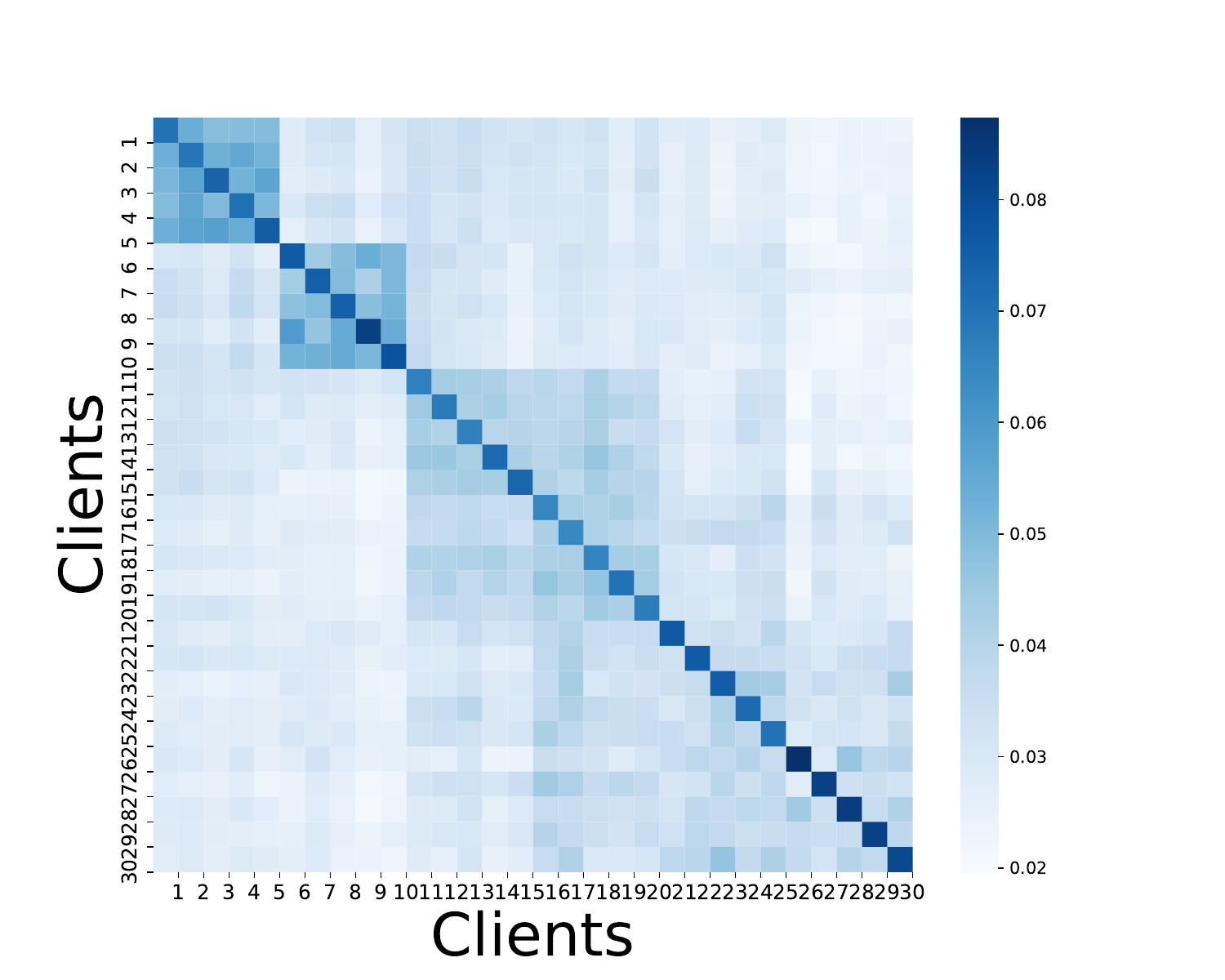}
        \caption{FedIIH of the 1st latent factor ($K=2$)}
        \label{fig_CiteSeer_04}
    \end{subfigure}
    \hfill
    \begin{subfigure}[t]{0.15\textwidth}
        \includegraphics[width=\linewidth]{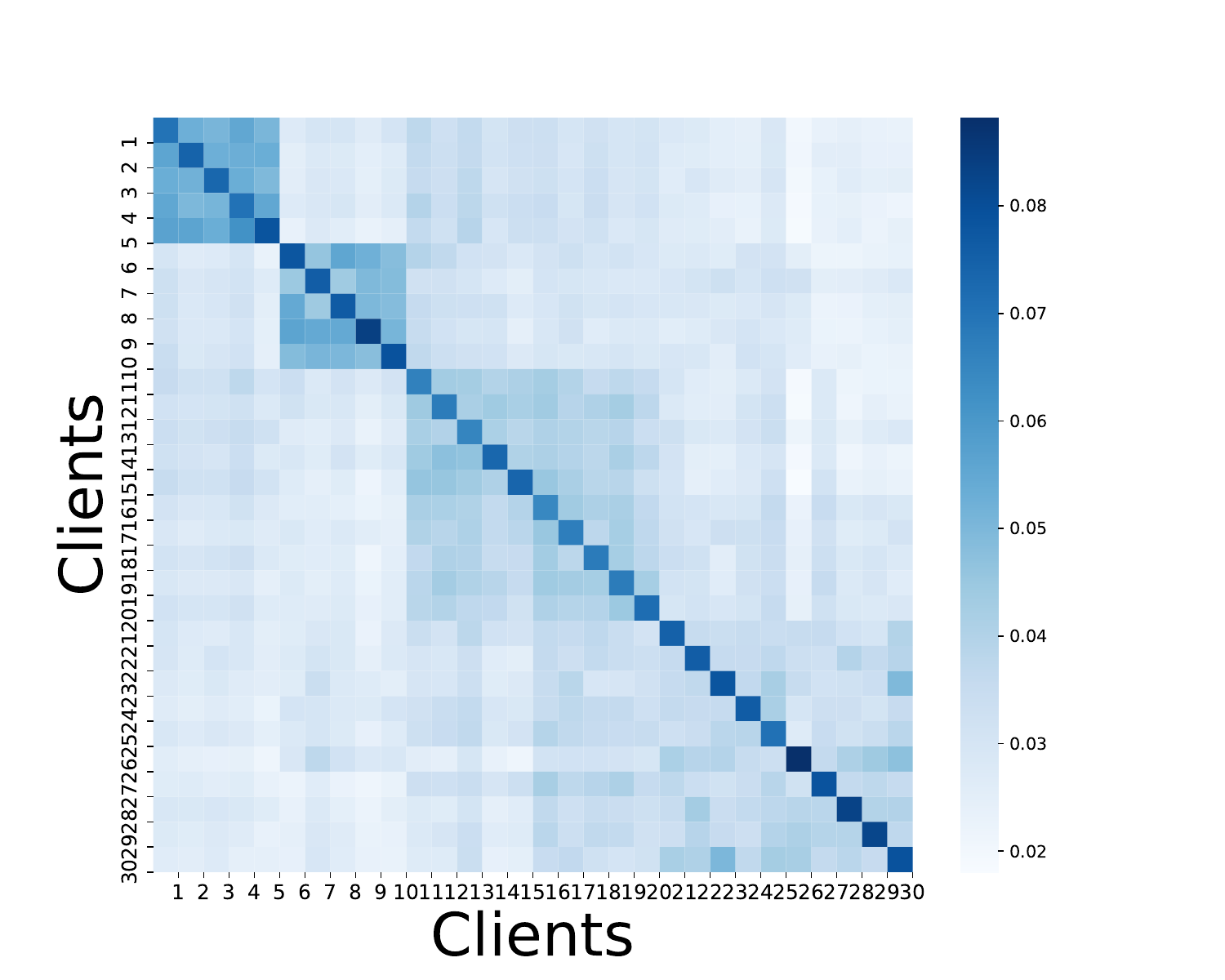}
        \caption{FedIIH of the 2nd latent factor ($K=2$)}
        \label{fig_CiteSeer_05}
    \end{subfigure}
    \caption{Similarity heatmaps on the \textit{CiteSeer} dataset in the overlapping setting with 30 clients.}
    \label{fig_CiteSeer_O}
\end{figure}

\begin{figure}[t]
    \centering
    \begin{subfigure}[t]{0.15\textwidth}
        \includegraphics[width=\linewidth]{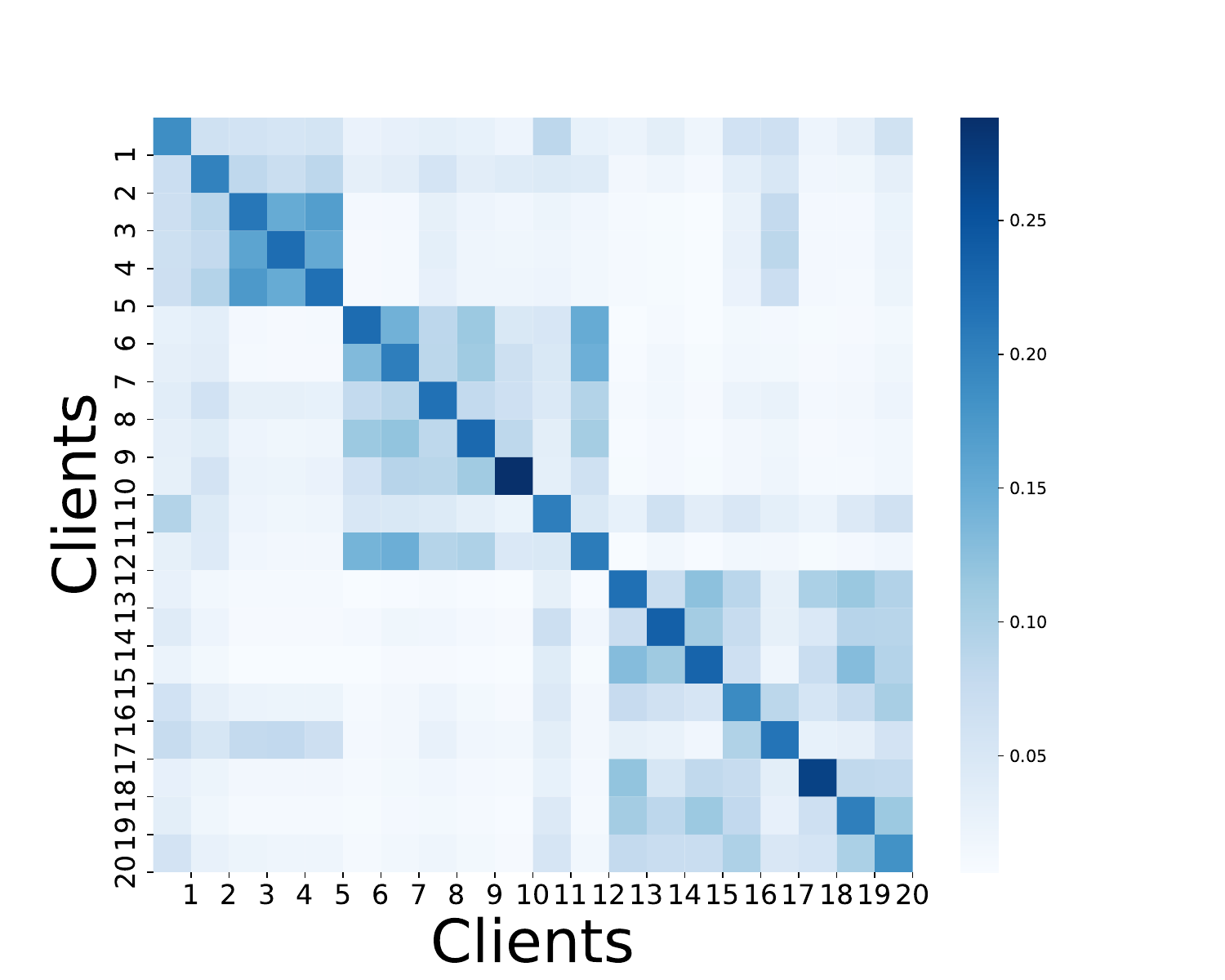}
        \caption{Distr. Sim.}
        \label{fig_PubMed_D1}
    \end{subfigure}%
    \hfill
    \begin{subfigure}[t]{0.15\textwidth}
        \includegraphics[width=\linewidth]{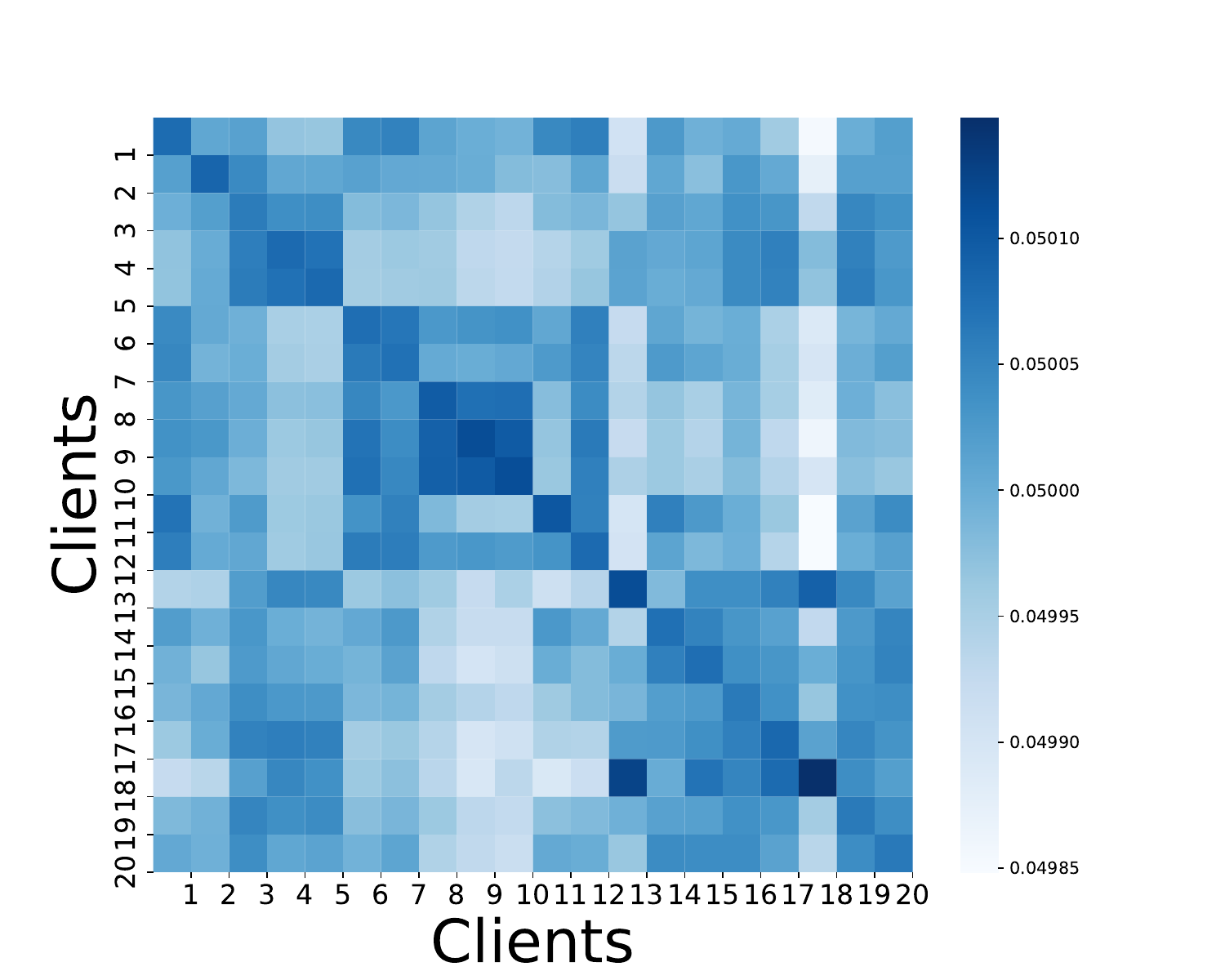}
        \caption{FED-PUB}
        \label{fig_PubMed_D2}
    \end{subfigure}%
    \hfill
    \begin{subfigure}[t]{0.15\textwidth}
        \includegraphics[width=\linewidth]{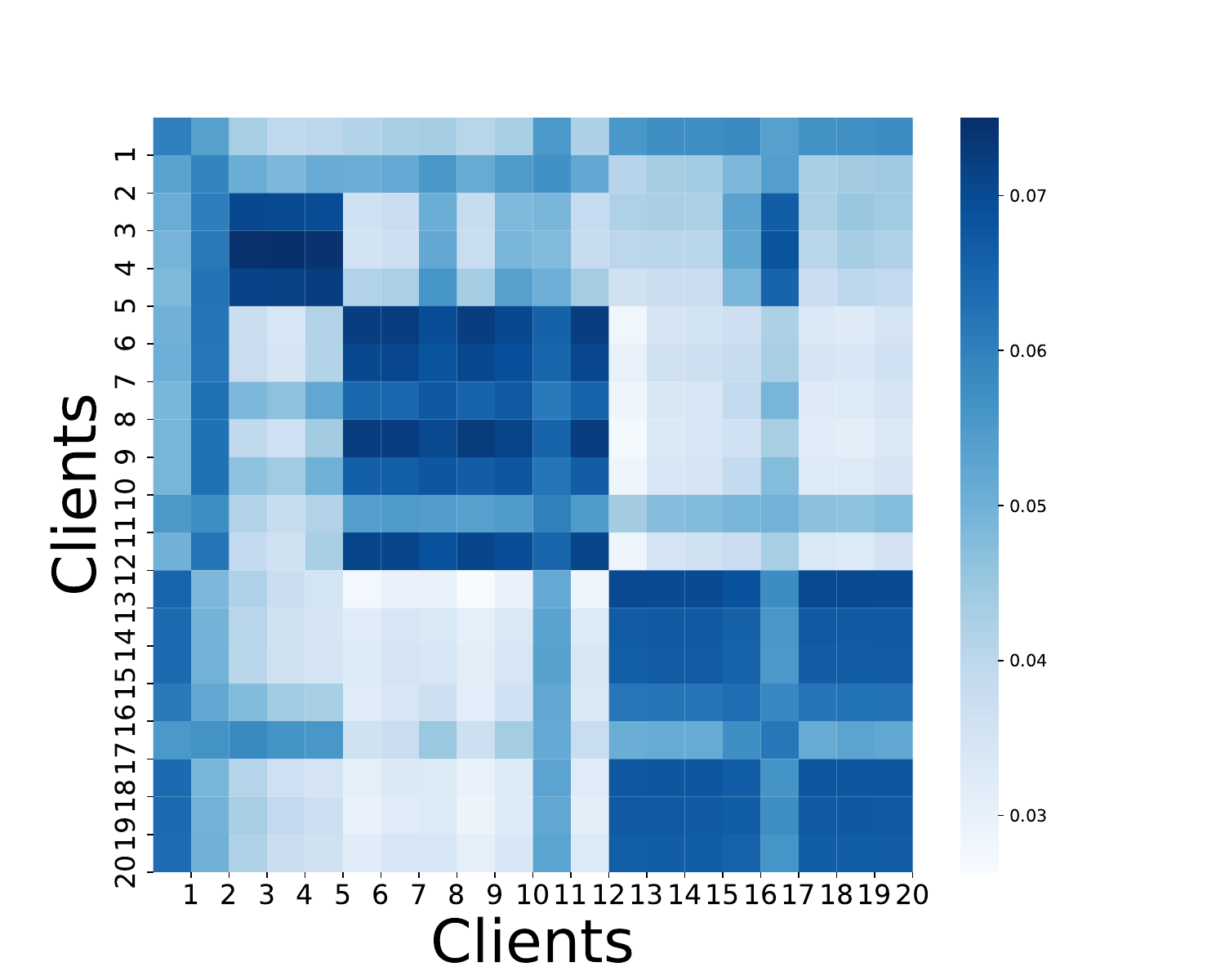}
        \caption{FedGTA}
        \label{fig_PubMed_D3}
    \end{subfigure}%
    \hfill
    \begin{subfigure}[t]{0.15\textwidth}
        \includegraphics[width=\linewidth]{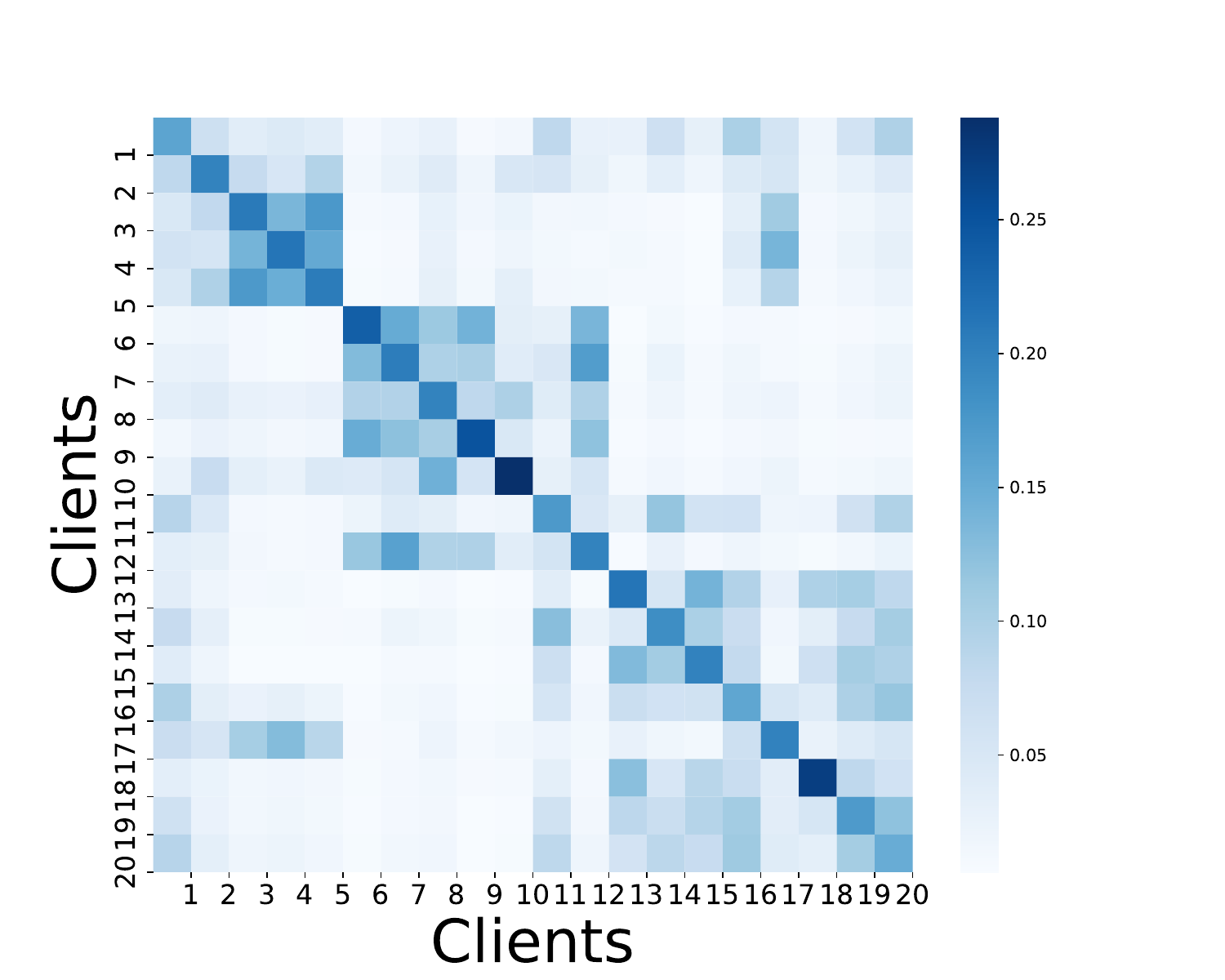}
        \caption{FedIIH of the 1st latent factor ($K=2$)}
        \label{fig_PubMed_D4}
    \end{subfigure}
    \hfill
    \begin{subfigure}[t]{0.15\textwidth}
        \includegraphics[width=\linewidth]{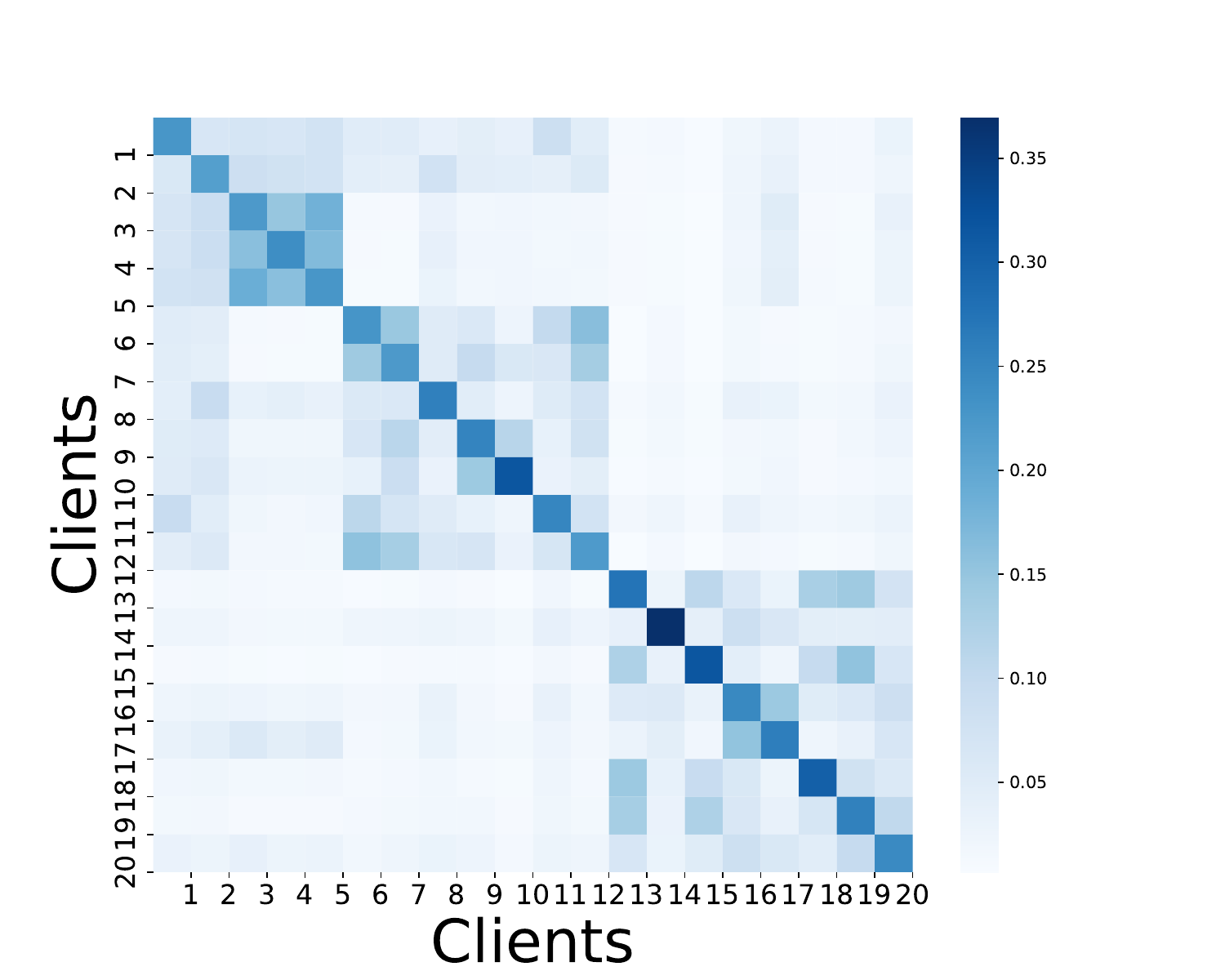}
        \caption{FedIIH of the 2nd latent factor ($K=2$)}
        \label{fig_PubMed_D5}
    \end{subfigure}
    \caption{Similarity heatmaps on the \textit{PubMed} dataset in the non-overlapping setting with 20 clients.}
    \label{fig_PubMed_D}
\end{figure}

\begin{figure}[t]
    \centering
    \begin{subfigure}[t]{0.15\textwidth}
        \includegraphics[width=\linewidth]{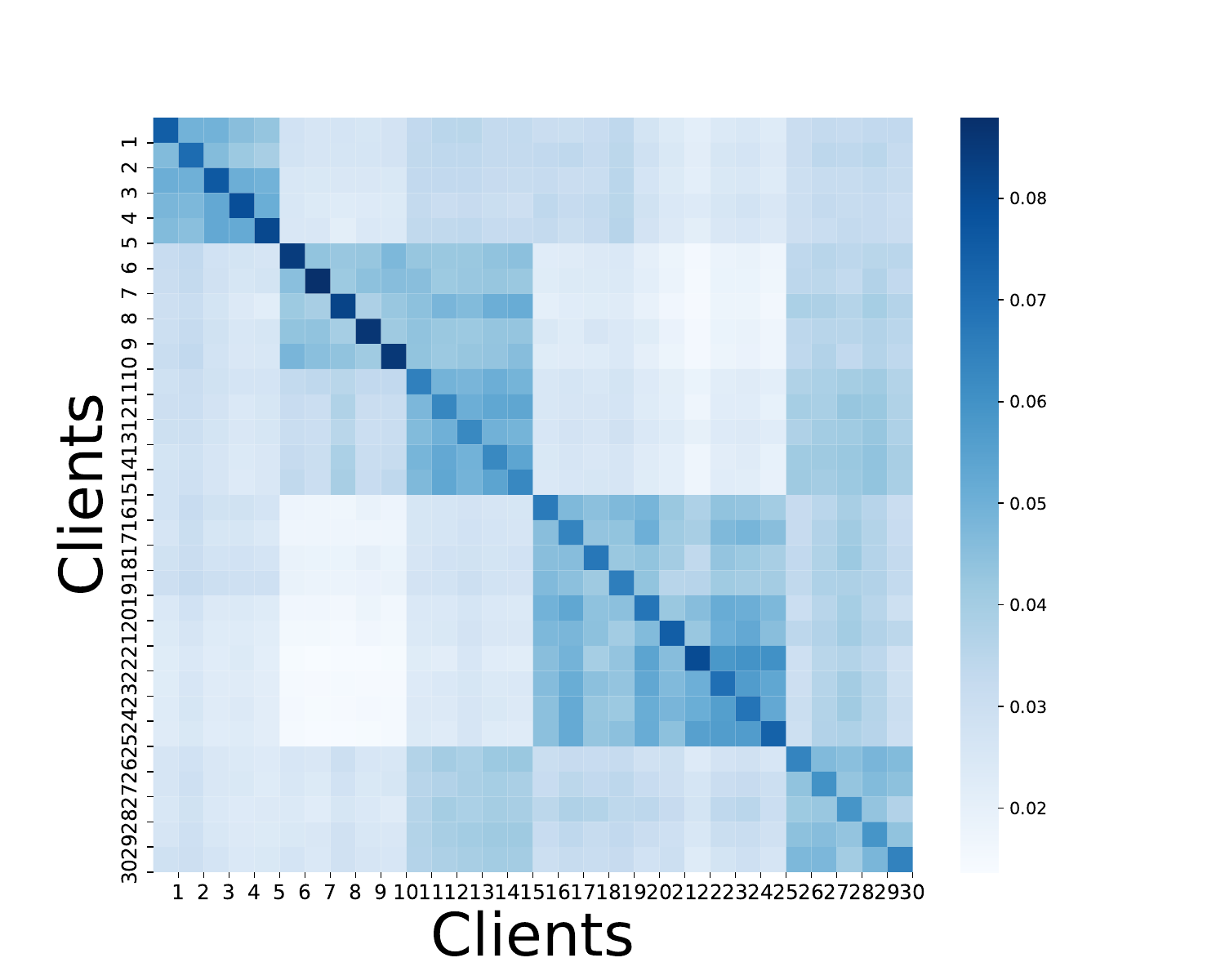}
        \caption{Distr. Sim.}
        \label{fig_PubMed_01}
    \end{subfigure}%
    \hfill
    \begin{subfigure}[t]{0.15\textwidth}
        \includegraphics[width=\linewidth]{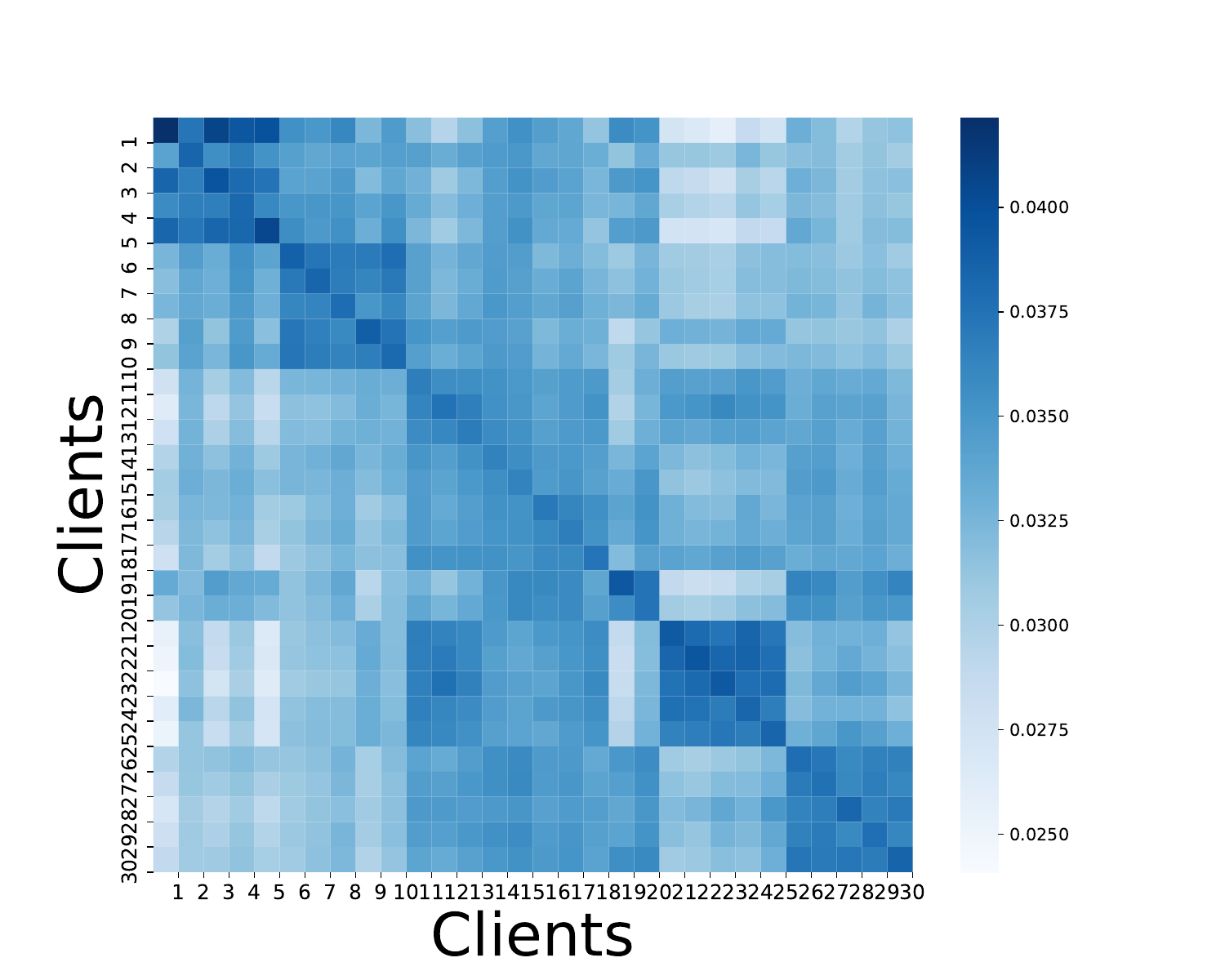}
        \caption{FED-PUB}
        \label{fig_PubMed_02}
    \end{subfigure}%
    \hfill
    \begin{subfigure}[t]{0.15\textwidth}
        \includegraphics[width=\linewidth]{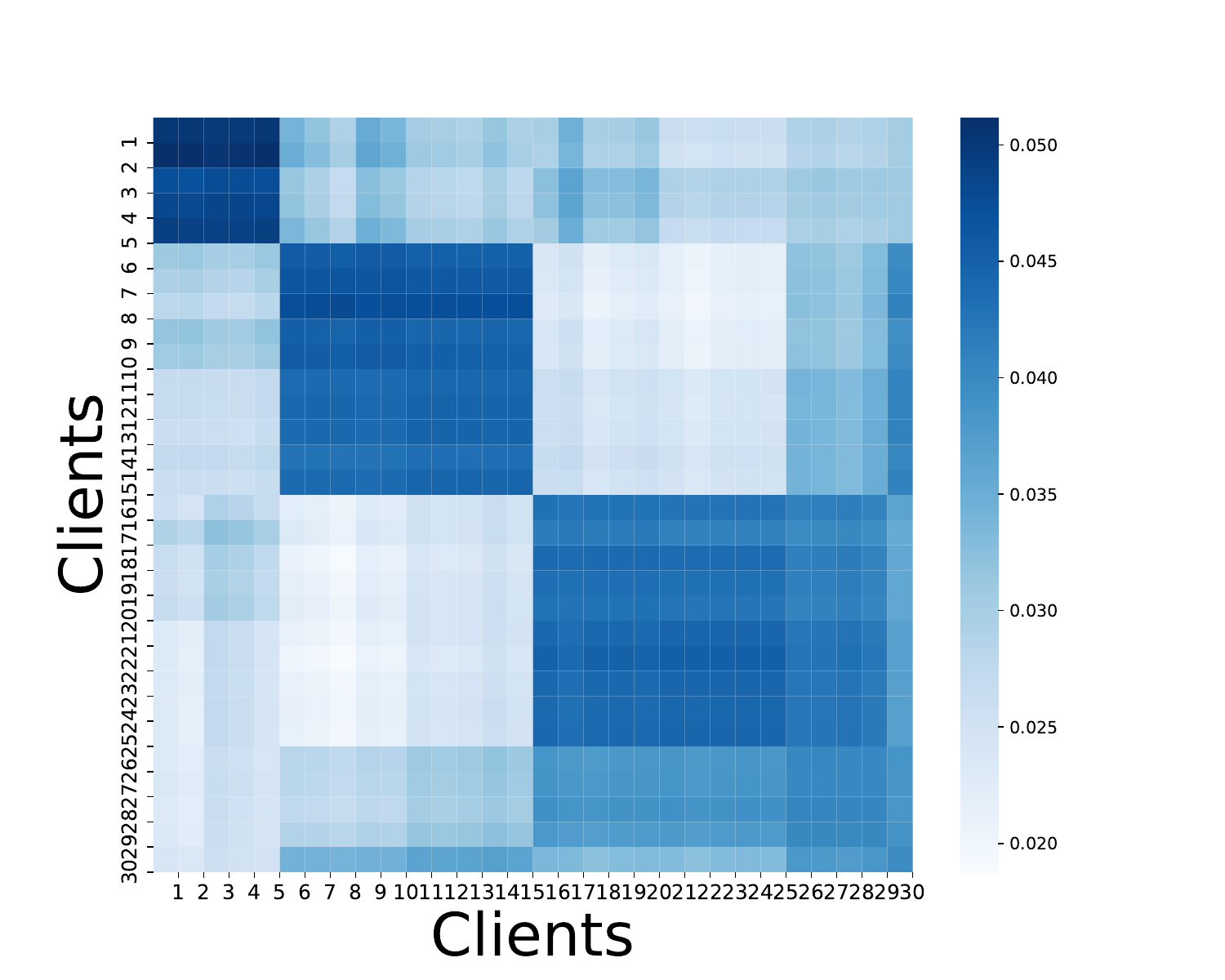}
        \caption{FedGTA}
        \label{fig_PubMed_03}
    \end{subfigure}%
    \hfill
    \begin{subfigure}[t]{0.15\textwidth}
        \includegraphics[width=\linewidth]{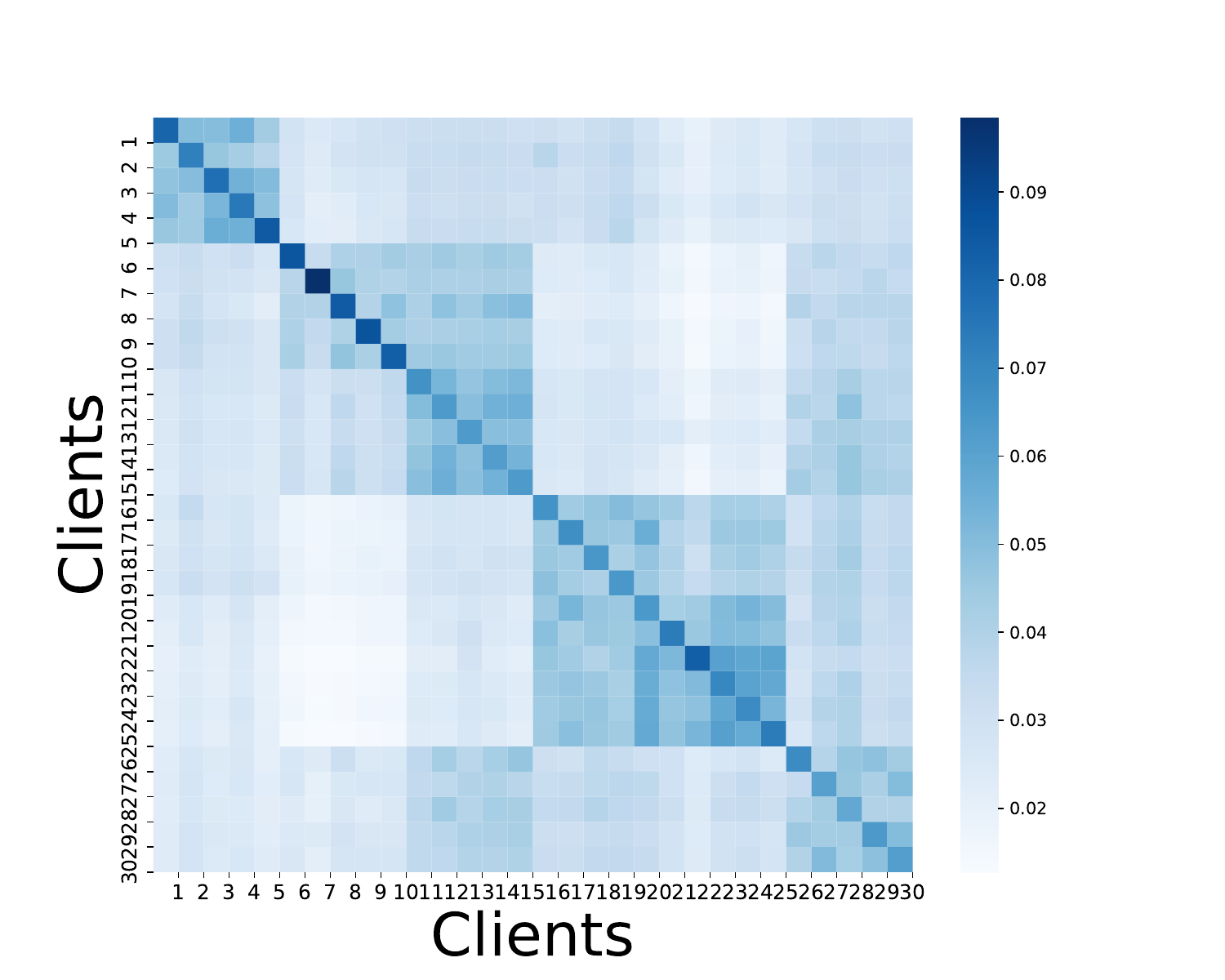}
        \caption{FedIIH of the 1st latent factor ($K=2$)}
        \label{fig_PubMed_04}
    \end{subfigure}
    \hfill
    \begin{subfigure}[t]{0.15\textwidth}
        \includegraphics[width=\linewidth]{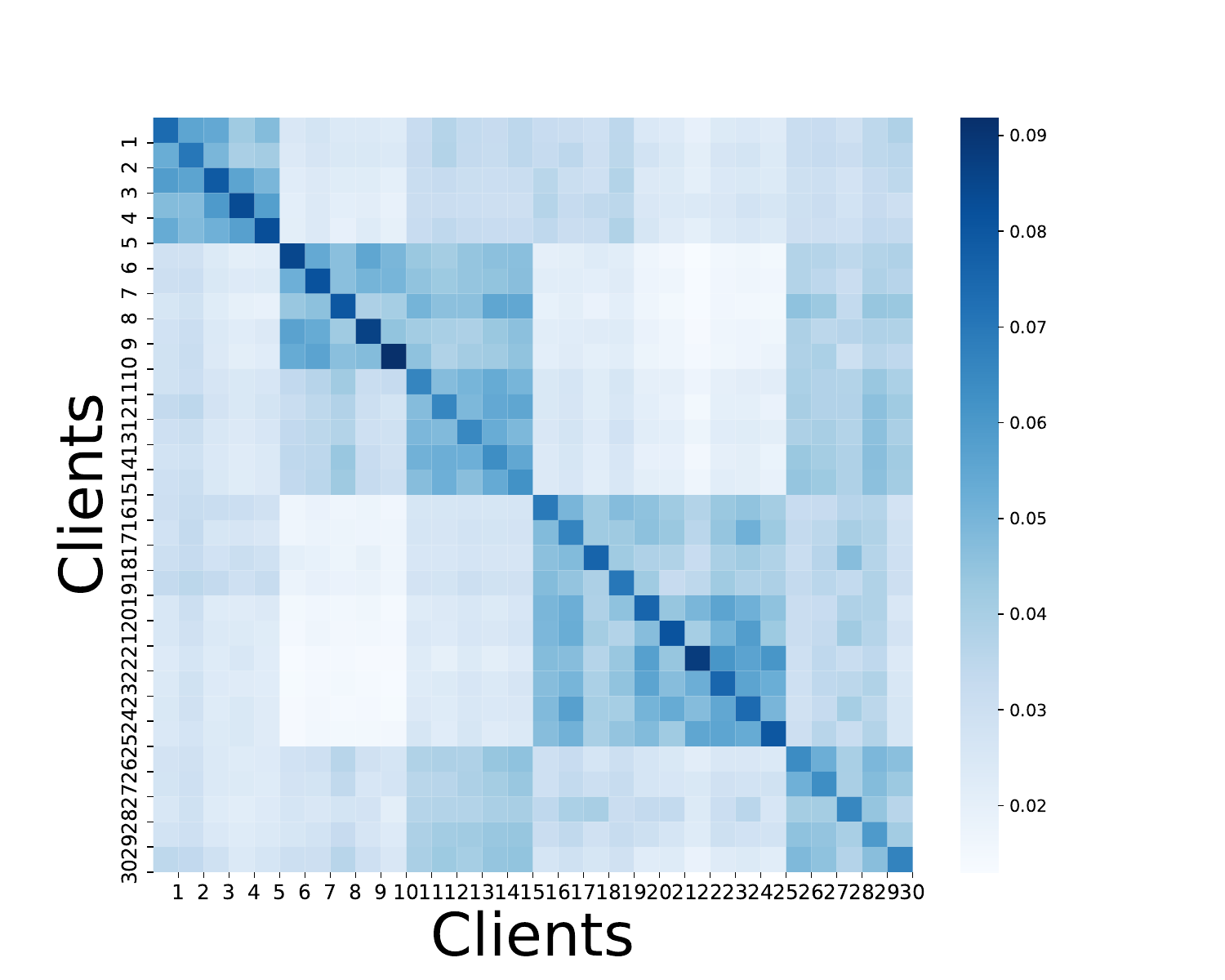}
        \caption{FedIIH of the 2nd latent factor ($K=2$)}
        \label{fig_PubMed_05}
    \end{subfigure}
    \caption{Similarity heatmaps on the \textit{PubMed} dataset in the overlapping setting with 30 clients.}
    \label{fig_PubMed_O}
\end{figure}

\begin{figure}[t]
    \centering
    \begin{subfigure}[t]{0.15\textwidth}
        \includegraphics[width=\linewidth]{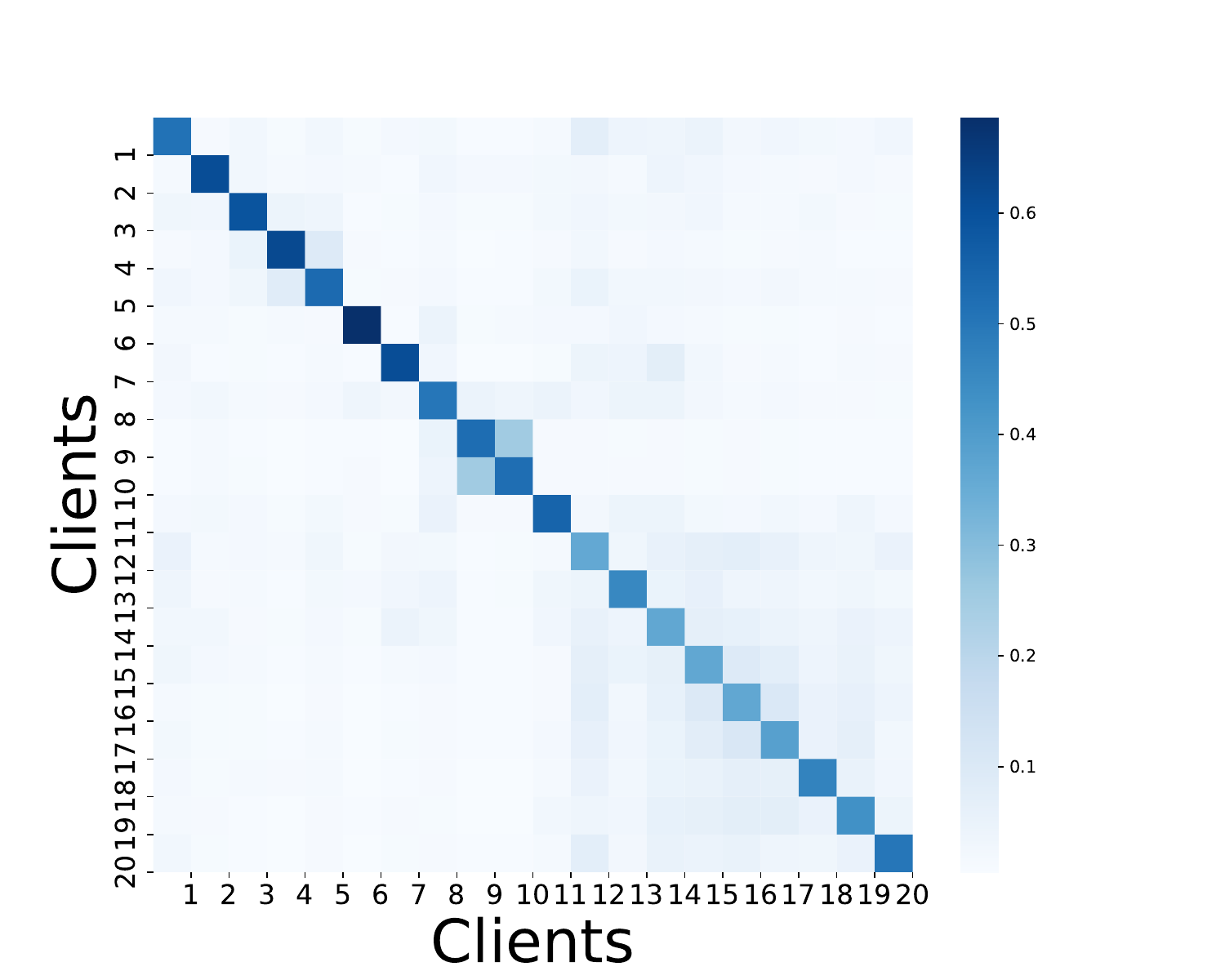}
        \caption{Distr. Sim.}
        \label{fig_Computers_D1}
    \end{subfigure}%
    \hfill
    \begin{subfigure}[t]{0.15\textwidth}
        \includegraphics[width=\linewidth]{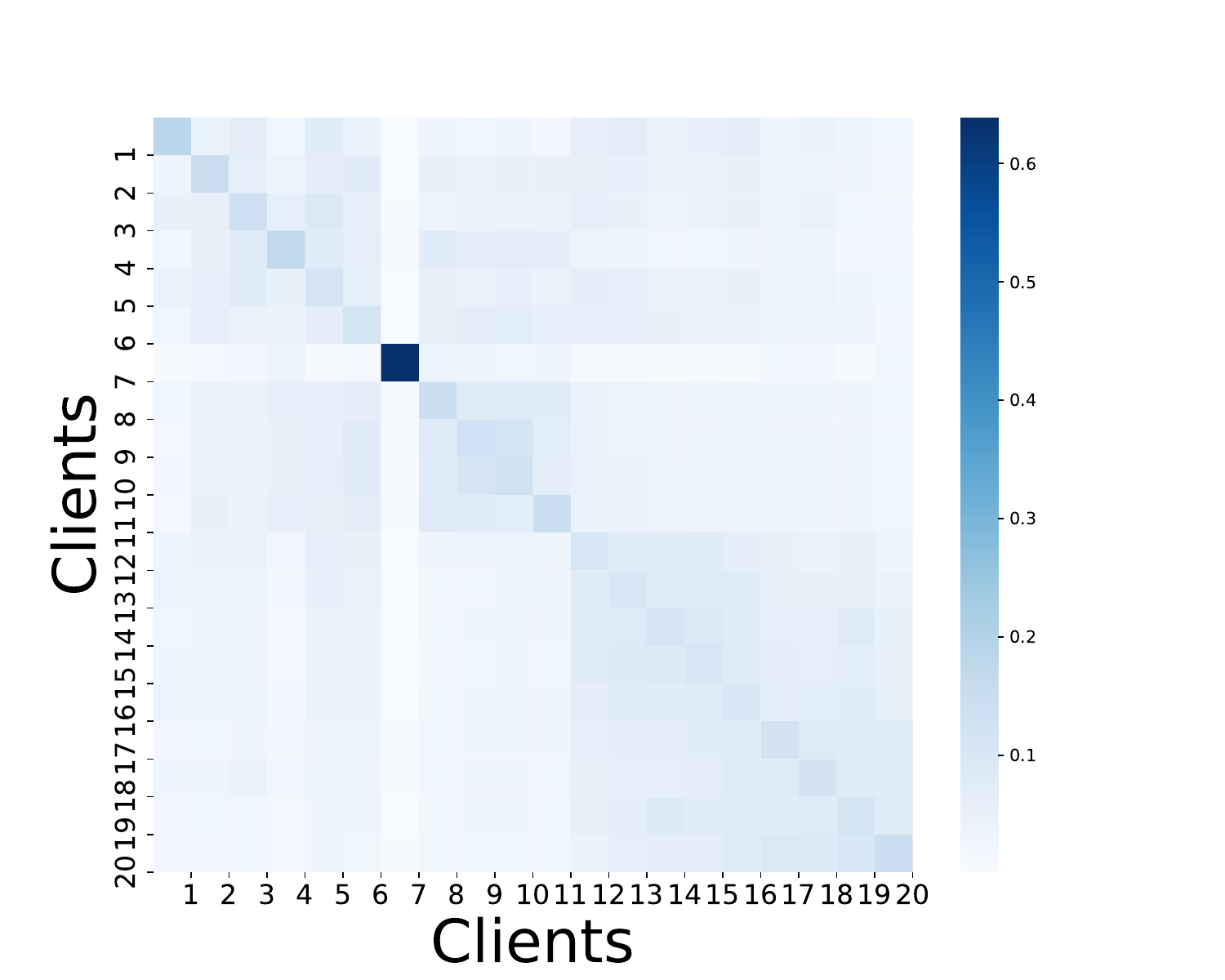}
        \caption{FED-PUB}
        \label{fig_Computers_D2}
    \end{subfigure}%
    \hfill
    \begin{subfigure}[t]{0.15\textwidth}
        \includegraphics[width=\linewidth]{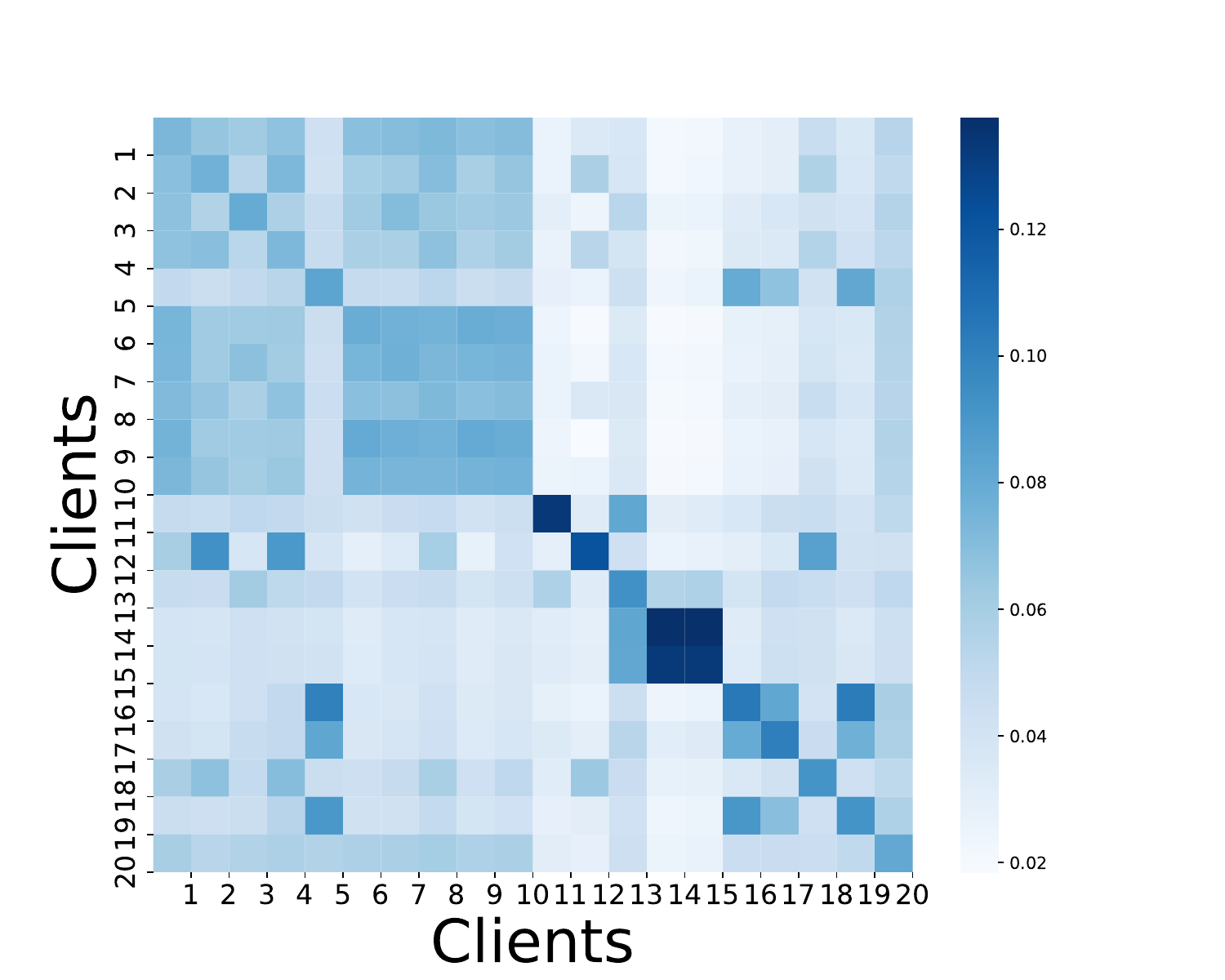}
        \caption{FedGTA}
        \label{fig_Computers_D3}
    \end{subfigure}%
    \hfill
    \begin{subfigure}[t]{0.15\textwidth}
        \includegraphics[width=\linewidth]{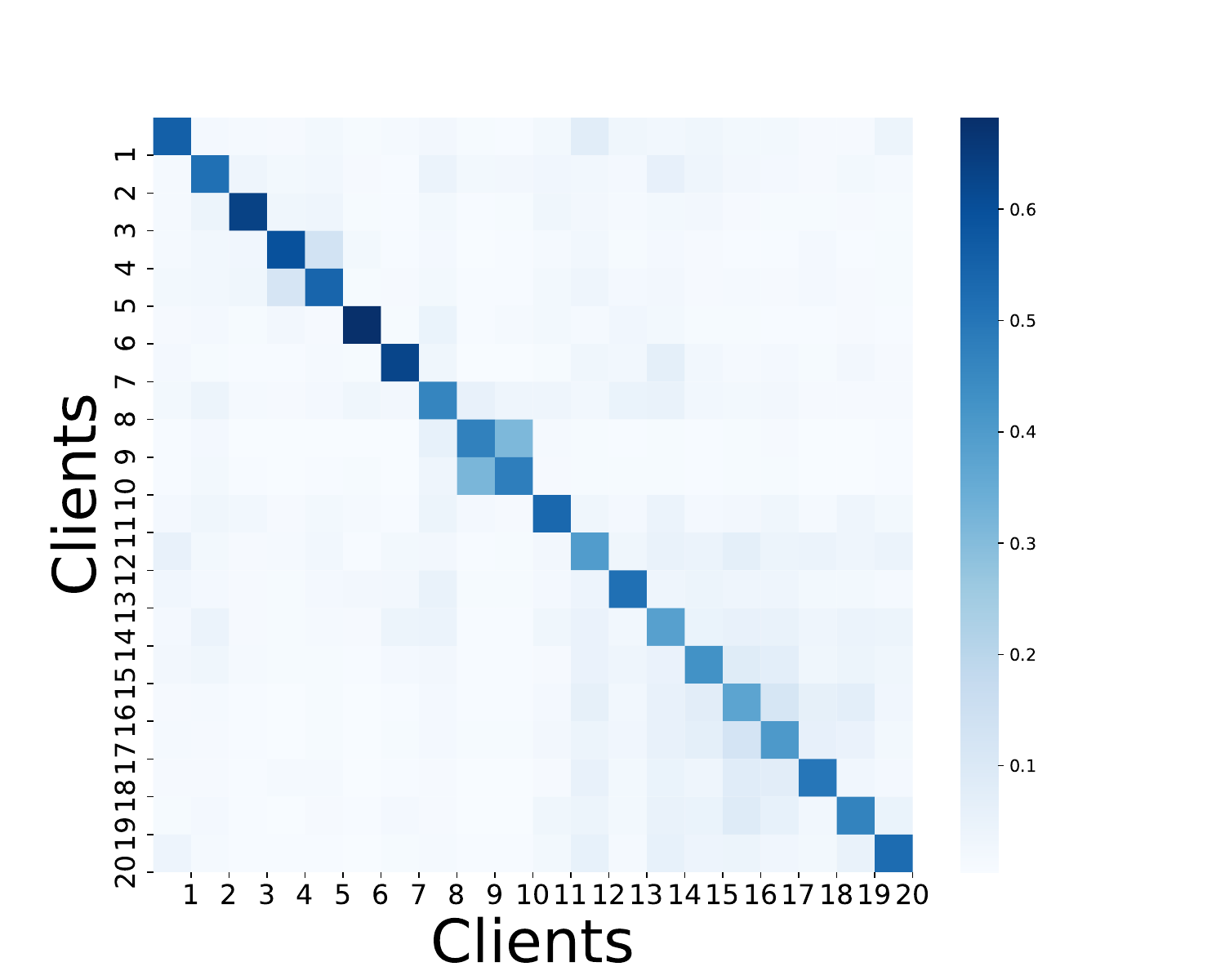}
        \caption{FedIIH of the 1st latent factor ($K=2$)}
        \label{fig_Computers_D4}
    \end{subfigure}
    \hfill
    \begin{subfigure}[t]{0.15\textwidth}
        \includegraphics[width=\linewidth]{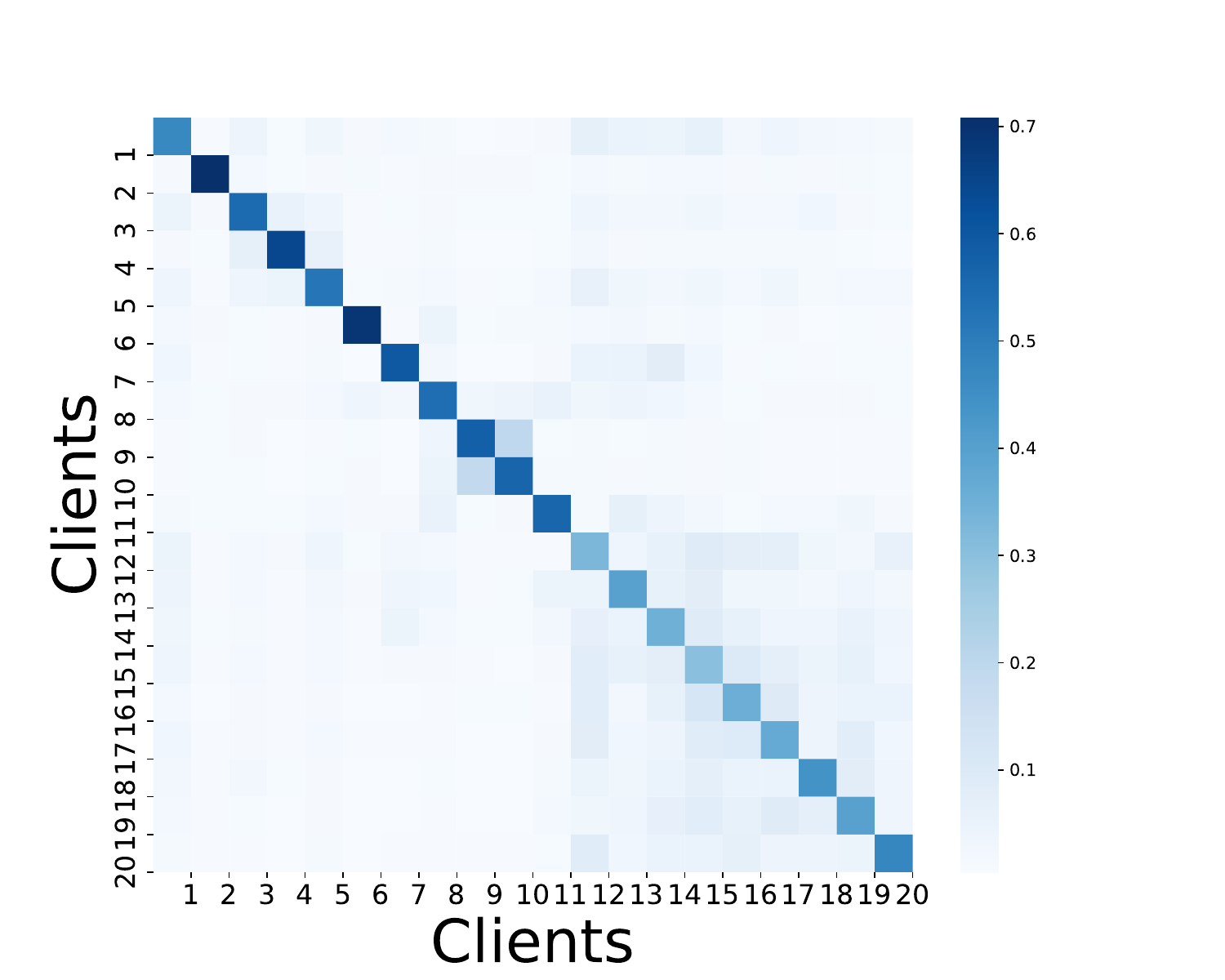}
        \caption{FedIIH of the 2nd latent factor ($K=2$)}
        \label{fig_Computers_D5}
    \end{subfigure}
    \caption{Similarity heatmaps on the \textit{Amazon-Computer} dataset in the non-overlapping setting with 20 clients.}
    \label{fig_Computers_D}
\end{figure}

\begin{figure}[t]
    \centering
    \begin{subfigure}[t]{0.15\textwidth}
        \includegraphics[width=\linewidth]{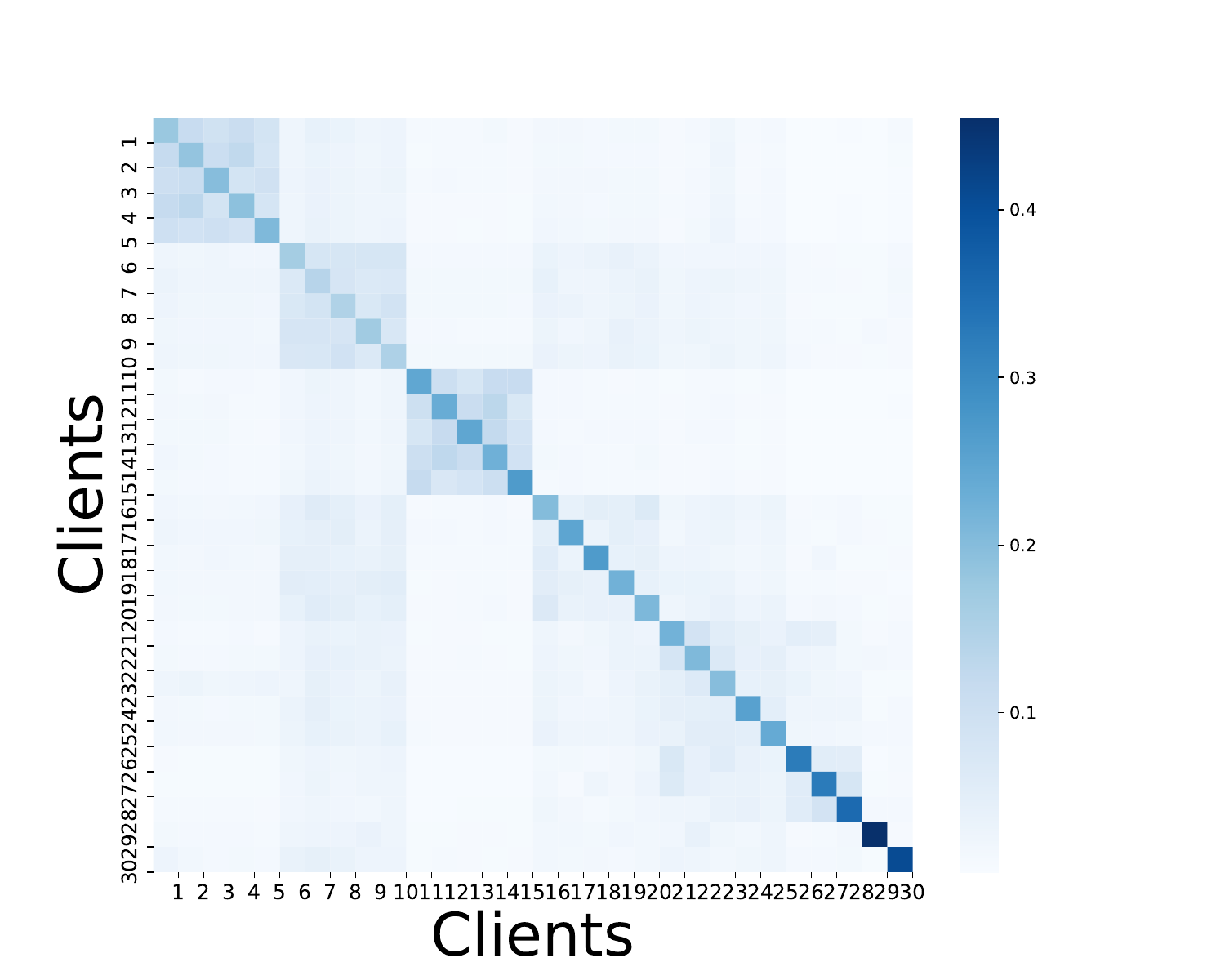}
        \caption{Distr. Sim.}
        \label{fig_Computers_01}
    \end{subfigure}%
    \hfill
    \begin{subfigure}[t]{0.15\textwidth}
        \includegraphics[width=\linewidth]{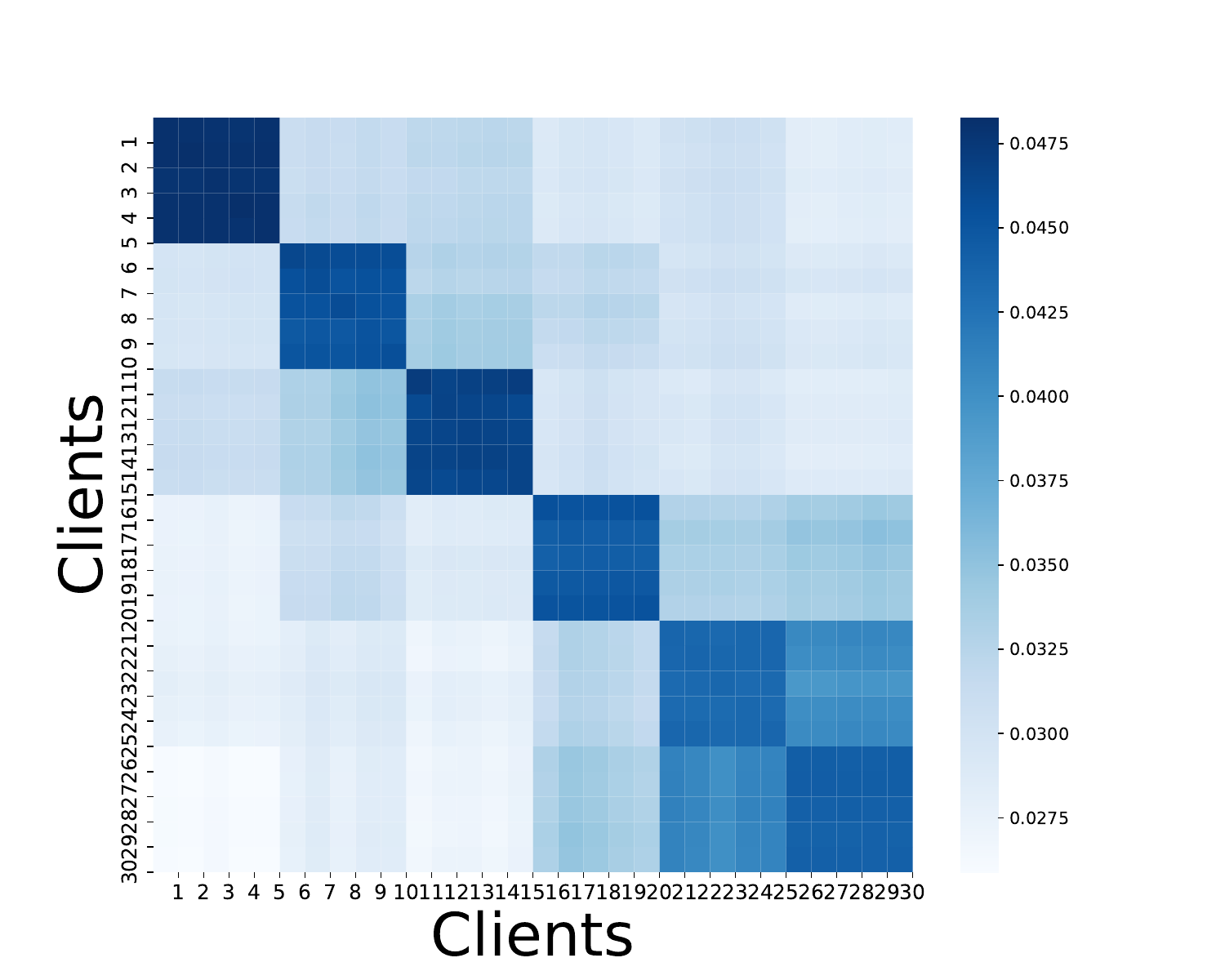}
        \caption{FED-PUB}
        \label{fig_Computers_02}
    \end{subfigure}%
    \hfill
    \begin{subfigure}[t]{0.15\textwidth}
        \includegraphics[width=\linewidth]{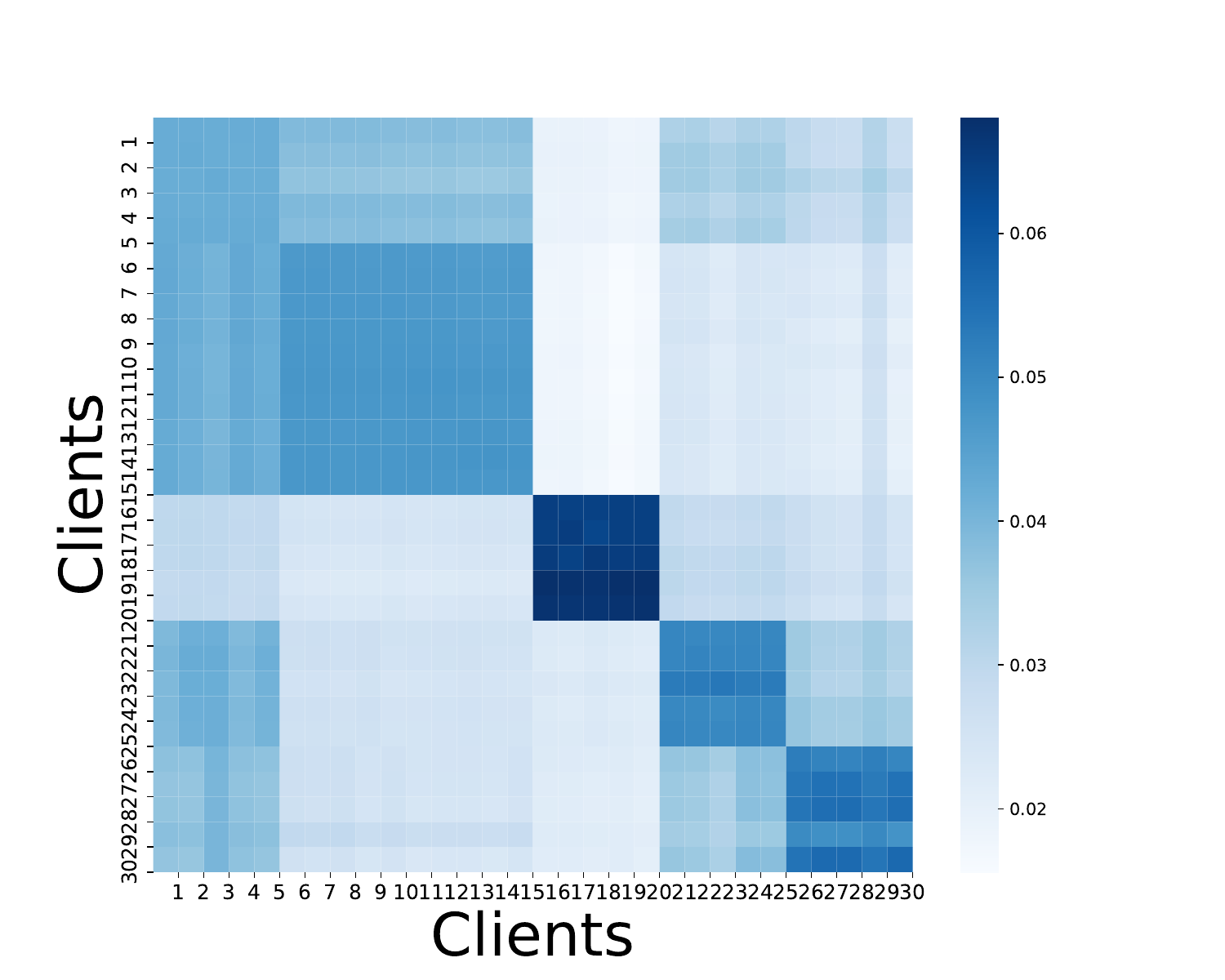}
        \caption{FedGTA}
        \label{fig_Computers_03}
    \end{subfigure}%
    \hfill
    \begin{subfigure}[t]{0.15\textwidth}
        \includegraphics[width=\linewidth]{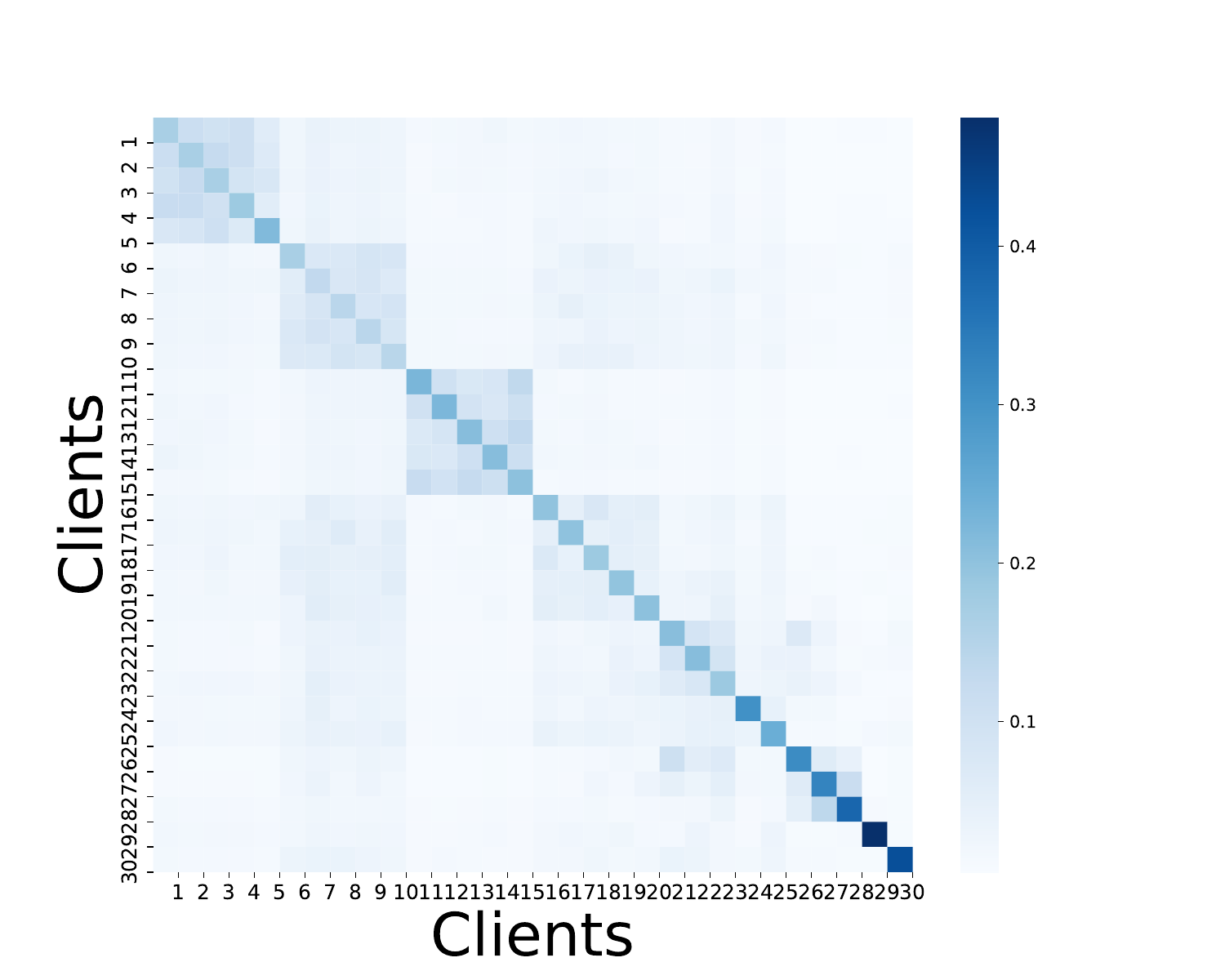}
        \caption{FedIIH of the 1st latent factor ($K=2$)}
        \label{fig_Computers_04}
    \end{subfigure}
    \hfill
    \begin{subfigure}[t]{0.15\textwidth}
        \includegraphics[width=\linewidth]{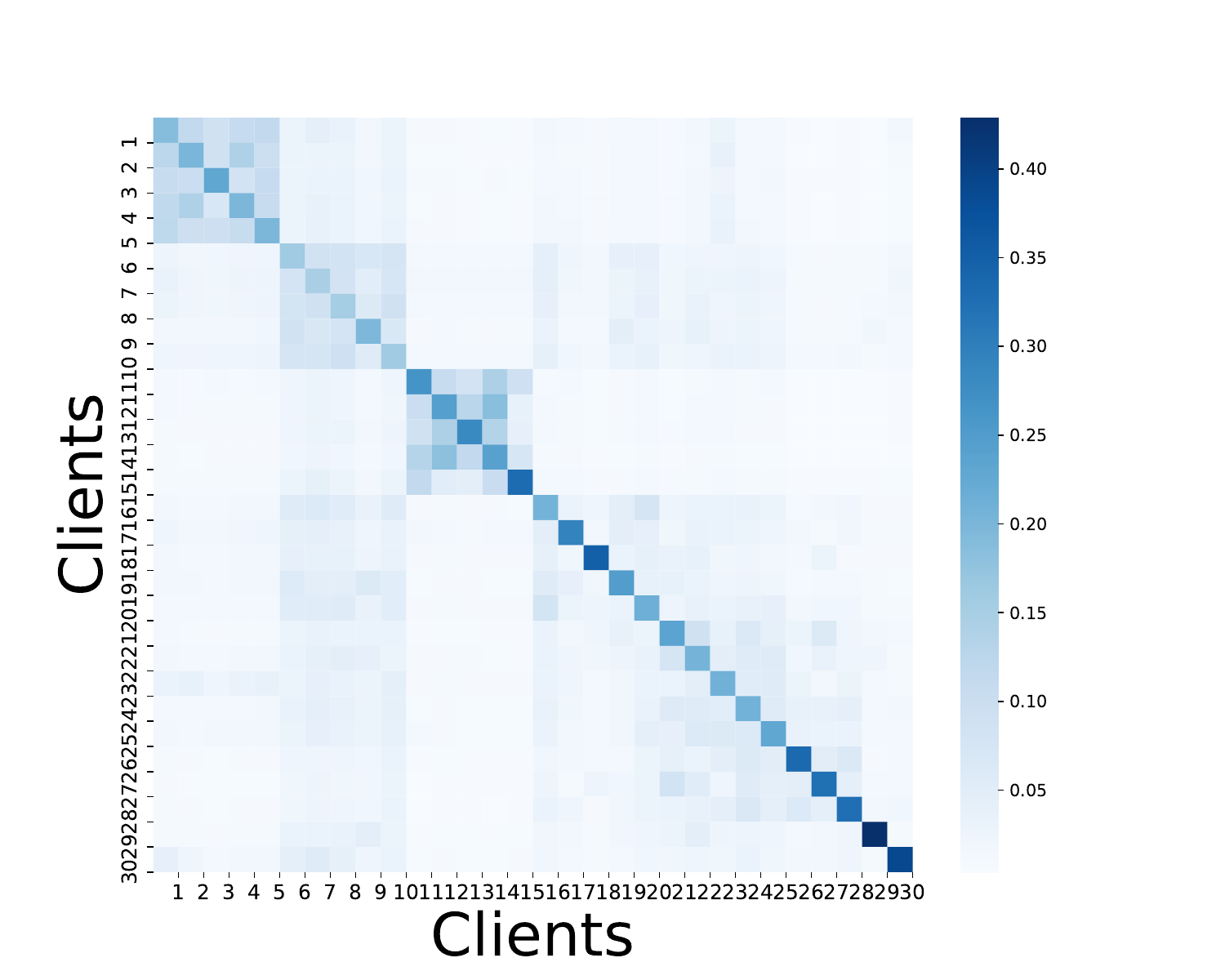}
        \caption{FedIIH of the 2nd latent factor ($K=2$)}
        \label{fig_Computers_05}
    \end{subfigure}
    \caption{Similarity heatmaps on the \textit{Amazon-Computer} dataset in the overlapping setting with 30 clients.}
    \label{fig_Computers_O}
\end{figure}

\begin{figure}[t]
    \centering
    \begin{subfigure}[t]{0.15\textwidth}
        \includegraphics[width=\linewidth]{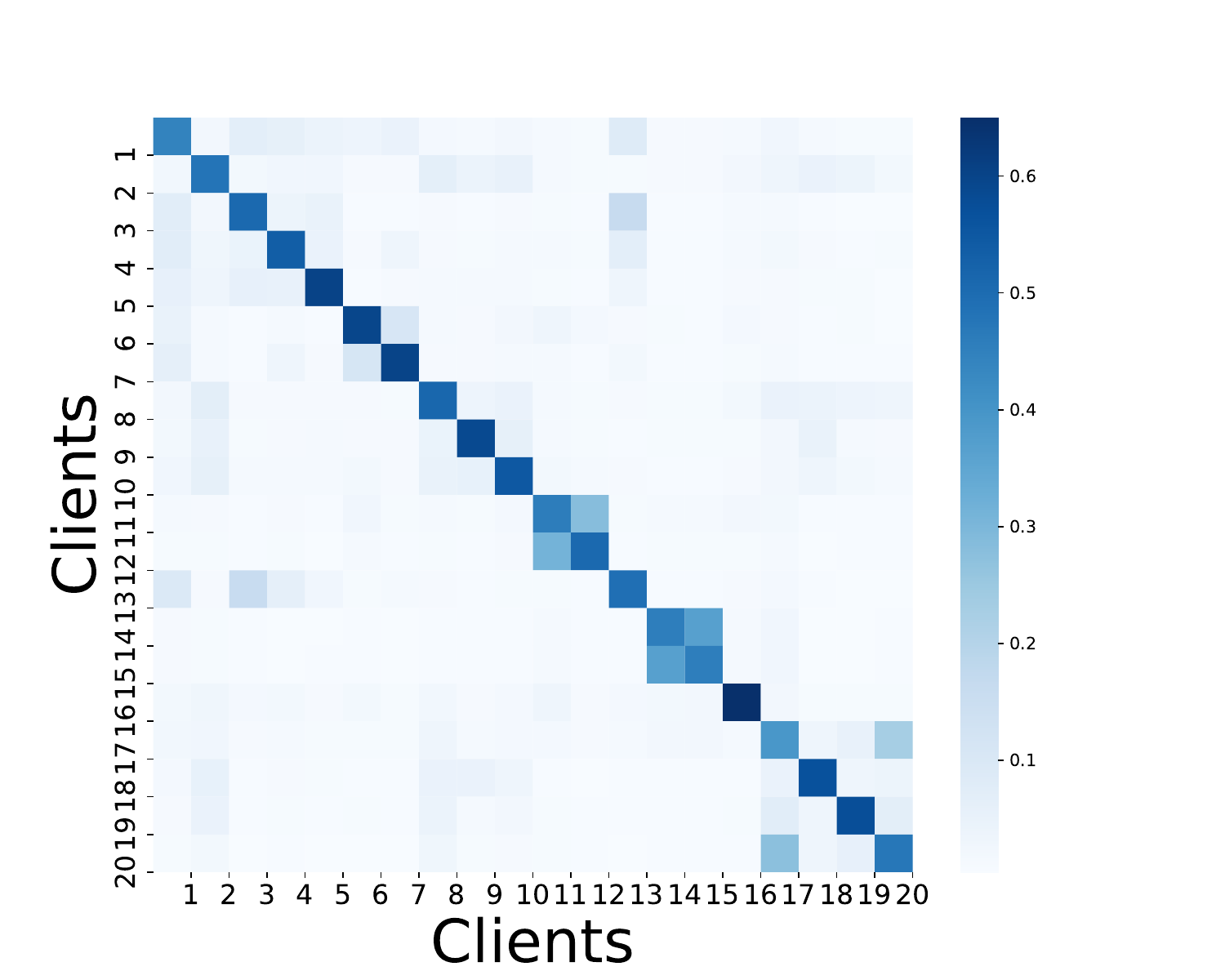}
        \caption{Distr. Sim.}
        \label{fig_Photo_D1}
    \end{subfigure}%
    \hfill
    \begin{subfigure}[t]{0.15\textwidth}
        \includegraphics[width=\linewidth]{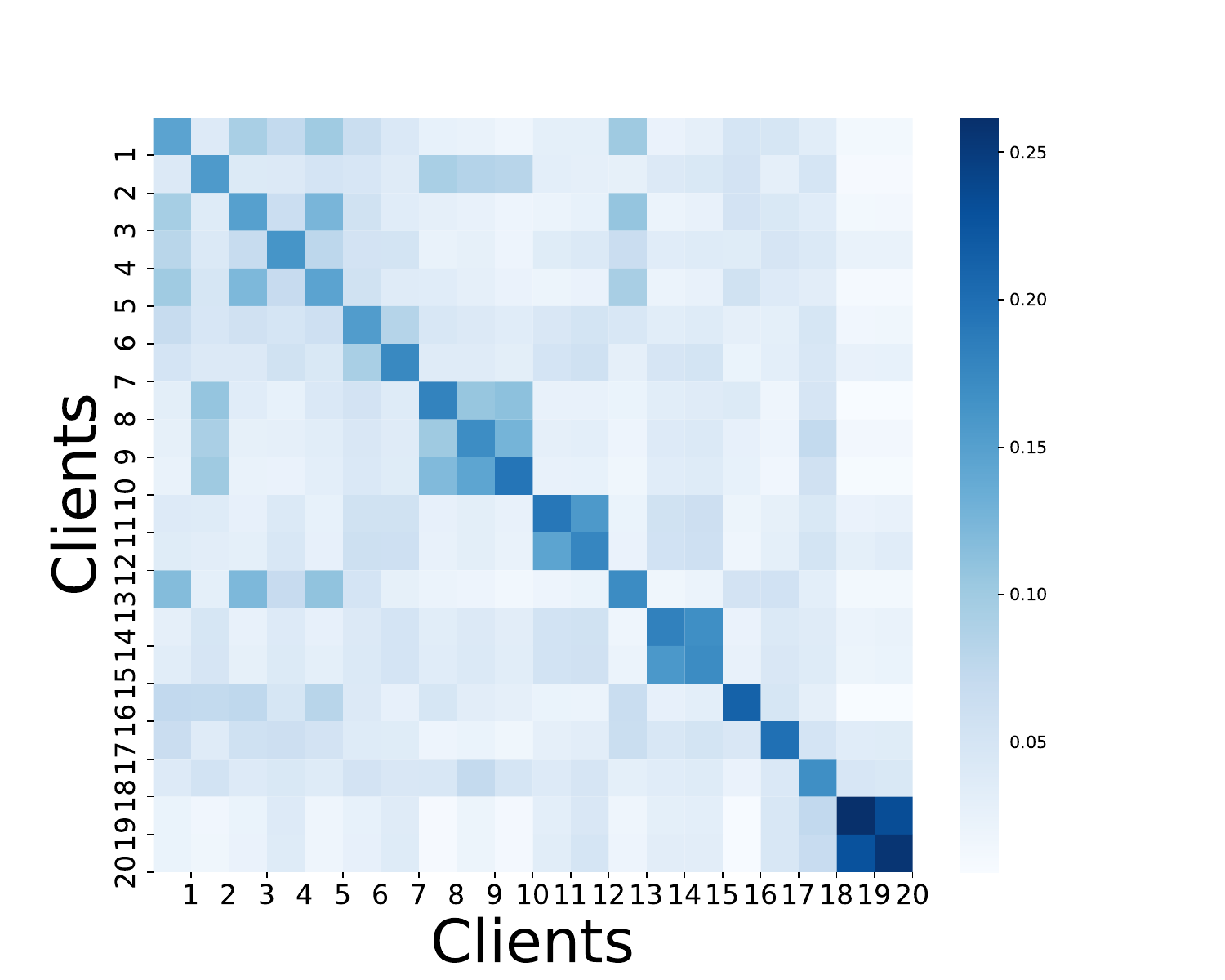}
        \caption{FED-PUB}
        \label{fig_Photo_D2}
    \end{subfigure}%
    \hfill
    \begin{subfigure}[t]{0.15\textwidth}
        \includegraphics[width=\linewidth]{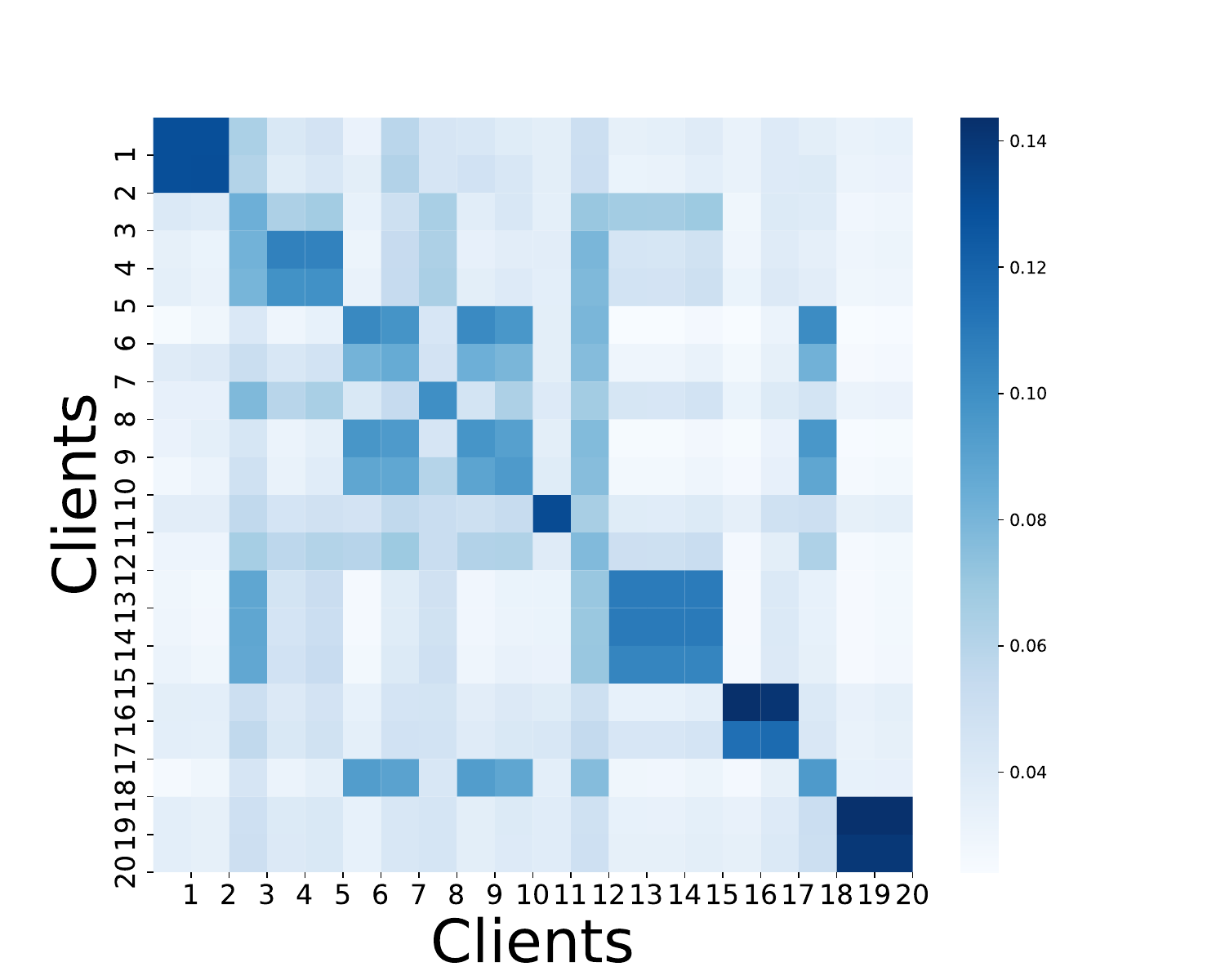}
        \caption{FedGTA}
        \label{fig_Photo_D3}
    \end{subfigure}%
    \hfill
    \begin{subfigure}[t]{0.15\textwidth}
        \includegraphics[width=\linewidth]{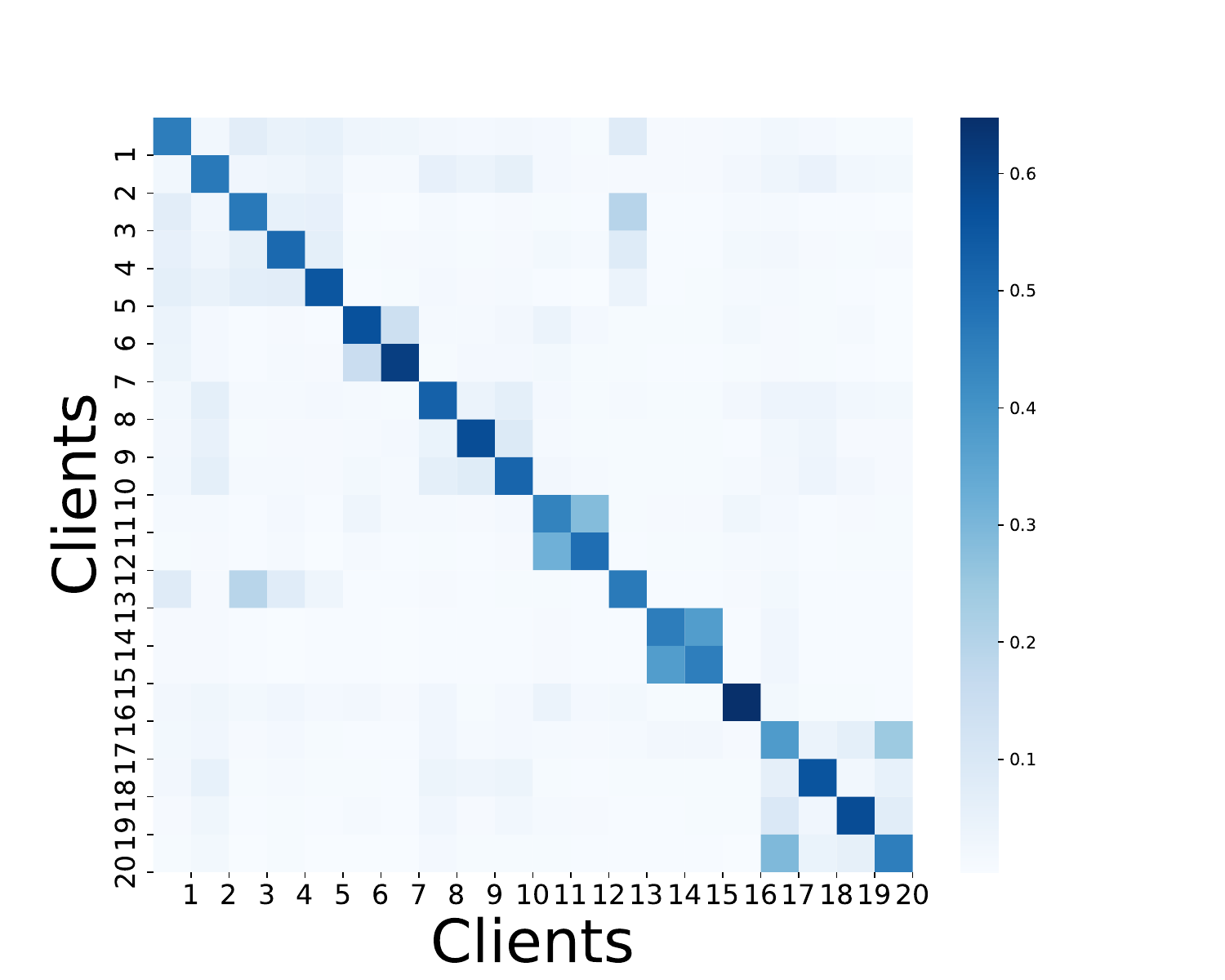}
        \caption{FedIIH of the 1st latent factor ($K=2$)}
        \label{fig_Photo_D4}
    \end{subfigure}
    \hfill
    \begin{subfigure}[t]{0.15\textwidth}
        \includegraphics[width=\linewidth]{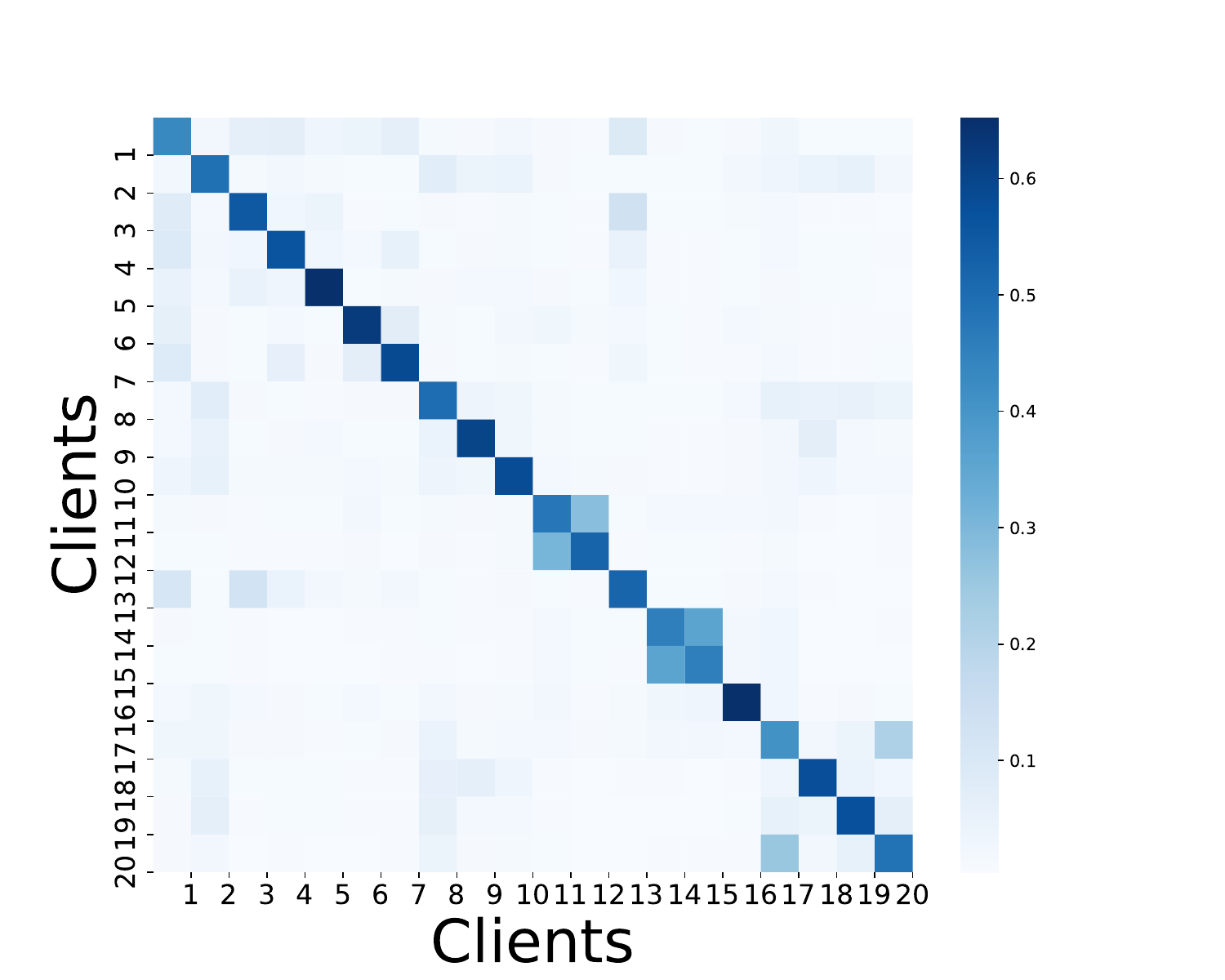}
        \caption{FedIIH of the 2nd latent factor ($K=2$)}
        \label{fig_Photo_D5}
    \end{subfigure}
    \caption{Similarity heatmaps on the \textit{Amazon-Photo} dataset in the non-overlapping setting with 20 clients.}
    \label{fig_Photo_D}
\end{figure}

\begin{figure}[t]
    \centering
    \begin{subfigure}[t]{0.15\textwidth}
        \includegraphics[width=\linewidth]{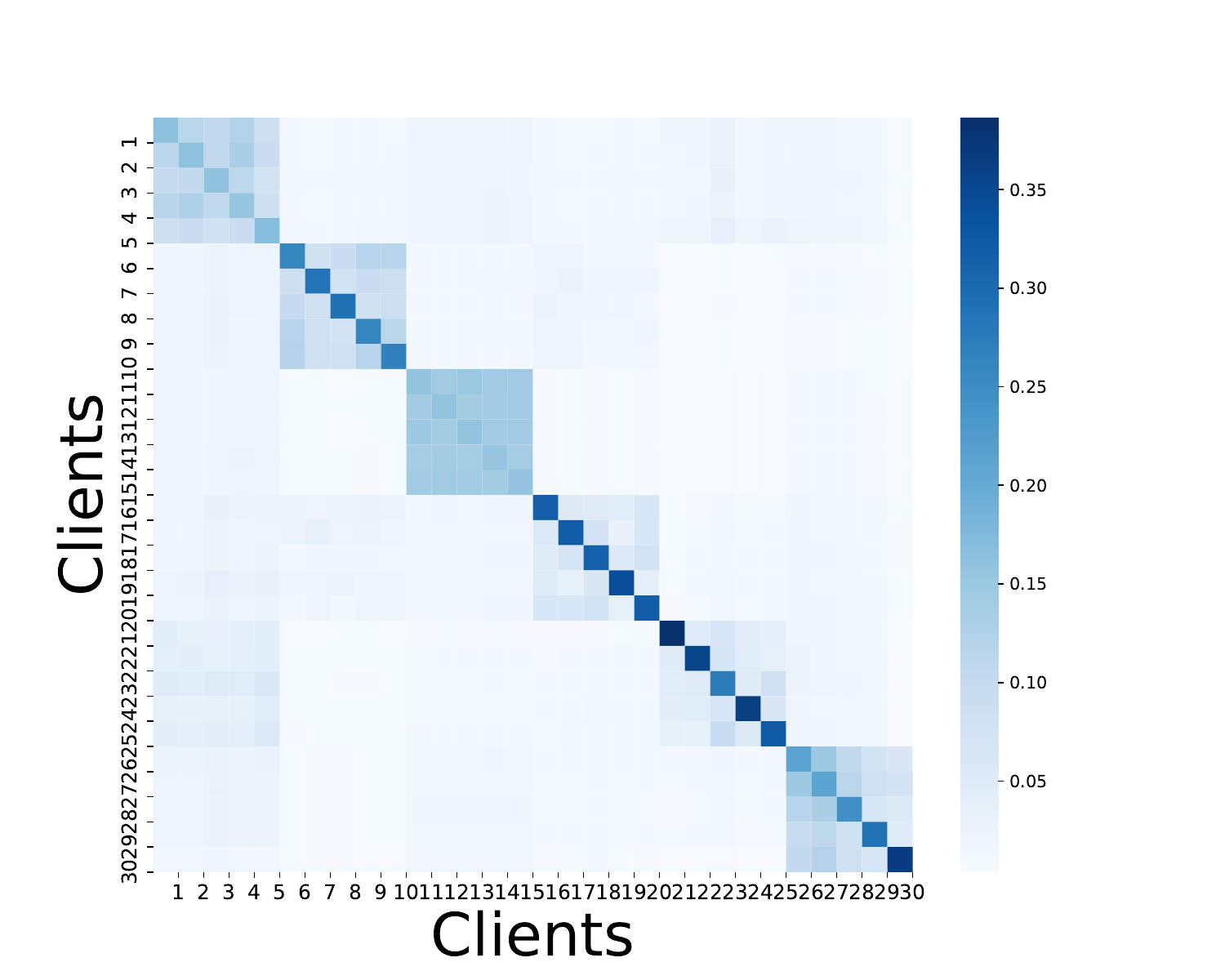}
        \caption{Distr. Sim.}
        \label{fig_Photo_01}
    \end{subfigure}%
    \hfill
    \begin{subfigure}[t]{0.15\textwidth}
        \includegraphics[width=\linewidth]{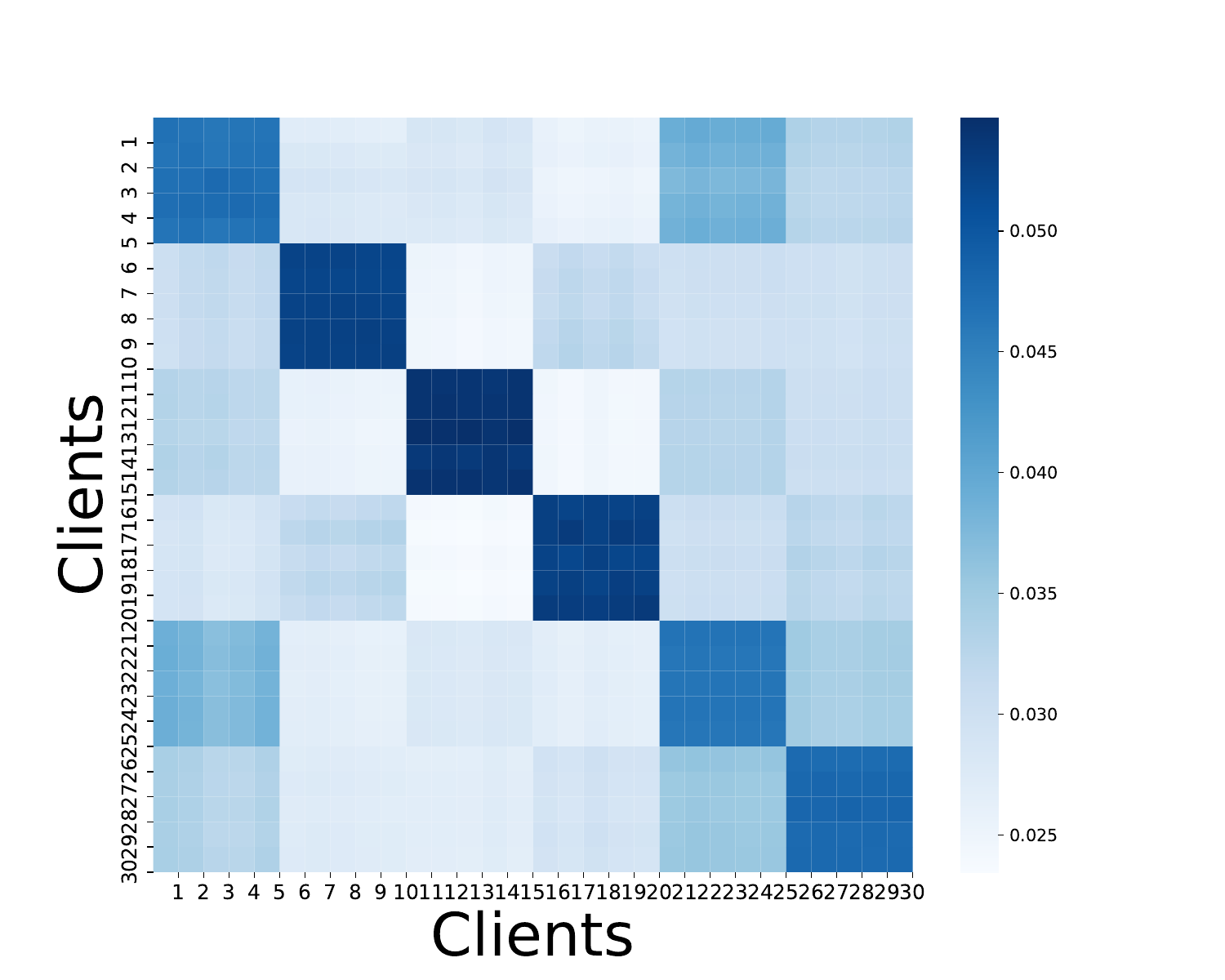}
        \caption{FED-PUB}
        \label{fig_Photo_02}
    \end{subfigure}%
    \hfill
    \begin{subfigure}[t]{0.15\textwidth}
        \includegraphics[width=\linewidth]{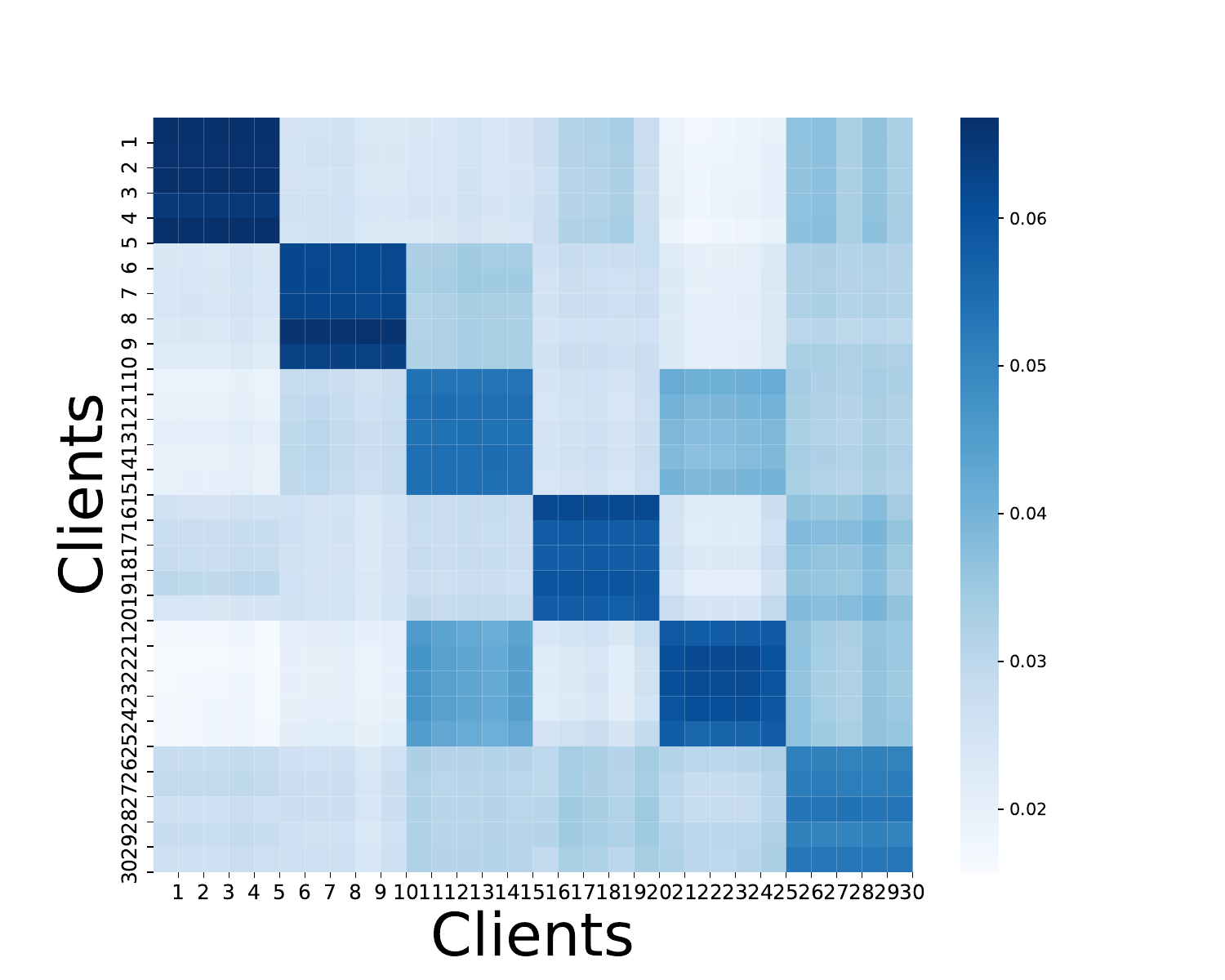}
        \caption{FedGTA}
        \label{fig_Photo_03}
    \end{subfigure}%
    \hfill
    \begin{subfigure}[t]{0.15\textwidth}
        \includegraphics[width=\linewidth]{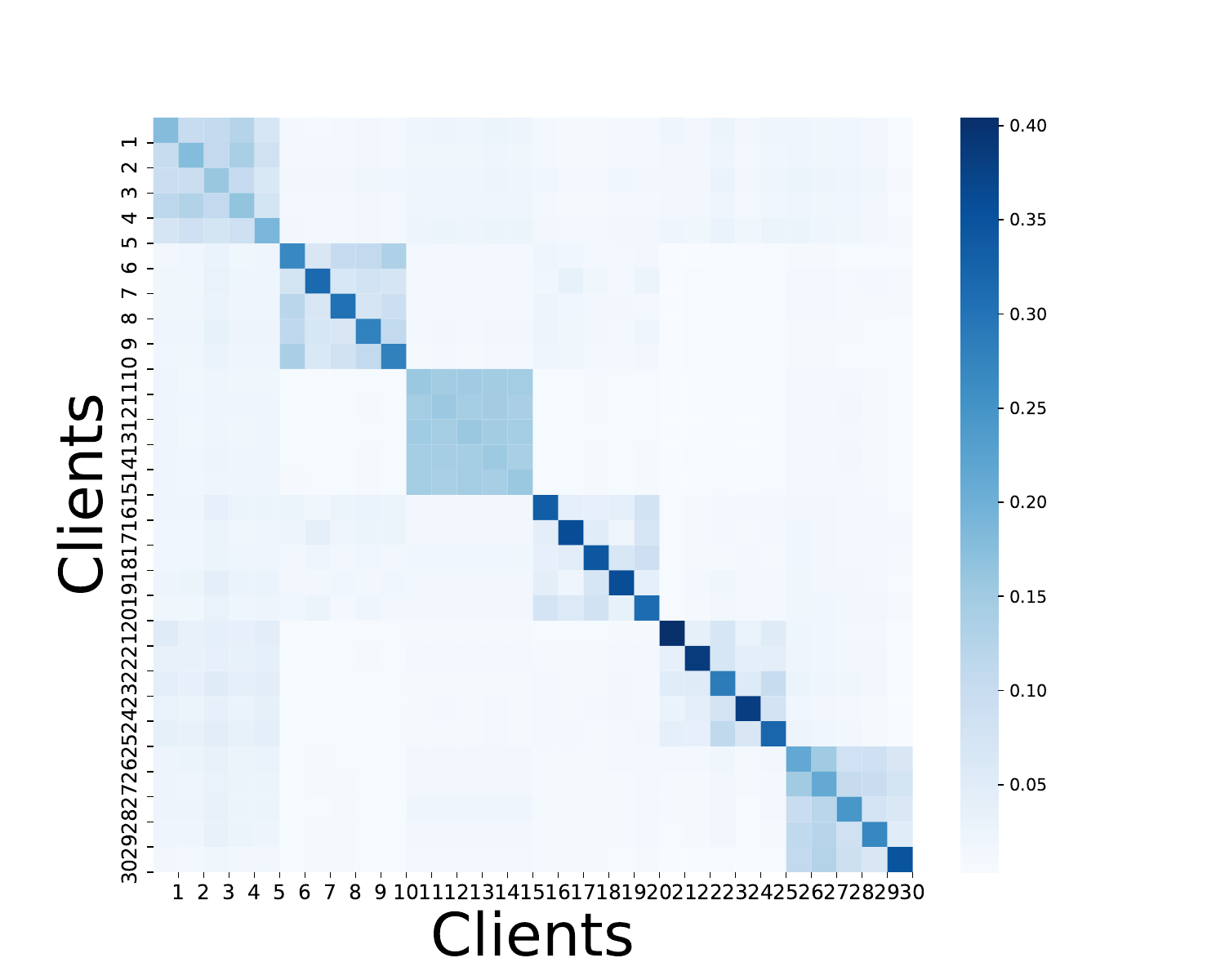}
        \caption{FedIIH of the 1st latent factor ($K=2$)}
        \label{fig_Photo_04}
    \end{subfigure}
    \hfill
    \begin{subfigure}[t]{0.15\textwidth}
        \includegraphics[width=\linewidth]{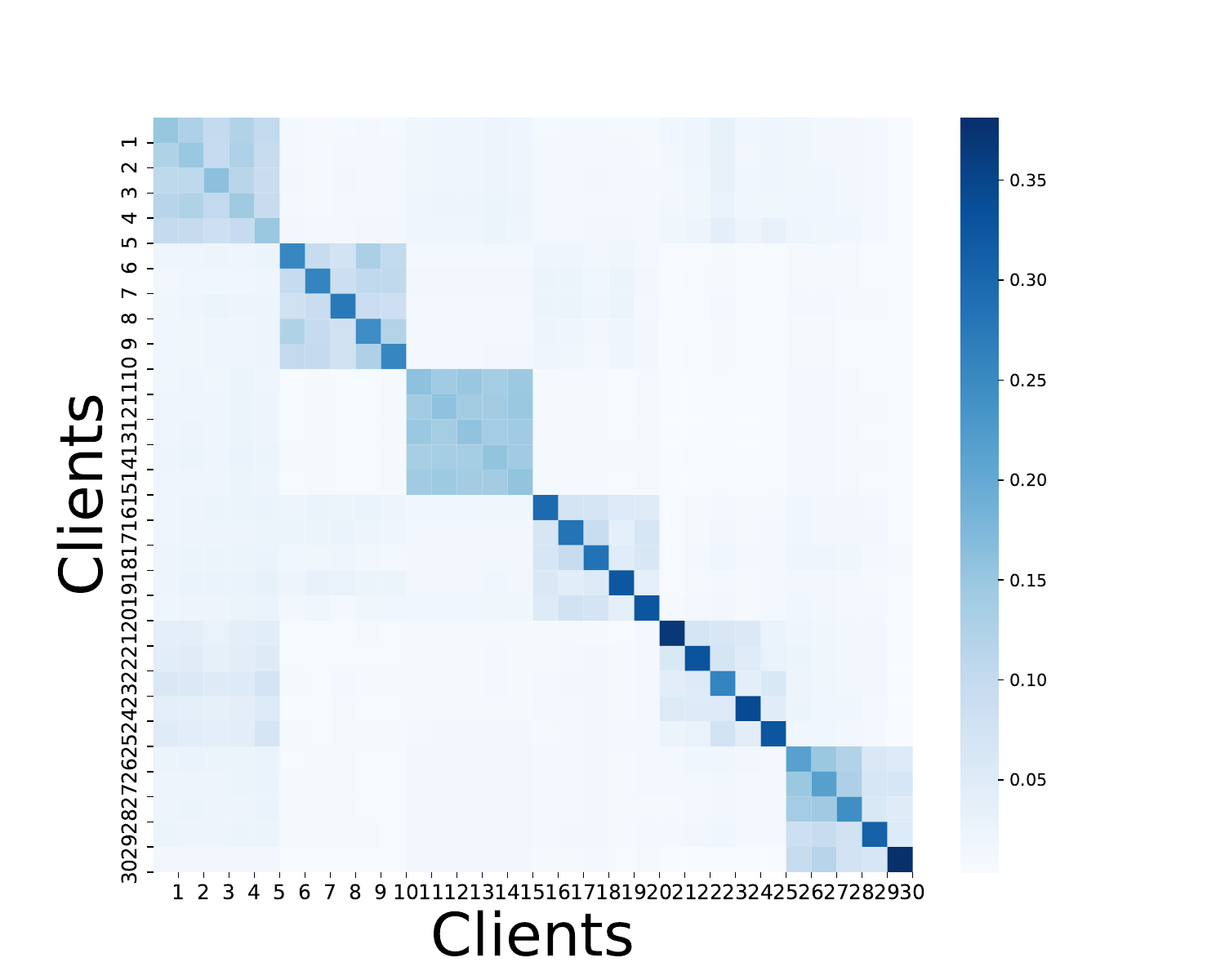}
        \caption{FedIIH of the 2nd latent factor ($K=2$)}
        \label{fig_Photo_05}
    \end{subfigure}
    \caption{Similarity heatmaps on the \textit{Amazon-Photo} dataset in the overlapping setting with 30 clients.}
    \label{fig_Photo_O}
\end{figure}

\begin{figure}[t]
    \centering
    \begin{subfigure}[t]{0.15\textwidth}
        \includegraphics[width=\linewidth]{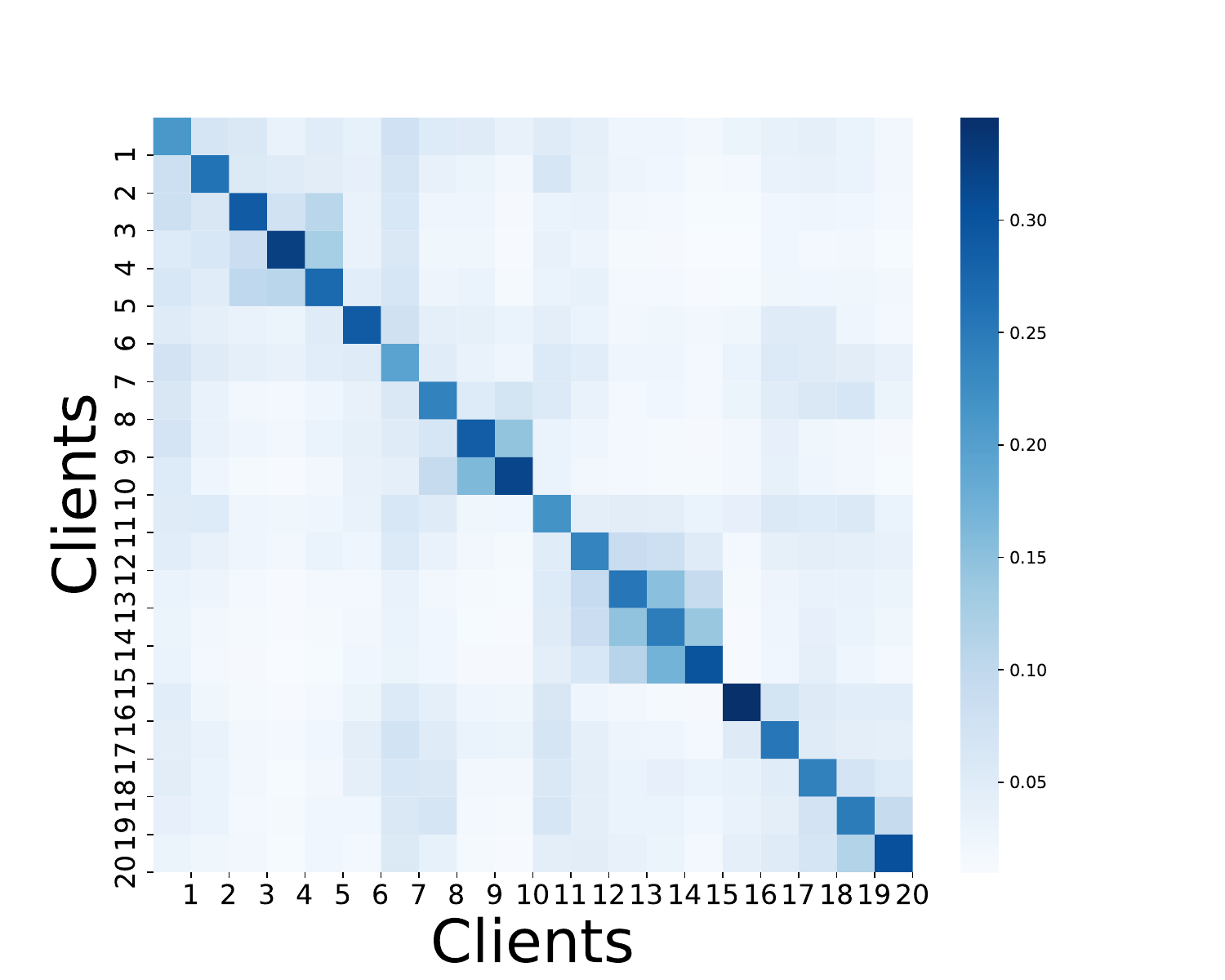}
        \caption{Distr. Sim.}
        \label{fig_ogbn-arxiv_D1}
    \end{subfigure}%
    \hfill
    \begin{subfigure}[t]{0.15\textwidth}
        \includegraphics[width=\linewidth]{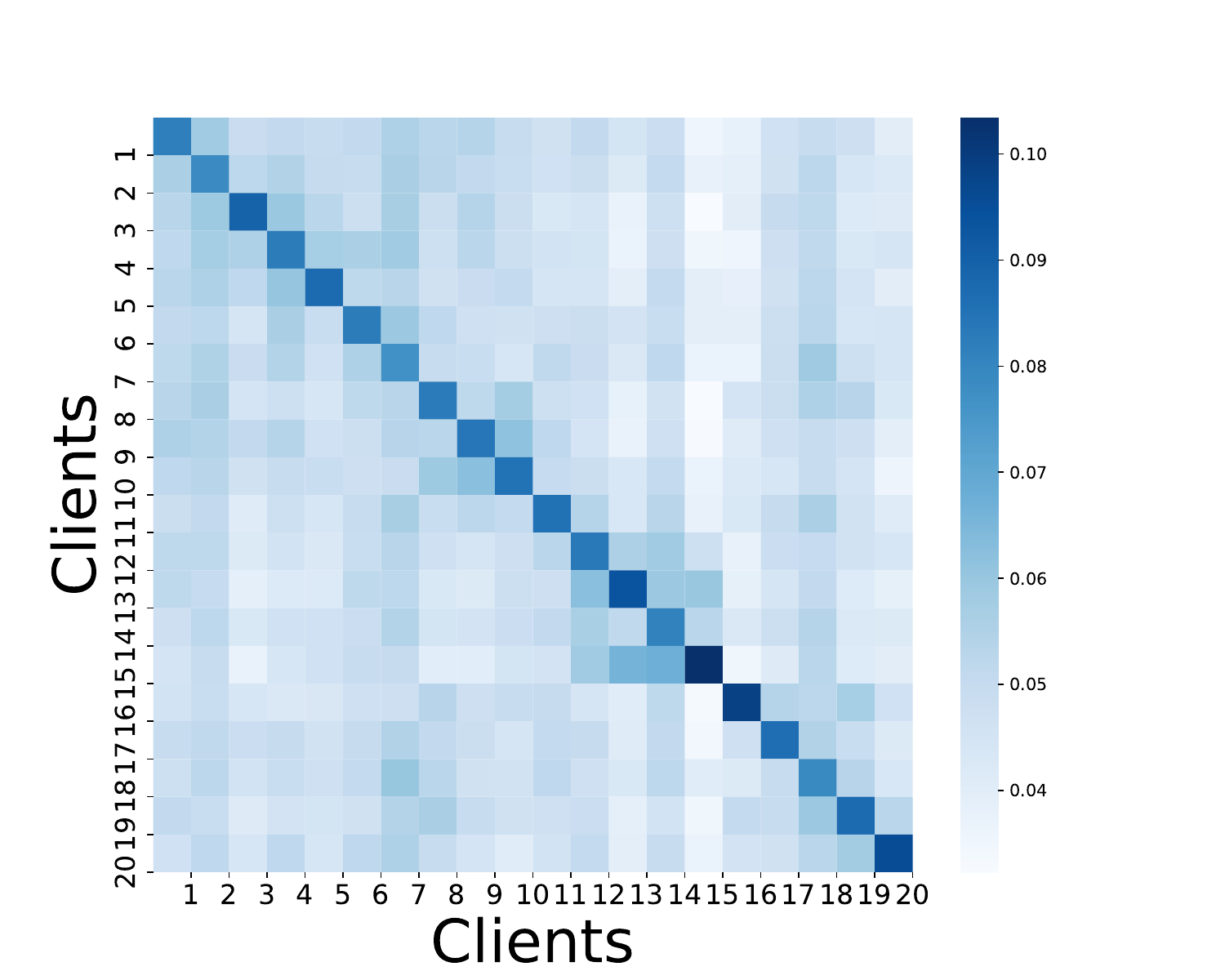}
        \caption{FED-PUB}
        \label{fig_ogbn-arxiv_D2}
    \end{subfigure}%
    \hfill
    \begin{subfigure}[t]{0.15\textwidth}
        \includegraphics[width=\linewidth]{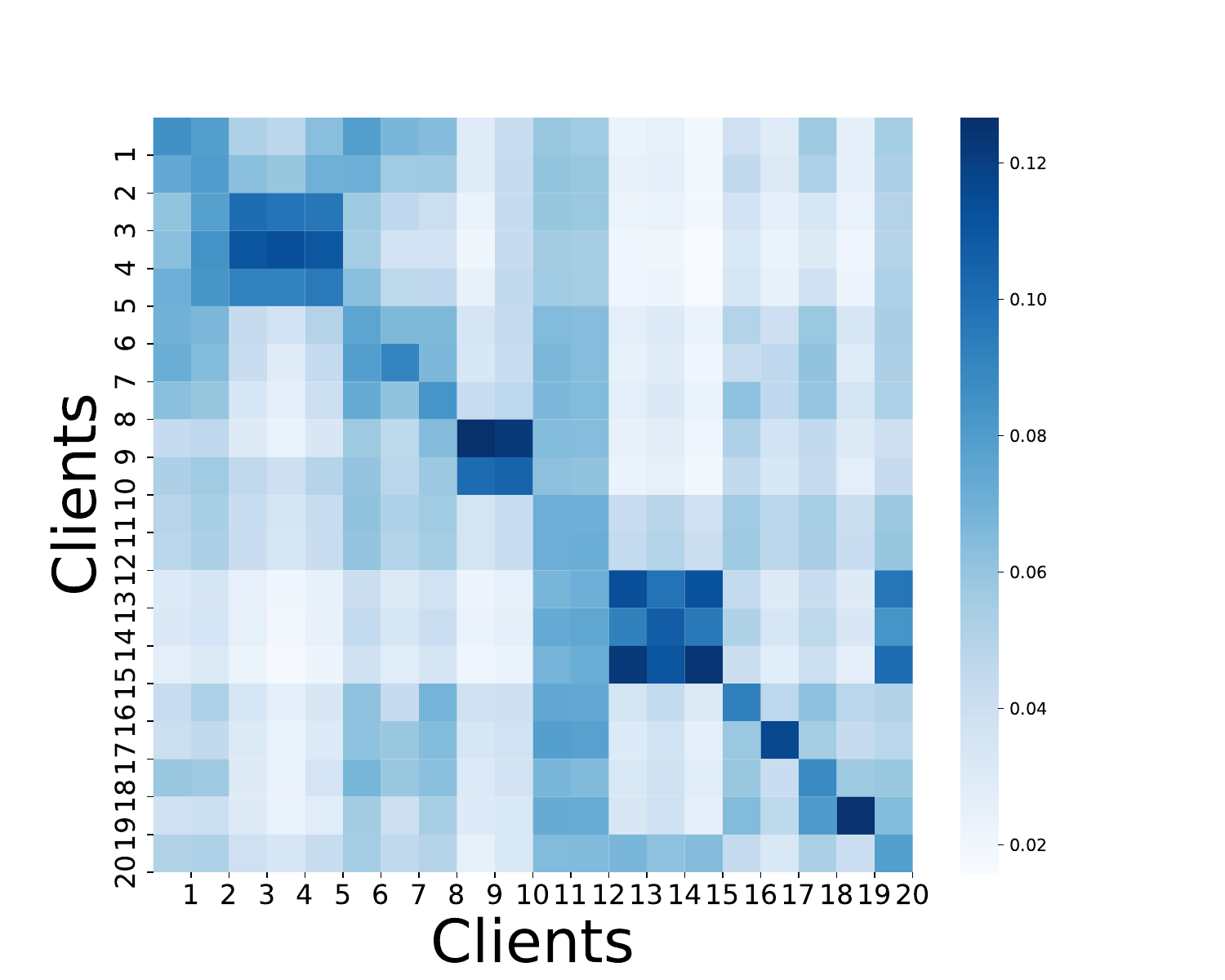}
        \caption{FedGTA}
        \label{fig_ogbn-arxiv_D3}
    \end{subfigure}%
    \hfill
    \begin{subfigure}[t]{0.15\textwidth}
        \includegraphics[width=\linewidth]{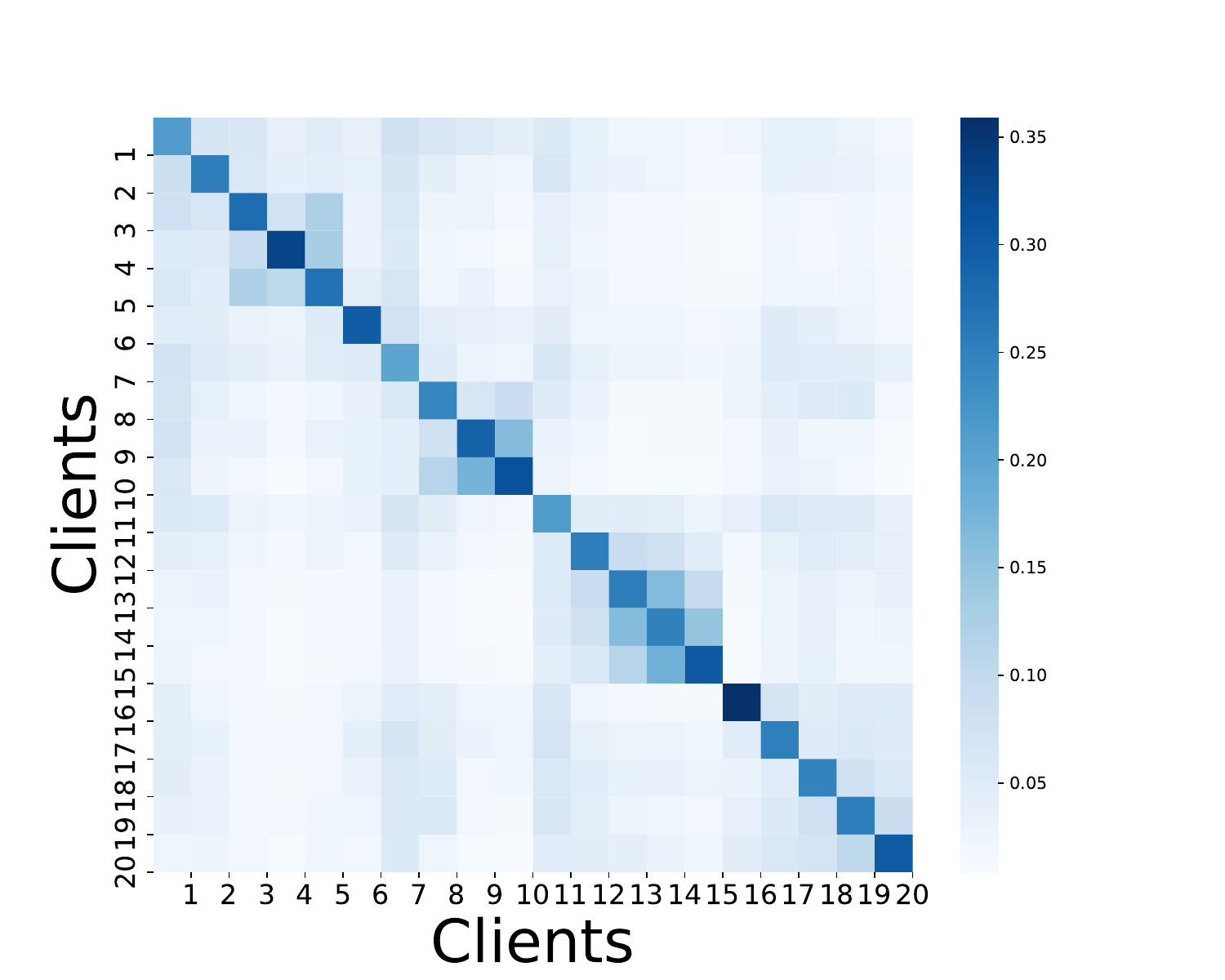}
        \caption{FedIIH of the 1st latent factor ($K=2$)}
        \label{fig_ogbn-arxiv_D4}
    \end{subfigure}
    \hfill
    \begin{subfigure}[t]{0.15\textwidth}
        \includegraphics[width=\linewidth]{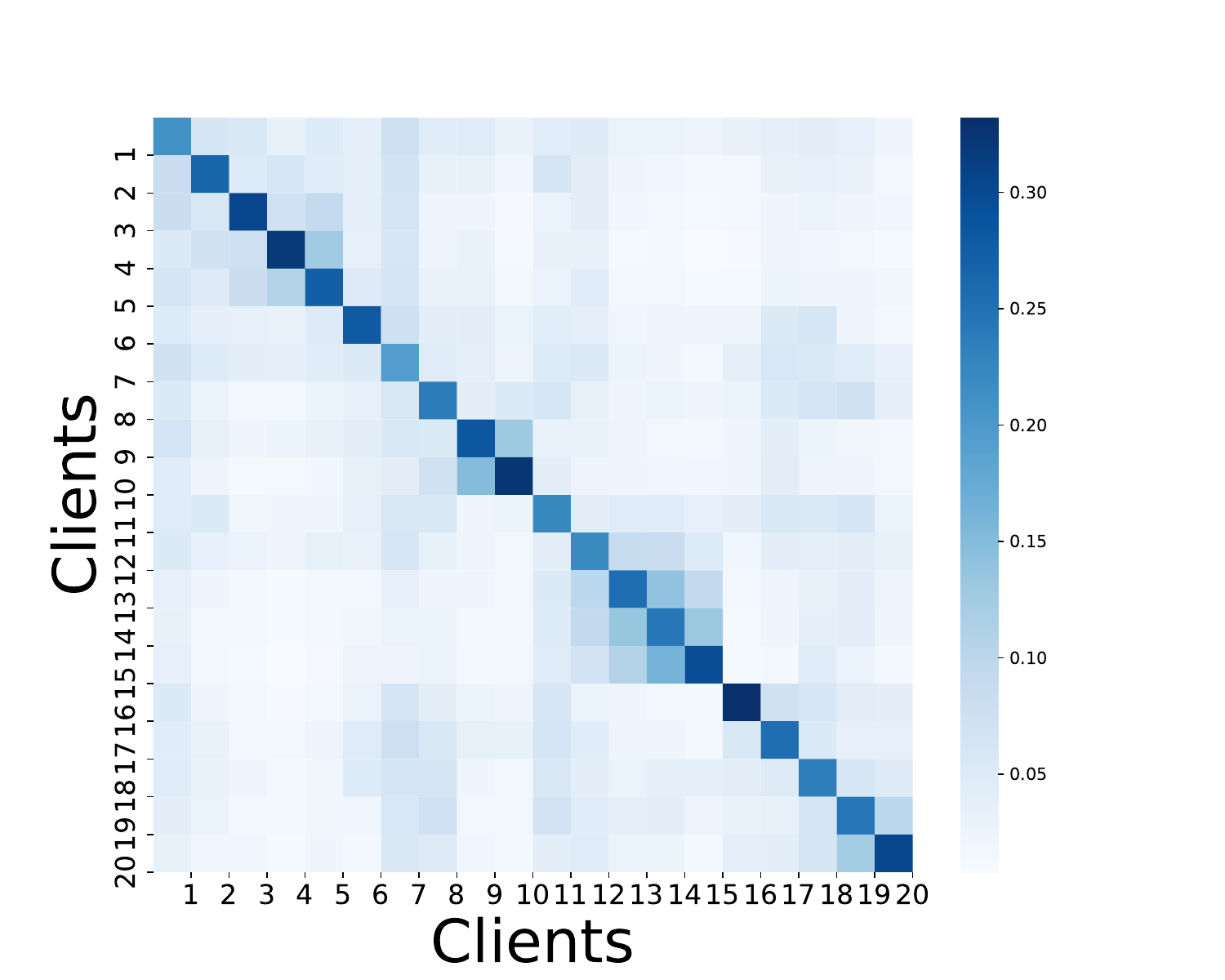}
        \caption{FedIIH of the 2nd latent factor ($K=2$)}
        \label{fig_ogbn-arxiv_D5}
    \end{subfigure}
    \caption{Similarity heatmaps on the \textit{ogbn-arxiv} dataset in the non-overlapping setting with 20 clients.}
    \label{fig_ogbn-arxiv_D}
\end{figure}

\begin{figure}[t]
    \centering
    \begin{subfigure}[t]{0.15\textwidth}
        \includegraphics[width=\linewidth]{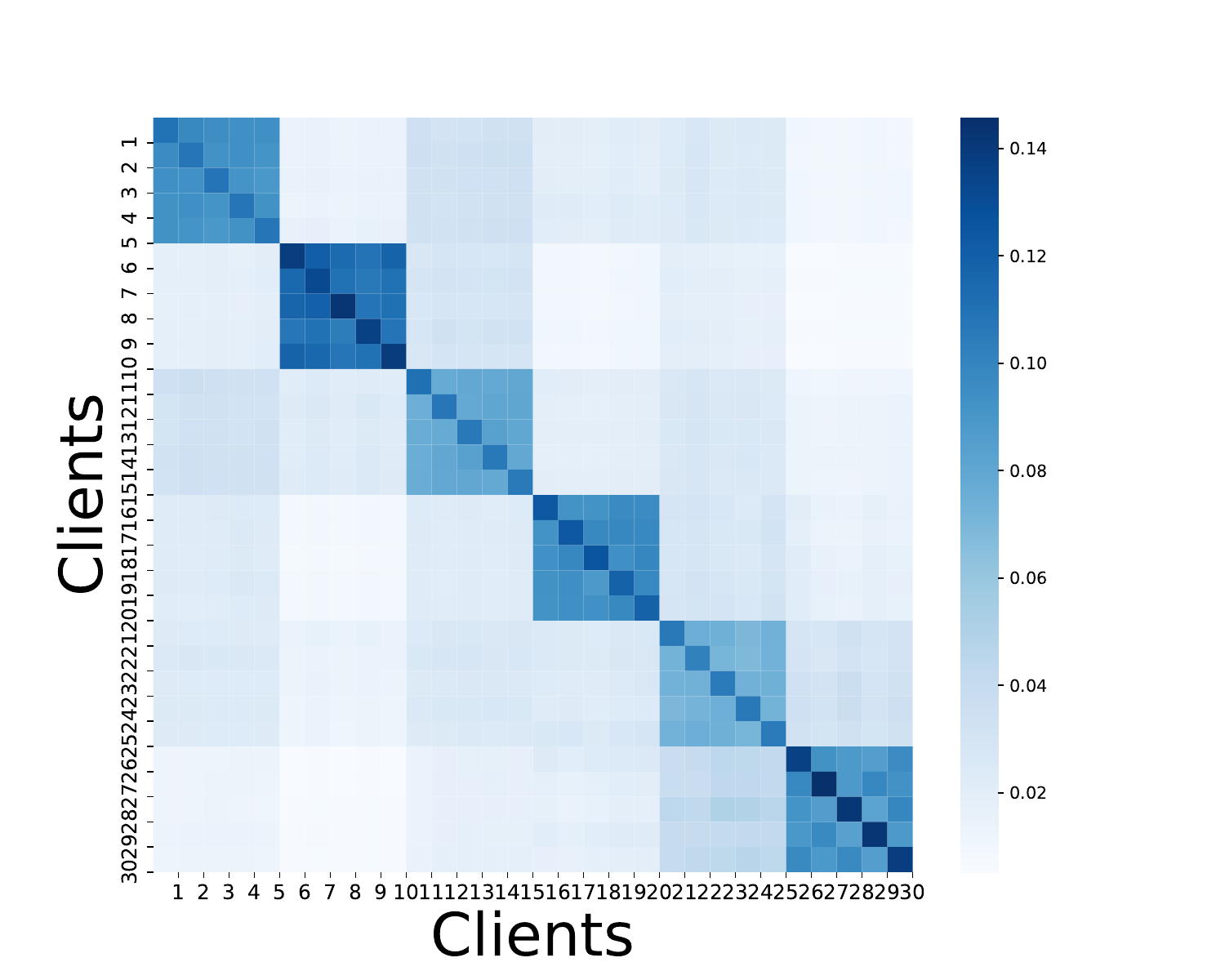}
        \caption{Distr. Sim.}
        \label{fig_ogbn-arxiv_01}
    \end{subfigure}%
    \hfill
    \begin{subfigure}[t]{0.15\textwidth}
        \includegraphics[width=\linewidth]{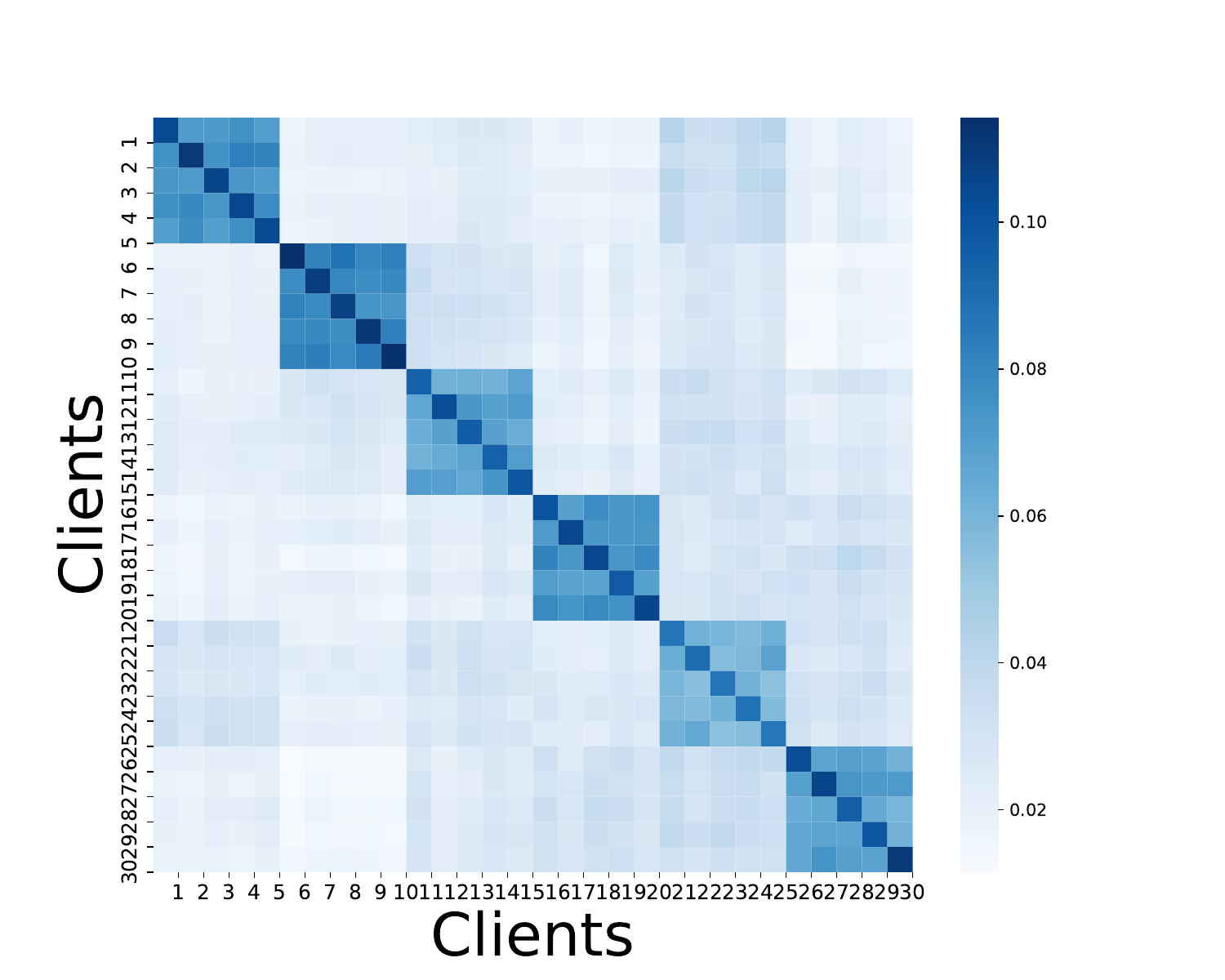}
        \caption{FED-PUB}
        \label{fig_ogbn-arxiv_02}
    \end{subfigure}%
    \hfill
    \begin{subfigure}[t]{0.15\textwidth}
        \includegraphics[width=\linewidth]{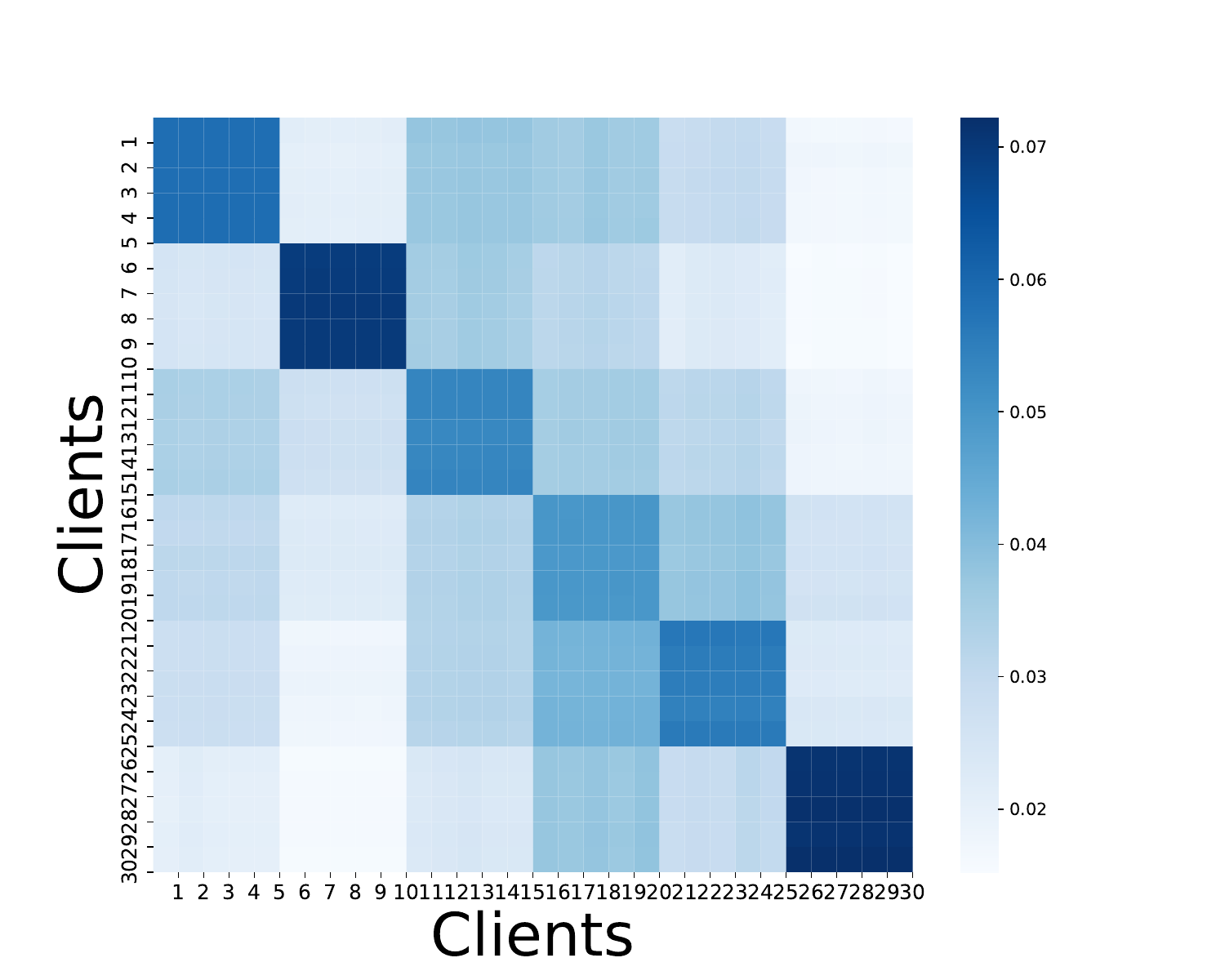}
        \caption{FedGTA}
        \label{fig_ogbn-arxiv_03}
    \end{subfigure}%
    \hfill
    \begin{subfigure}[t]{0.15\textwidth}
        \includegraphics[width=\linewidth]{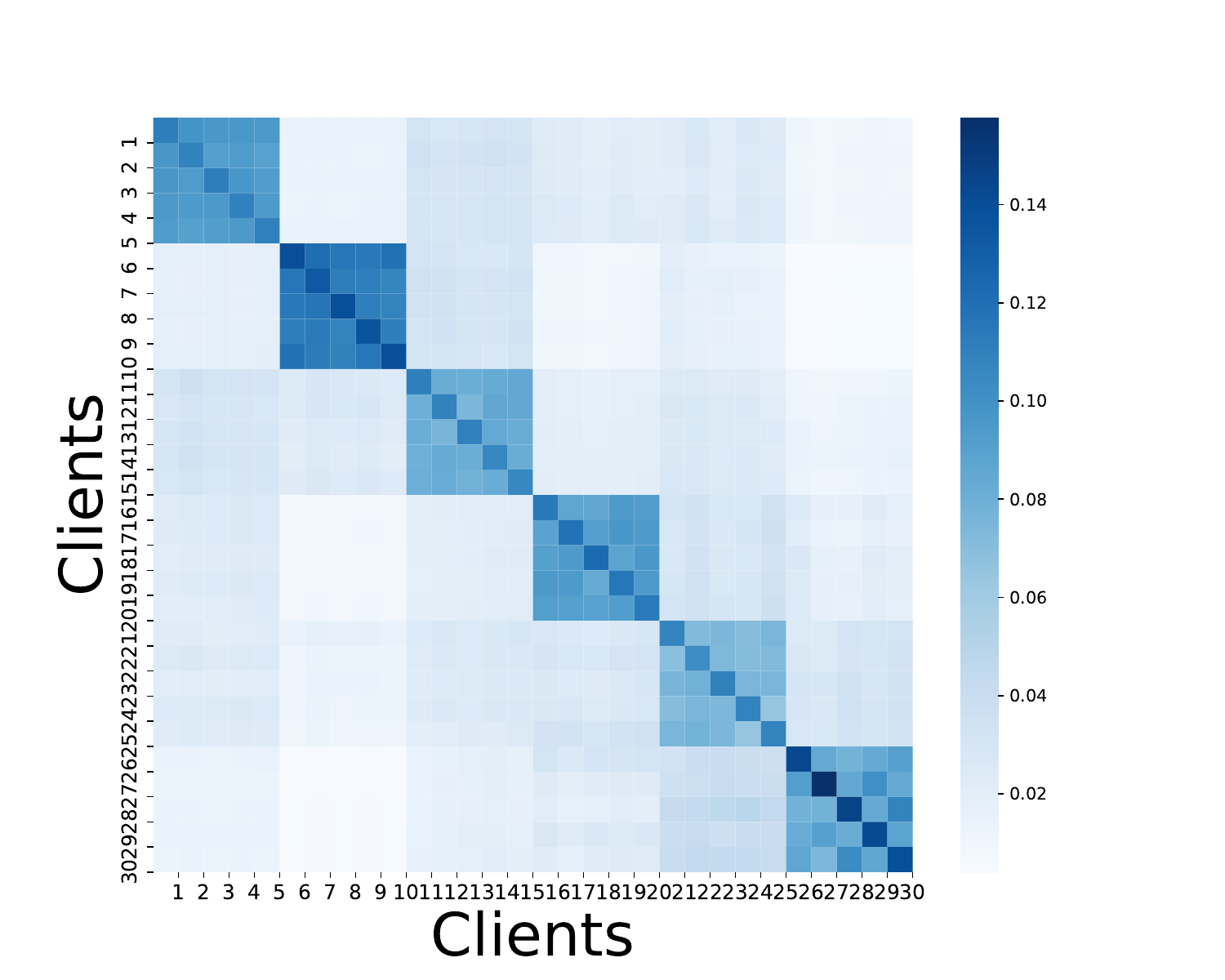}
        \caption{FedIIH of the 1st latent factor ($K=2$)}
        \label{fig_ogbn-arxiv_04}
    \end{subfigure}
    \hfill
    \begin{subfigure}[t]{0.15\textwidth}
        \includegraphics[width=\linewidth]{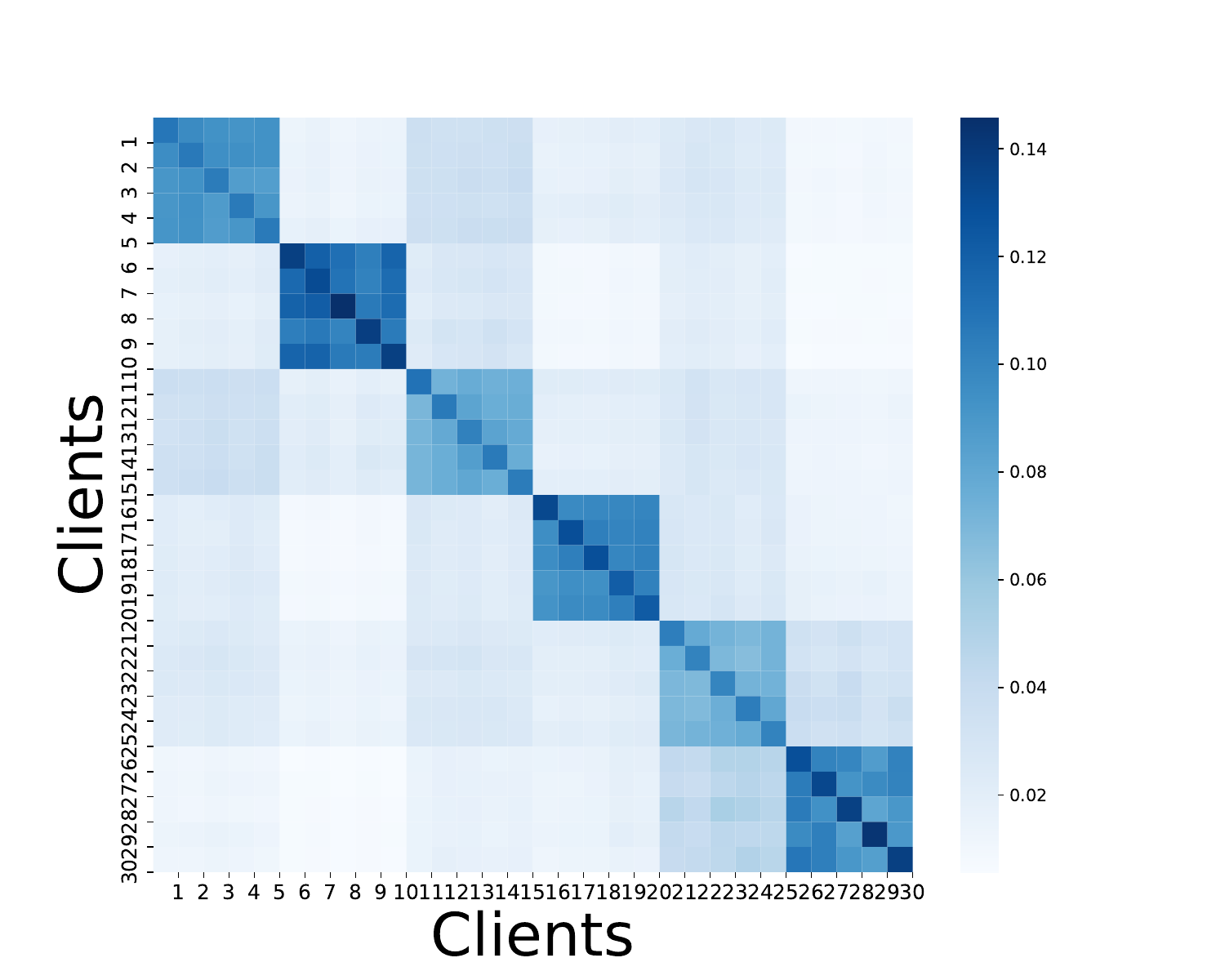}
        \caption{FedIIH of the 2nd latent factor ($K=2$)}
        \label{fig_ogbn-arxiv_05}
    \end{subfigure}
    \caption{Similarity heatmaps on the \textit{ogbn-arxiv} dataset in the overlapping setting with 30 clients.}
    \label{fig_ogbn-arxiv_O}
\end{figure}

\begin{figure}[t]
    \centering
    \begin{subfigure}[t]{0.15\textwidth}
        \includegraphics[width=\linewidth]{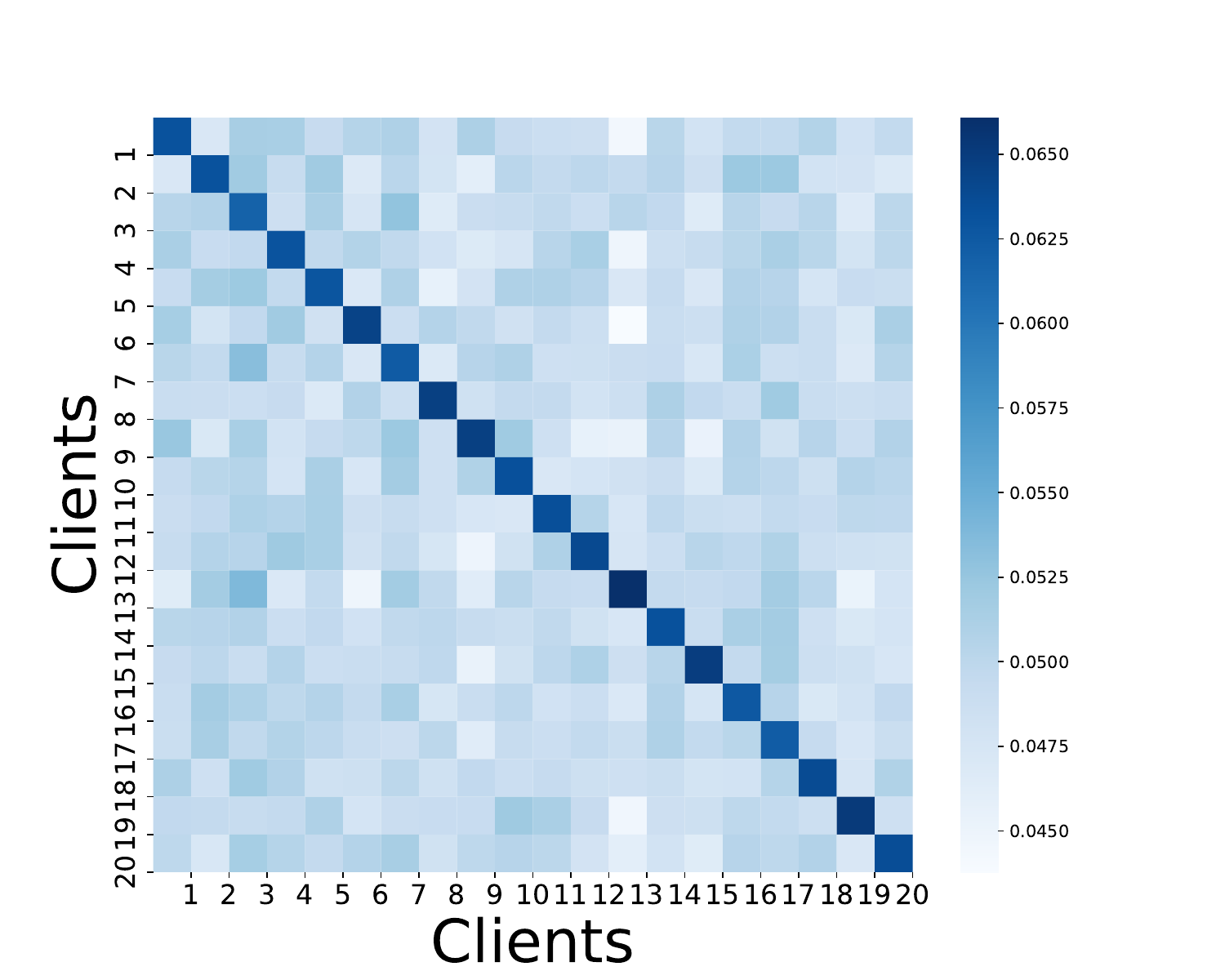}
        \caption{Distr. Sim.}
        \label{fig_Roman-empire_D1}
    \end{subfigure}%
    \hfill
    \begin{subfigure}[t]{0.15\textwidth}
        \includegraphics[width=\linewidth]{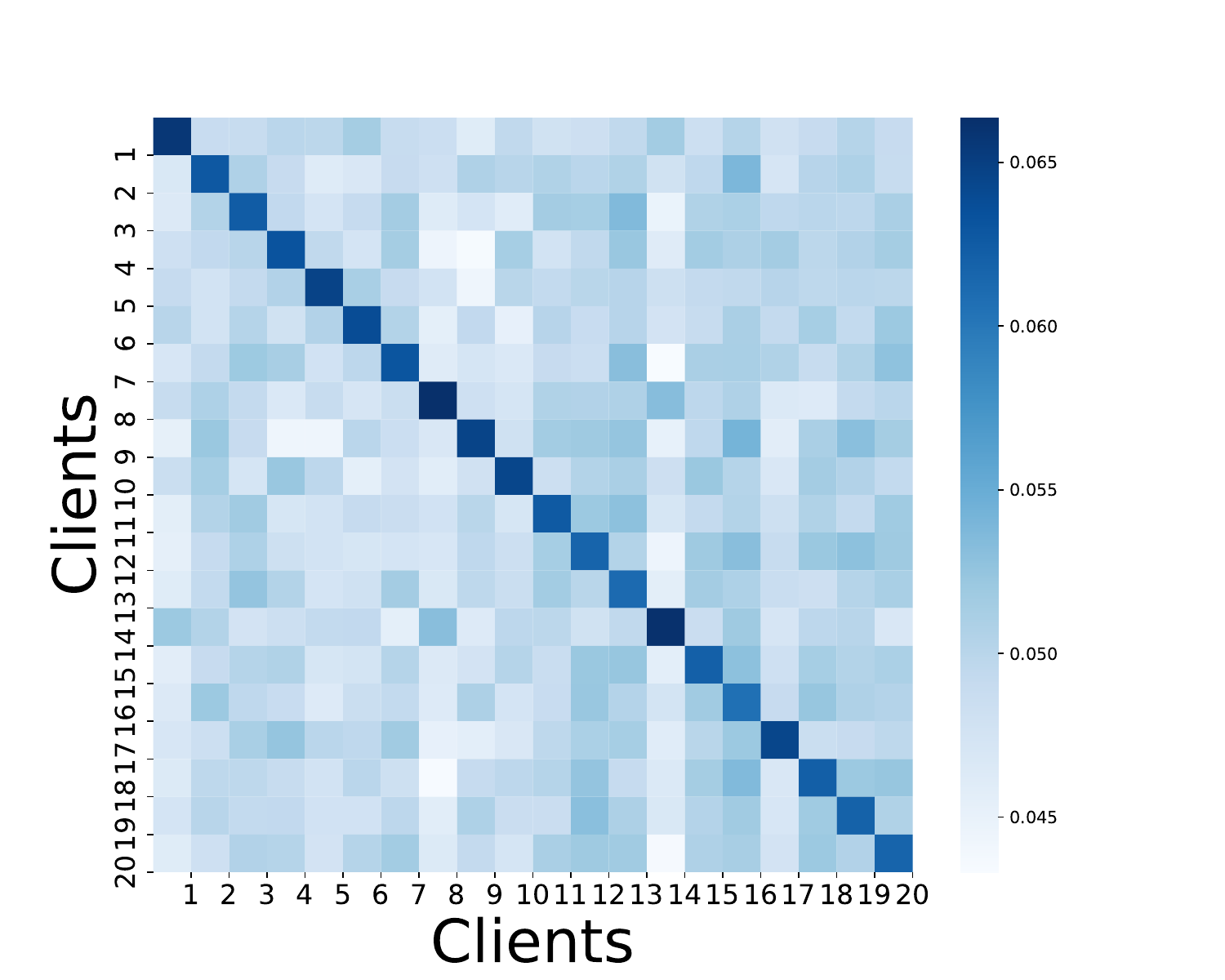}
        \caption{FED-PUB}
        \label{fig_Roman-empire_D2}
    \end{subfigure}%
    \hfill
    \begin{subfigure}[t]{0.15\textwidth}
        \includegraphics[width=\linewidth]{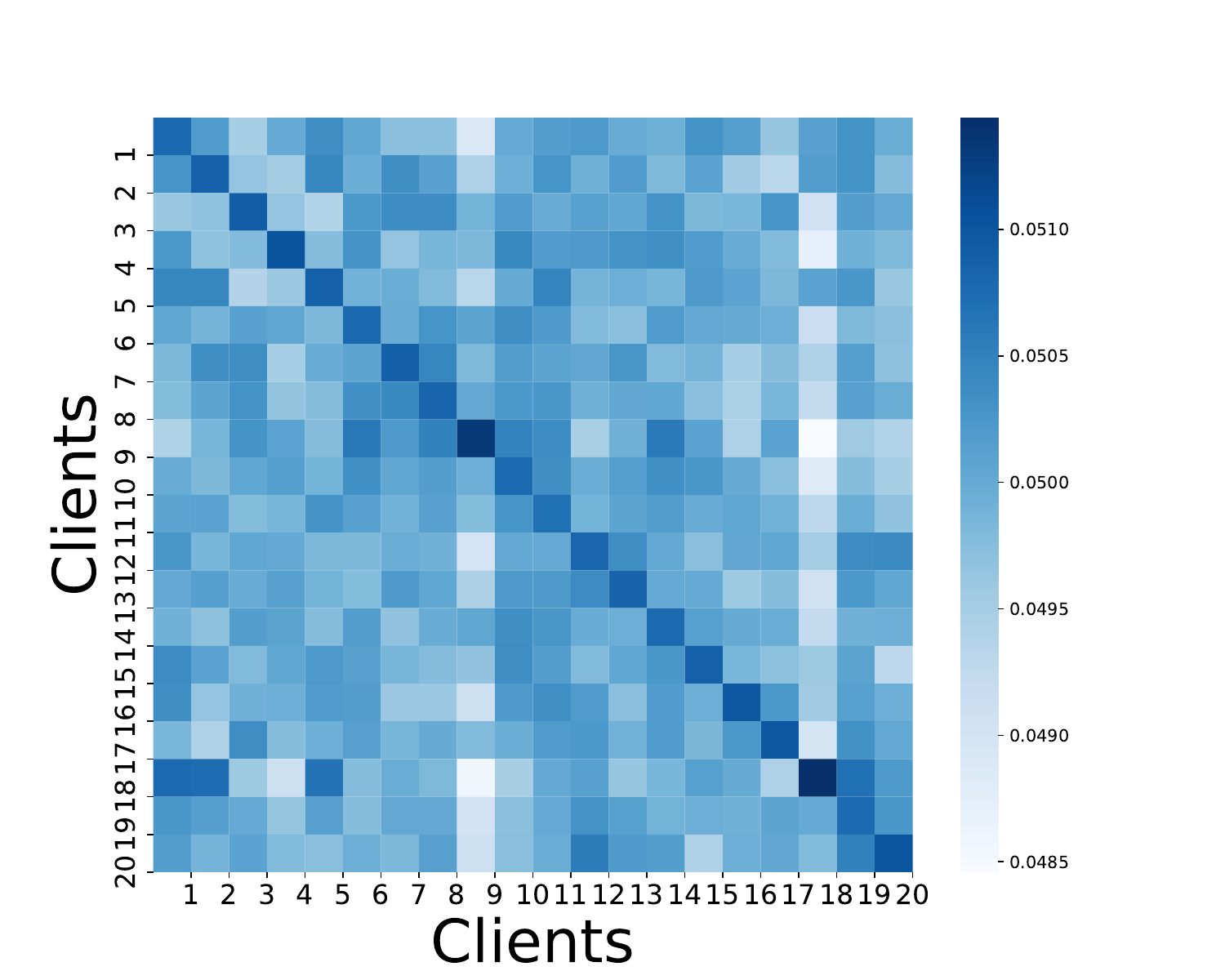}
        \caption{FedGTA}
        \label{fig_Roman-empire_D3}
    \end{subfigure}%
    \hfill
    \begin{subfigure}[t]{0.15\textwidth}
        \includegraphics[width=\linewidth]{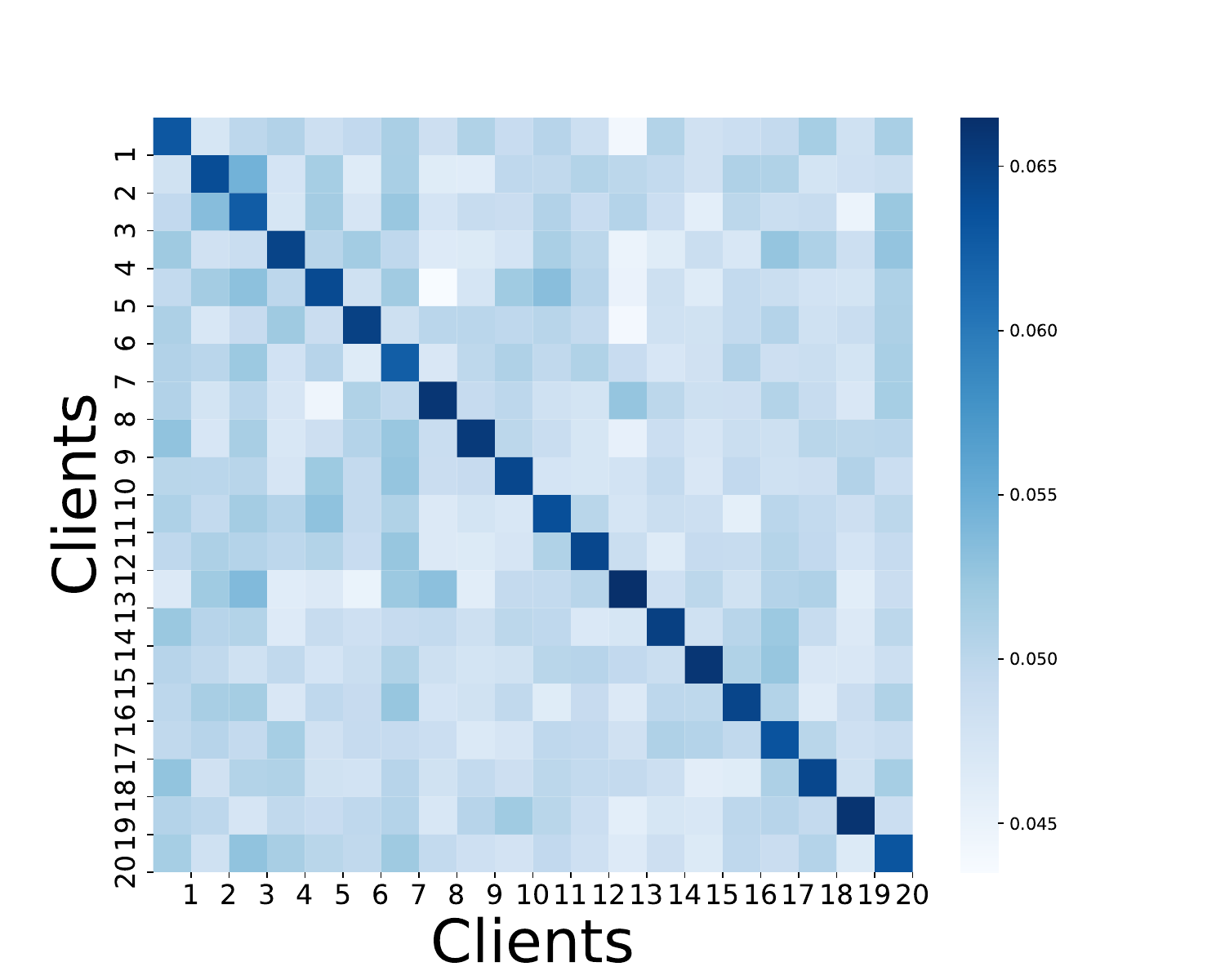}
        \caption{FedIIH of the 1st latent factor ($K=2$)}
        \label{fig_Roman-empire_D4}
    \end{subfigure}
    \hfill
    \begin{subfigure}[t]{0.15\textwidth}
        \includegraphics[width=\linewidth]{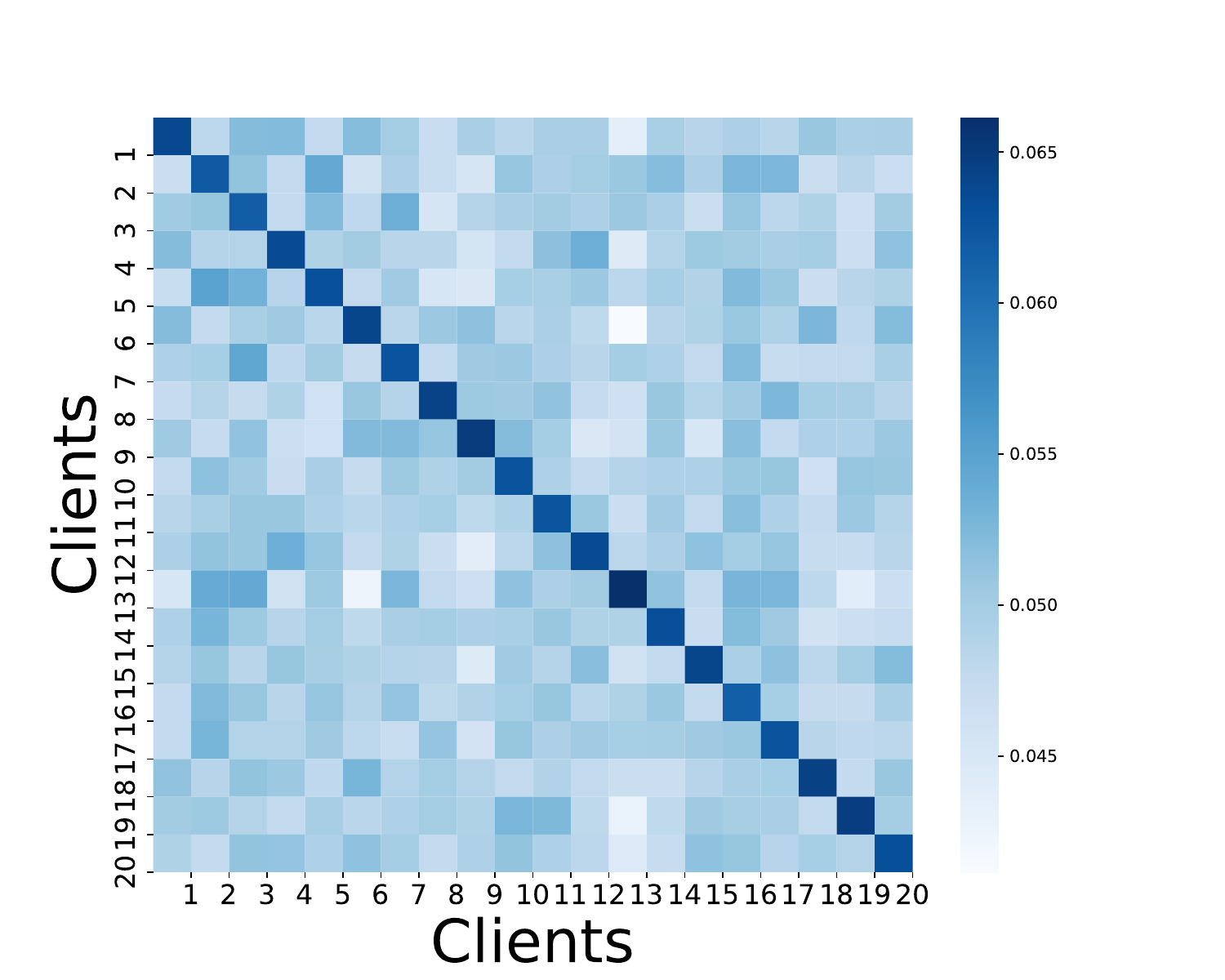}
        \caption{FedIIH of the 2nd latent factor ($K=2$)}
        \label{fig_Roman-empire_D5}
    \end{subfigure}
    \caption{Similarity heatmaps on the \textit{Roman-empire} dataset in the non-overlapping setting with 20 clients.}
    \label{fig_Roman-empire_D}
\end{figure}

\begin{figure}[t]
    \centering
    \begin{subfigure}[t]{0.15\textwidth}
        \includegraphics[width=\linewidth]{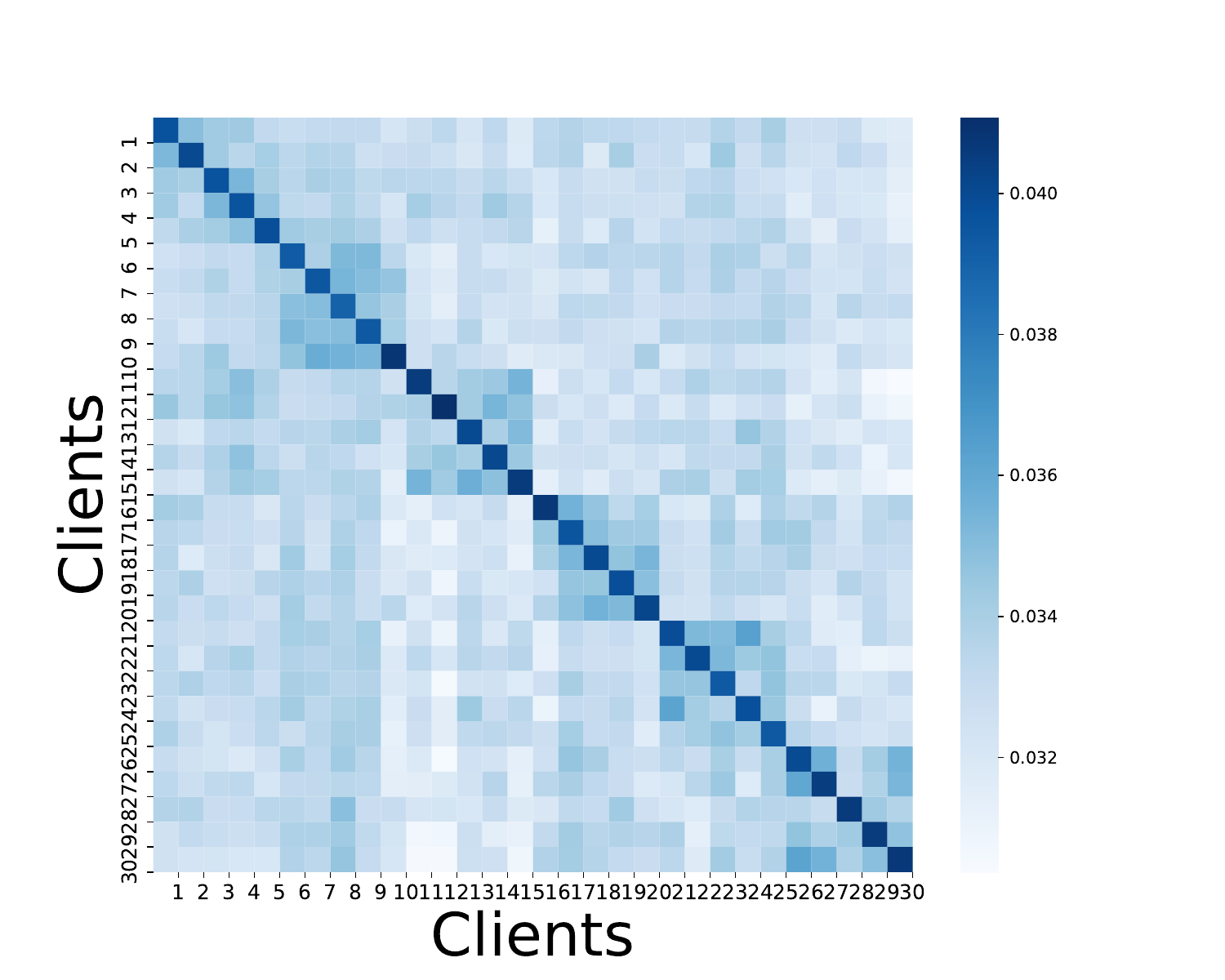}
        \caption{Distr. Sim.}
        \label{fig_Roman-empire_01}
    \end{subfigure}%
    \hfill
    \begin{subfigure}[t]{0.15\textwidth}
        \includegraphics[width=\linewidth]{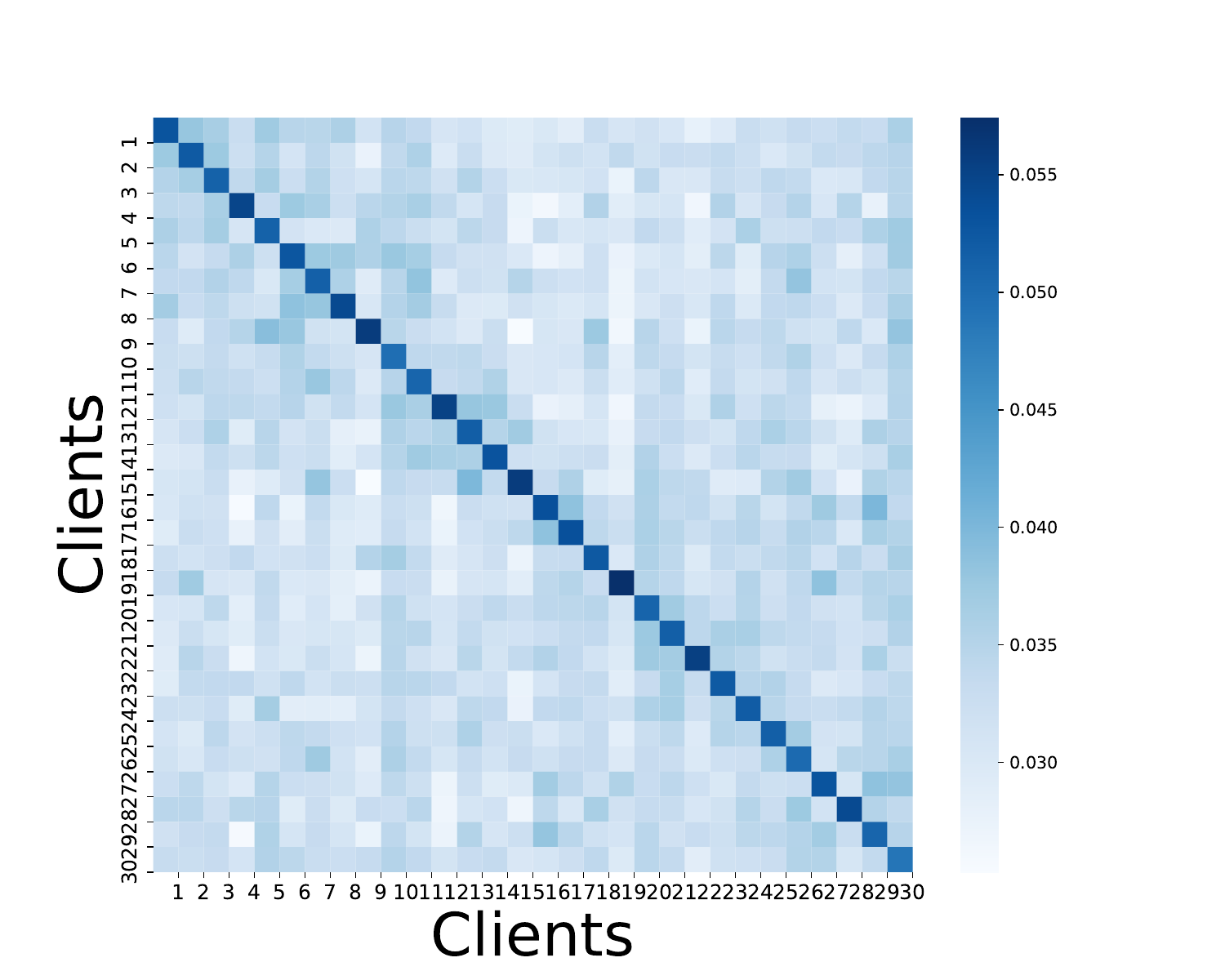}
        \caption{FED-PUB}
        \label{fig_Roman-empire_02}
    \end{subfigure}%
    \hfill
    \begin{subfigure}[t]{0.15\textwidth}
        \includegraphics[width=\linewidth]{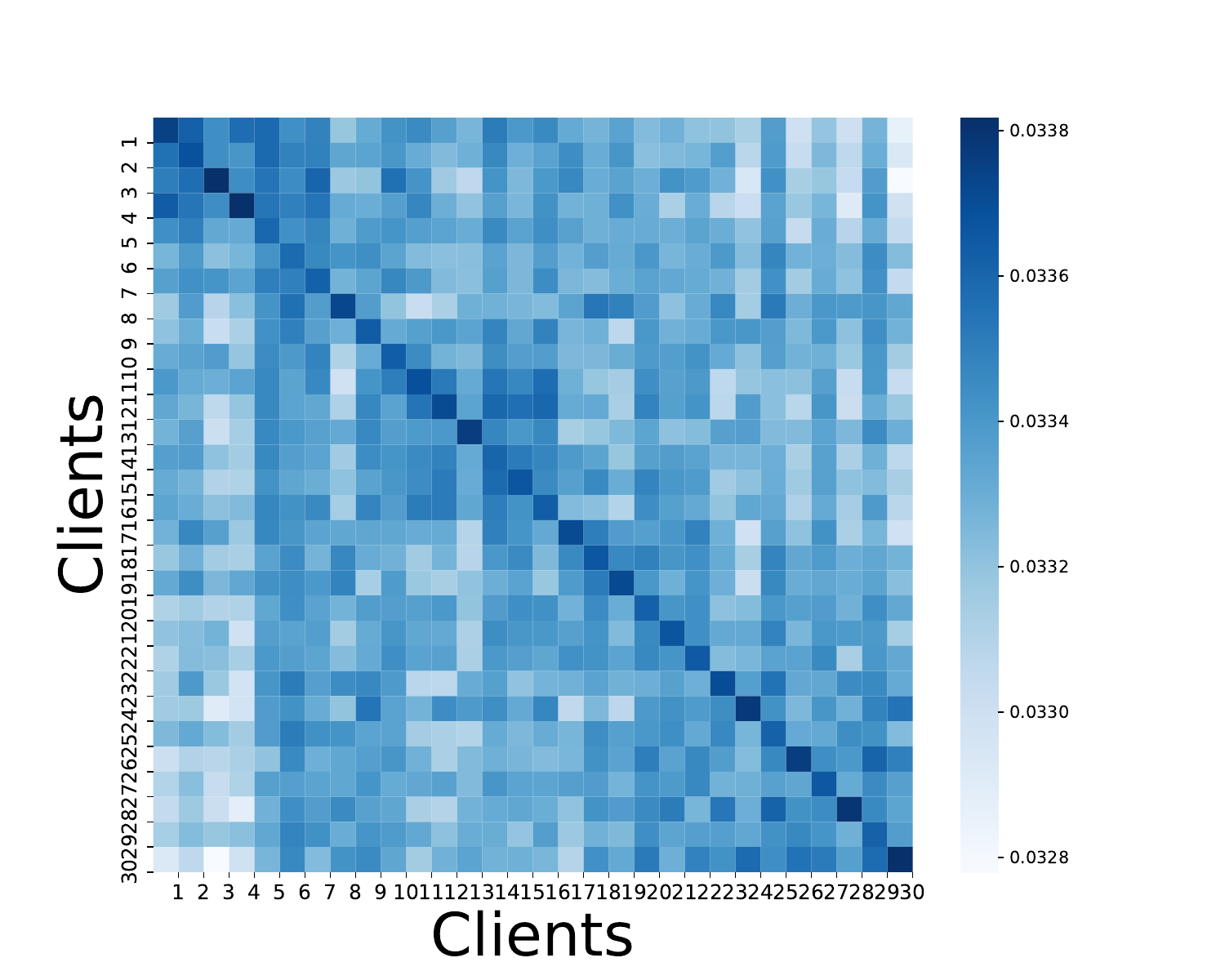}
        \caption{FedGTA}
        \label{fig_Roman-empire_03}
    \end{subfigure}%
    \hfill
    \begin{subfigure}[t]{0.15\textwidth}
        \includegraphics[width=\linewidth]{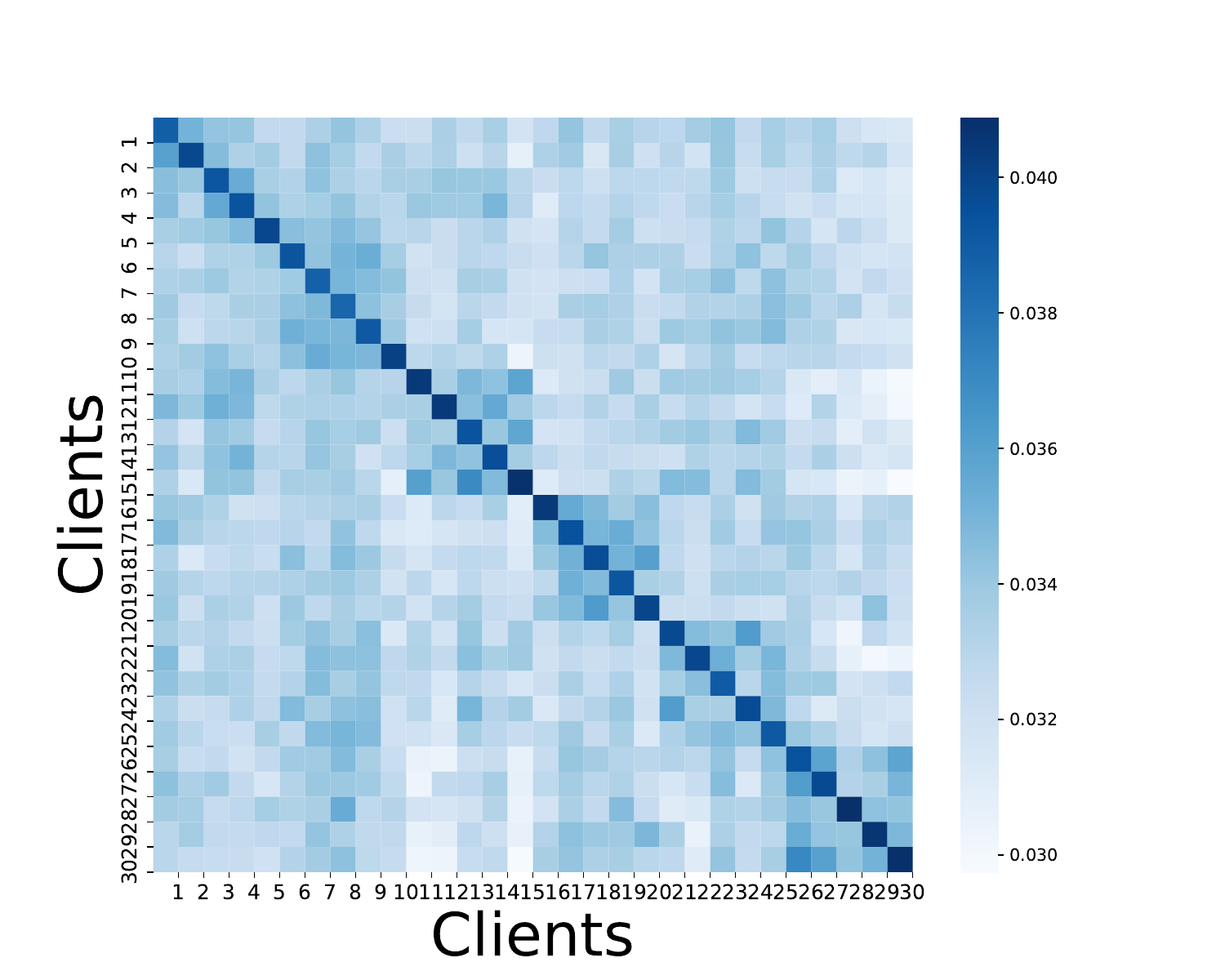}
        \caption{FedIIH of the 1st latent factor ($K=2$)}
        \label{fig_Roman-empire_04}
    \end{subfigure}
    \hfill
    \begin{subfigure}[t]{0.15\textwidth}
        \includegraphics[width=\linewidth]{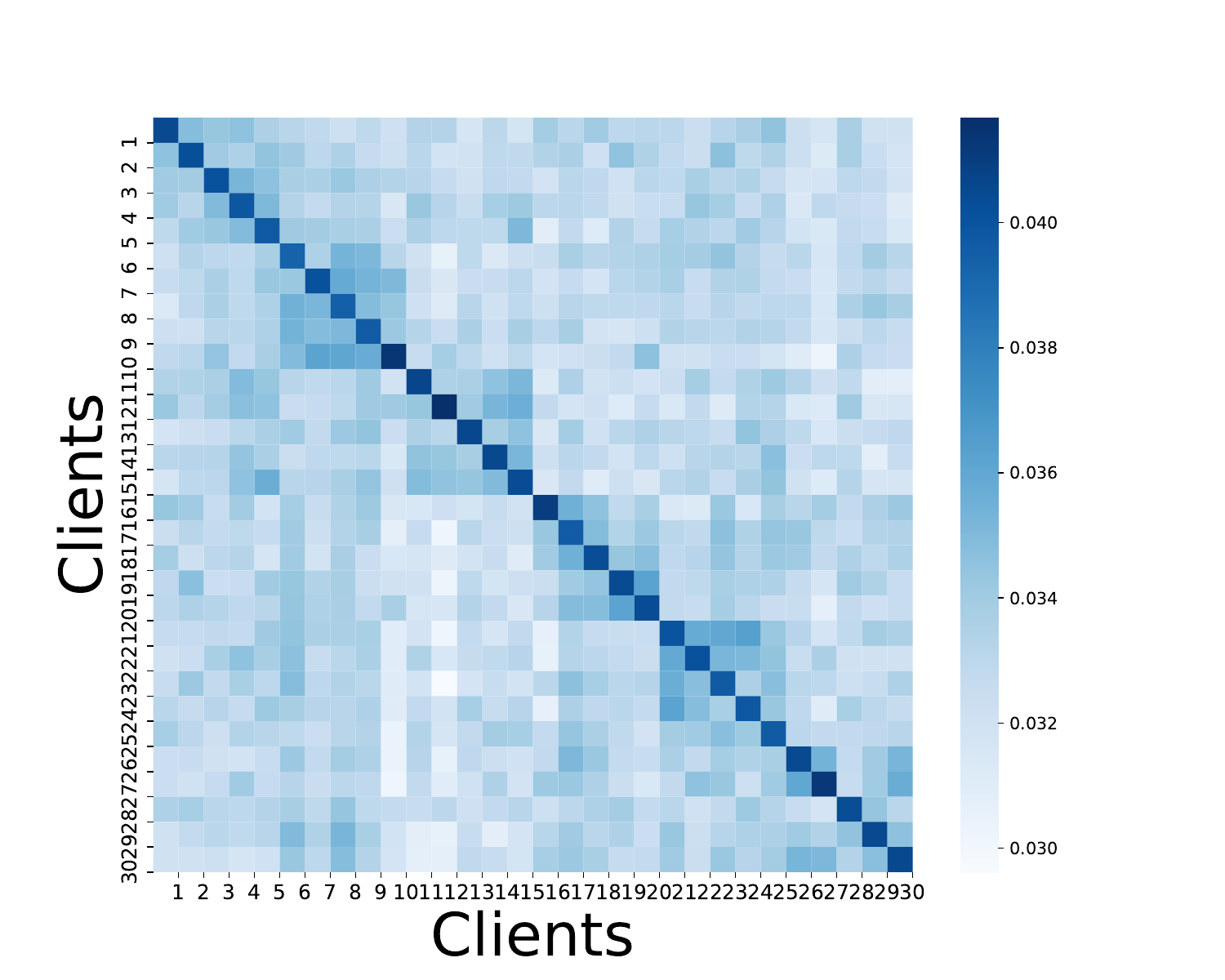}
        \caption{FedIIH of the 2nd latent factor ($K=2$)}
        \label{fig_Roman-empire_05}
    \end{subfigure}
    \caption{Similarity heatmaps on the \textit{Roman-empire} dataset in the overlapping setting with 30 clients.}
    \label{fig_Roman-empire_O}
\end{figure}

\begin{figure}[t]
    \centering
    \begin{subfigure}[t]{0.15\textwidth}
        \includegraphics[width=\linewidth]{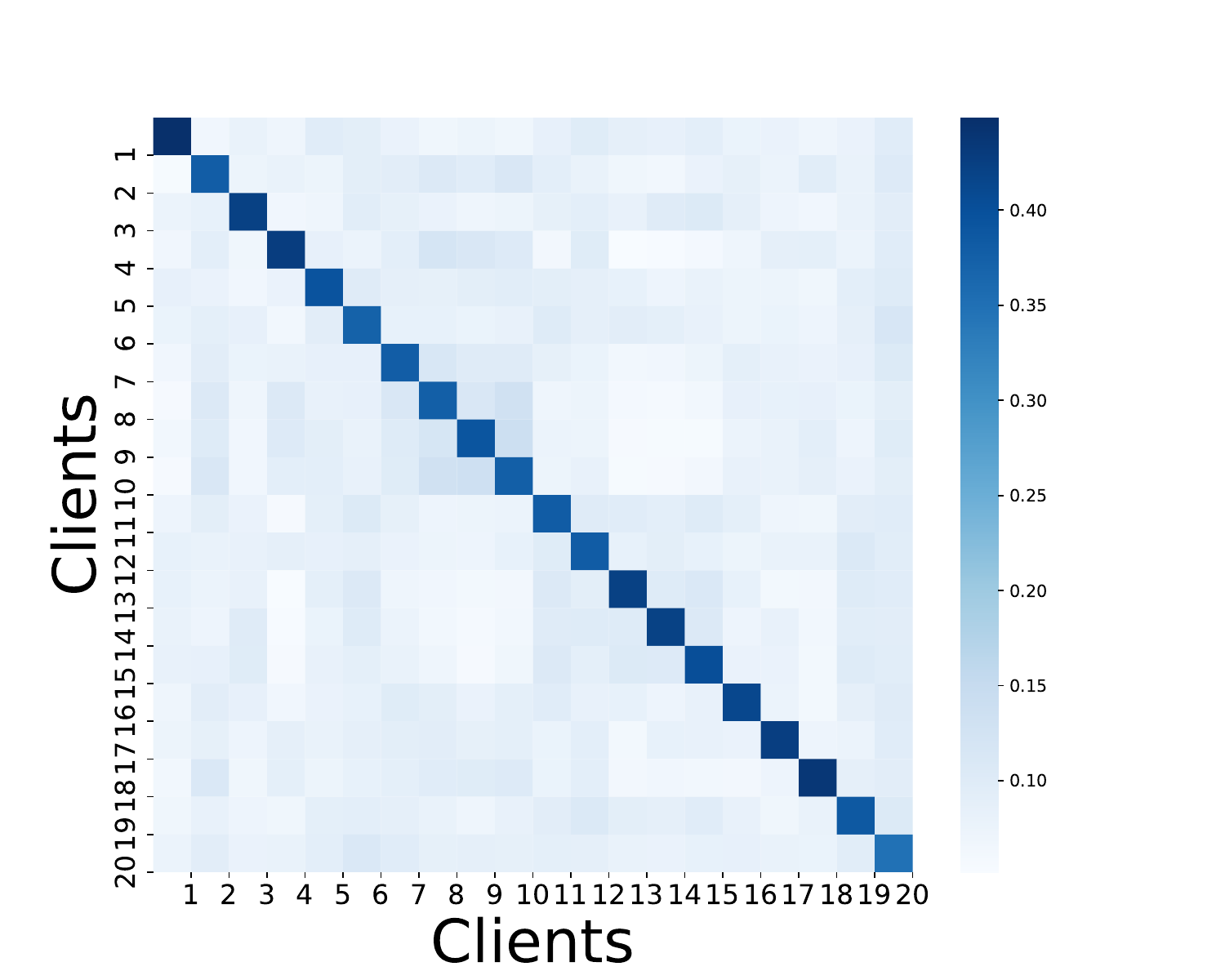}
        \caption{Distr. Sim.}
        \label{fig_Amazon-ratings_D1}
    \end{subfigure}%
    \hfill
    \begin{subfigure}[t]{0.15\textwidth}
        \includegraphics[width=\linewidth]{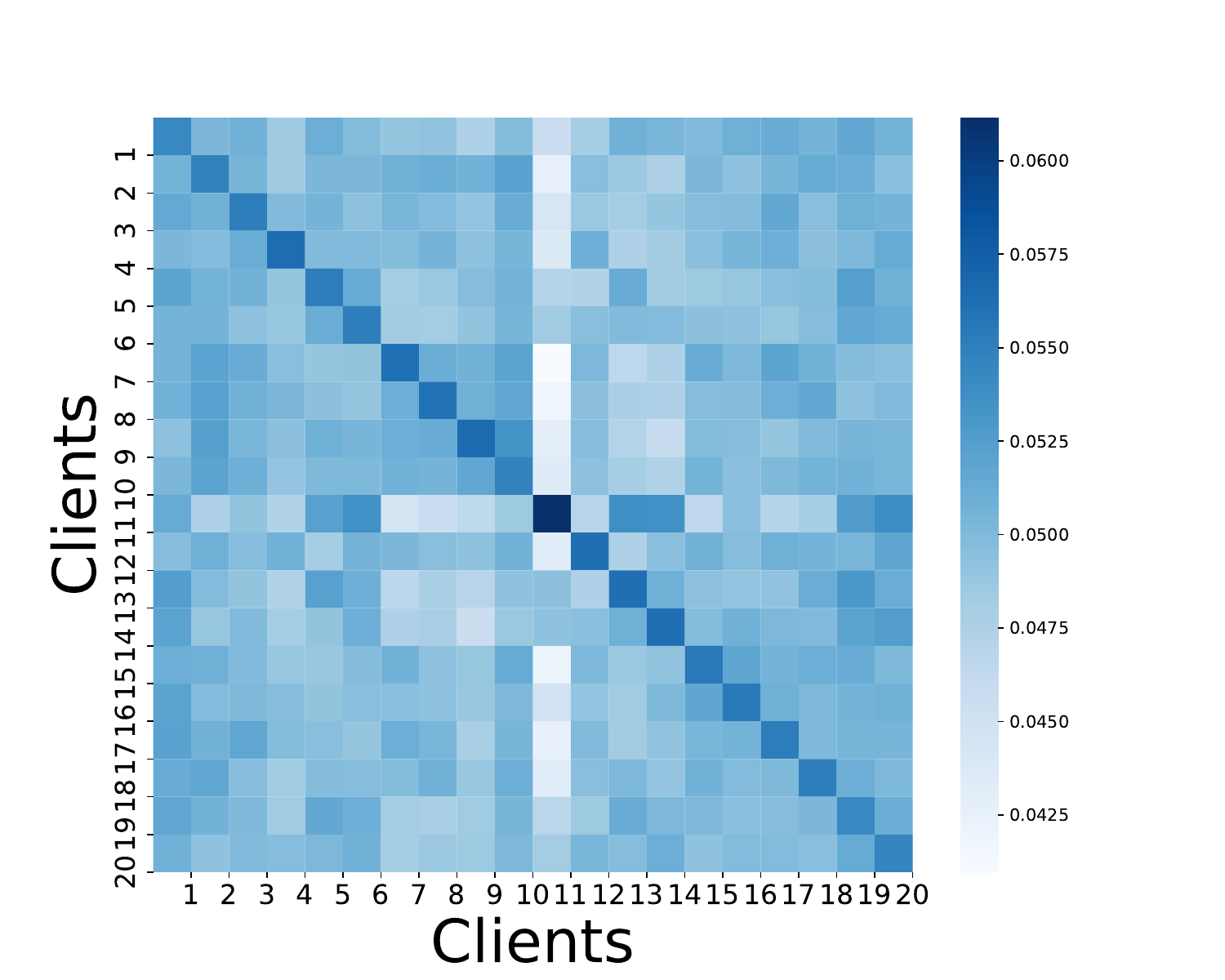}
        \caption{FED-PUB}
        \label{fig_Amazon-ratings_D2}
    \end{subfigure}%
    \hfill
    \begin{subfigure}[t]{0.15\textwidth}
        \includegraphics[width=\linewidth]{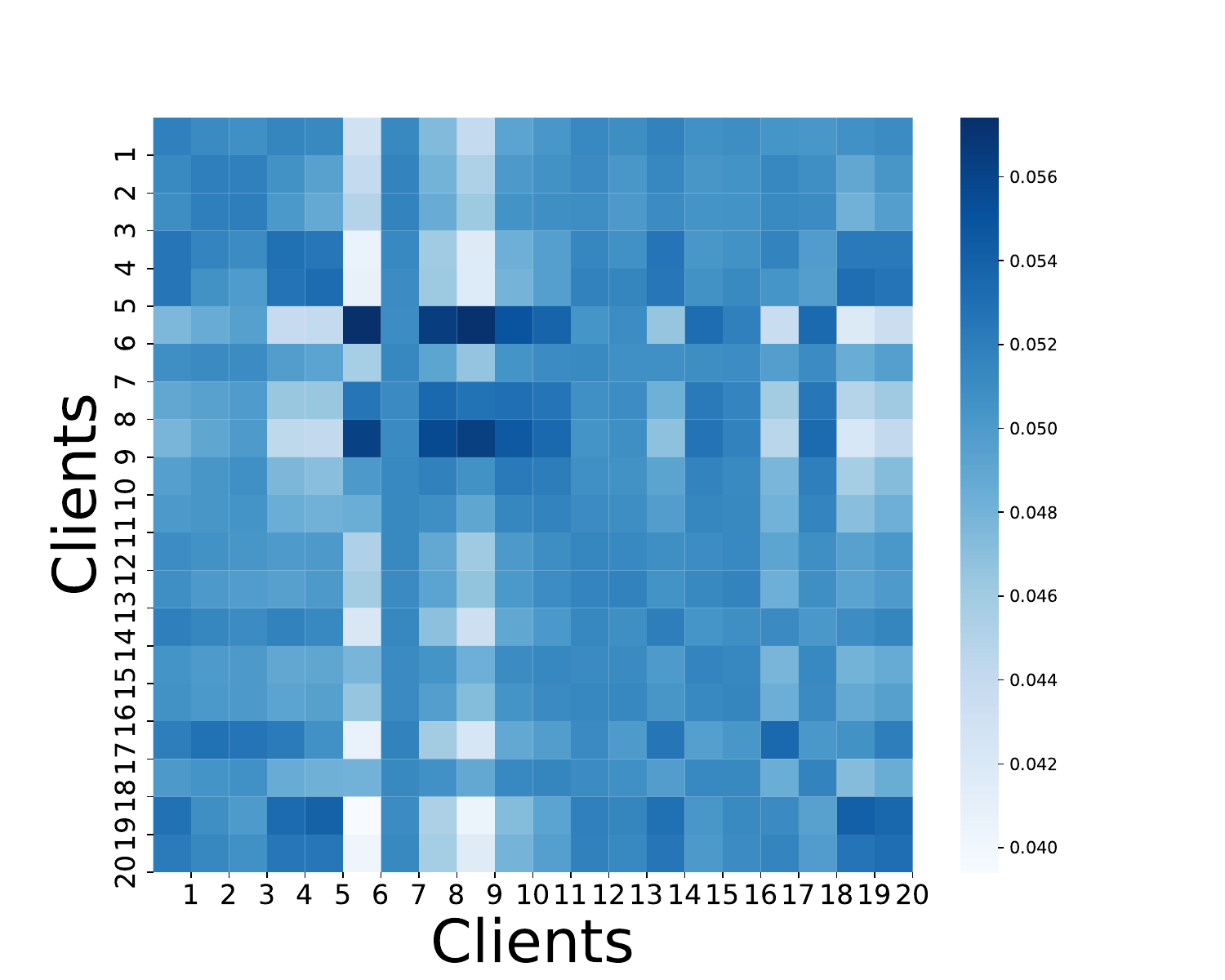}
        \caption{FedGTA}
        \label{fig_Amazon-ratings_D3}
    \end{subfigure}%
    \hfill
    \begin{subfigure}[t]{0.15\textwidth}
        \includegraphics[width=\linewidth]{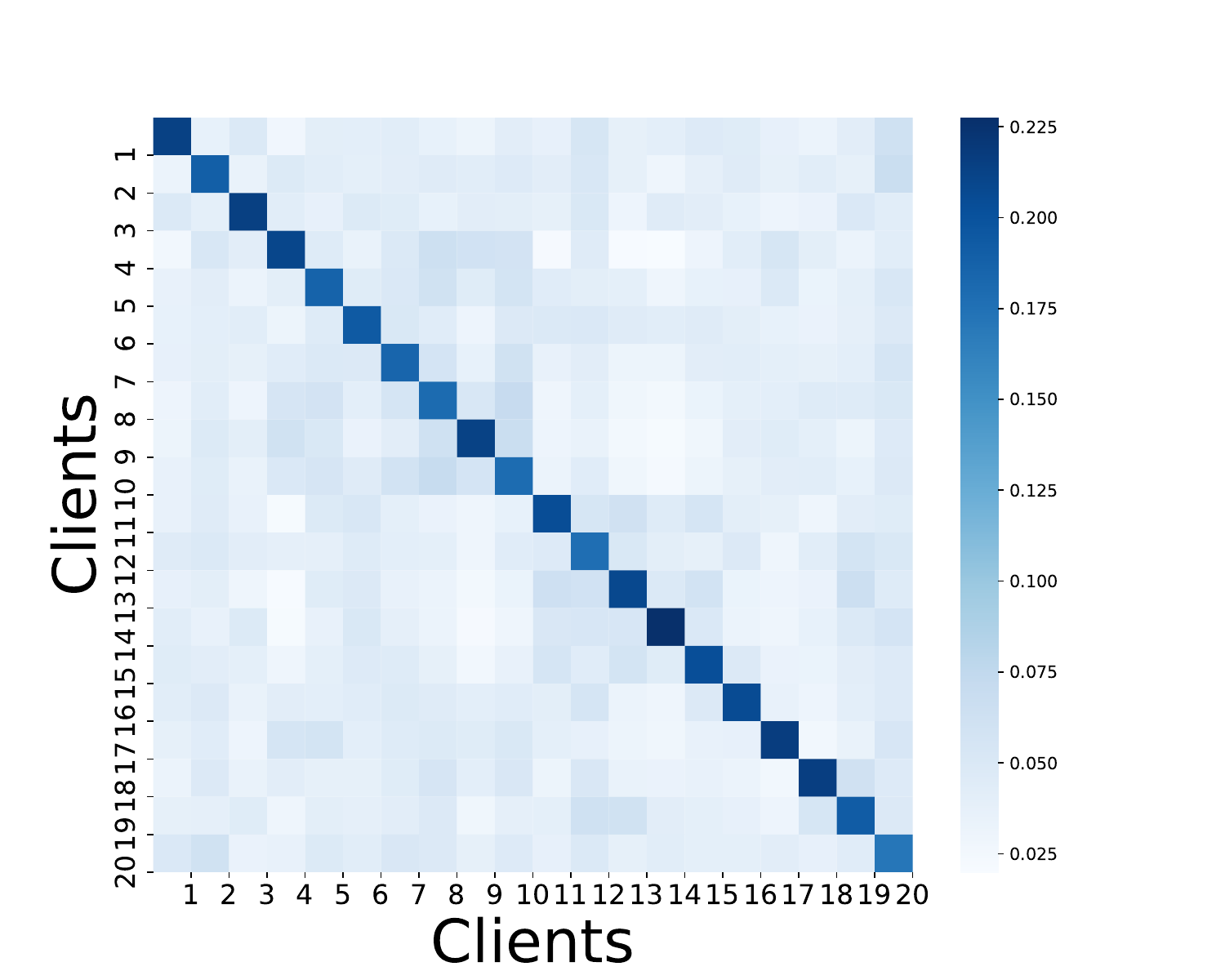}
        \caption{FedIIH of the 1st latent factor ($K=2$)}
        \label{fig_Amazon-ratings_D4}
    \end{subfigure}
    \hfill
    \begin{subfigure}[t]{0.15\textwidth}
        \includegraphics[width=\linewidth]{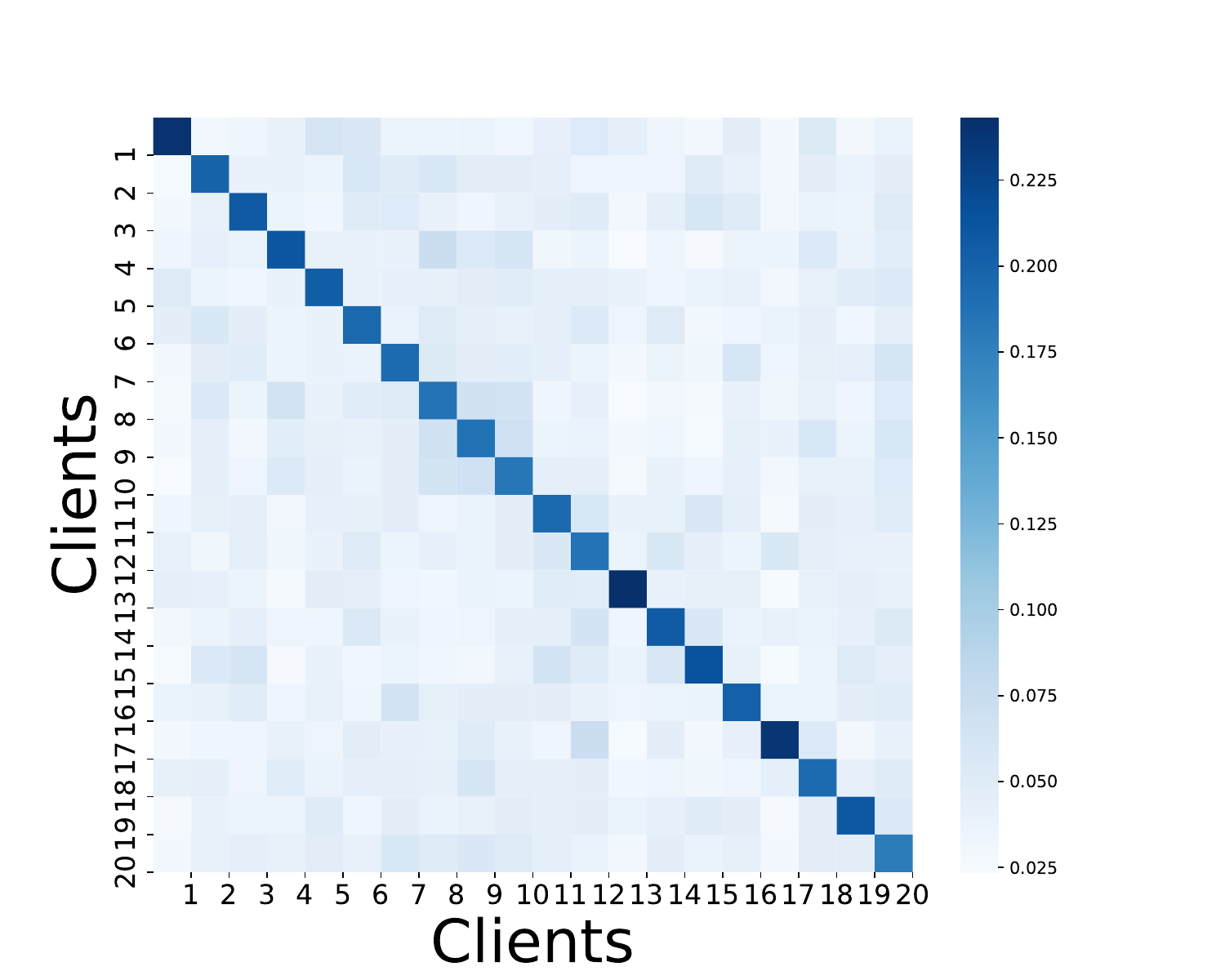}
        \caption{FedIIH of the 2nd latent factor ($K=2$)}
        \label{fig_Amazon-ratings_D5}
    \end{subfigure}
    \caption{Similarity heatmaps on the \textit{Amazon-ratings} dataset in the non-overlapping setting with 20 clients.}
    \label{fig_Amazon-ratings_D}
\end{figure}

\begin{figure}[t]
    \centering
    \begin{subfigure}[t]{0.15\textwidth}
        \includegraphics[width=\linewidth]{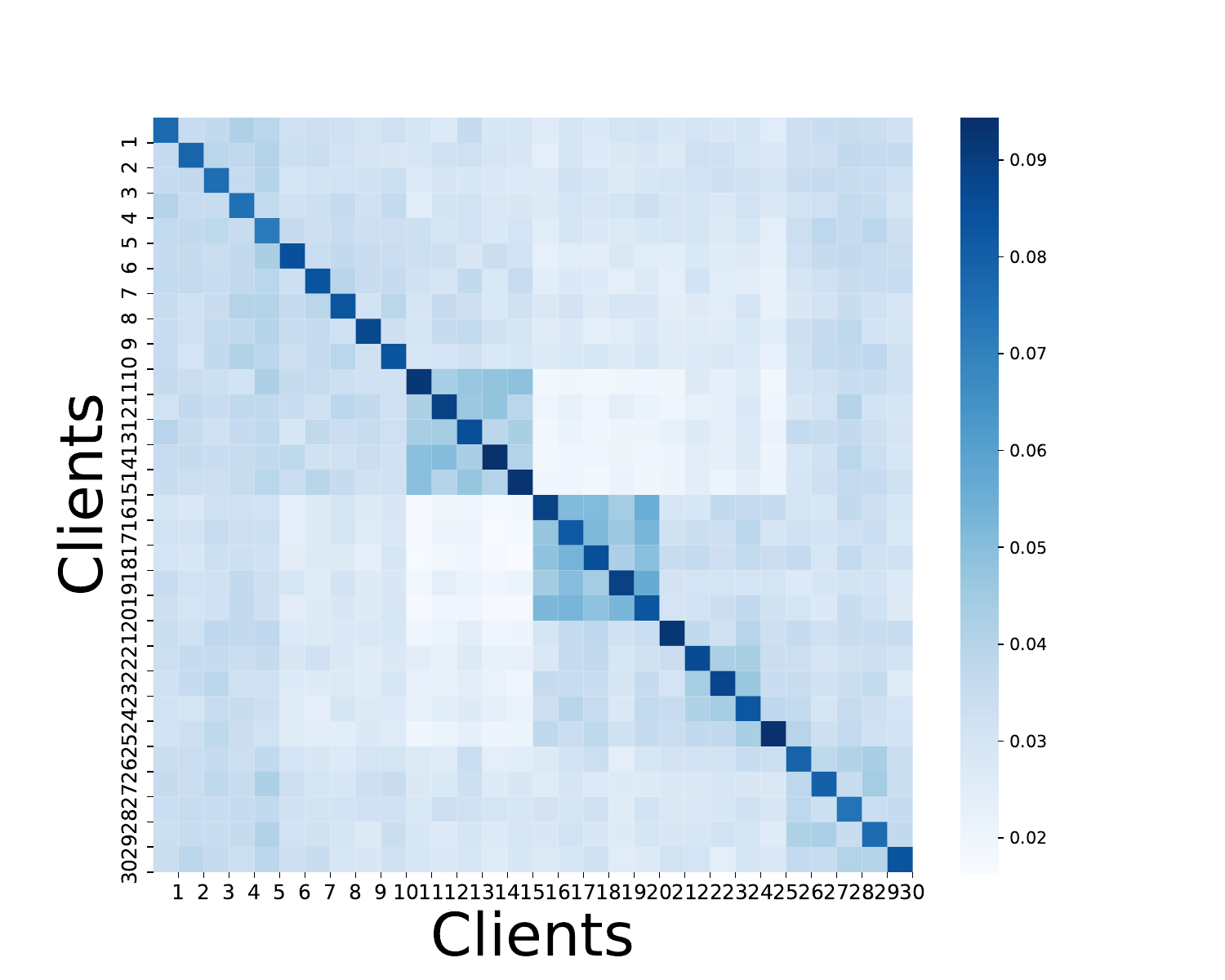}
        \caption{Distr. Sim.}
        \label{fig_Amazon-ratings_01}
    \end{subfigure}%
    \hfill
    \begin{subfigure}[t]{0.15\textwidth}
        \includegraphics[width=\linewidth]{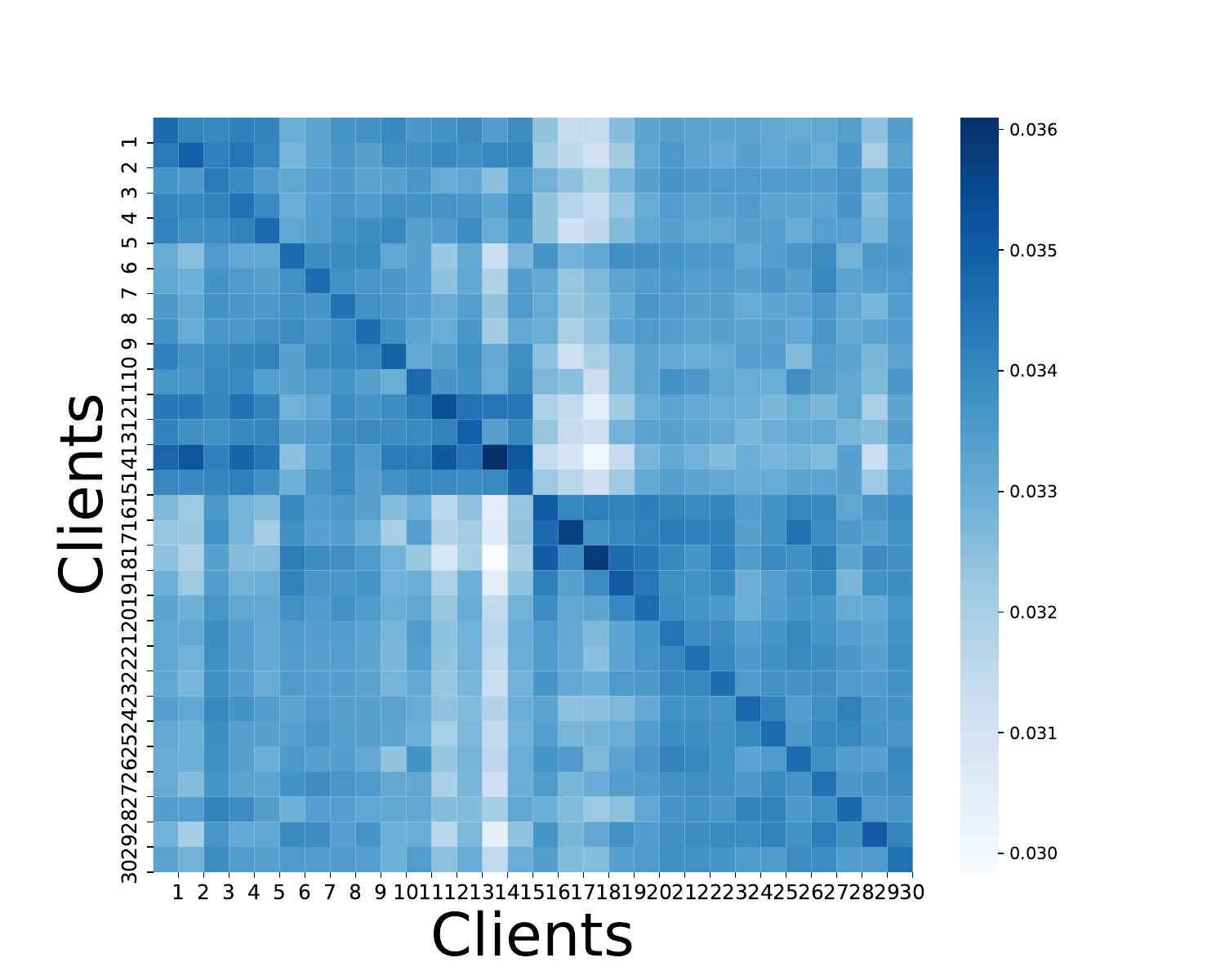}
        \caption{FED-PUB}
        \label{fig_Amazon-ratings_02}
    \end{subfigure}%
    \hfill
    \begin{subfigure}[t]{0.15\textwidth}
        \includegraphics[width=\linewidth]{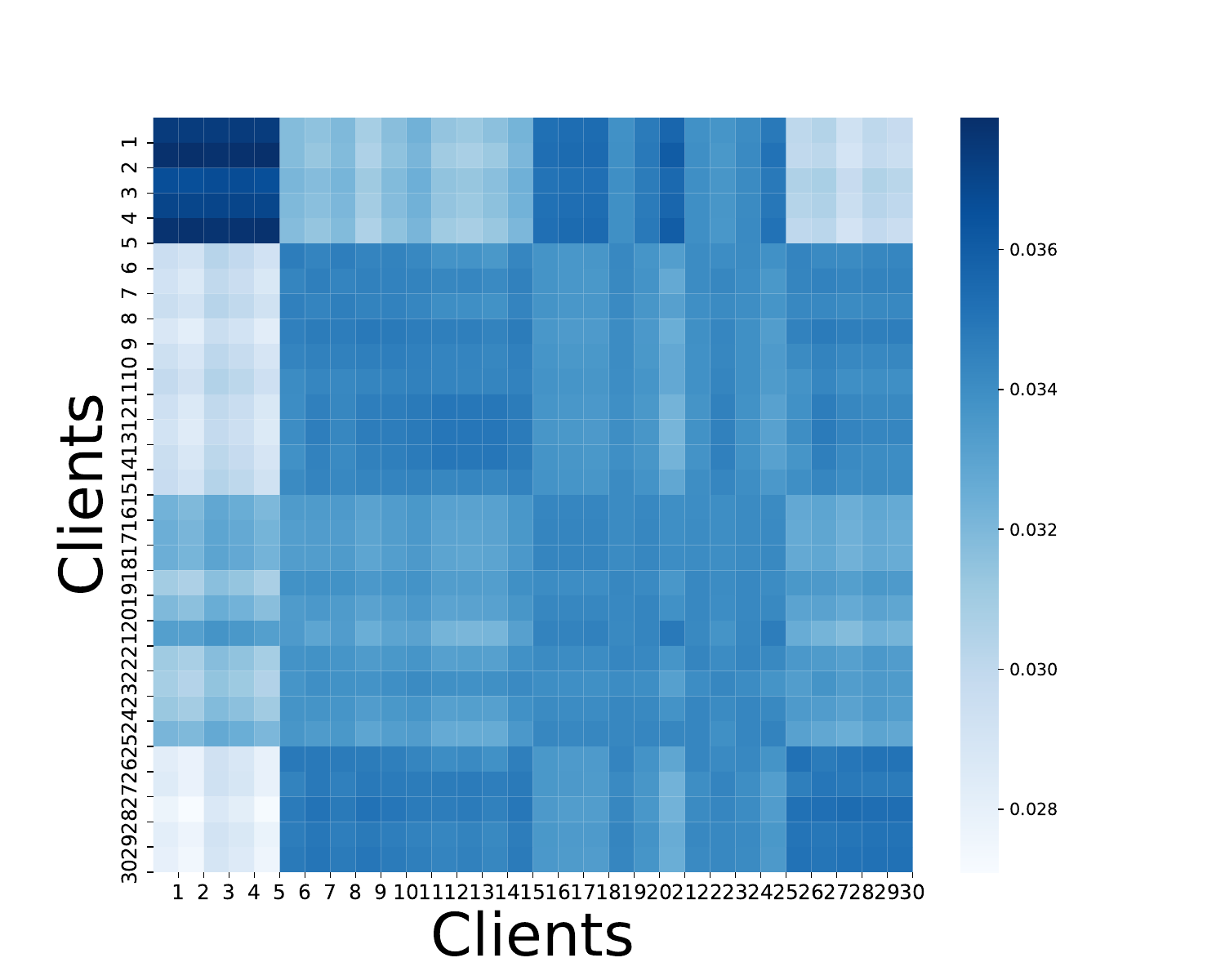}
        \caption{FedGTA}
        \label{fig_Amazon-ratings_03}
    \end{subfigure}%
    \hfill
    \begin{subfigure}[t]{0.15\textwidth}
        \includegraphics[width=\linewidth]{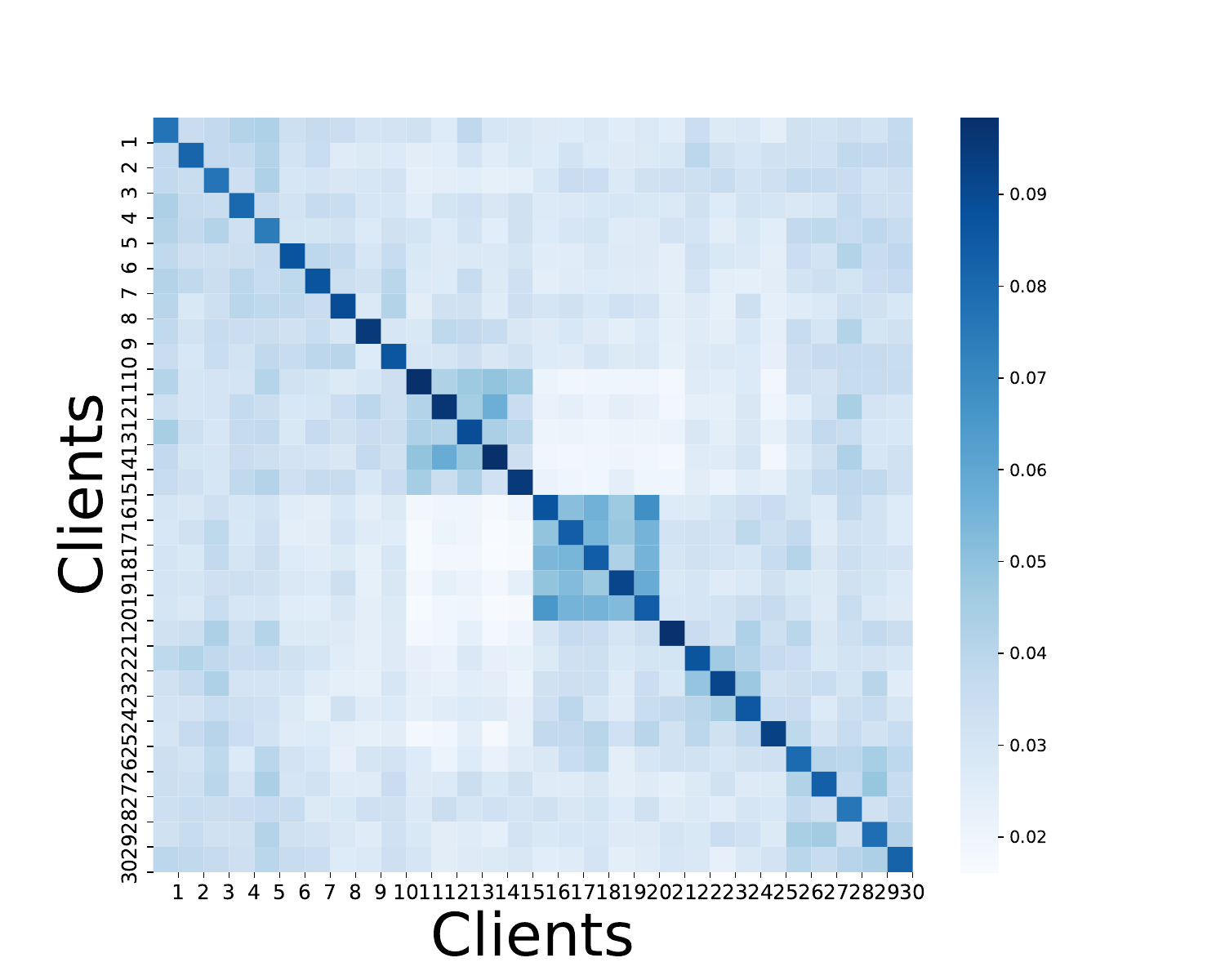}
        \caption{FedIIH of the 1st latent factor ($K=2$)}
        \label{fig_Amazon-ratings_04}
    \end{subfigure}
    \hfill
    \begin{subfigure}[t]{0.15\textwidth}
        \includegraphics[width=\linewidth]{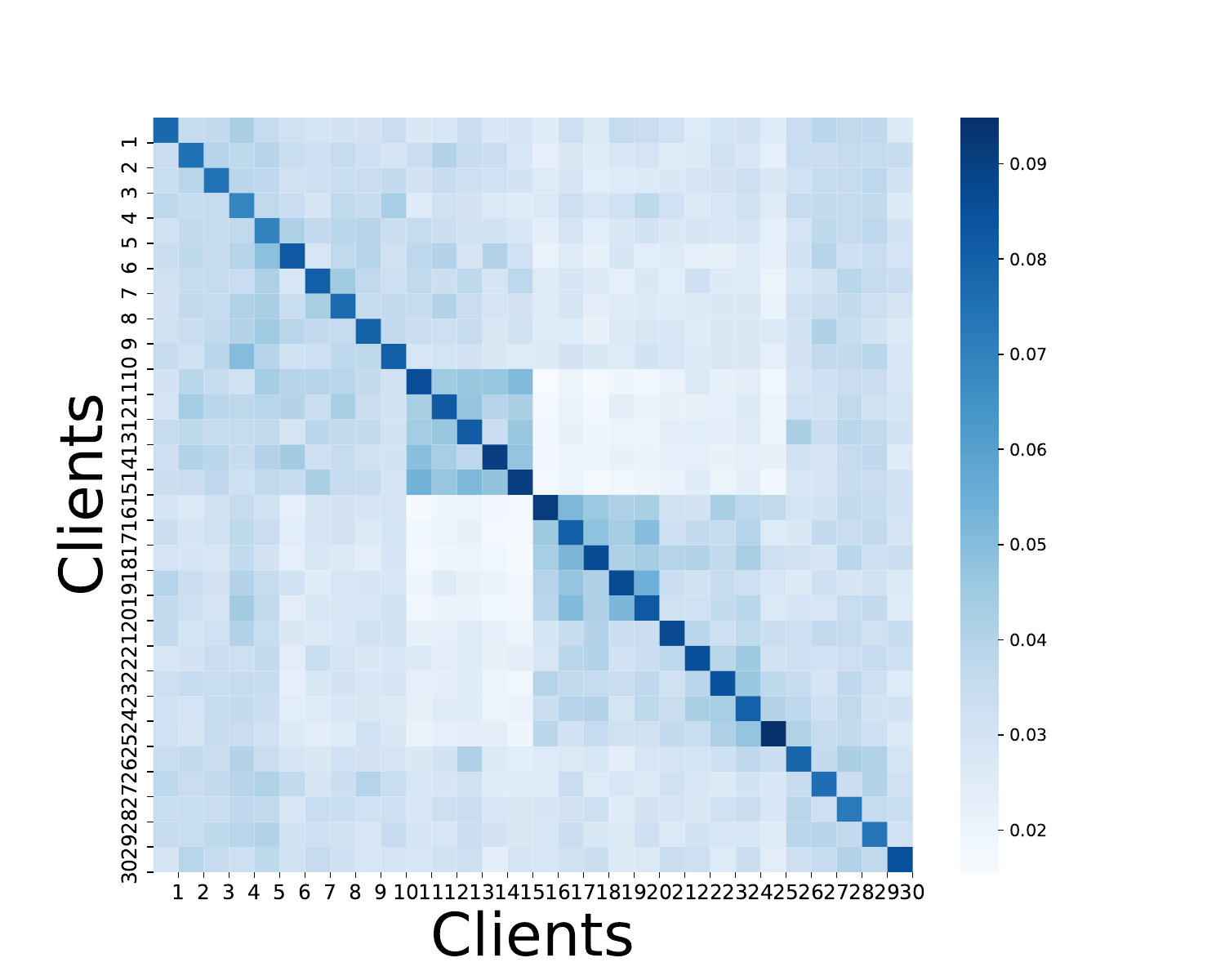}
        \caption{FedIIH of the 2nd latent factor ($K=2$)}
        \label{fig_Amazon-ratings_05}
    \end{subfigure}
    \caption{Similarity heatmaps on the \textit{Amazon-ratings} dataset in the overlapping setting with 30 clients.}
    \label{fig_Amazon-ratings_O}
\end{figure}

\begin{figure}[t]
    \centering
    \begin{subfigure}[t]{0.15\textwidth}
        \includegraphics[width=\linewidth]{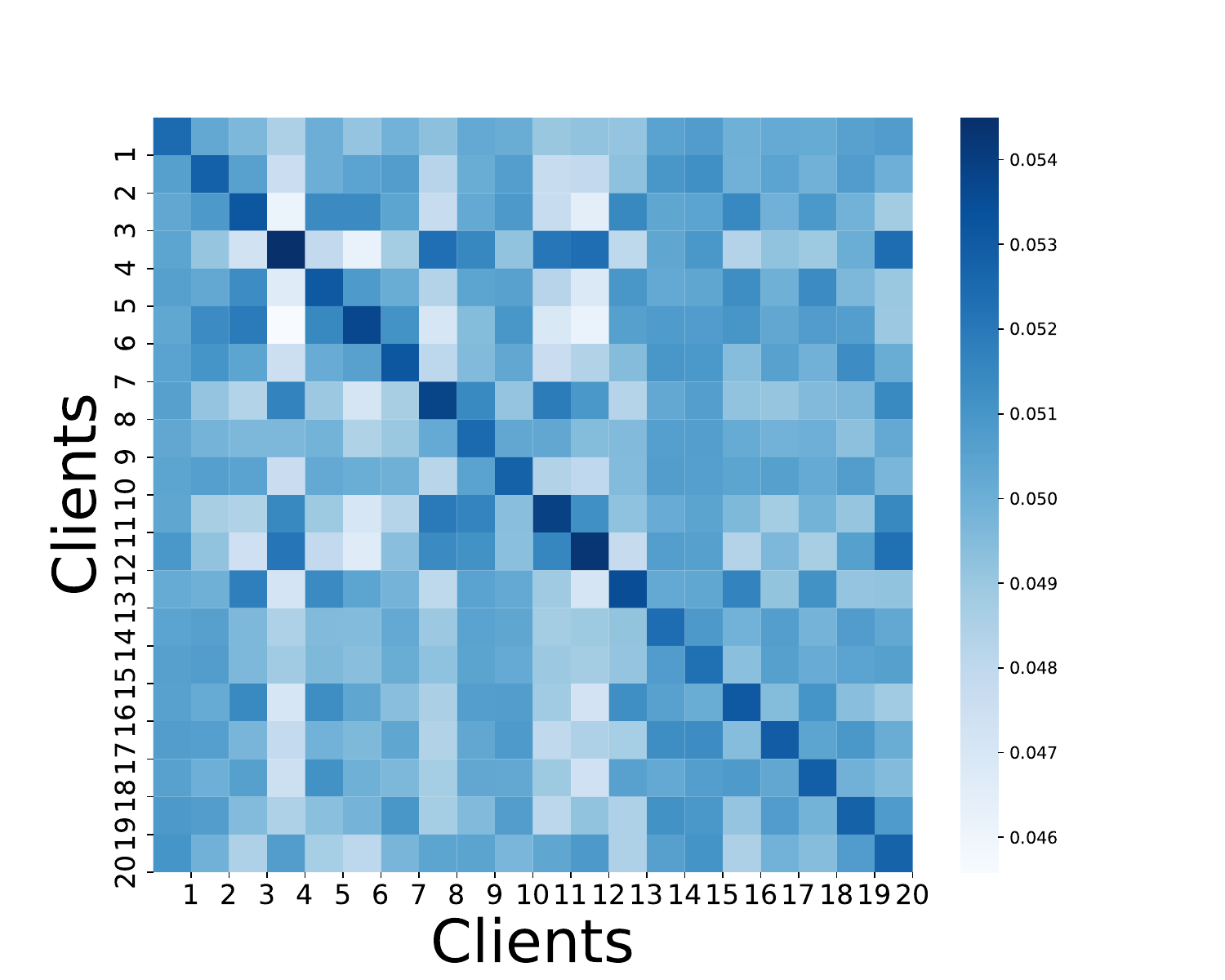}
        \caption{Distr. Sim.}
        \label{fig_Minesweeper_D1}
    \end{subfigure}%
    \hfill
    \begin{subfigure}[t]{0.15\textwidth}
        \includegraphics[width=\linewidth]{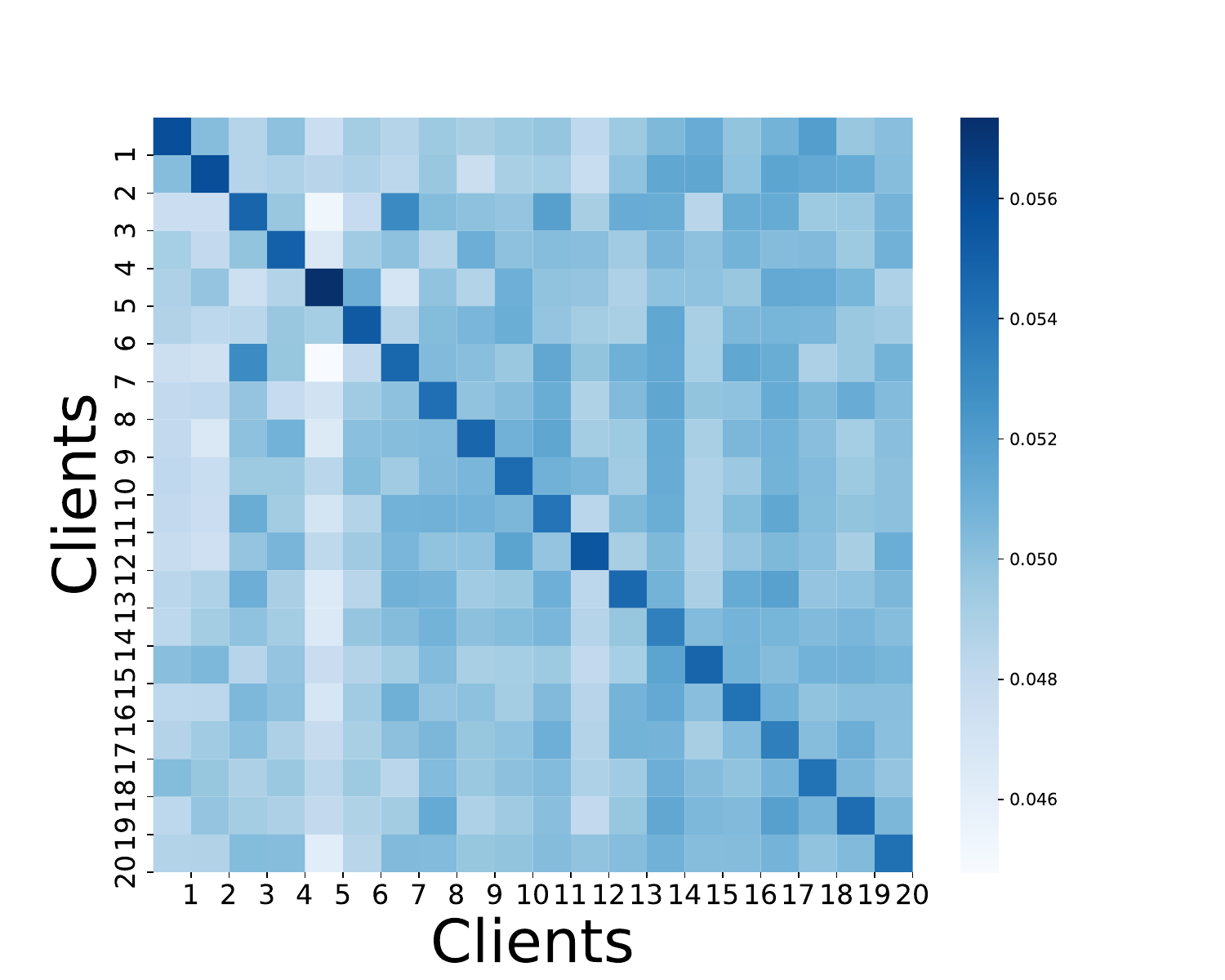}
        \caption{FED-PUB}
        \label{fig_Minesweeper_D2}
    \end{subfigure}%
    \hfill
    \begin{subfigure}[t]{0.15\textwidth}
        \includegraphics[width=\linewidth]{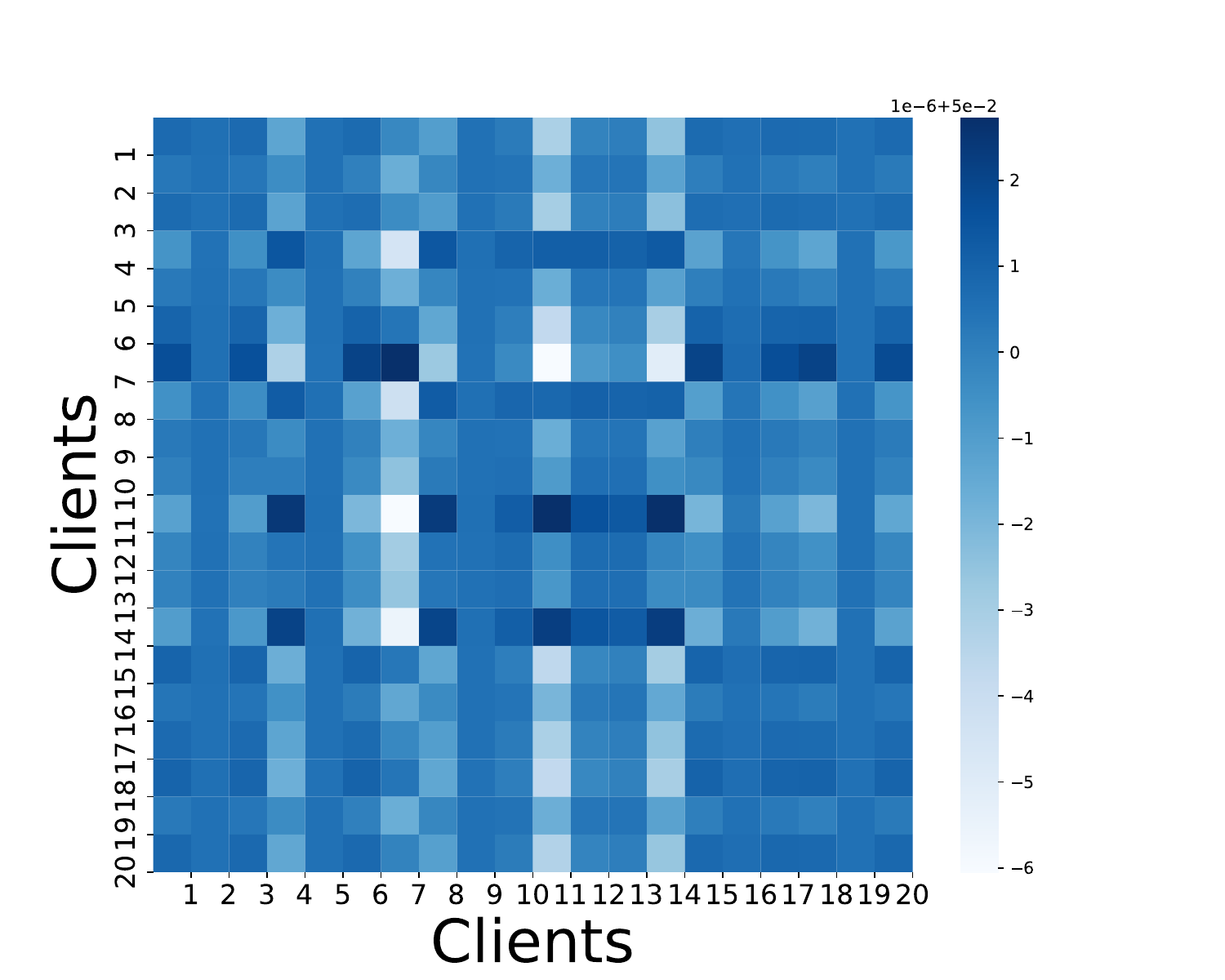}
        \caption{FedGTA}
        \label{fig_Minesweeper_D3}
    \end{subfigure}%
    \hfill
    \begin{subfigure}[t]{0.15\textwidth}
        \includegraphics[width=\linewidth]{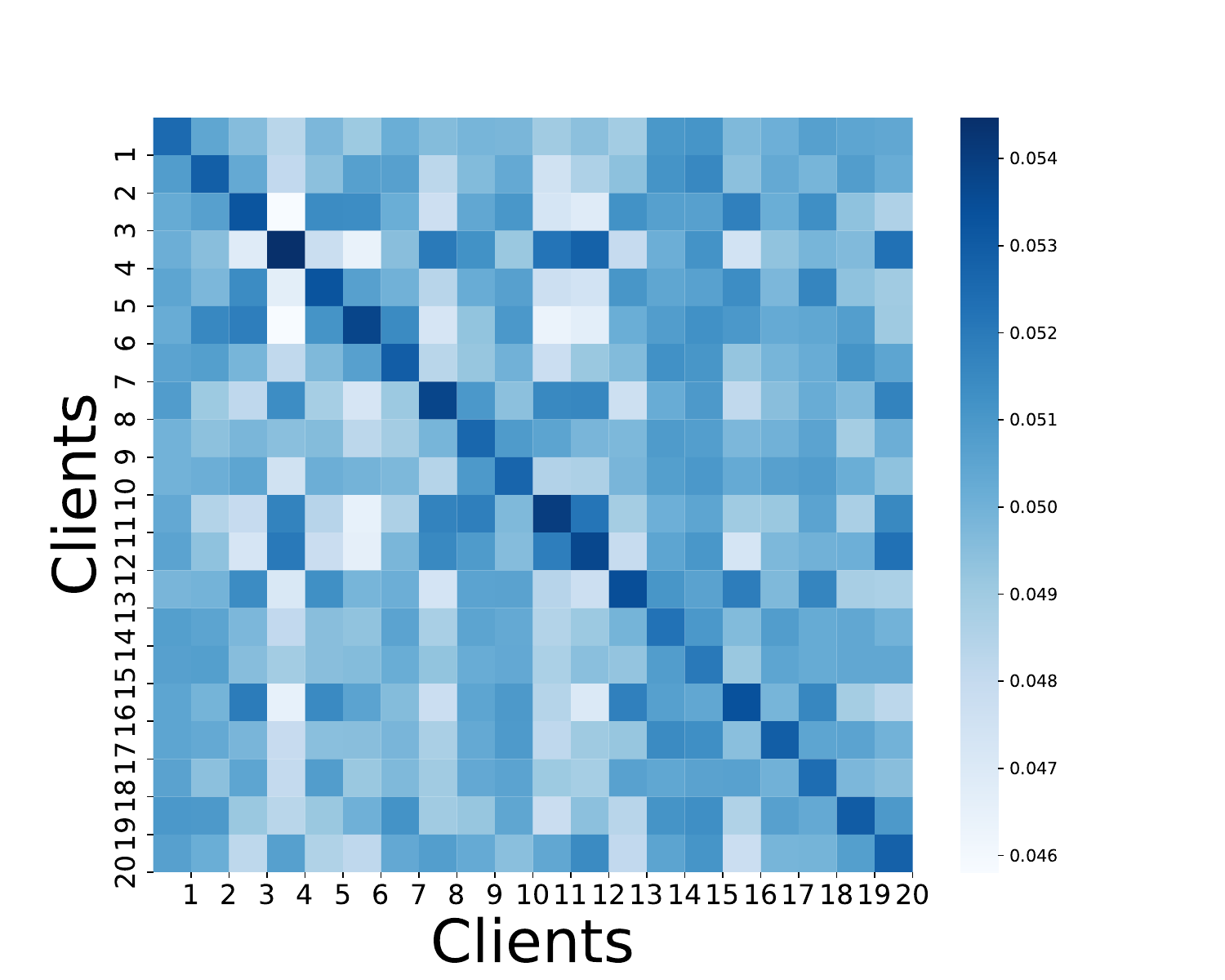}
        \caption{FedIIH of the 1st latent factor ($K=2$)}
        \label{fig_Minesweeper_D4}
    \end{subfigure}
    \hfill
    \begin{subfigure}[t]{0.15\textwidth}
        \includegraphics[width=\linewidth]{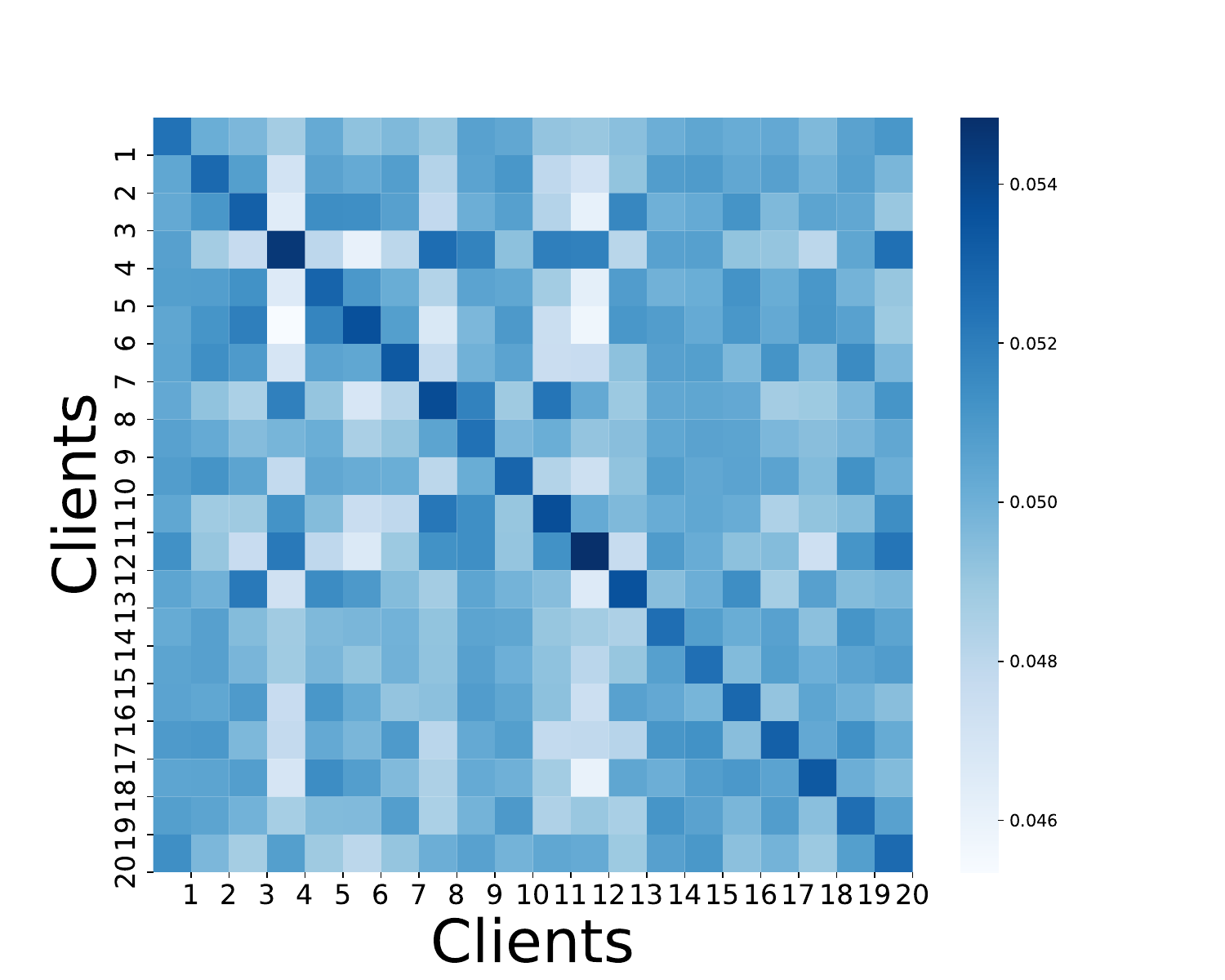}
        \caption{FedIIH of the 2nd latent factor ($K=2$)}
        \label{fig_Minesweeper_D5}
    \end{subfigure}
    \caption{Similarity heatmaps on the \textit{Minesweeper} dataset in the non-overlapping setting with 20 clients.}
    \label{fig_Minesweeper_D}
\end{figure}

\begin{figure}[t]
    \centering
    \begin{subfigure}[t]{0.15\textwidth}
        \includegraphics[width=\linewidth]{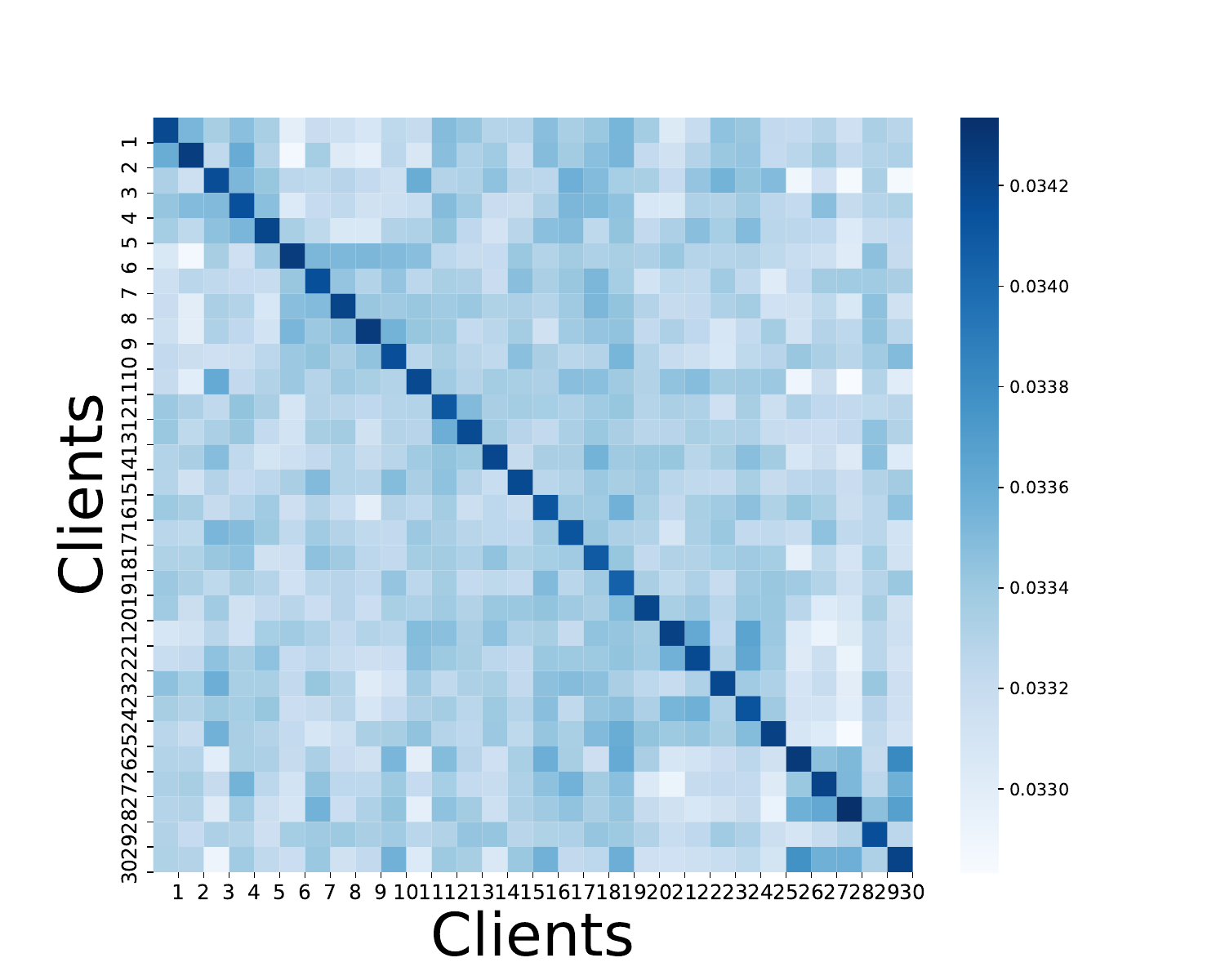}
        \caption{Distr. Sim.}
        \label{fig_Minesweeper_01}
    \end{subfigure}%
    \hfill
    \begin{subfigure}[t]{0.15\textwidth}
        \includegraphics[width=\linewidth]{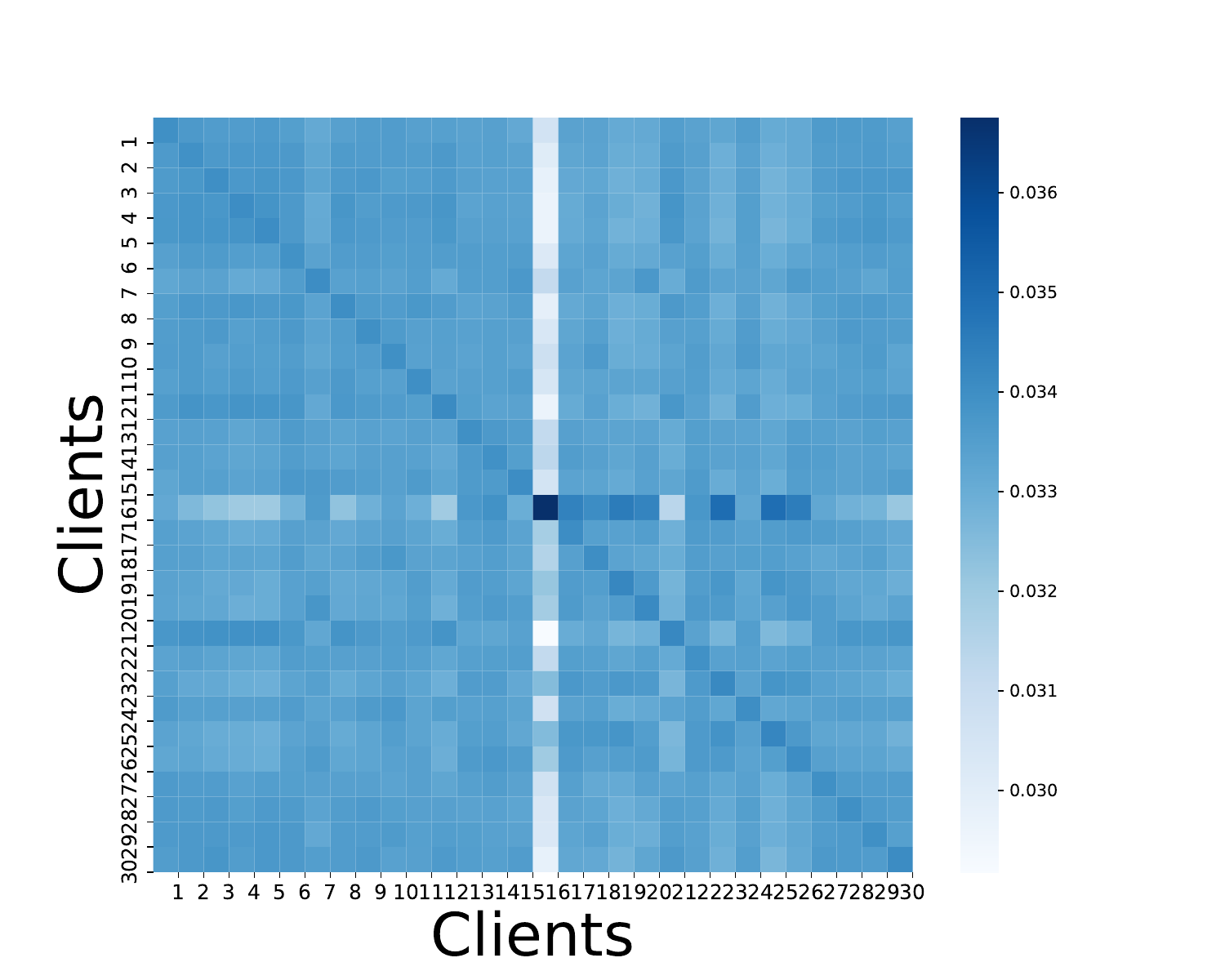}
        \caption{FED-PUB}
        \label{fig_Minesweeper_02}
    \end{subfigure}%
    \hfill
    \begin{subfigure}[t]{0.15\textwidth}
        \includegraphics[width=\linewidth]{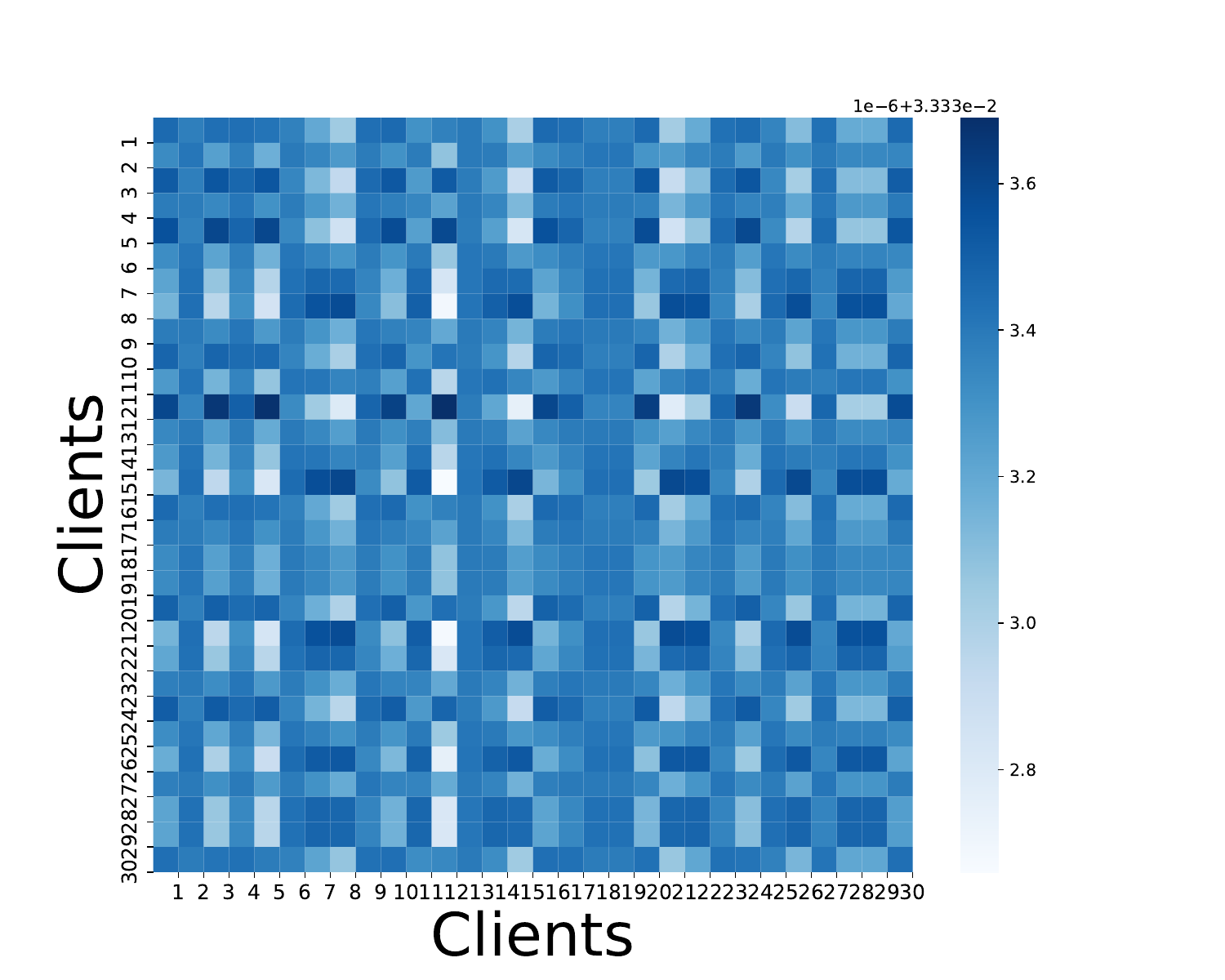}
        \caption{FedGTA}
        \label{fig_Minesweeper_03}
    \end{subfigure}%
    \hfill
    \begin{subfigure}[t]{0.15\textwidth}
        \includegraphics[width=\linewidth]{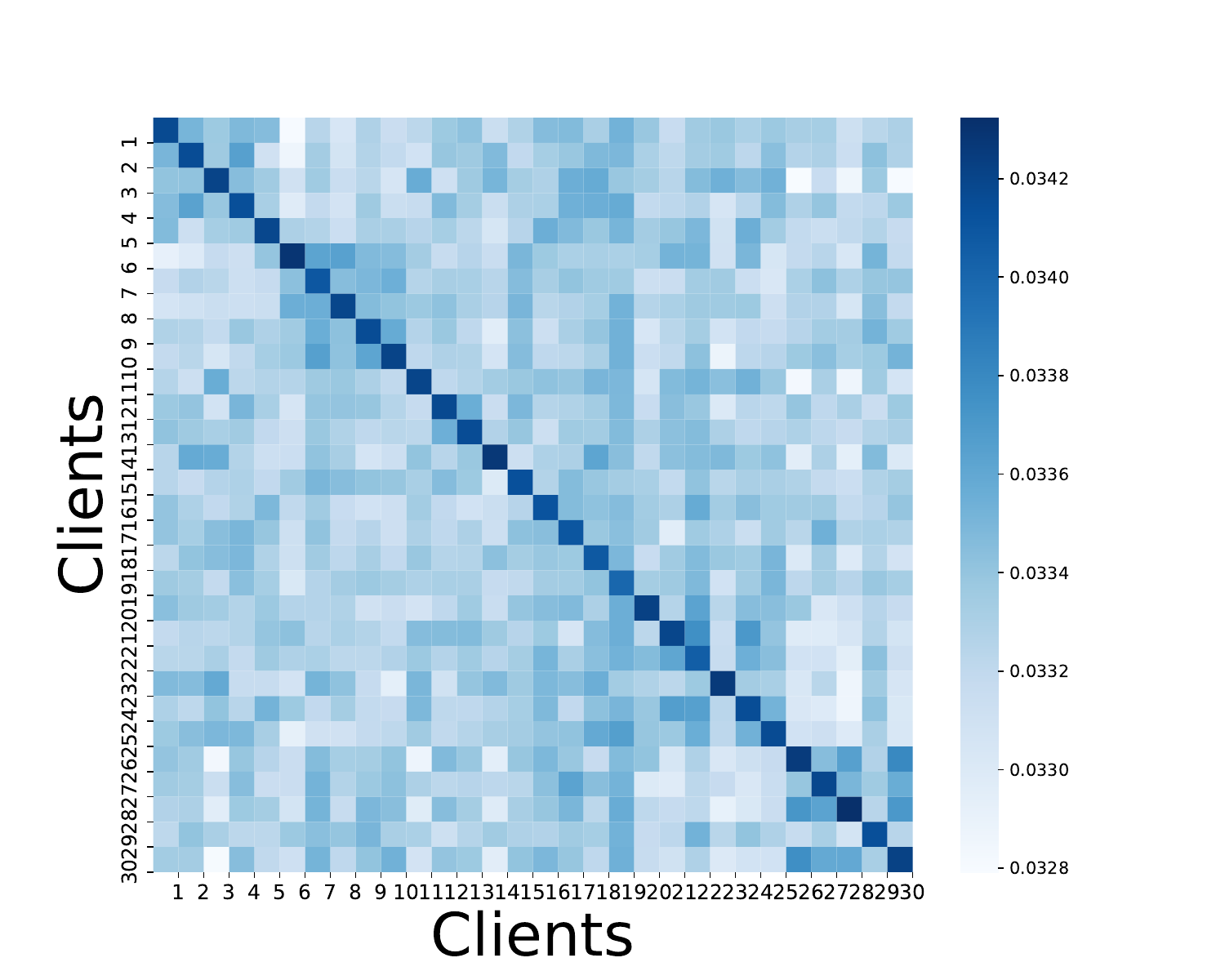}
        \caption{FedIIH of the 1st latent factor ($K=2$)}
        \label{fig_Minesweeper_04}
    \end{subfigure}
    \hfill
    \begin{subfigure}[t]{0.15\textwidth}
        \includegraphics[width=\linewidth]{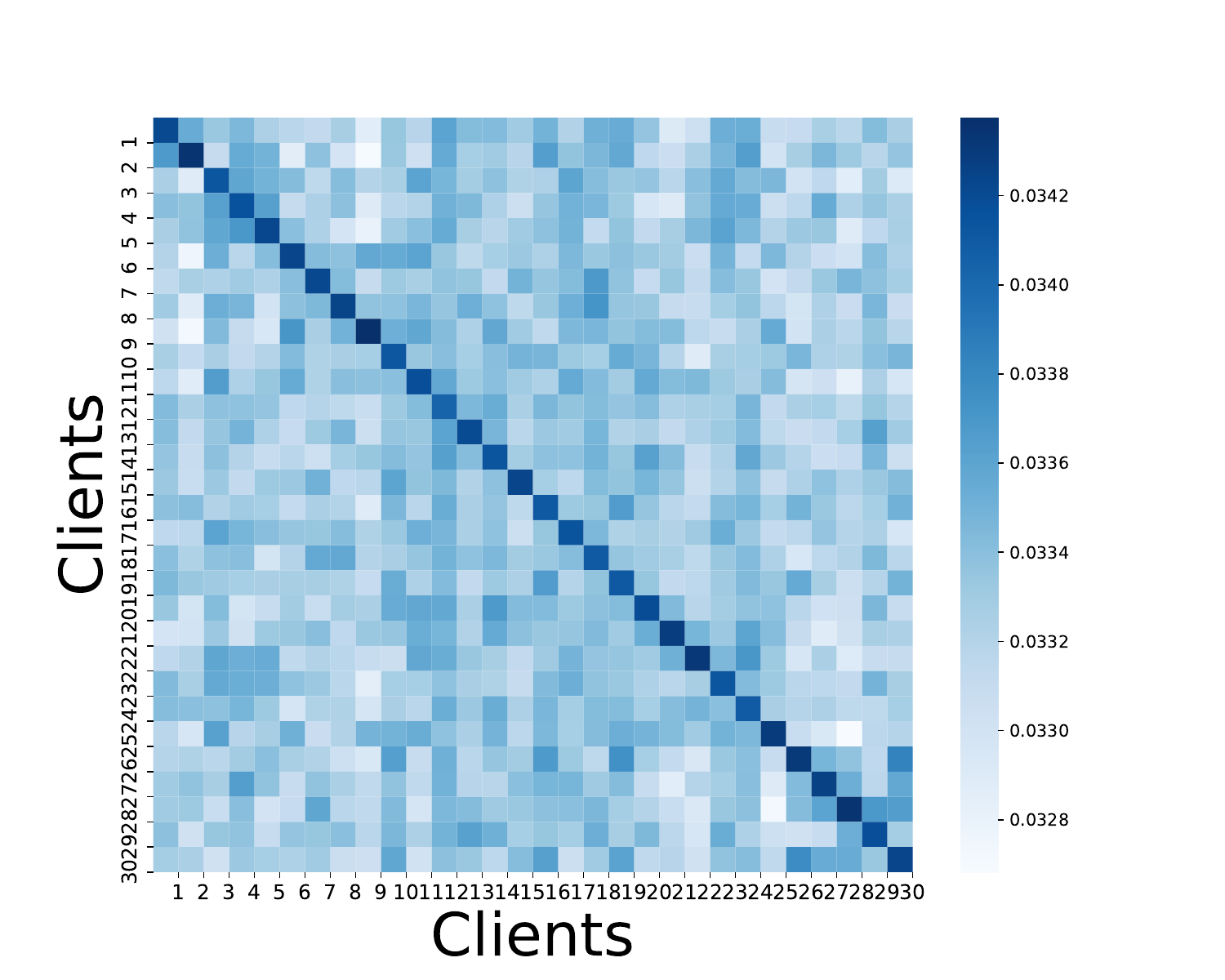}
        \caption{FedIIH of the 2nd latent factor ($K=2$)}
        \label{fig_Minesweeper_05}
    \end{subfigure}
    \caption{Similarity heatmaps on the \textit{Minesweeper} dataset in the overlapping setting with 30 clients.}
    \label{fig_Minesweeper_O}
\end{figure}

\begin{figure}[t]
    \centering
    \begin{subfigure}[t]{0.15\textwidth}
        \includegraphics[width=\linewidth]{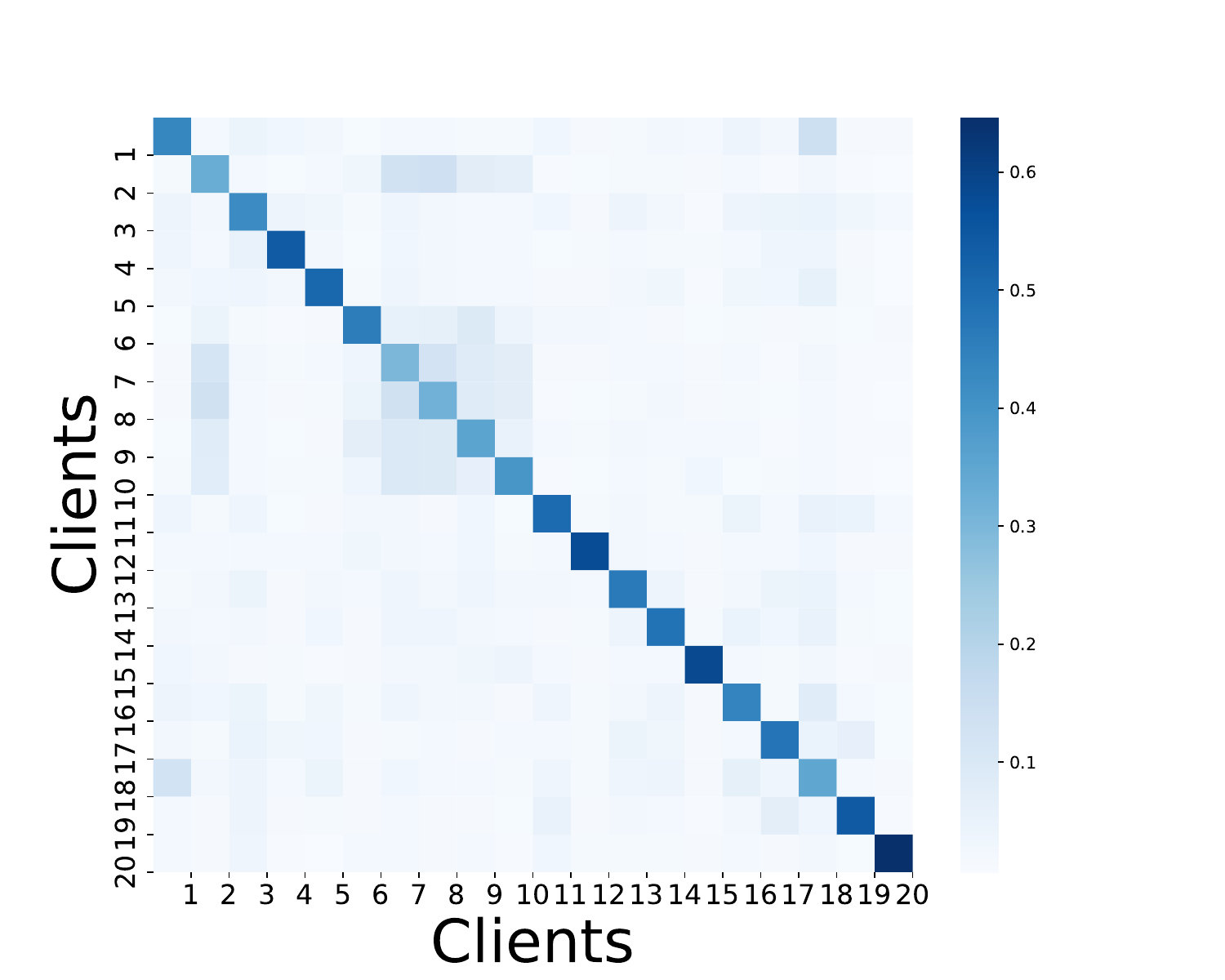}
        \caption{Distr. Sim.}
        \label{fig_Tolokers_D1}
    \end{subfigure}%
    \hfill
    \begin{subfigure}[t]{0.15\textwidth}
        \includegraphics[width=\linewidth]{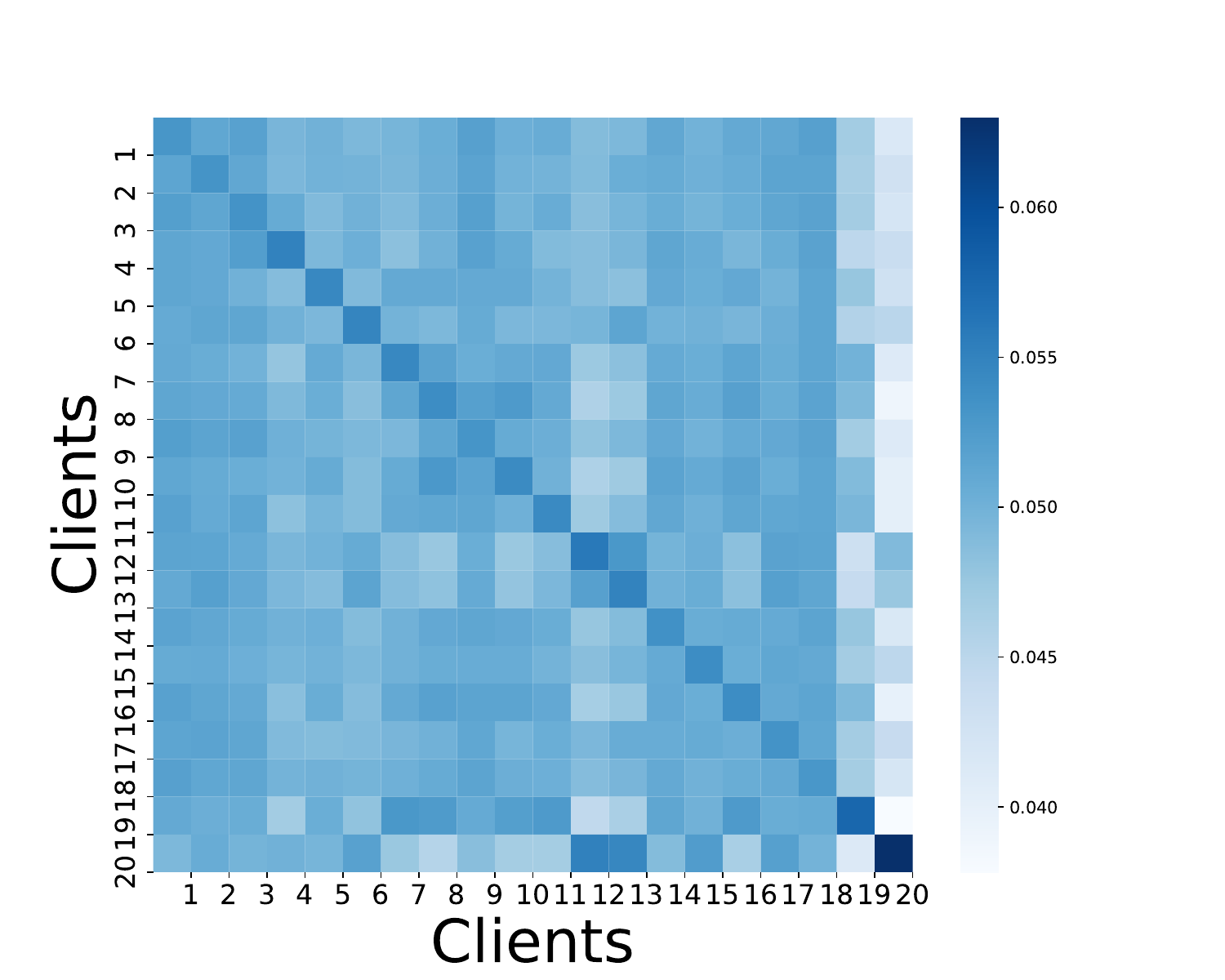}
        \caption{FED-PUB}
        \label{fig_Tolokers_D2}
    \end{subfigure}%
    \hfill
    \begin{subfigure}[t]{0.15\textwidth}
        \includegraphics[width=\linewidth]{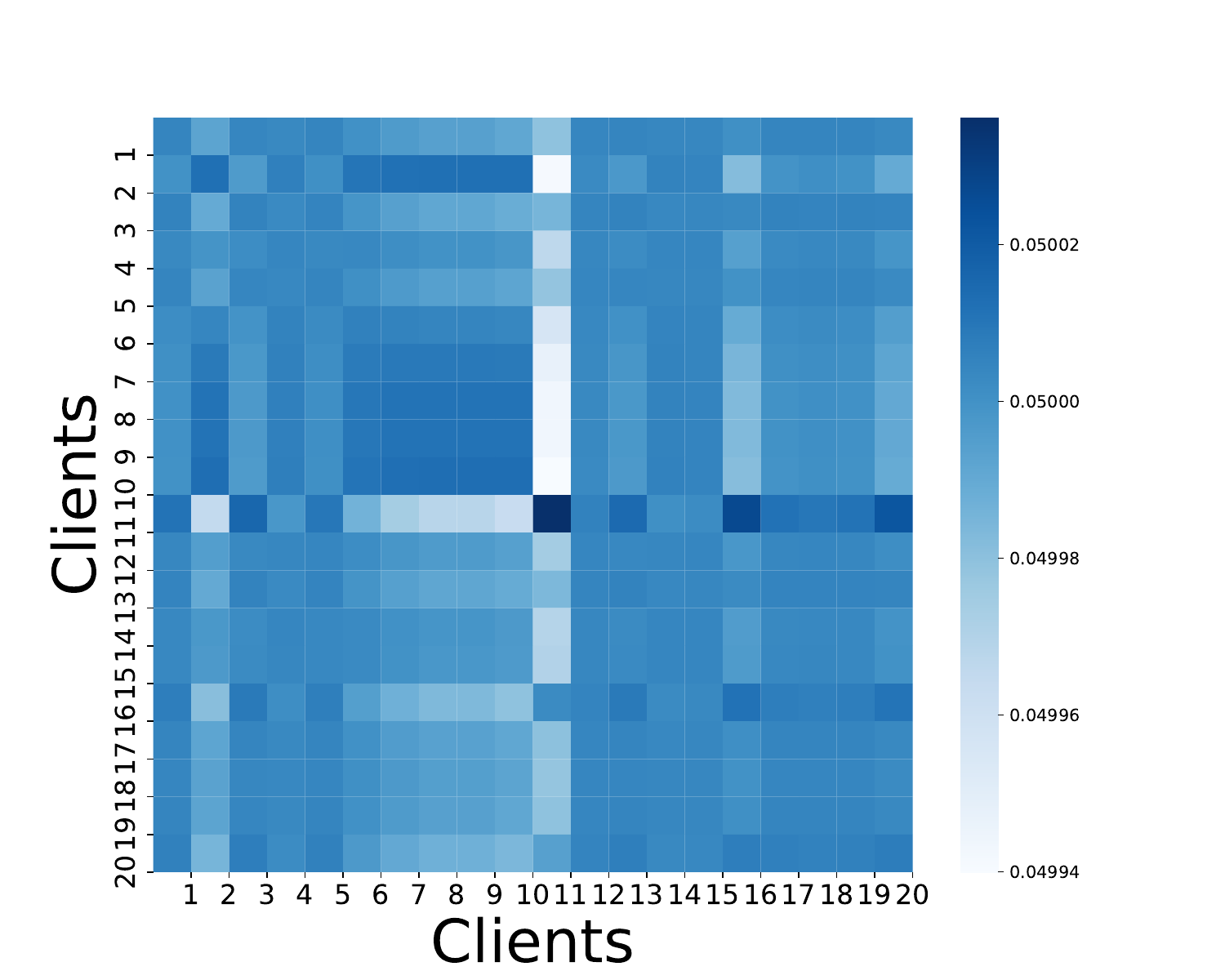}
        \caption{FedGTA}
        \label{fig_Tolokers_D3}
    \end{subfigure}%
    \hfill
    \begin{subfigure}[t]{0.15\textwidth}
        \includegraphics[width=\linewidth]{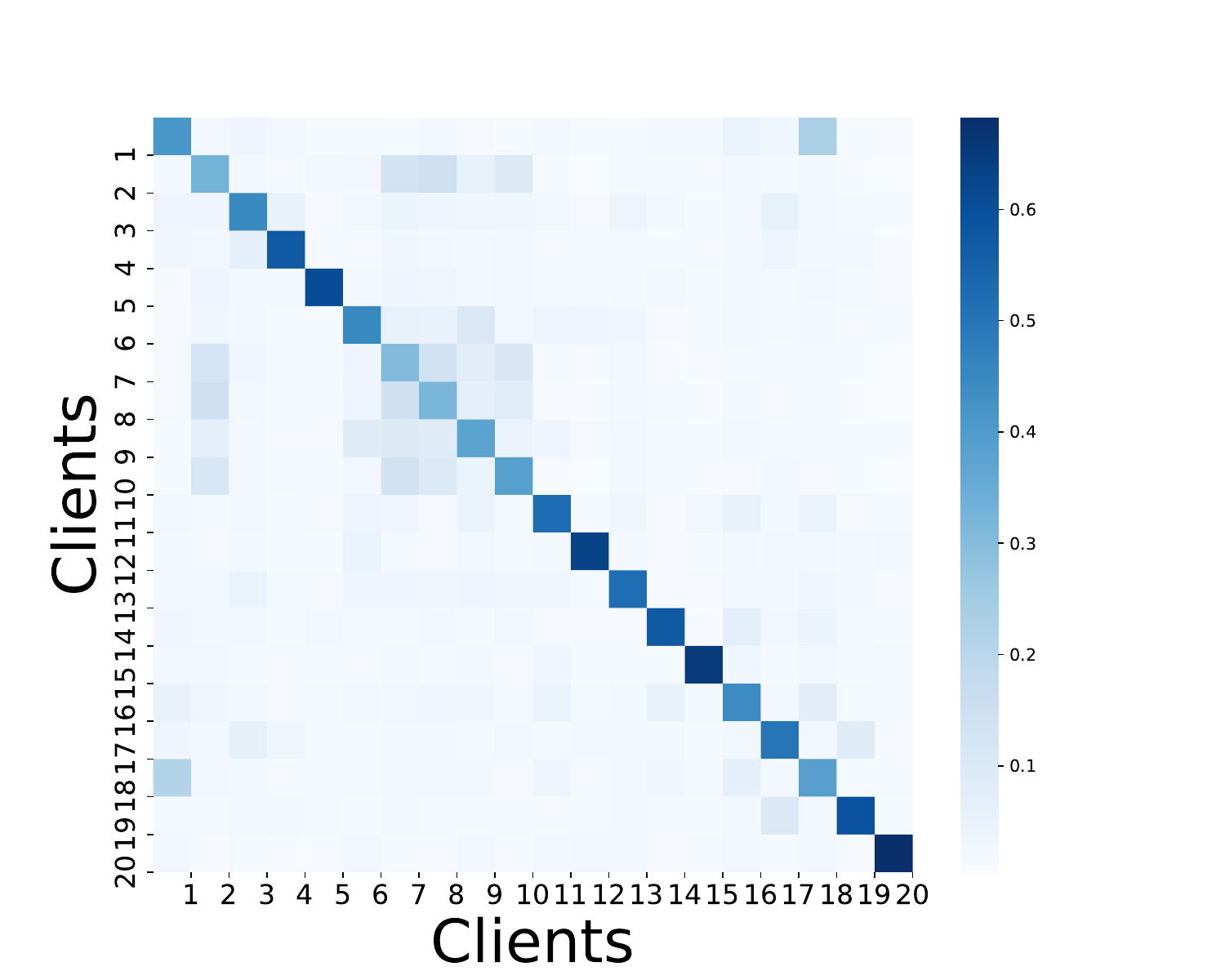}
        \caption{FedIIH of the 1st latent factor ($K=2$)}
        \label{fig_Tolokers_D4}
    \end{subfigure}
    \hfill
    \begin{subfigure}[t]{0.15\textwidth}
        \includegraphics[width=\linewidth]{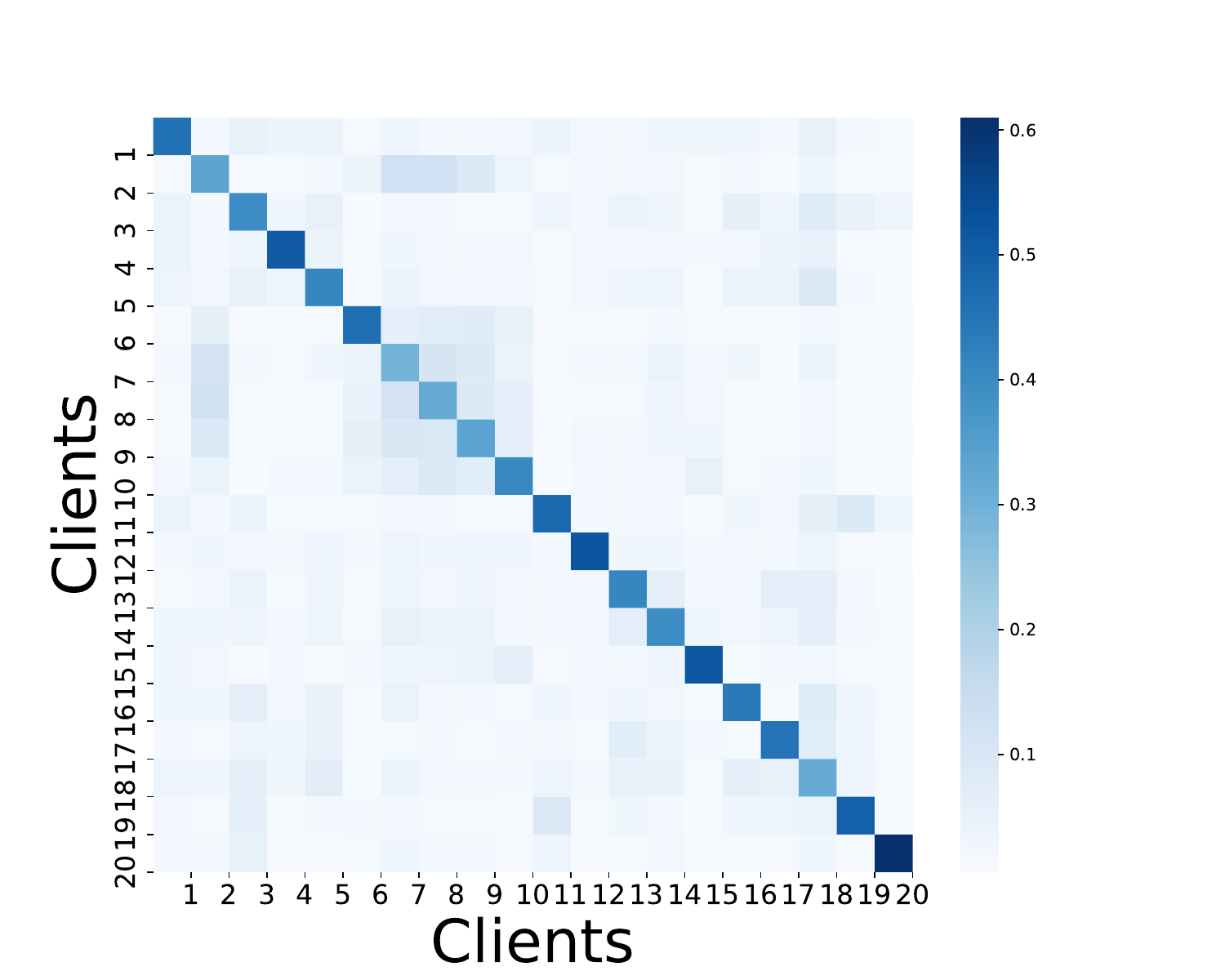}
        \caption{FedIIH of the 2nd latent factor ($K=2$)}
        \label{fig_Tolokers_D5}
    \end{subfigure}
    \caption{Similarity heatmaps on the \textit{Tolokers} dataset in the non-overlapping setting with 20 clients.}
    \label{fig_Tolokers_D}
\end{figure}

\begin{figure}[t]
    \centering
    \begin{subfigure}[t]{0.15\textwidth}
        \includegraphics[width=\linewidth]{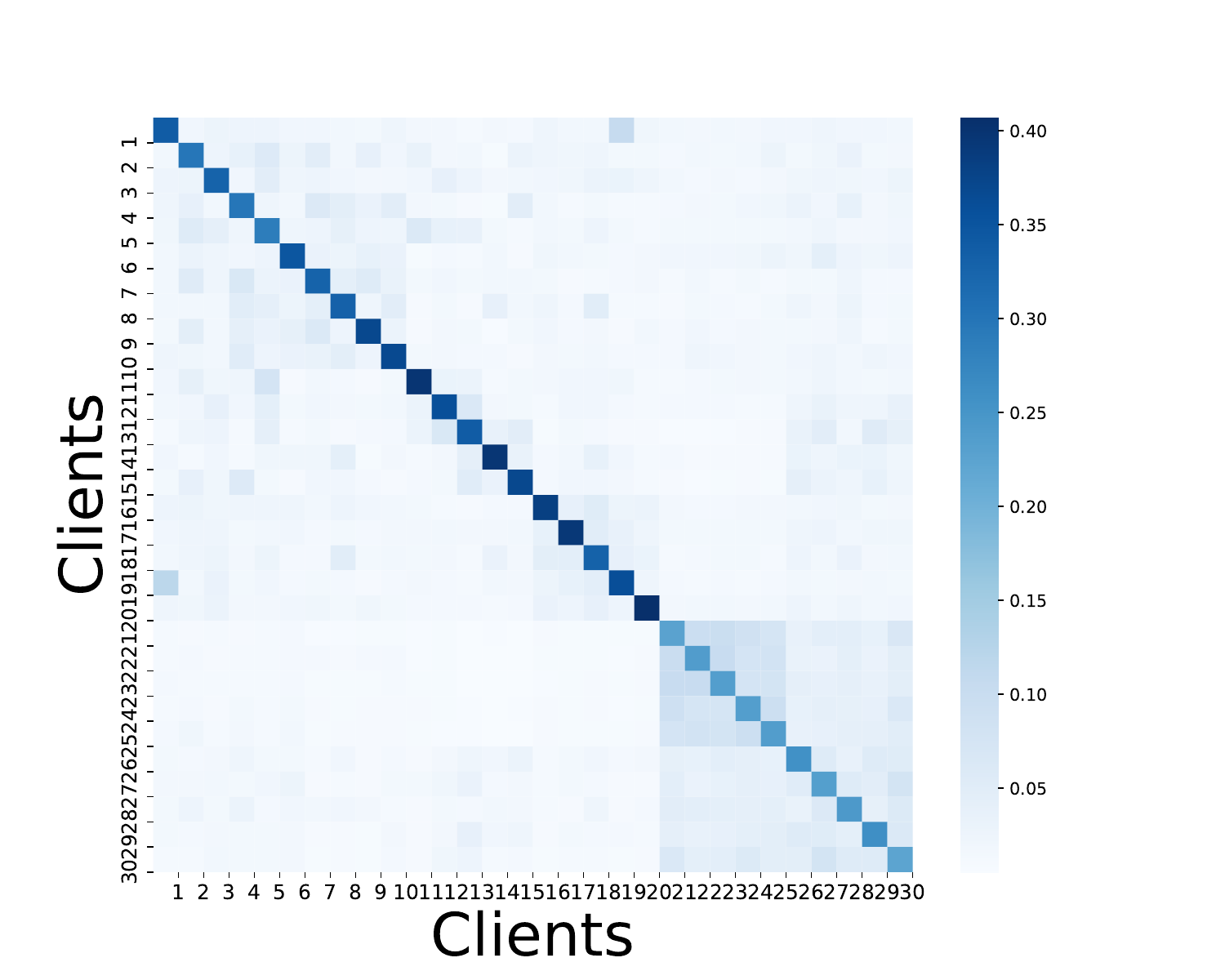}
        \caption{Distr. Sim.}
        \label{fig_Tolokers_01}
    \end{subfigure}%
    \hfill
    \begin{subfigure}[t]{0.15\textwidth}
        \includegraphics[width=\linewidth]{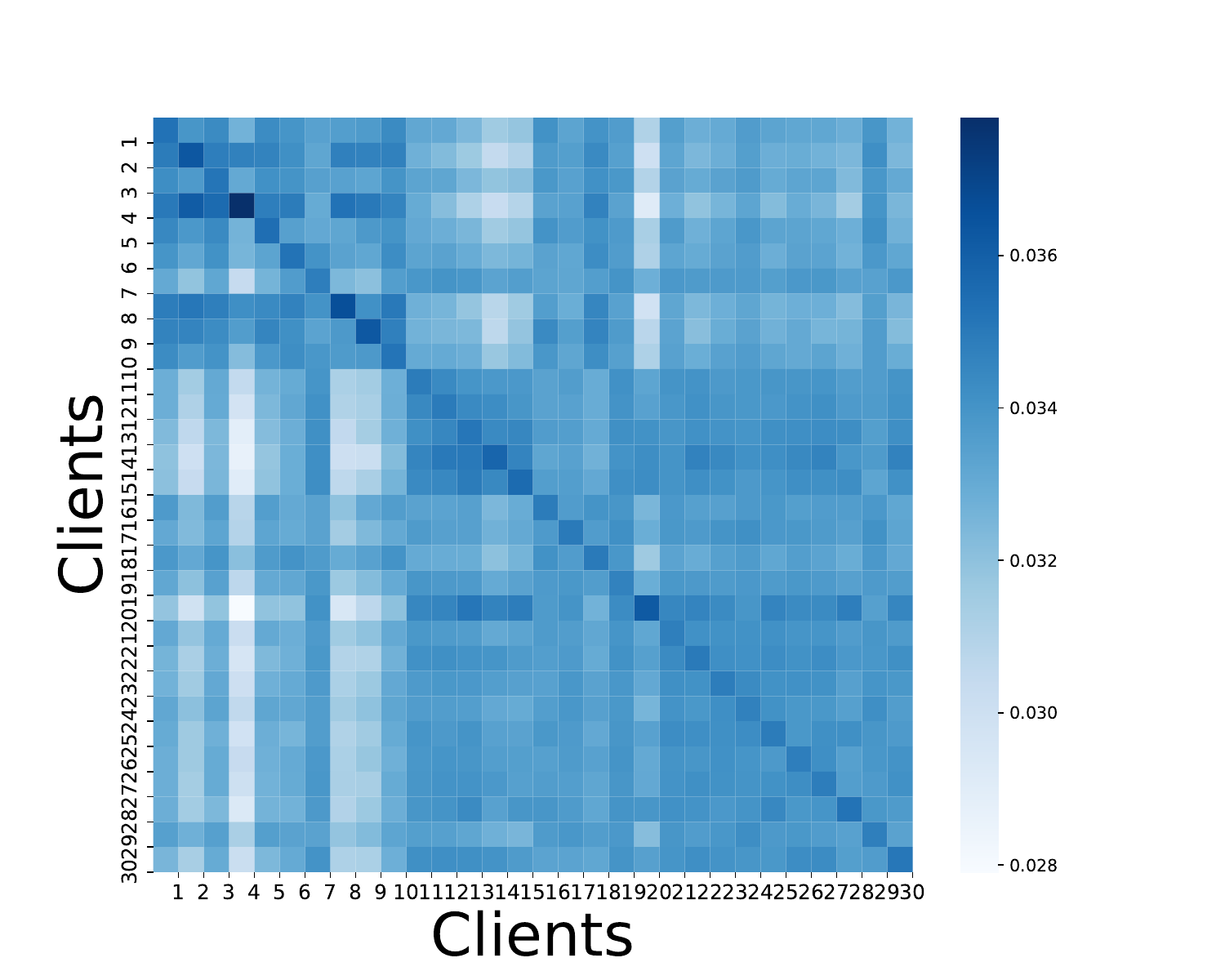}
        \caption{FED-PUB}
        \label{fig_Tolokers_02}
    \end{subfigure}%
    \hfill
    \begin{subfigure}[t]{0.15\textwidth}
        \includegraphics[width=\linewidth]{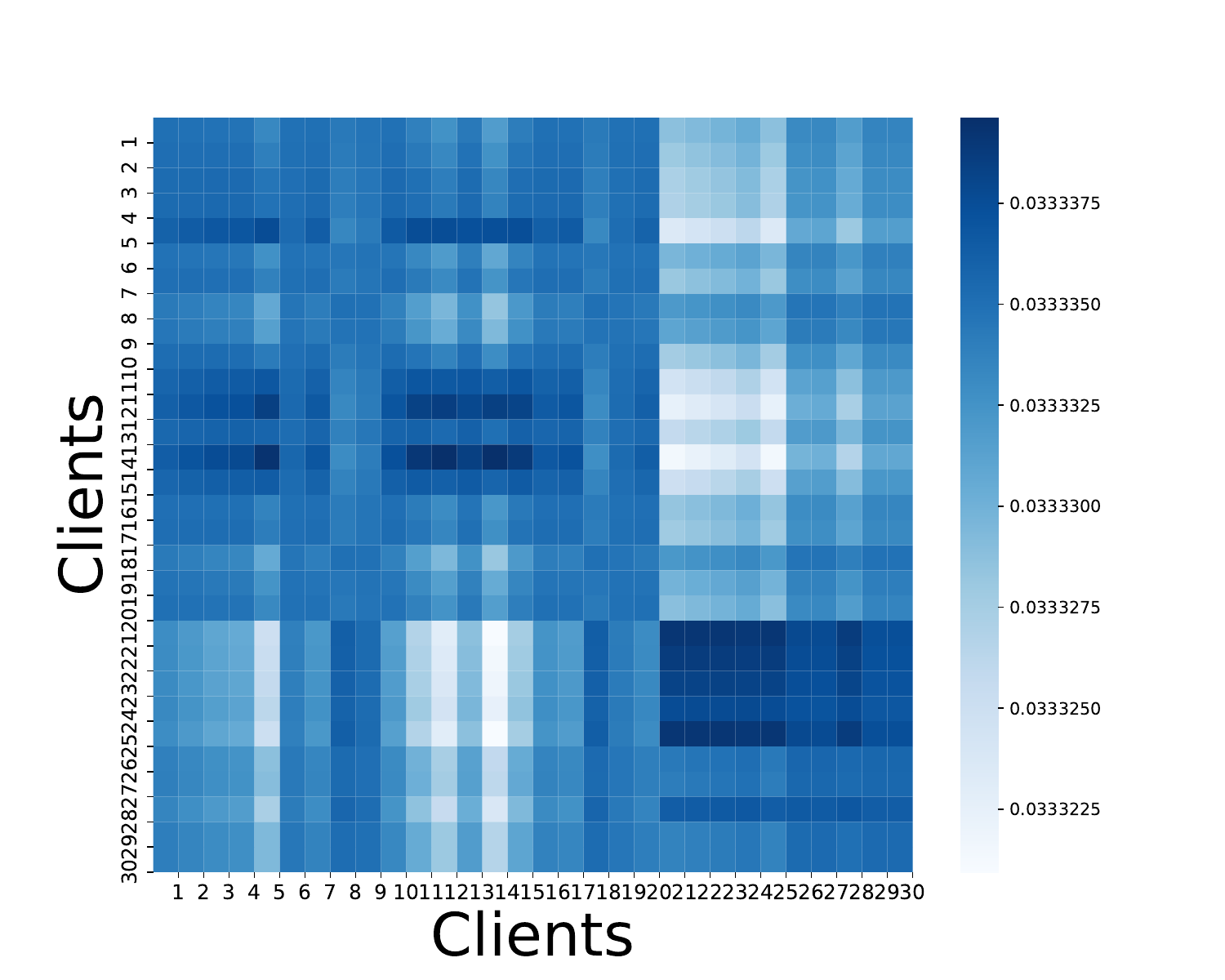}
        \caption{FedGTA}
        \label{fig_Tolokers_03}
    \end{subfigure}%
    \hfill
    \begin{subfigure}[t]{0.15\textwidth}
        \includegraphics[width=\linewidth]{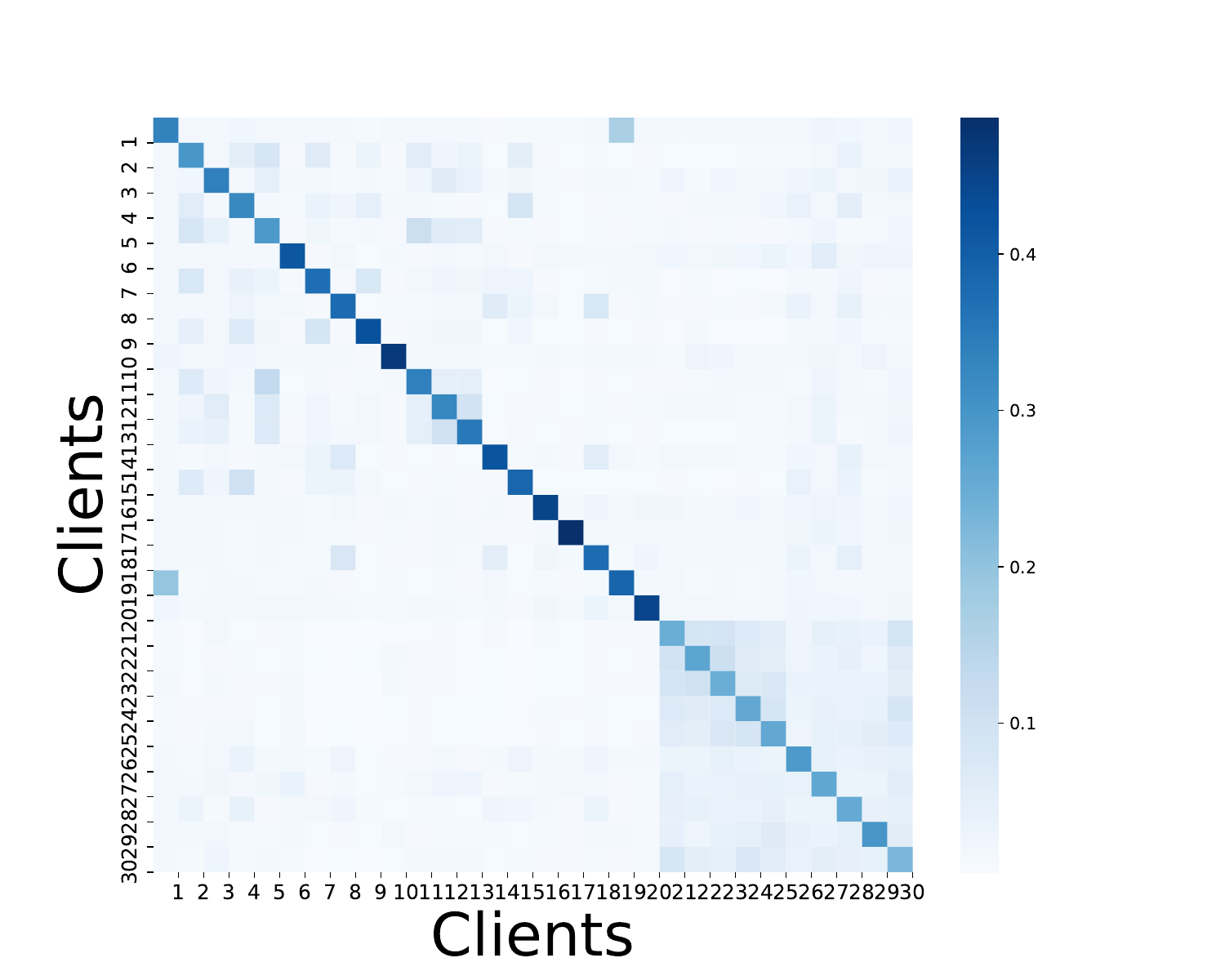}
        \caption{FedIIH of the 1st latent factor ($K=2$)}
        \label{fig_Tolokers_04}
    \end{subfigure}
    \hfill
    \begin{subfigure}[t]{0.15\textwidth}
        \includegraphics[width=\linewidth]{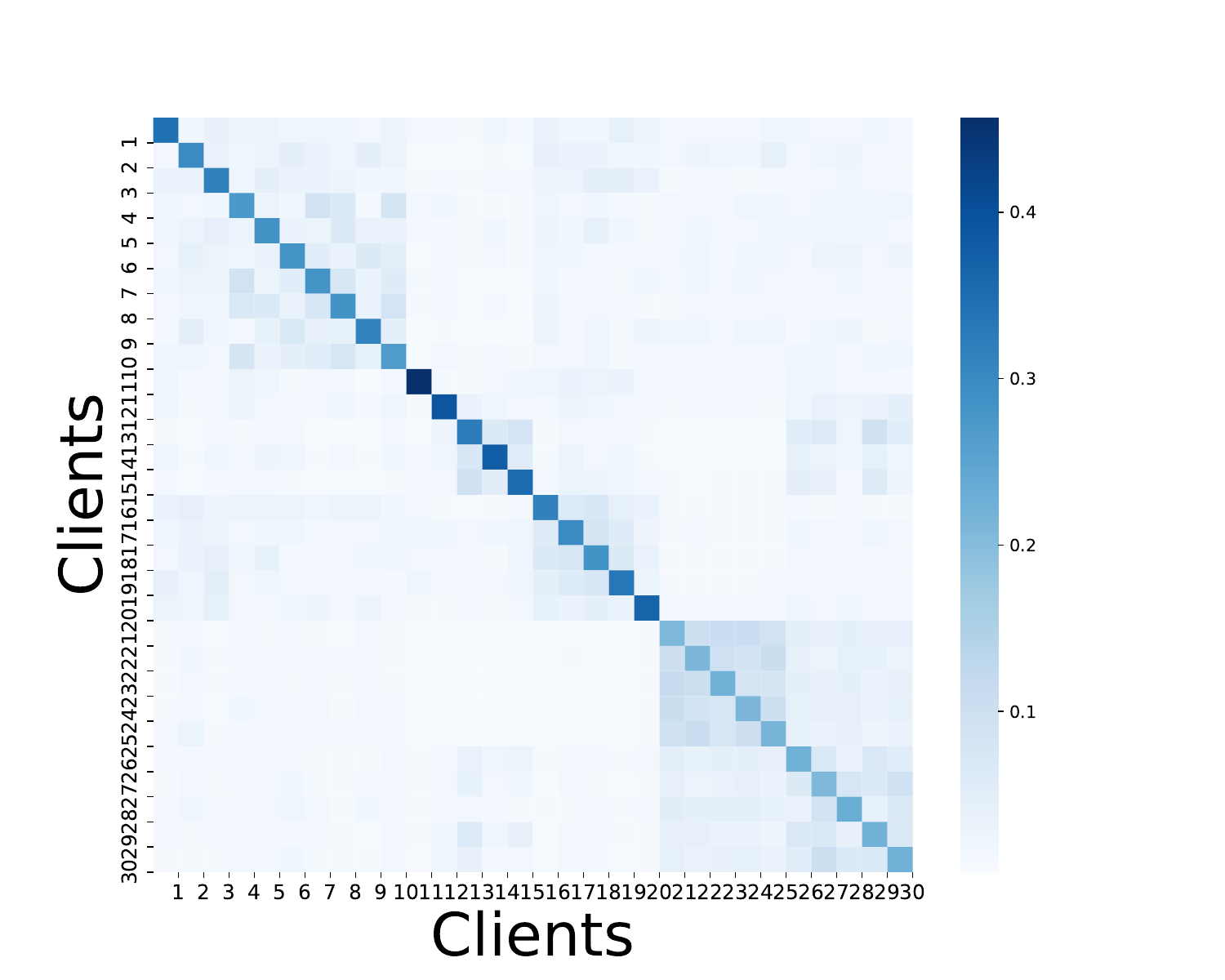}
        \caption{FedIIH of the 2nd latent factor ($K=2$)}
        \label{fig_Tolokers_05}
    \end{subfigure}
    \caption{Similarity heatmaps on the \textit{Tolokers} dataset in the overlapping setting with 30 clients.}
    \label{fig_Tolokers_O}
\end{figure}

\begin{figure}[t]
    \centering
    \begin{subfigure}[t]{0.15\textwidth}
        \includegraphics[width=\linewidth]{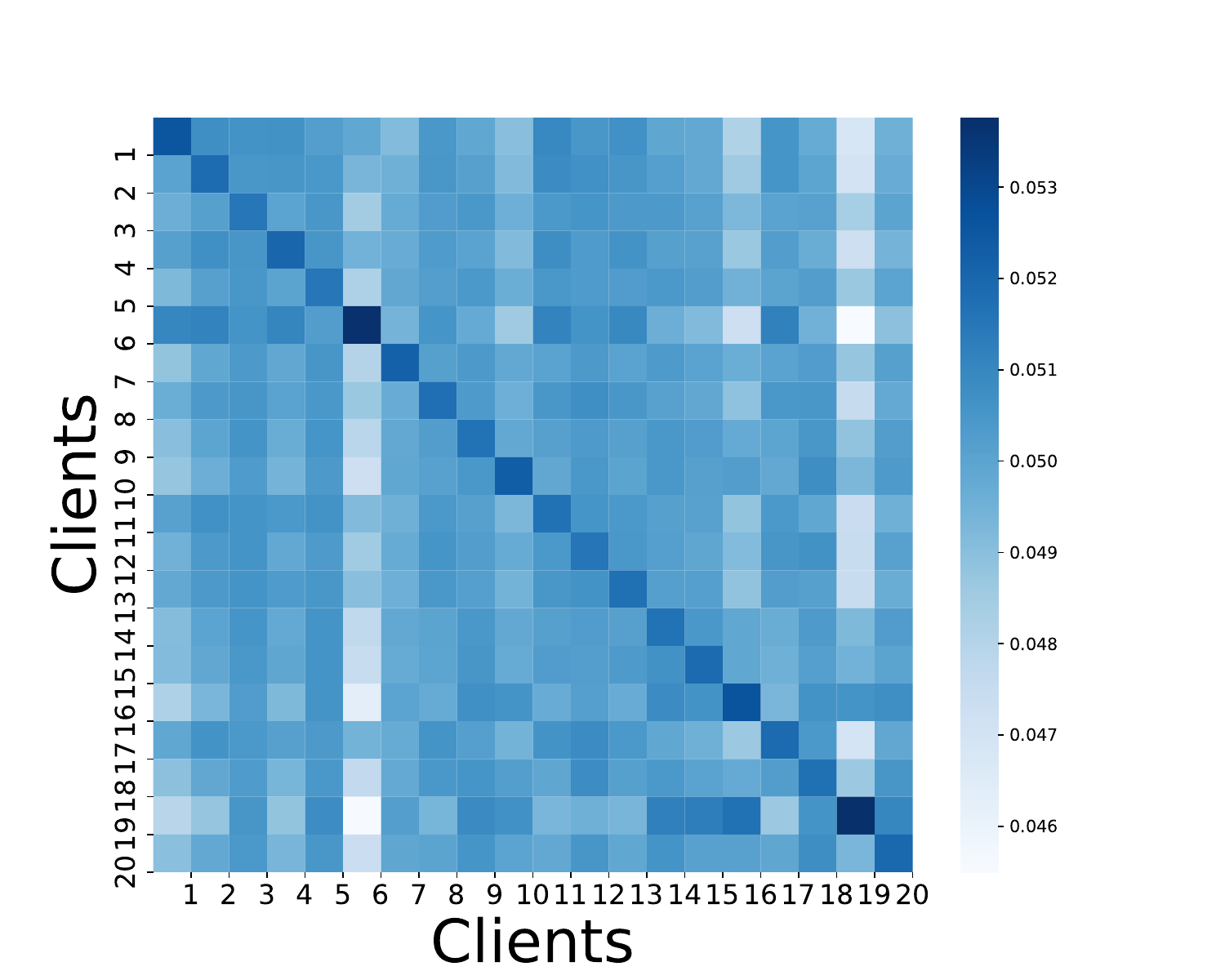}
        \caption{Distr. Sim.}
        \label{fig_Questions_D1}
    \end{subfigure}%
    \hfill
    \begin{subfigure}[t]{0.15\textwidth}
        \includegraphics[width=\linewidth]{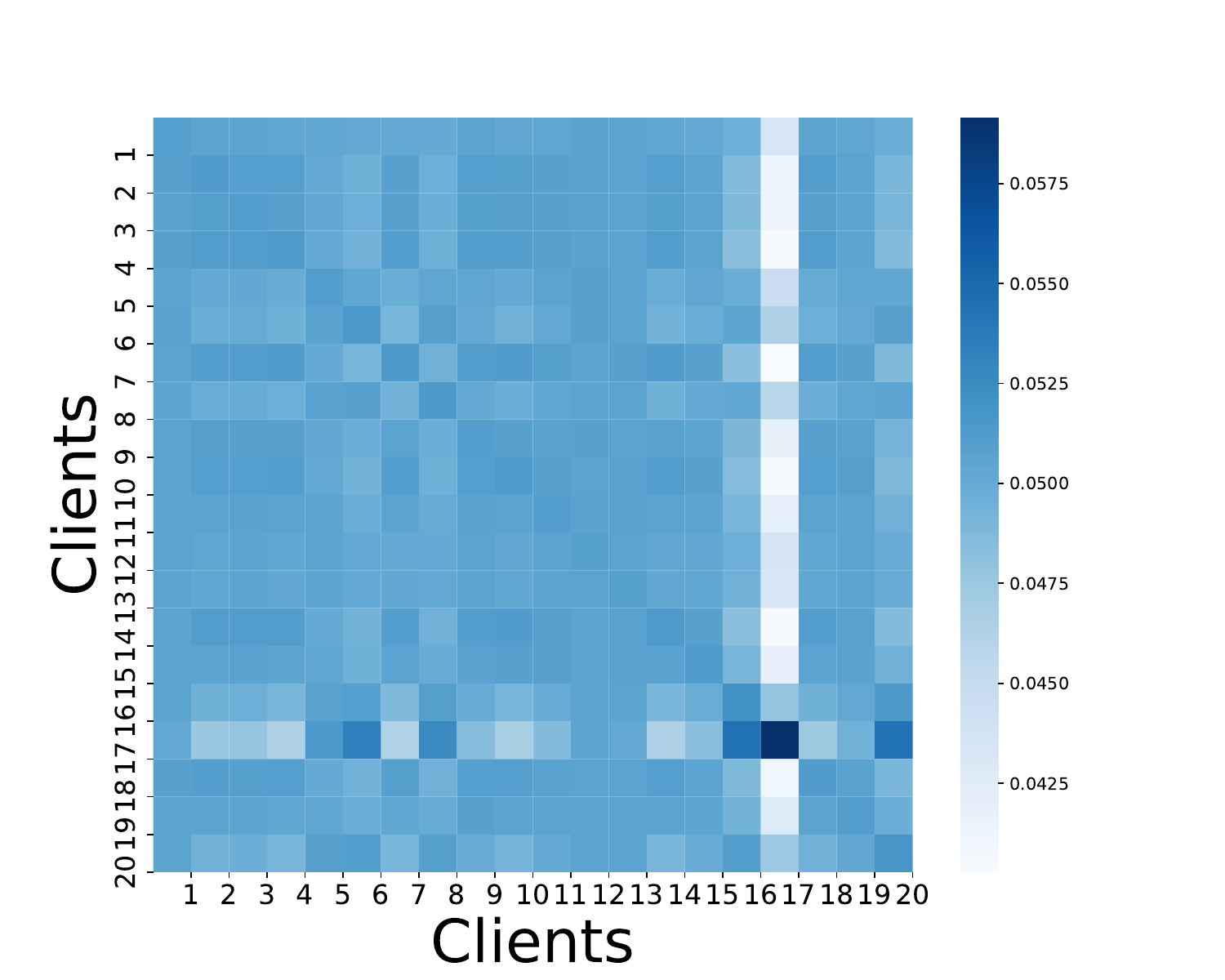}
        \caption{FED-PUB}
        \label{fig_Questions_D2}
    \end{subfigure}%
    \hfill
    \begin{subfigure}[t]{0.15\textwidth}
        \includegraphics[width=\linewidth]{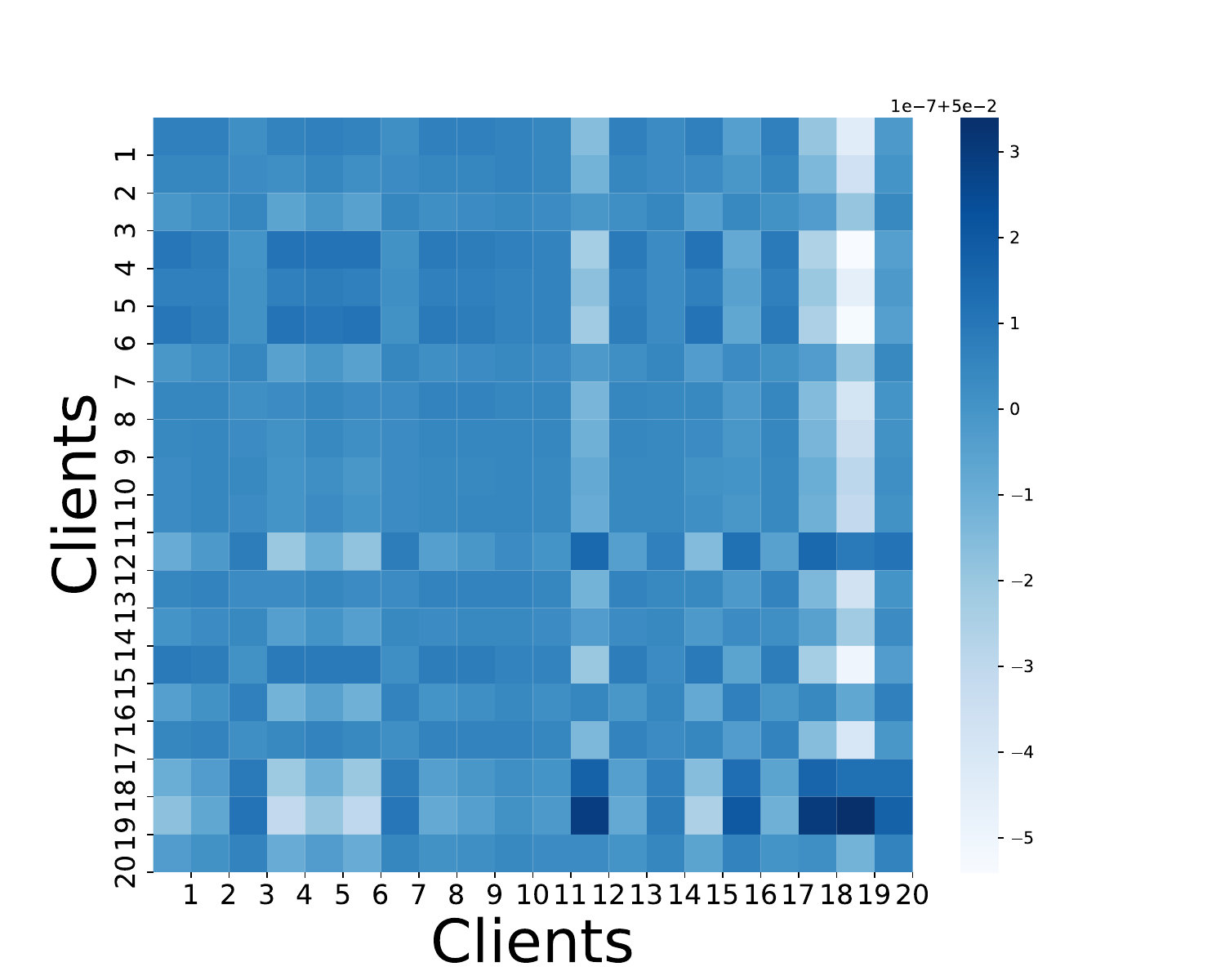}
        \caption{FedGTA}
        \label{fig_Questions_D3}
    \end{subfigure}%
    \hfill
    \begin{subfigure}[t]{0.15\textwidth}
        \includegraphics[width=\linewidth]{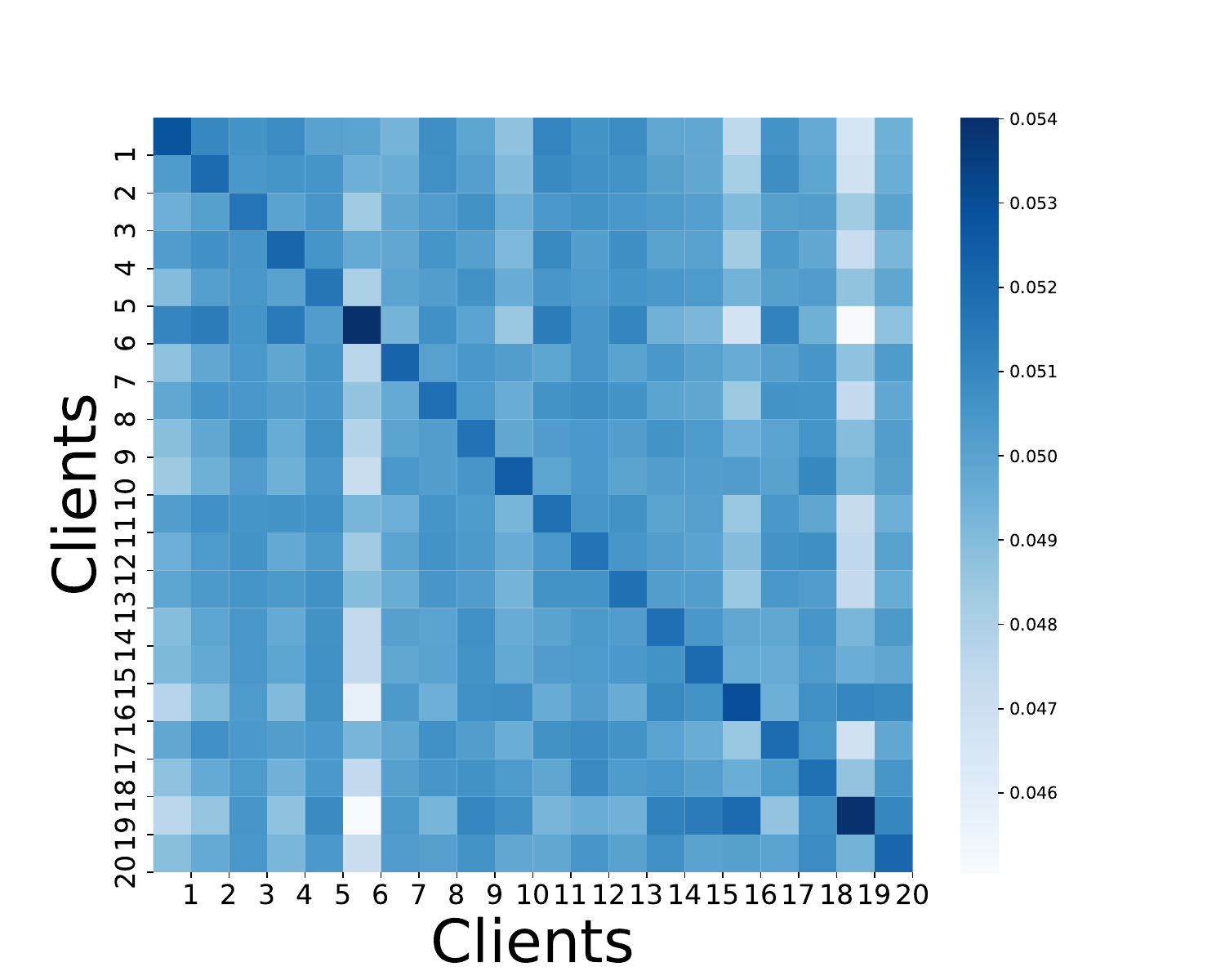}
        \caption{FedIIH of the 1st latent factor ($K=2$)}
        \label{fig_Questions_D4}
    \end{subfigure}
    \hfill
    \begin{subfigure}[t]{0.15\textwidth}
        \includegraphics[width=\linewidth]{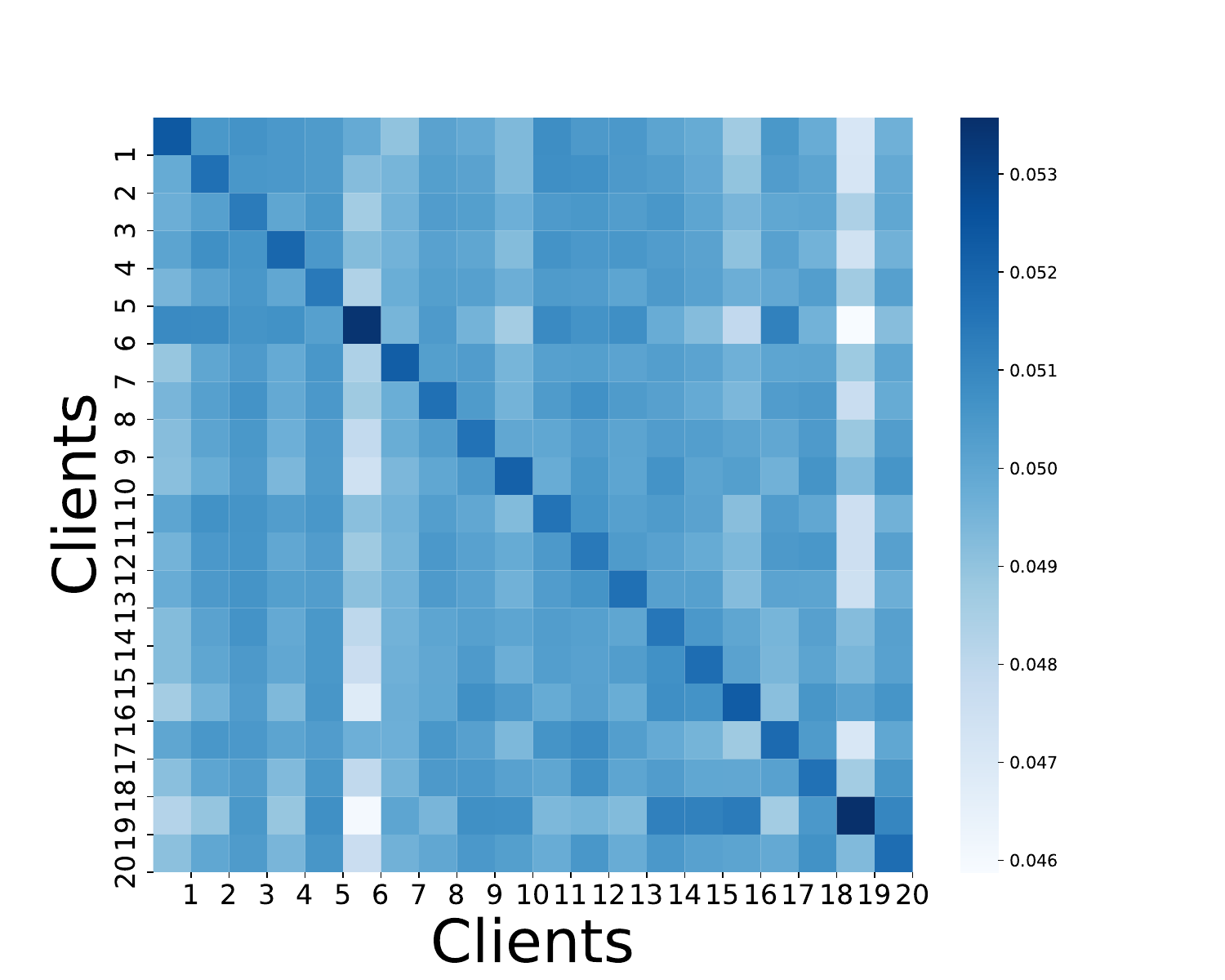}
        \caption{FedIIH of the 2nd latent factor ($K=2$)}
        \label{fig_Questions_D5}
    \end{subfigure}
    \caption{Similarity heatmaps on the \textit{Questions} dataset in the non-overlapping setting with 20 clients.}
    \label{fig_Questions_D}
\end{figure}

\begin{figure}[t]
    \centering
    \begin{subfigure}[t]{0.15\textwidth}
        \includegraphics[width=\linewidth]{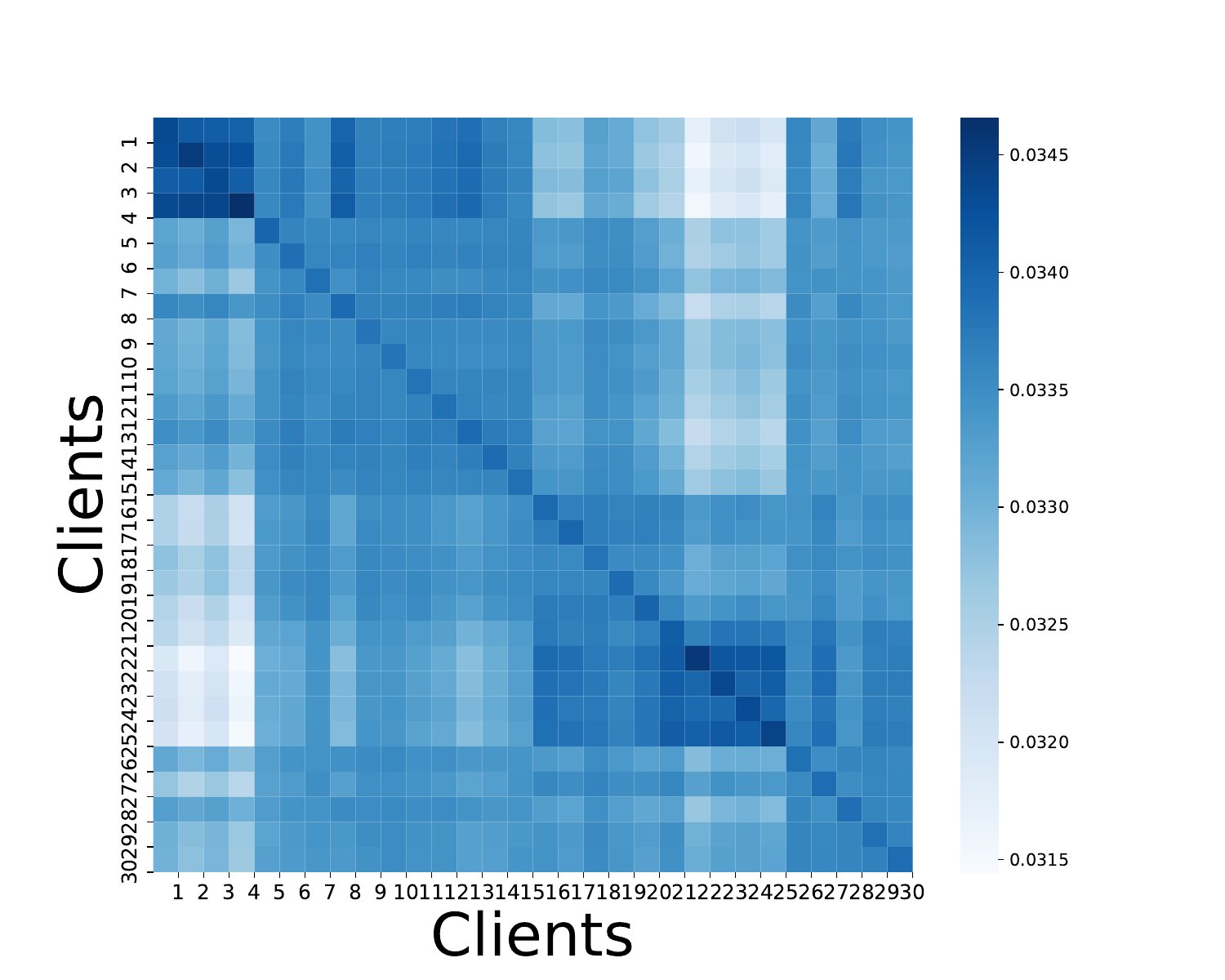}
        \caption{Distr. Sim.}
        \label{fig_Questions_01}
    \end{subfigure}%
    \hfill
    \begin{subfigure}[t]{0.15\textwidth}
        \includegraphics[width=\linewidth]{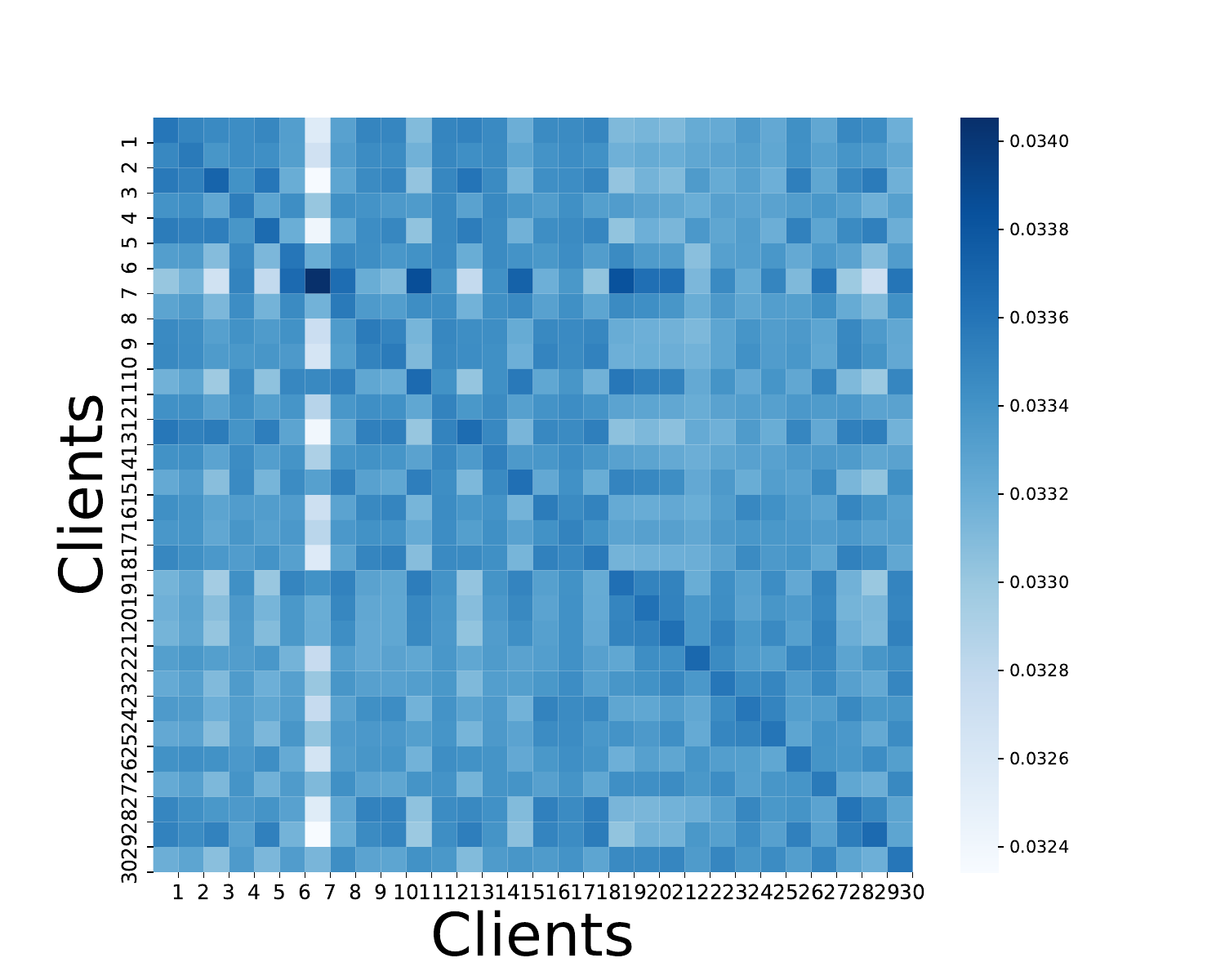}
        \caption{FED-PUB}
        \label{fig_Questions_02}
    \end{subfigure}%
    \hfill
    \begin{subfigure}[t]{0.15\textwidth}
        \includegraphics[width=\linewidth]{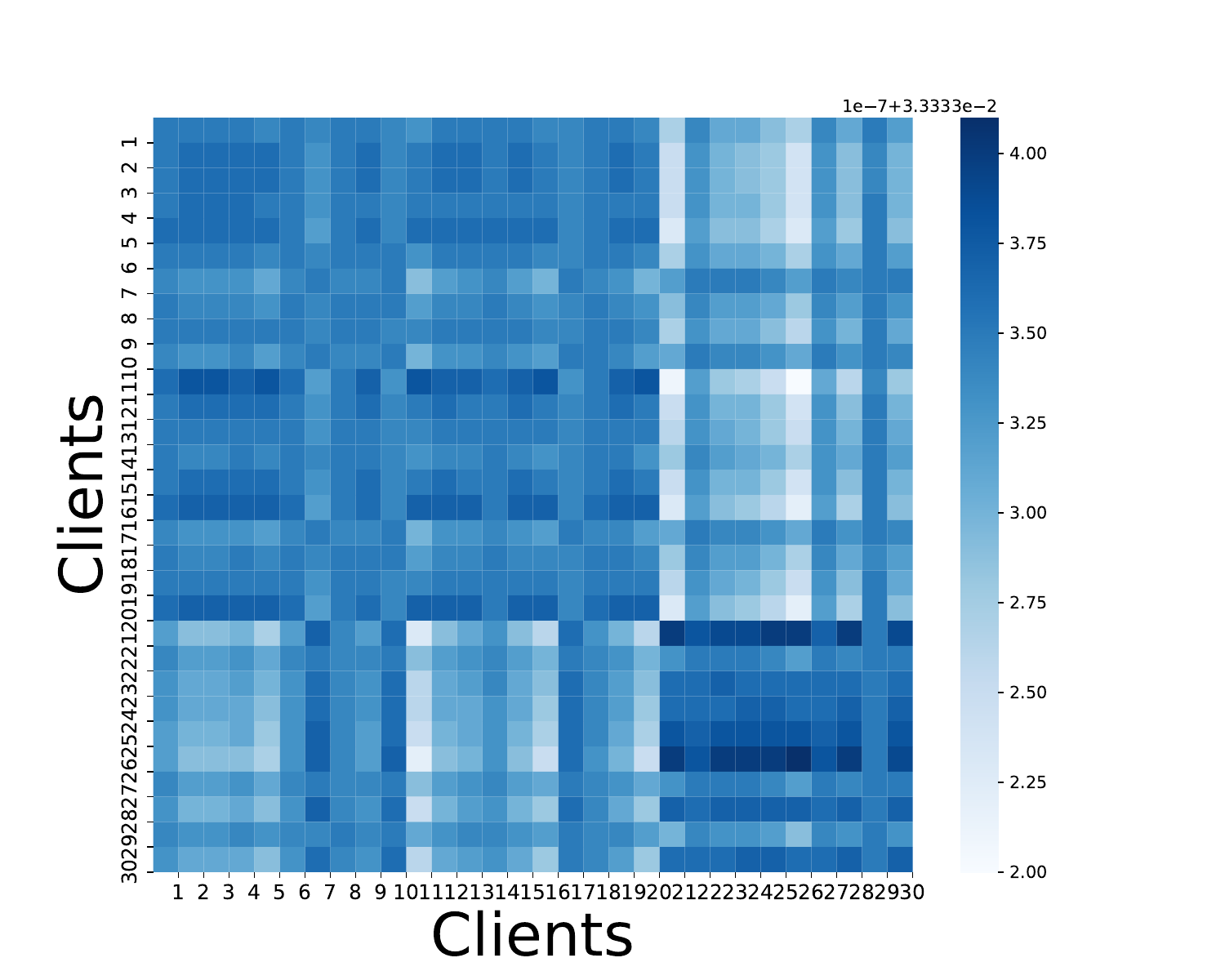}
        \caption{FedGTA}
        \label{fig_Questions_03}
    \end{subfigure}%
    \hfill
    \begin{subfigure}[t]{0.15\textwidth}
        \includegraphics[width=\linewidth]{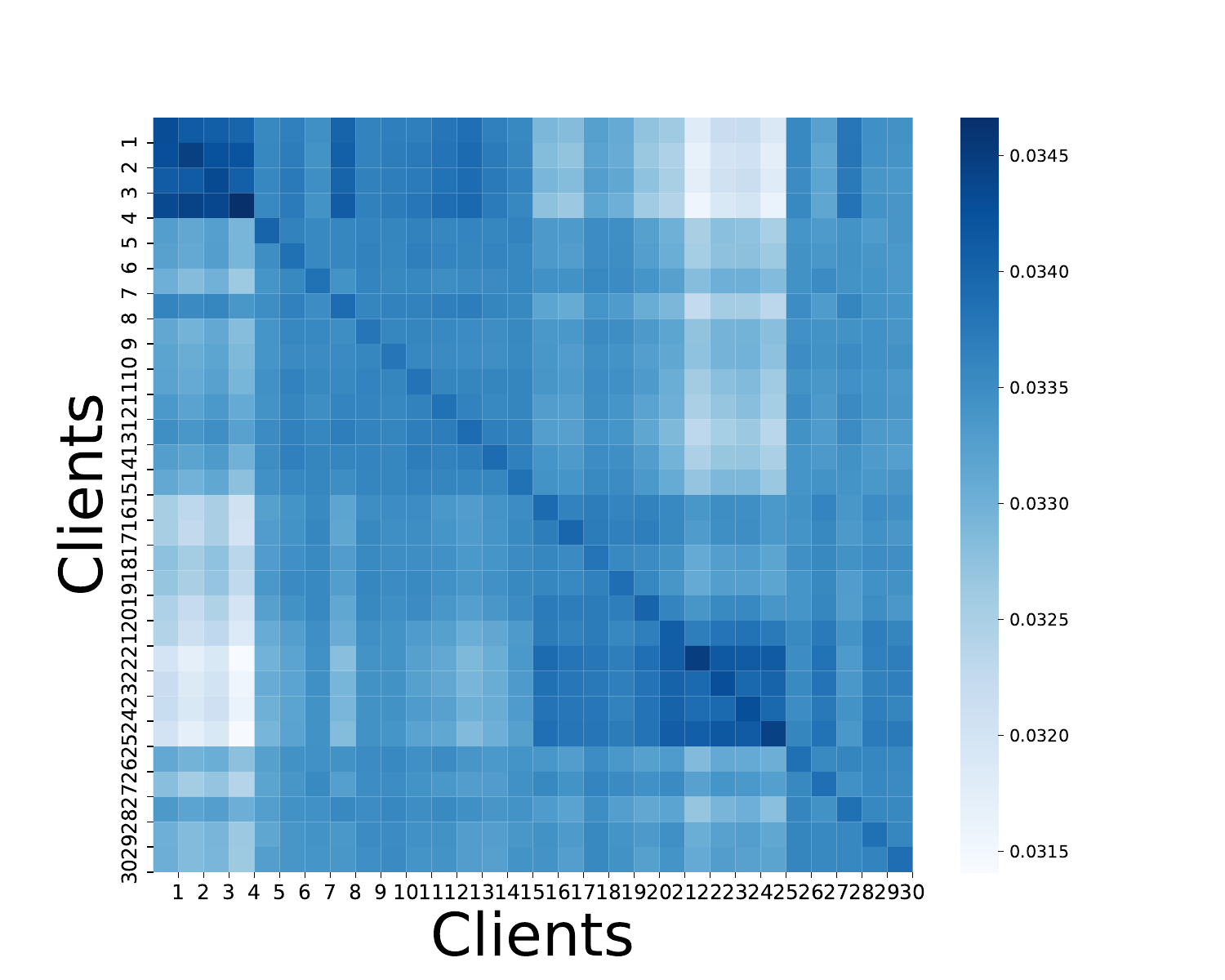}
        \caption{FedIIH of the 1st latent factor ($K=2$)}
        \label{fig_Questions_04}
    \end{subfigure}
    \hfill
    \begin{subfigure}[t]{0.15\textwidth}
        \includegraphics[width=\linewidth]{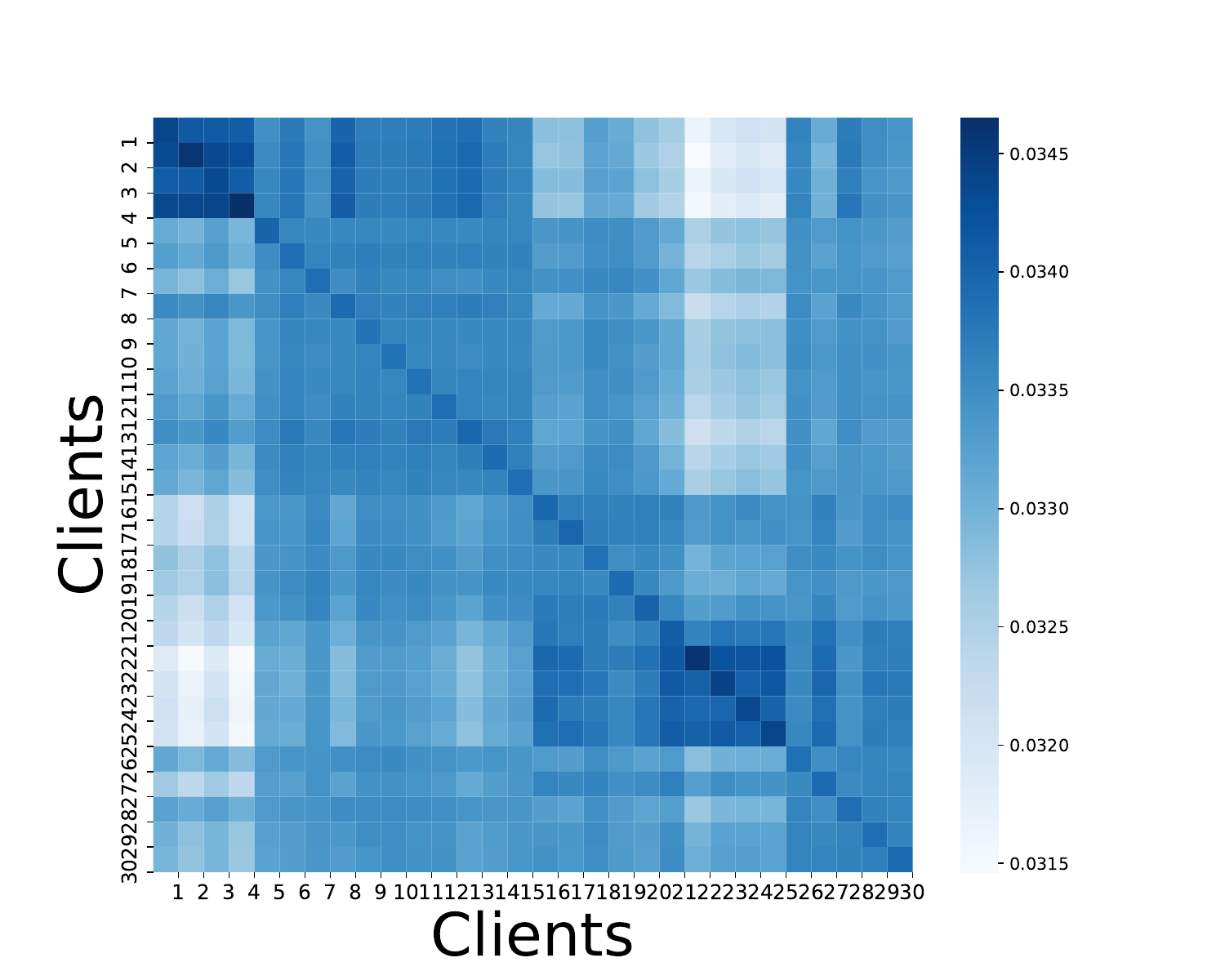}
        \caption{FedIIH of the 2nd latent factor ($K=2$)}
        \label{fig_Questions_05}
    \end{subfigure}
    \caption{Similarity heatmaps on the \textit{Questions} dataset in the overlapping setting with 30 clients.}
    \label{fig_Questions_O}
\end{figure}

\subsection{K.3 Case Study of Different Latent Factors}
\label{case_study}
\begin{figure}
    \centering
    \begin{subfigure}[t]{0.24\textwidth}
        \includegraphics[width=\linewidth]{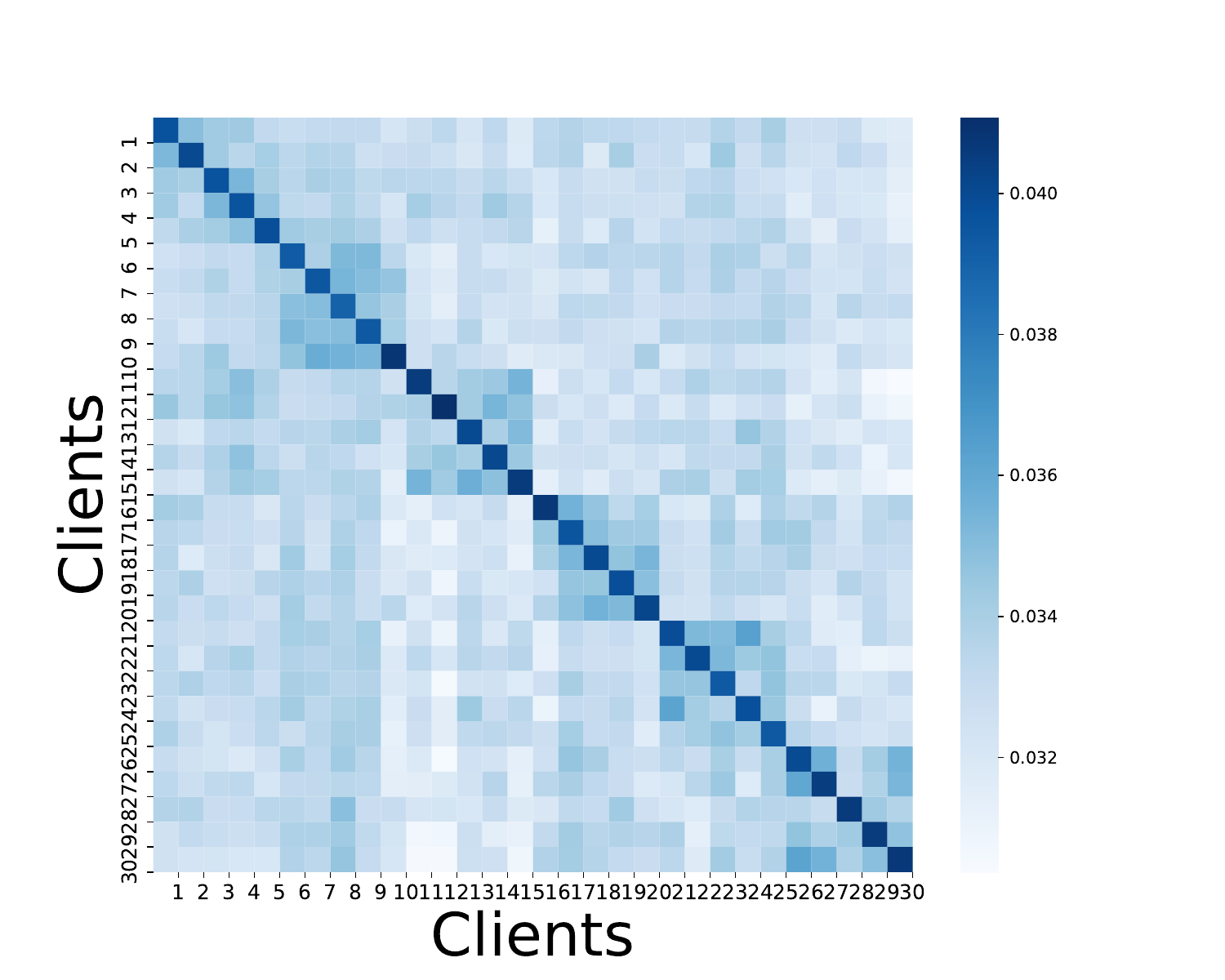}
        \caption{Distr. Sim.}
        \label{fig_case_1}
    \end{subfigure}%
    \hfill
    \begin{subfigure}[t]{0.24\textwidth}
        \includegraphics[width=\linewidth]{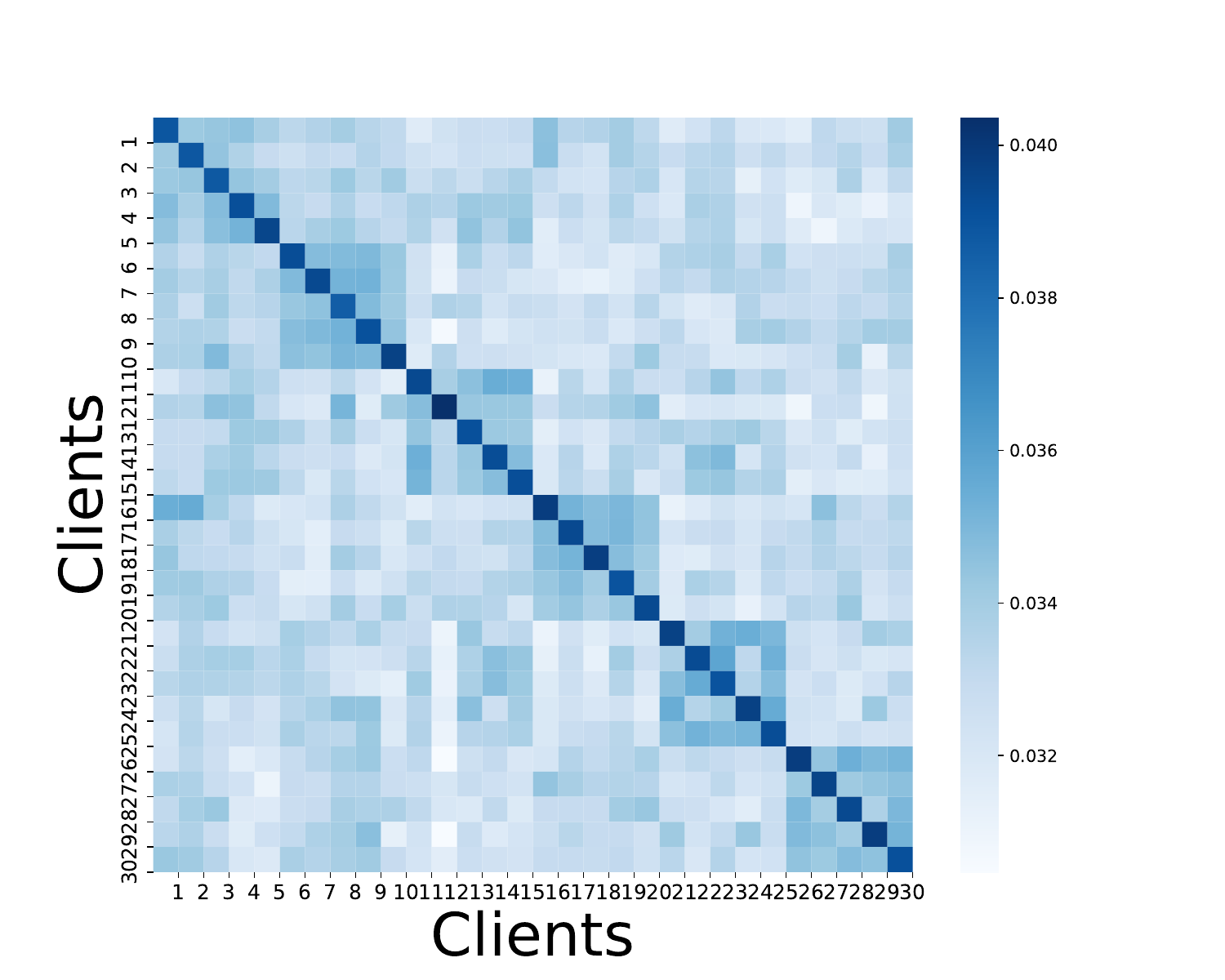}
        \caption{FedIIH of the 1st latent factor ($K=1$)}
        \label{fig_case_2}
    \end{subfigure}%
    \hfill
    \begin{subfigure}[t]{0.24\textwidth}
        \includegraphics[width=\linewidth]{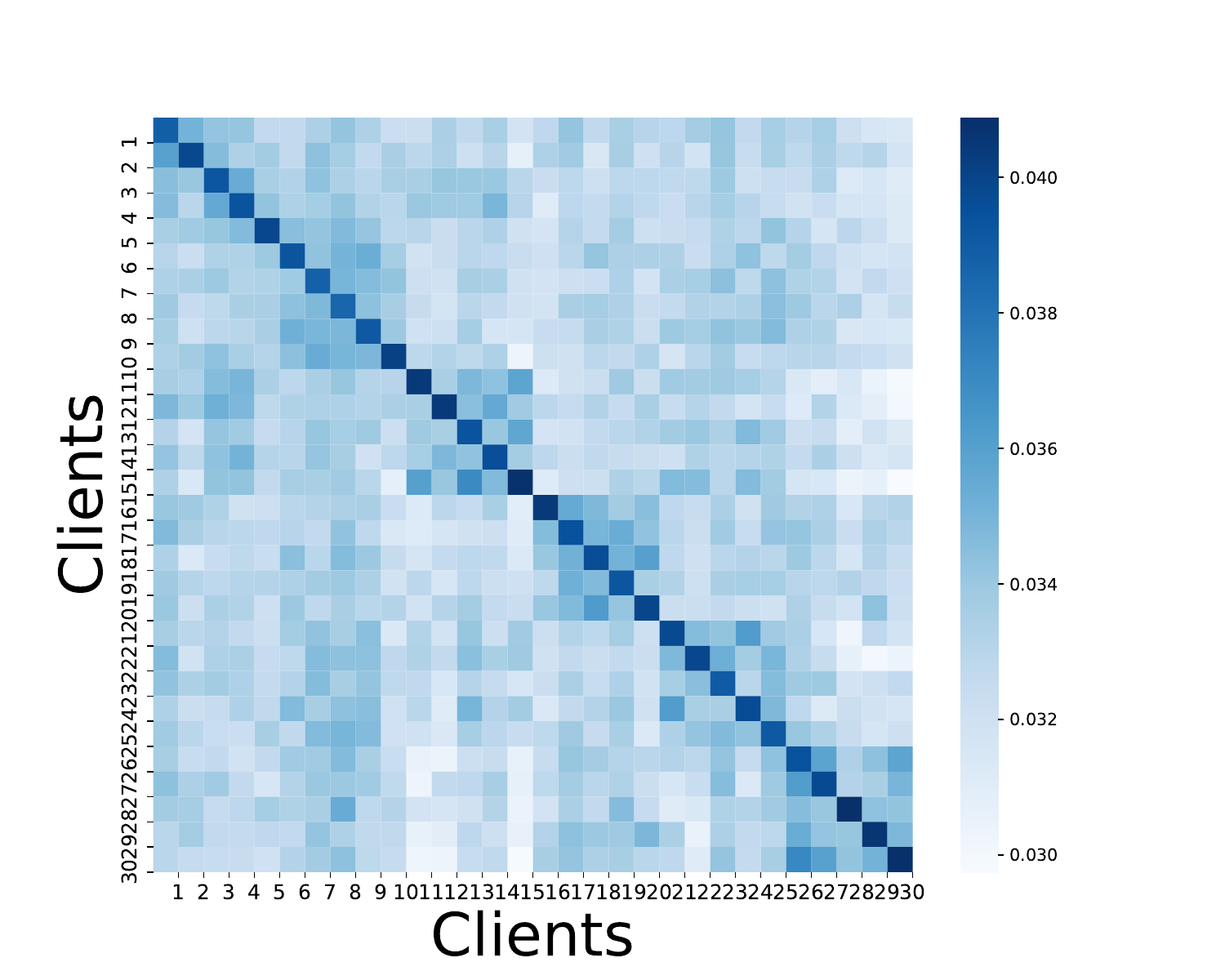}
        \caption{FedIIH of the 1st latent factor ($K=2$)}
        \label{fig_case_3}
    \end{subfigure}
    \hfill
    \begin{subfigure}[t]{0.24\textwidth}
        \includegraphics[width=\linewidth]{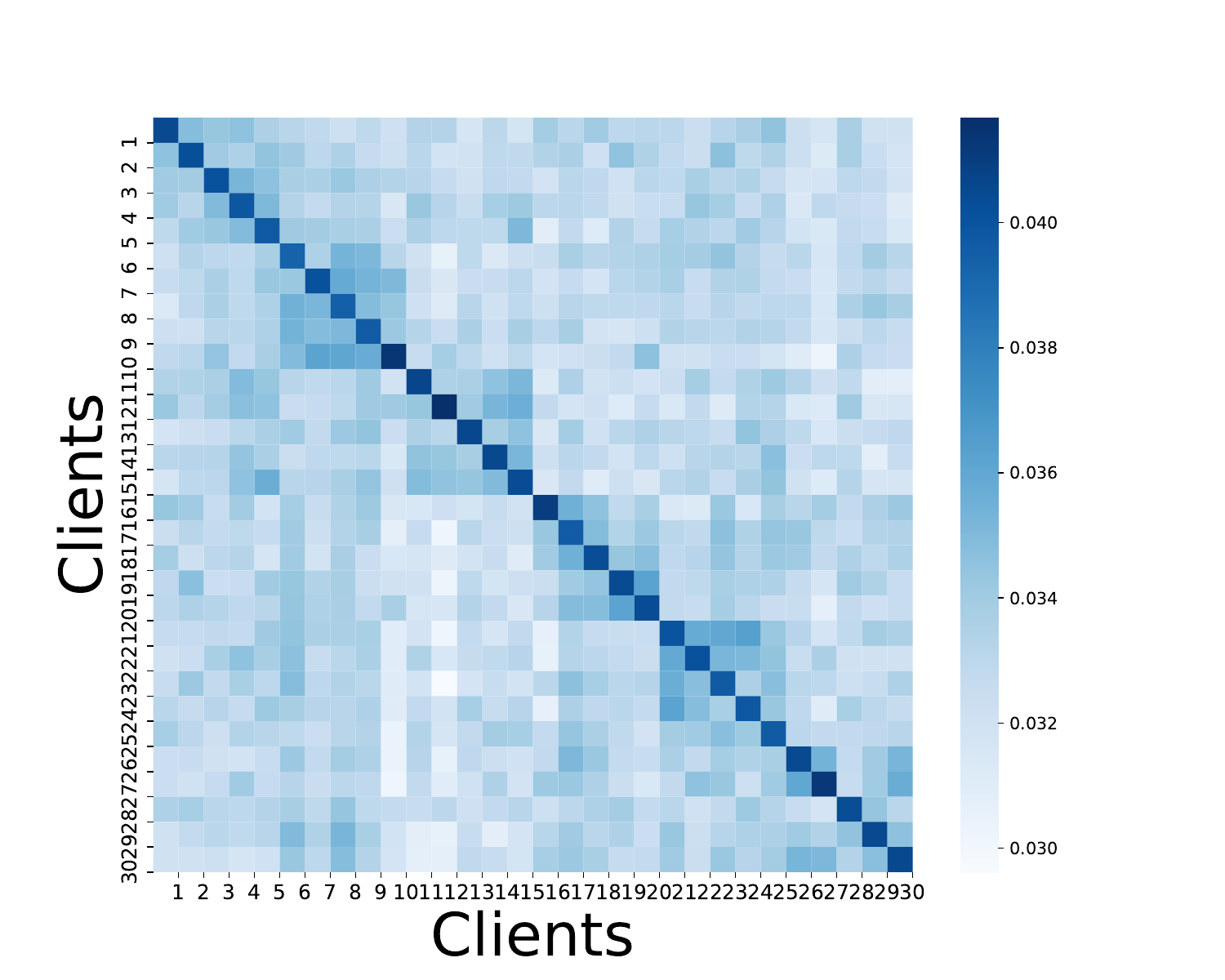}
        \caption{FedIIH of the 2nd latent factor ($K=2$)}
        \label{fig_case_4}
    \end{subfigure}

    \begin{subfigure}[t]{0.24\textwidth}
        \includegraphics[width=\linewidth]{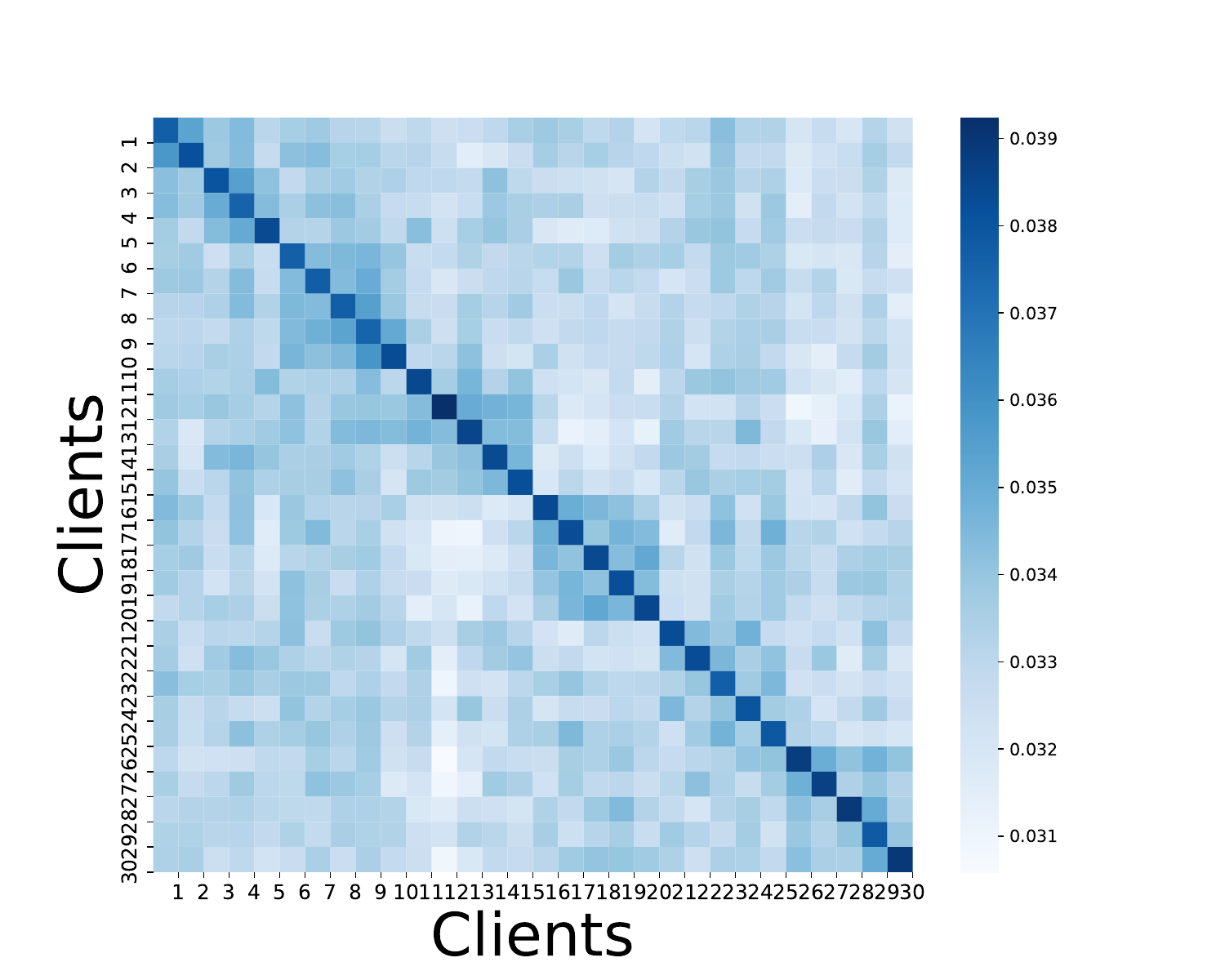}
        \caption{FedIIH of the 1st latent factor ($K=4$)}
        \label{fig_case_5}
    \end{subfigure}
    \hfill
    \begin{subfigure}[t]{0.24\textwidth}
        \includegraphics[width=\linewidth]{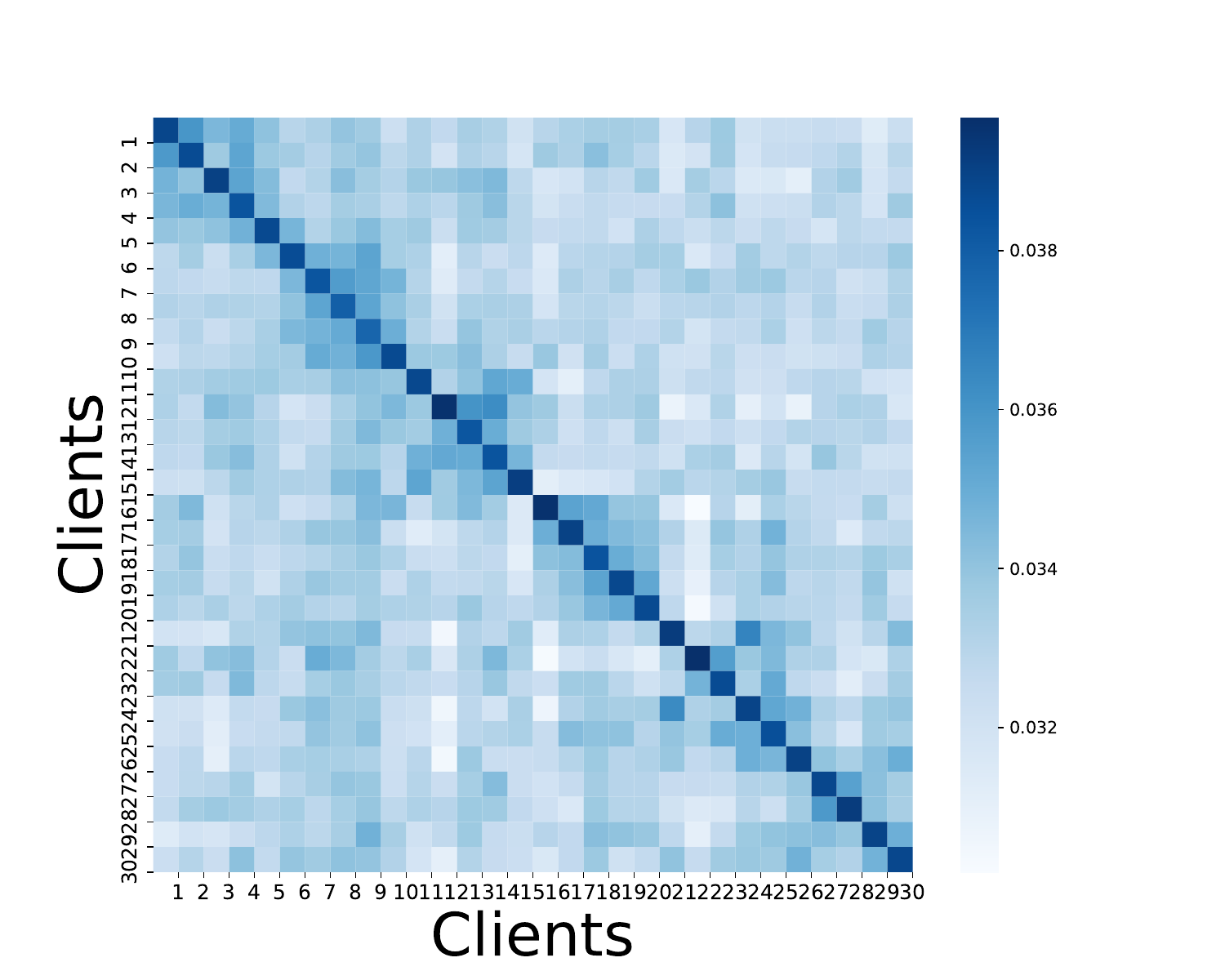}
        \caption{FedIIH of the 2nd latent factor ($K=4$)}
        \label{fig_case_6}
    \end{subfigure}
    \hfill
    \begin{subfigure}[t]{0.24\textwidth}
        \includegraphics[width=\linewidth]{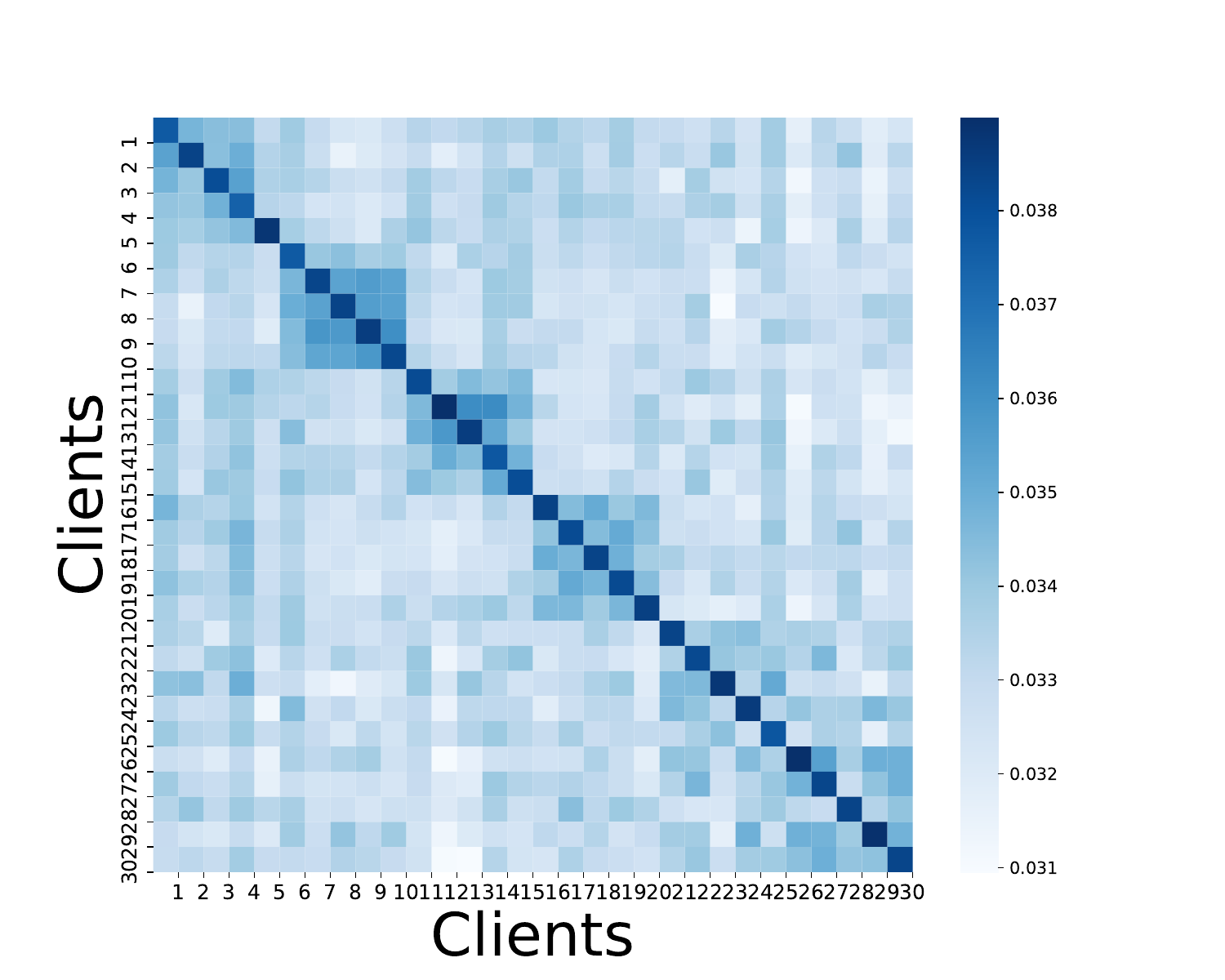}
        \caption{FedIIH of the 3rd latent factor ($K=4$)}
        \label{fig_case_7}
    \end{subfigure}
    \hfill
    \begin{subfigure}[t]{0.24\textwidth}
        \includegraphics[width=\linewidth]{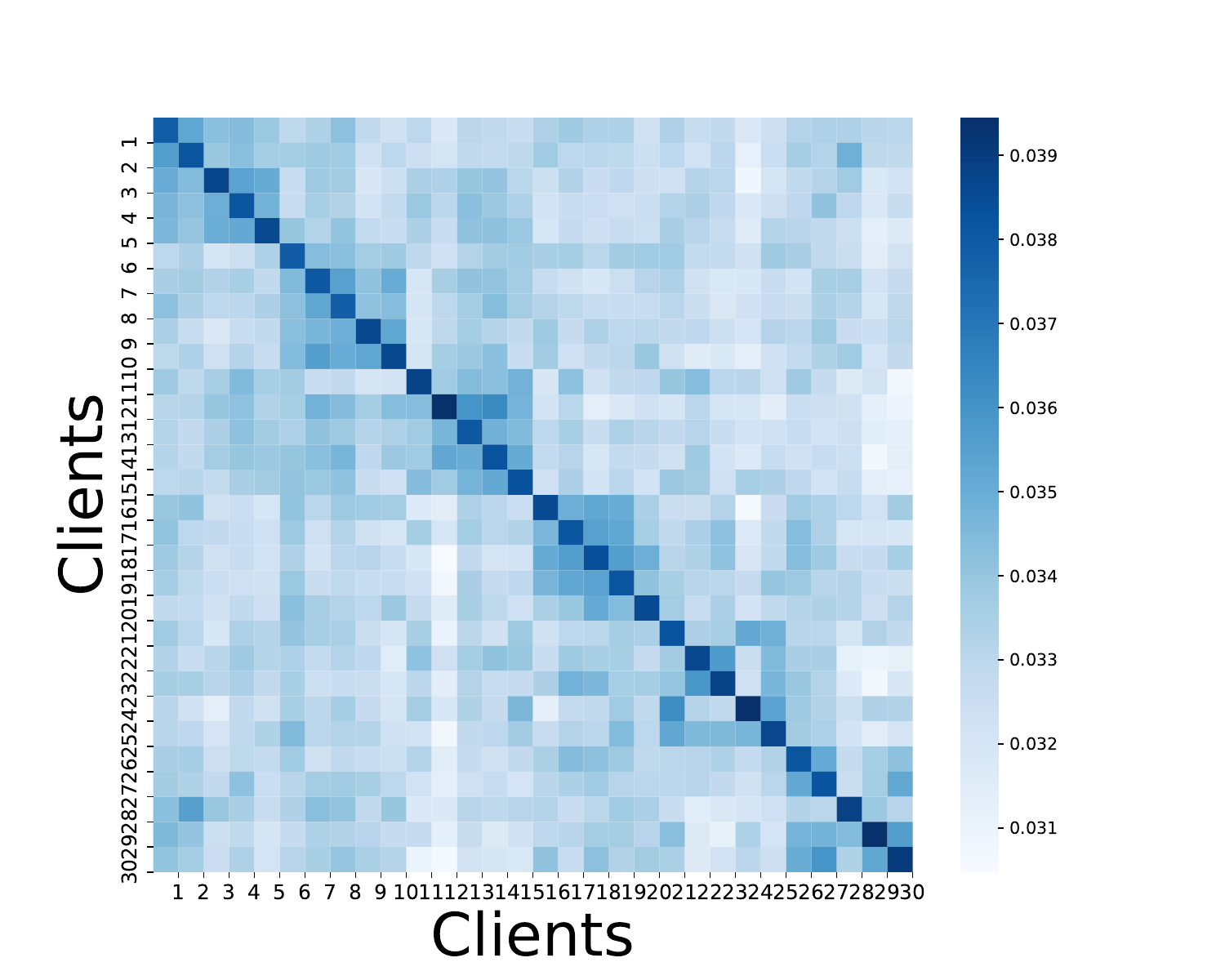}
        \caption{FedIIH of the 4th latent factor ($K=4$)}
        \label{fig_case_8}
    \end{subfigure}

    \caption{Similarity heatmaps on the \textit{Roman-empire} dataset in the overlapping setting with 30 clients. There are indeed differences between different similarity heatmaps when $K$ is set to 1, 2, and 4.}
    \label{fig_case_study}
\end{figure}

Since our proposed FedIIH disentangles the subgraph into several latent factors, one may ask what is the difference between the similarity heatmaps under different latent factors. To illustrate this, we take the \textit{Roman-empire} dataset as an example to perform a case study. As demonstrated in Fig.~\ref{fig_K_7}, the accuracies of our FedIIH obviously increase as $K$ changes from 1, to 2, to 4. Consequently, in Fig.~\ref{fig_case_study}, we present the similarity heatmaps of FedIIH on the \textit{Roman-empire} dataset when $K$ is set to 1, 2, and 4, respectively. From Fig.~\ref{fig_case_study}, it can be observed that there are indeed differences between different similarity heatmaps. Although these differences may seem trivial, they have a very large impact on the separate federation and thus on the final performance.

\subsection{K.4 Disentangled Latent Factors}
Here we provide some insights into the interpretability of the disentangled latent factors. Disentangled latent factors, as widely explored in~\cite{pmlrv97ma19a, NEURIPS2021_b6cda17a, guo2022learning}, are used to explore the reasons why a node is connected to others. In other words, the interpretability of the disentangled latent factors can be considered as the relationship from a given node to one of its neighbors. For example, a user in a social graph is connected to others for various different reasons, such as families, hobbies, studies, and work. Each disentangled latent factor is capable of capturing mutually exclusive information. For example, the correlation plot of DisenGCN on the eight-factor synthetic graph dataset (Figure 3 in~\cite{pmlrv97ma19a}) clearly shows eight diagonal blocks, verifying that the latent factors have indeed been successfully disentangled.

One might ask: Is it possible that the parameter positions corresponding to the same disentangled latent factor differ between clients? We would like to clarify that the parameter positions corresponding to the same latent factor usually remain the same in the disentangled GNNs of different clients. Taking the DisenGCN as an example, there are two crucial processes: node feature projection (Eq.~\eqref{eq1}) and neighborhood routing mechanism (Eq.~\eqref{eq_A_1} and Eq.~\eqref{eq_A_2}). According to Eq.~\eqref{eq1}, we can find that the position of parameters corresponding to the $k$-th latent factor (\textit{i.e.}, $\mathbf{W}^k$) is determined and then fixed by the node feature $\mathbf{x}^i$. Since the distributions of $\mathbf{x}^i$ are similar in different clients, the parameter positions corresponding to $K$ latent factors on each client are determined and then fixed in the same way. Moreover, the parameter positions corresponding to $K$ latent factors are not changed by the neighborhood routing mechanism, because there are no learnable parameters in the neighborhood routing mechanism. Therefore, the parameter positions corresponding to the same latent factor usually remain the same. This can be verified by the experiments, where the standard deviations of our FedIIH are quite small on different datasets (see Tab. 1, 2, 3, and 4).

\subsection{K.5 Disentangled Graph Neural Networks}
The current implementation of our proposed FedIIH is based on the existing method DisenGCN~\cite{pmlrv97ma19a}, which is used to instantiate the inference network in our HVGAE. Instantiation in variational inference is very common in many existing approaches~\cite{ma2019flexible, wang2020graph, wan2021contrastive}. For example, ~\cite{ma2019flexible} and \cite{wang2020graph} use the GCN and Graph Attention neTworks (GAT) to instantiate their inference networks, respectively.

Although many disentangled graph neural networks~\cite{pmlrv97ma19a, NEURIPS2021_b6cda17a, guo2022learning} can be chosen flexibly, we directly choose a simple but popular model (\textit{i.e.}, DisenGCN). This is because DisenGCN is a pioneering work in the field. Moreover, our proposed FedIIH is flexible since other disentangled graph neural networks~\cite{NEURIPS2021_b6cda17a, guo2022learning} can easily be used.

\subsection{K.6 Differences Between FED-PUB and Our FedIIH}
First, in FED-PUB~\cite{baek2023personalized}, each client simply feeds the randomly generated graph to the local model and sends the output to the server. In stark contrast, our FedIIH differs noticeably in that it uses HVGAE to disentangle the subgraph into multiple latent factors and accurately infer the distribution of the subgraph data. Then, each client sends this inferred data distribution to the server.

Second, in FED-PUB, the server only measures the similarities of the clients by computing the cosine similarities of the outputs of local models. Conversely, in our FedIIH, the server specifically measures the similarities by computing the JS divergences of the inferred subgraph data distributions, therefore providing a more accurate and stable measure of client similarities.

In summary, our FedIIH differs noticeably from FED-PUB, as FED-PUB mainly focuses on computing the cosine similarities based on the outputs of local models. Unlike FED-PUB, we measure the similarities of clients by computing the JS divergences of the inferred subgraph data distribution.

\subsection{K.7 Reasons for Setting $\tau$ to 10}
\label{why_tau}
On one hand, $\tau$ is recommended by FED-PUB~\cite{baek2023personalized} to be set to 10. On the other hand, according to the experimental results of grid search, the performance can achieve the best result when $\tau$ is around 10. Tab.~\ref{tau_10} provides the accuracies when using different values of $\tau$.

\begin{table}[t]
    \centering
    \scriptsize
    \caption{The hyperparameter sensitivity analysis of $\tau$. The best results are shown in \textbf{bold}.}
    \label{tau_10}
    \scalebox{0.95}{
    \begin{tabular}{ccccc}
    \hline
    $\tau$ & \begin{tabular}[c]{@{}c@{}}\textit{Cora} Non-overlapping\\ 10 Clients\end{tabular} & \begin{tabular}[c]{@{}c@{}}\textit{Cora} Overlapping\\ 30 Clients\end{tabular} & \begin{tabular}[c]{@{}c@{}}\textit{Roman-empire} Non-overlapping\\ 10 Clients\end{tabular} & \begin{tabular}[c]{@{}c@{}}\textit{Roman-empire} Overlapping\\ 30 Clients\end{tabular} \\ \hline
    1      & 81.58$\pm$0.15                                                                     & 76.74$\pm$0.34                                                                 & 66.12$\pm$0.25                                                                             & 63.15$\pm$0.15                                                                         \\
    2      & 81.61$\pm$0.12                                                                     & 76.52$\pm$0.16                                                                 & 66.41$\pm$0.34                                                                             & 63.23$\pm$0.12                                                                         \\
    3      & 81.76$\pm$0.16                                                                     & 76.69$\pm$0.20                                                                 & 66.34$\pm$0.32                                                                             & 63.19$\pm$0.25                                                                         \\
    4      & 81.80$\pm$0.17                                                                     & 76.72$\pm$0.23                                                                 & 66.40$\pm$0.36                                                                             & 63.29$\pm$0.30                                                                         \\
    5      & 81.65$\pm$0.05                                                                     & 76.66$\pm$0.31                                                                 & 66.36$\pm$0.27                                                                             & 63.26$\pm$0.31                                                                         \\
    6      & 81.82$\pm$0.10                                                                     & 76.75$\pm$0.26                                                                 & 66.21$\pm$0.19                                                                             & 63.28$\pm$0.24                                                                         \\
    7      & 81.75$\pm$0.11                                                                     & 76.80$\pm$0.15                                                                 & 66.37$\pm$0.33                                                                             & 63.13$\pm$0.15                                                                         \\
    8      & 81.81$\pm$0.18                                                                     & 76.79$\pm$0.25                                                                 & 66.36$\pm$0.40                                                                             & 63.22$\pm$0.22                                                                         \\
    9      & 81.82$\pm$0.14                                                                     & 76.74$\pm$0.26                                                                 & 66.34$\pm$0.21                                                                             & 63.30$\pm$0.15                                                                         \\
    10     & \textbf{81.85$\pm$0.09}                                                            & \textbf{76.82$\pm$0.24}                                                        & \textbf{66.44$\pm$0.28}                                                                    & \textbf{63.32$\pm$0.06}                                                                \\ \hline
    \end{tabular}
    }
    \end{table}

\subsection{K.8 Why do local model outputs not accurately reflect the distribution of subgraph?}
Most existing methods~\cite{baek2023personalized, li2023fedgta, zhang2024fedgt} compute the inter-subgraph similarities based on the simplex outputs of local models. However, we argue that the outputs of local models cannot accurately reveal the whole distribution of subgraph data. Here we provide the reason.

According to the universal approximation theorem of neural networks~\cite{bruel2020universal}, even if two neural networks (\textit{e.g.}, two local models on different clients) have different inputs, they can still produce the completely same output. Therefore, the outputs of local models cannot accurately reveal the whole distribution of subgraph data. This can be verified by experiments. In Fig.~3, the similarity heatmap of FED-PUB (Fig.~3b) is quite different from the ground truth (Fig.~3a), which means that the calculated similarities based on the local model outputs cannot accurately reflect the overall distribution of subgraph data.

\section{L. Efficiency Analysis}
In this section, we present the spatial and temporal complexity of our proposed FedIIH and that of two baseline methods (\textit{i.e.}, FedAvg and FED-PUB) on the client and server sides, respectively. Furthermore, we also compare the training time of our proposed FedIIH and that of two baseline methods (\textit{i.e.}, FedAvg and FED-PUB).

\subsection{L.1 Spatial Complexity}
First, the spatial complexity of our FedIIH on each client side and that of two baseline methods (\textit{i.e.}, FedAvg and FED-PUB) are both $\mathcal{O}(n_m \times d + L \times d^2)$, where $n_m$, $d$, and $L$ denote the node number, feature dimensionality, and layer number, respectively. Therefore, the spatial complexity of our FedIIH on each client is the same as that of two baseline methods.
 
Second, the spatial complexity of our method on the server side is $\mathcal{O}\big( M \times d \times ( L \times d^2 + K \times M)\big)$, where $M$ and $K$ denote the number of clients and the number of disentangled latent factors, respectively. The spatial complexity of the baseline method (\textit{i.e.}, FED-PUB) on each client is $\mathcal{O}\big( M \times d \times (  L \times d^2 + M)\big)$. By comparing the spatial complexity of our FedIIH with that of the baseline method (\textit{i.e.}, FED-PUB), we find that the only difference is that the third factor in FedIIH's spatial complexity is $L \times d^2 + K \times M$ while the third factor in FED-PUB's spatial complexity is $L \times d^2 + M$. Since $K \leq 10$ (as mentioned in the Appendix I.5), we can find that the spatial complexity of our FedIIH on the server side is similar to that of the baseline method. 

\subsection{L.2 Temporal Complexity}
First, the temporal complexity of HVGAE on each client is $\mathcal{O}\big(n_m \times d \times (L + n_m + d)\big)$. Since the model deployed on each client is actually a HVGAE, the total temporal complexity of our FedIIH on each client is $\mathcal{O}\big(n_m \times d \times (L + n_m + d)\big)$. The temporal complexity of two baseline methods (\textit{i.e.}, FedAvg and FED-PUB) on each client is $\mathcal{O}\big(n_m \times d \times (L + L  \times d + d)\big)$. By comparing the temporal complexity of our FedIIH with that of two baseline methods (\textit{i.e.}, FedAvg and FED-PUB), the only difference is that the third factor in FedIIH's temporal complexity is $L + n_m + d$ while the third factor in their temporal complexity is $L + L  \times d + d$. Since $n_m \leq L \times d$ in most situations (can be verified in the Appendix I.5), we can find that the temporal complexity of our FedIIH on the client side is similar to that of two baseline methods (\textit{i.e.}, FedAvg and FED-PUB). Therefore, we can find that the introduction of HVGAE does not increase too much computational overhead of our FedIIH.

Second, the temporal complexity of the divergence computation on the server is $\mathcal{O}(K \times M^2 \times d)$, and the total temporal complexity of our FedIIH on the server side is $\mathcal{O}\big( M \times d \times (  K \times M + L \times d)\big)$. The temporal complexity of the baseline method (\textit{i.e.}, FED-PUB) on the server side is $\mathcal{O}\big(M \times d \times (M + L \times d)\big)$. By comparing the temporal complexity of our FedIIH with that of the baseline method (\textit{i.e.}, FED-PUB), we find that the only difference is that the third factor in FedIIH's temporal complexity is $K \times M + L \times d$ while the third factor in FED-PUB's temporal complexity is $M + L \times d$. Since $K \leq 10$ (as mentioned in the Appendix I.5), we can find that the temporal complexity of our FedIIH on the server side is similar to that of the baseline method (\textit{i.e.}, FED-PUB).

\subsection{L.3 Training Time}
We report the training time of one communication round of our FedIIH and two baseline methods (\textit{i.e.}, FedAvg and FED-PUB). The experimental platform is shown in the Appendix I.1. We conduct experiments on a homophilic graph dataset (\textit{i.e.}, \textit{Cora}) and a heterophilic graph dataset (\textit{i.e.}, \textit{Roman-empire}). As shown in Tab.~\ref{training_time}, we can find that our FedIIH is more efficient in practice than the baseline methods (\textit{i.e.}, FedAvg and FED-PUB).

\begin{table}[t]
    \centering
    \scriptsize
    \caption{The training time of one communication round of our FedIIH and two baseline methods.}
    \label{training_time}
    \scalebox{0.999}{
    \begin{tabular}{lcccccc}
    \hline
                  & \multicolumn{3}{c}{\textit{Cora} Nonoverlapping}         & \multicolumn{3}{c}{\textit{Cora} Overlapping}         \\ \cline{2-7} 
    Methods       & 5 Clients         & 10 Clients        & 20 Clients       & 10 Clients       & 30 Clients      & 50 Clients       \\ \hline
    FedAvg        & 20.81s            & 24.90s            & 59.58s           & 30.63s           & 72.55s          & 141.82s          \\
    FED-PUB       & 22.04s            & 27.34s            & 60.31s           & 33.46s           & 80.54s          & 147.84s          \\
    FedIIH (Ours) & \textbf{19.57s}   & \textbf{22.76s}   & \textbf{56.09s}  & \textbf{19.67s}  & \textbf{65.49s} & \textbf{139.03s} \\ \hline
                  & \multicolumn{3}{c}{\textit{Roman-empire} Nonoverlapping} & \multicolumn{3}{c}{\textit{Roman-empire} Overlapping} \\ \cline{2-7} 
    Methods       & 5 Clients         & 10 Clients        & 20 Clients       & 10 Clients       & 30 Clients      & 50 Clients       \\ \hline
    FedAvg        & 18.25s            & 29.85s            & 55.21s           & 28.30s           & 79.12s          & 127.61s          \\
    FED-PUB       & 18.81s            & 28.03s            & 61.75s           & 28.45s           & 83.07s          & 133.05s          \\
    FedIIH (Ours) & \textbf{17.45s}   & \textbf{24.90s}   & \textbf{47.01s}  & \textbf{28.19s}  & \textbf{61.17s} & \textbf{100.10s} \\ \hline
    \end{tabular}
        }
    \end{table}

\section{M. Robustness Analysis}
In this section, we analyze the robustness of our FedIIH in terms of client sparsity and hyperparameter sensitivity, respectively.

\subsection{M.1 Client Sparsity Analysis}
\begin{figure}[]
    \centering
    \begin{subfigure}[t]{0.6\textwidth}
        \includegraphics[width=\linewidth]{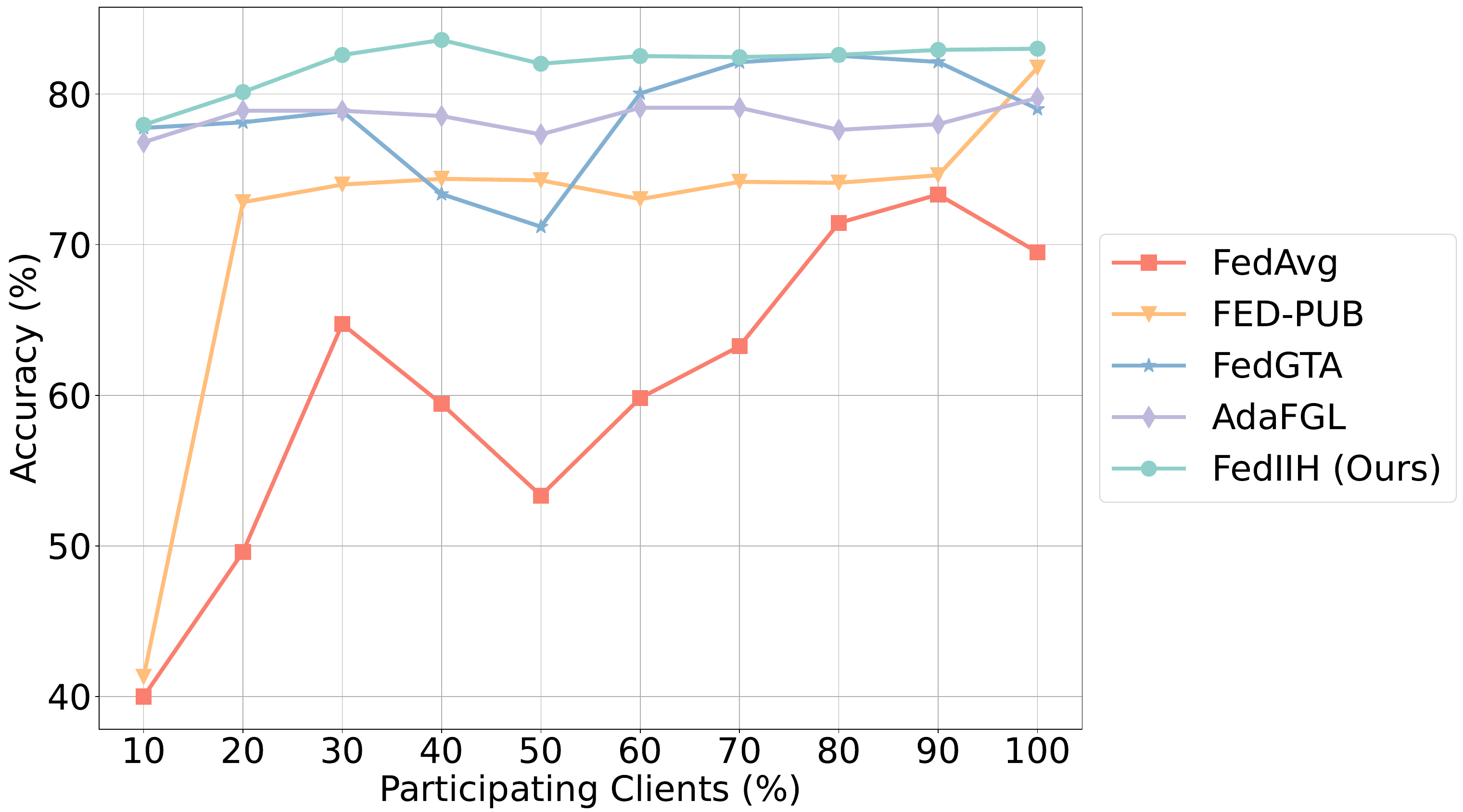}
        \caption{\textit{Cora} Non-overlapping 20 Clients} 
    \end{subfigure}%
    \hfill
    \begin{subfigure}[t]{0.6\textwidth}
        \includegraphics[width=\linewidth]{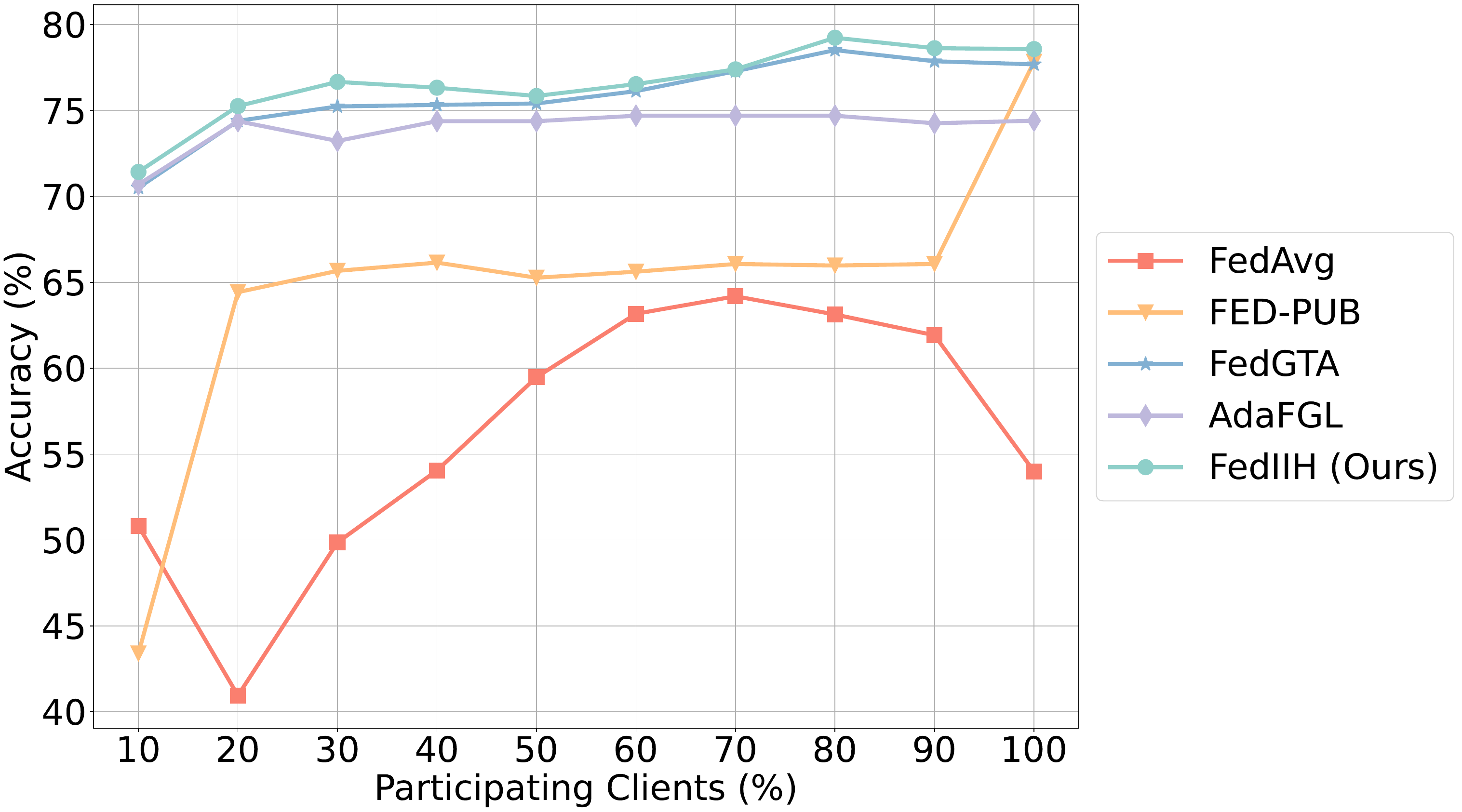}
        \caption{\textit{Cora} Overlapping 50 Clients} 
    \end{subfigure}%
    \hfill
    \begin{subfigure}[t]{0.6\textwidth}
        \includegraphics[width=\linewidth]{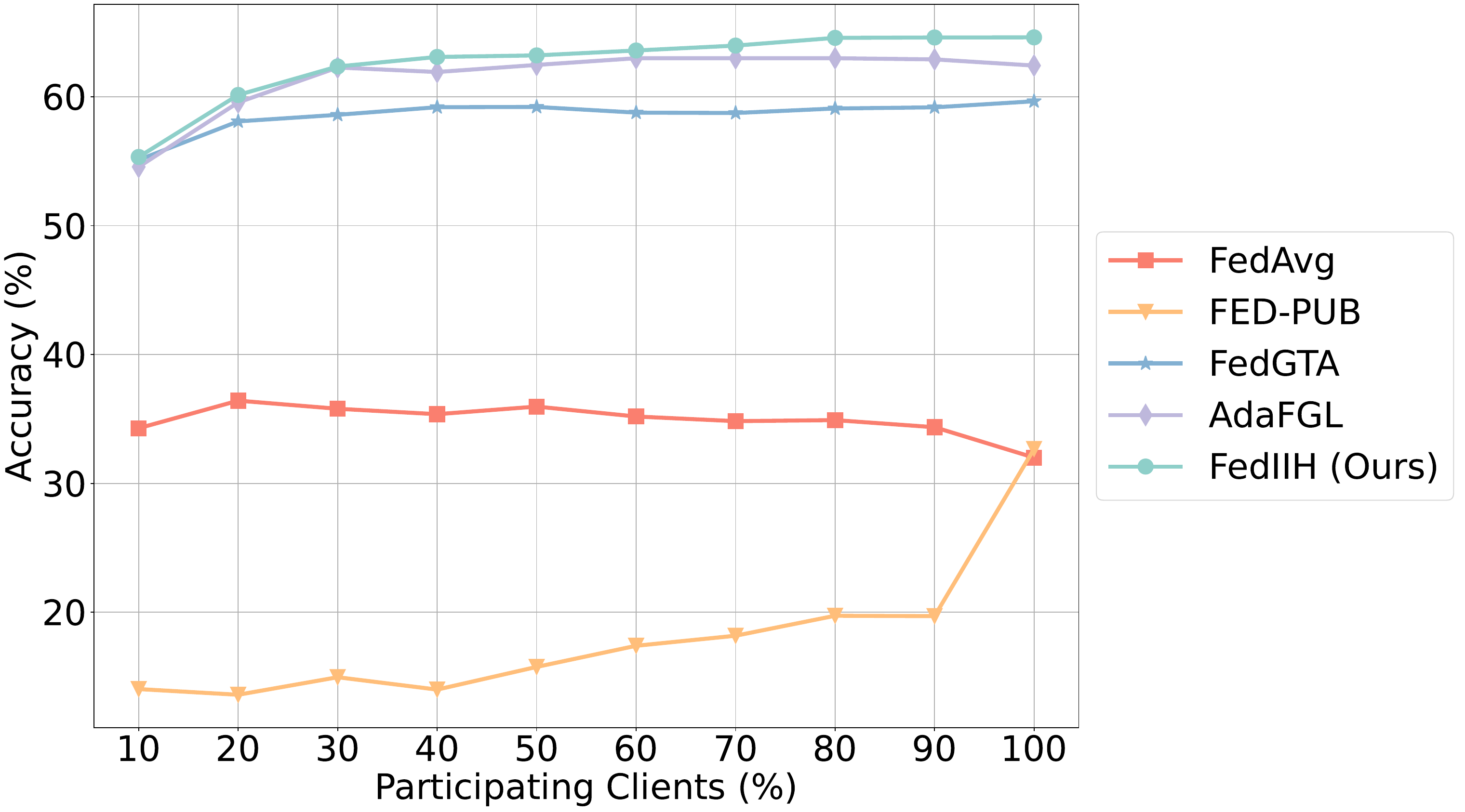}
        \caption{\textit{Roman-empire} Non-overlapping 20 Clients} 
    \end{subfigure}%
    \hfill
    \begin{subfigure}[t]{0.6\textwidth}
        \includegraphics[width=\linewidth]{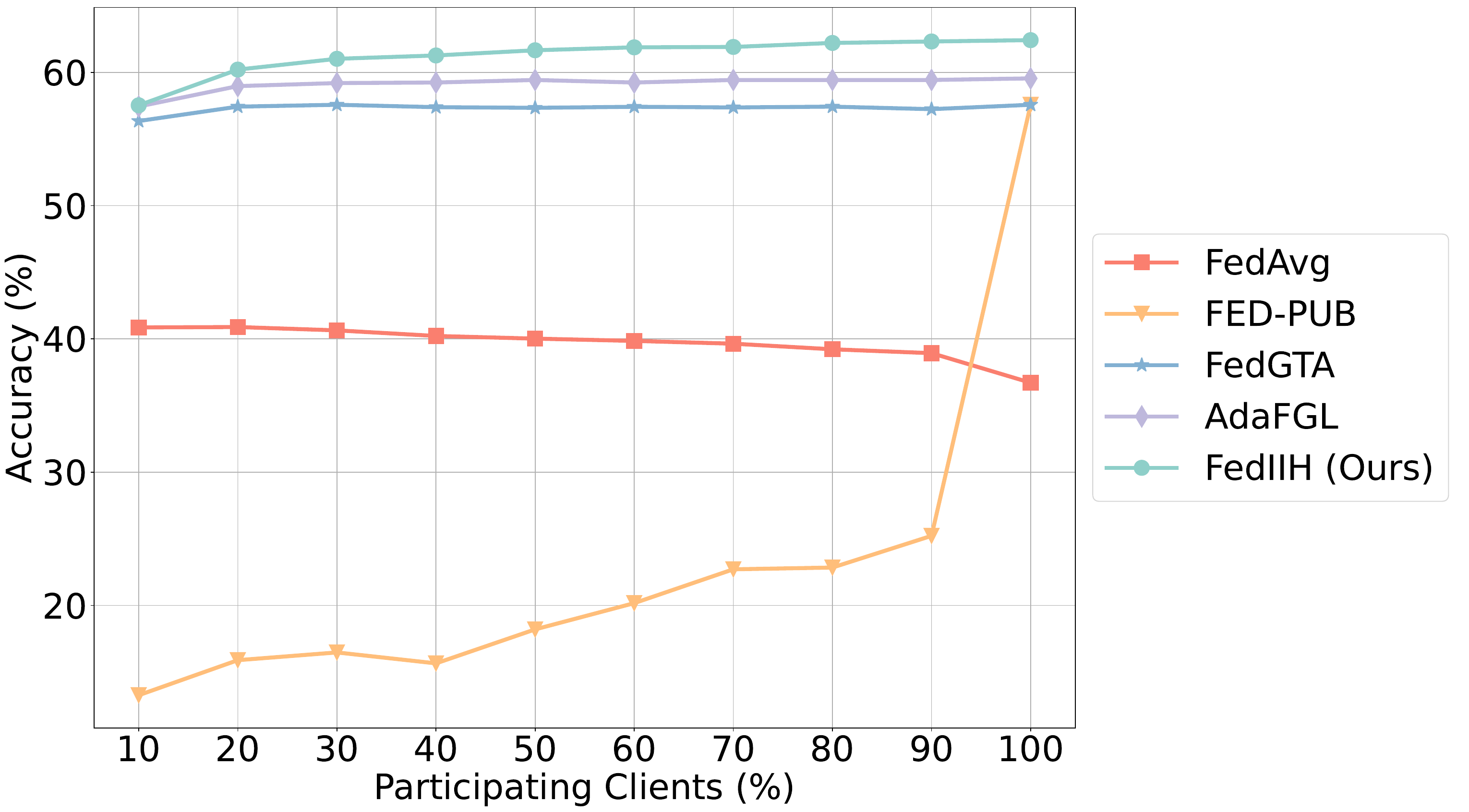}
        \caption{\textit{Roman-empire} Overlapping 50 Clients} 
    \end{subfigure}
    \caption{Performances with different percentages of participating clients.}
    \label{fig_client_sparse}
\end{figure}

Robustness to client sparsity is a critical aspect in evaluating the effectiveness of a similarity-based personalized federated optimization strategy. Here we conduct experiments on a homophilic graph dataset (\textit{i.e.}, \textit{Cora}) and a heterophilic graph dataset (\textit{i.e.}, \textit{Roman-empire}) as the percentage of participating clients increases from 10\% to 100\%. As shown in Fig.~\ref{fig_client_sparse}, the experimental results clearly show that the performance of our FedIIH is more robust and stable than baseline methods, and is not too affected when the number of participating clients decreases.

\subsection{M.2 Hyperparameter Sensitivity Analysis}
\label{complete_Hyperparameter}
There are four vital hyperparameters (\textit{i.e.}, number of latent factors, number of neighborhood routing layers, number of neighborhood routing iterations, and $\tau$) in our proposed FedIIH. Here we perform experiments to analyze the hyperparameter sensitivity.

First, as shown in the Appendix J.2, the variation in performance under different numbers of latent factors (\textit{i.e.}, $K$) is small. Second, as shown in Tab.~\ref{hyperparameter_1},  Tab.~\ref{hyperparameter_2}, and Tab.~\ref{hyperparameter_3}, the variations in performance under different numbers of neighborhood routing layers, different numbers of neighborhood routing iterations, and different values of $\tau$ are all small. These experimental results clearly demonstrate that the performances of our proposed FedIIH are very stable within a given range of hyperparameters, therefore the hyperparameter of our FedIIH can be easily tuned in practical use.

\begin{table}[t]
    \centering
    \scriptsize
\caption{The hyperparameter sensitivity analysis of the number of neighborhood routing layers.}
\label{hyperparameter_1}
\scalebox{0.95}{
    \begin{tabular}{ccccc}
    \hline
    \begin{tabular}[c]{@{}c@{}}\# neighborhood\\ routing layers\end{tabular} & \begin{tabular}[c]{@{}c@{}}\textit{Cora} Non-overlapping\\ 10 Clients\end{tabular} & \begin{tabular}[c]{@{}c@{}}\textit{Cora} Overlapping\\ 30 Clients\end{tabular} & \begin{tabular}[c]{@{}c@{}}\textit{Roman-empire} Non-overlapping\\ 10 Clients\end{tabular} & \begin{tabular}[c]{@{}c@{}}\textit{Roman-empire} Overlapping\\ 30 Clients\end{tabular} \\ \hline
    1                                                                        & 81.19$\pm$0.27                                                                     & 76.18$\pm$0.46                                                                 & 66.44$\pm$0.28                                                                             & 63.32$\pm$0.06                                                                         \\
    2                                                                        & 81.41$\pm$0.15                                                                     & 76.34$\pm$0.31                                                                 & 66.16$\pm$0.41                                                                             & 63.15$\pm$0.15                                                                         \\
    3                                                                        & 81.46$\pm$0.12                                                                     & 76.48$\pm$0.15                                                                 & 66.23$\pm$0.36                                                                             & 63.12$\pm$0.21                                                                         \\
    4                                                                        & 81.85$\pm$0.09                                                            & 76.67$\pm$0.18                                                                 & 66.15$\pm$0.46                                                                    & 63.15$\pm$0.42                                                                         \\
    5                                                                        & 81.62$\pm$0.08                                                                     & 76.82$\pm$0.24                                                                 & 66.11$\pm$0.55                                                                             & 63.10$\pm$0.24                                                                         \\
    6                                                                        & 81.37$\pm$0.24                                                                     & 76.46$\pm$0.45                                                                 & 66.08$\pm$0.39                                                                             & 63.04$\pm$0.38                                                                         \\ \hline
    max - min                                                                & 0.66                                                                               & 0.64                                                                           & 0.36                                                                                       & 0.28                                                                                   \\ \hline
    \end{tabular}
}
    \end{table}

    \begin{table}[t]
        \centering
        \scriptsize
        \caption{The hyperparameter sensitivity analysis of the number of neighborhood routing iterations.}
        \label{hyperparameter_2}
        \scalebox{0.95}{
        \begin{tabular}{ccccc}
        \hline
        \begin{tabular}[c]{@{}c@{}}\# neighborhood\\ routing iterations\end{tabular} & \begin{tabular}[c]{@{}c@{}}\textit{Cora} Non-overlapping\\ 10 Clients\end{tabular} & \begin{tabular}[c]{@{}c@{}}\textit{Cora} Overlapping\\ 30 Clients\end{tabular} & \begin{tabular}[c]{@{}c@{}}\textit{Roman-empire} Non-overlapping\\ 10 Clients\end{tabular} & \begin{tabular}[c]{@{}c@{}}\textit{Roman-empire} Overlapping\\ 30 Clients\end{tabular} \\ \hline
        2                                                                            & 81.24$\pm$0.50                                                                     & 76.22$\pm$0.30                                                                 & 66.04$\pm$0.43                                                                             & 63.01$\pm$0.45                                                                         \\
        3                                                                            & 81.38$\pm$0.45                                                                     & 76.31$\pm$0.44                                                                 & 66.10$\pm$0.46                                                                             & 63.10$\pm$0.34                                                                         \\
        4                                                                            & 81.45$\pm$0.41                                                                     & 76.37$\pm$0.45                                                                 & 66.16$\pm$0.44                                                                             & 63.18$\pm$0.29                                                                         \\
        5                                                                            & 81.57$\pm$0.35                                                                     & 76.45$\pm$0.41                                                                 & 66.15$\pm$0.35                                                                             & 63.24$\pm$0.17                                                                         \\
        6                                                                            & 81.85$\pm$0.09                                                                     & 76.82$\pm$0.24                                                                 & 66.44$\pm$0.28                                                                             & 63.32$\pm$0.06                                                                         \\
        7                                                                            & 81.36$\pm$0.42                                                                     & 76.68$\pm$0.39                                                                 & 66.24$\pm$0.29                                                                             & 63.12$\pm$0.22                                                                         \\ \hline
        max - min                                                                    & 0.61                                                                               & 0.60                                                                           & 0.40                                                                                       & 0.31                                                                                   \\ \hline
        \end{tabular}
        }
        \end{table}

        \begin{table}[t]
            \centering
            \scriptsize
            \caption{The hyperparameter sensitivity analysis of $\tau$.}
            \label{hyperparameter_3}
            \scalebox{0.95}{
            \begin{tabular}{ccccc}
            \hline
            $\tau$    & \begin{tabular}[c]{@{}c@{}}\textit{Cora} Non-overlapping\\ 10 Clients\end{tabular} & \begin{tabular}[c]{@{}c@{}}\textit{Cora} Overlapping\\ 30 Clients\end{tabular} & \begin{tabular}[c]{@{}c@{}}\textit{Roman-empire} Non-overlapping\\ 10 Clients\end{tabular} & \begin{tabular}[c]{@{}c@{}}\textit{Roman-empire} Overlapping\\ 30 Clients\end{tabular} \\ \hline
            1         & 81.58$\pm$0.15                                                                     & 76.74$\pm$0.34                                                                 & 66.12$\pm$0.25                                                                             & 63.15$\pm$0.15                                                                         \\
            2         & 81.61$\pm$0.12                                                                     & 76.52$\pm$0.16                                                                 & 66.41$\pm$0.34                                                                             & 63.23$\pm$0.12                                                                         \\
            3         & 81.76$\pm$0.16                                                                     & 76.69$\pm$0.20                                                                 & 66.34$\pm$0.32                                                                             & 63.19$\pm$0.25                                                                         \\
            4         & 81.80$\pm$0.17                                                                     & 76.72$\pm$0.23                                                                 & 66.40$\pm$0.36                                                                             & 63.29$\pm$0.30                                                                         \\
            5         & 81.65$\pm$0.05                                                                     & 76.66$\pm$0.31                                                                 & 66.36$\pm$0.27                                                                             & 63.26$\pm$0.31                                                                         \\
            6         & 81.82$\pm$0.10                                                                     & 76.75$\pm$0.26                                                                 & 66.21$\pm$0.19                                                                             & 63.28$\pm$0.24                                                                         \\
            7         & 81.75$\pm$0.11                                                                     & 76.80$\pm$0.15                                                                 & 66.37$\pm$0.33                                                                             & 63.13$\pm$0.15                                                                         \\
            8         & 81.81$\pm$0.18                                                                     & 76.79$\pm$0.25                                                                 & 66.36$\pm$0.40                                                                             & 63.22$\pm$0.22                                                                         \\
            9         & 81.82$\pm$0.14                                                                     & 76.74$\pm$0.26                                                                 & 66.34$\pm$0.21                                                                             & 63.30$\pm$0.15                                                                         \\
            10        & 81.85$\pm$0.09                                                                     & 76.82$\pm$0.24                                                                 & 66.44$\pm$0.28                                                                             & 63.32$\pm$0.06                                                                         \\ \hline
            max - min & 0.27                                                                               & 0.30                                                                           & 0.32                                                                                       & 0.19                                                                                   \\ \hline
            \end{tabular}
            }
            \end{table}


\section{N. Broader Impact}
\label{broader_impact}
Our work could have the following positive impacts: (1) We provide a new method for GFL to deal with the inter-intra heterogeneity. (2) Our proposed method can greatly improve the performance of GFL on homophilic and heterophilic graph datasets in both non-overlapping and overlapping settings.

The proposed method can be used for both good and bad, similar to many other FL methods. Note that most existing methods and our proposal are not immune to such misuse. We believe that such a problem can be solved in the future, although we do not have an optimal solution.

In summary, we believe that our proposed method can benefit society because many important real-world graphs face inter-intra heterogeneity when performing the GFL. Therefore, they can benefit from our proposed FedIIH.

\section*{Reproducibility Checklist}

\begin{enumerate}

    \item[] 1. This paper includes a conceptual outline and/or pseudocode description of AI methods introduced. (yes/partial/no/NA)
    \item[] Answer: \answerYes{} 
    \item[] Justification: Our paper includes a conceptual outline and pseudocode description of methods in this Appendix.

    \item[] 2. This paper clearly delineates statements that are opinions, hypothesis, and speculation from objective facts and results. (yes/no)
    \item[] Answer: \answerYes{} 
    \item[] Justification: First, all the formulas and proofs in the paper are numbered and cross-referenced. Second, due to the space limitation, the detailed proof of ELBO is presented in the Appendix C. Third, we provide the extensive experimental results.

    \item[] 3. This paper provides well marked pedagogical references for less-familiare readers to gain background necessary to replicate the paper. (yes/no)
    \item[] Answer: \answerYes{} 
    \item[] Justification: We provide the detailed preliminaries.

    \item[] 4. Does this paper make theoretical contributions? (yes/no)
    \item[] Answer: \answerYes{} 
    \item[] Justification: This paper make theoretical contributions related to the ELBO.
    
    \item[] 5. All assumptions and restrictions are stated clearly and formally. (yes/partial/no)
    \item[] Answer: \answerYes{} 
    \item[] Justification: This paper clearly and formally states all assumptions and restrictions.
    
    \item[] 6. All novel claims are stated formally (e.g., in theorem statements). (yes/partial/no)
    \item[] Answer: \answerYes{} 
    \item[] Justification: This paper clearly and formally states all all novel claims.

    \item[] 7. Proofs of all novel claims are included. (yes/partial/no)
    \item[] Answer: \answerYes{} 
    \item[] Justification: We provide the proofs in this Appendix.

    \item[] 8. Proof sketches or intuitions are given for complex and/or novel results. (yes/partial/no)
    \item[] Answer: \answerYes{} 
    \item[] Justification: We provide the proofs in this Appendix.

    \item[] 9. Appropriate citations to theoretical tools used are given. (yes/partial/no)
    \item[] Answer: \answerYes{} 
    \item[] Justification: We provide all the citations.

    \item[] 10. All theoretical claims are demonstrated empirically to hold. (yes/partial/no/NA)
    \item[] Answer: \answerYes{} 
    \item[] Justification: The experimental results verifiy all theoretical claims.
    
    \item[] 11. All experimental code used to eliminate or disprove claims is included. (yes/no/NA)
    \item[] Answer: \answerYes{} 
    \item[] Justification: Our code is available at \url{https://github.com/blgpb/FedIIH}. We also provide a README.md file to describe the detailed instructions on how to replicate the results.
    
    \item[] 12. Does this paper rely on one or more datasets? (yes/no)
    \item[] Answer: \answerYes{} 
    \item[] Justification: We use several datasets.
    
    \item[] 13. A motivation is given for why the experiments are conducted on the selected datasets (yes/partial/no/NA)
    \item[] Answer: \answerYes{} 
    \item[] Justification: We explain it in the paper.

    \item[] 14. All novel datasets introduced in this paper are included in a data appendix. (yes/partial/no/NA)
    \item[] Answer: \answerYes{} 
    \item[] Justification: We provide them in this Appendix.
    
    \item[] 15. All novel datasets introduced in this paper will be made publicly available upon publication of the paper with a license that allows free usage for research purposes. (yes/partial/no/NA)
    \item[] Answer: \answerYes{} 
    \item[] Justification: We provide them in this Appendix.

    \item[] 16. All datasets drawn from the existing literature (potentially including authors’ own previously published work) are accompanied by appropriate citations. (yes/no/NA)
    \item[] Answer: \answerYes{} 
    \item[] Justification: We cite all the papers that produced the datasets.

    \item[] 17. All datasets drawn from the existing literature (potentially including authors’ own previously published work) are publicly available. (yes/partial/no/NA)
    \item[] Answer: \answerYes{} 
    \item[] Justification: All datasets are publicly available.
    
    \item[] 18. Does this paper include computational experiments? (yes/no)
    \item[] Answer: \answerYes{} 
    \item[] Justification: We provide them in the paper.

    \item[] 19. Any code required for pre-processing data is included in the appendix. (yes/partial/no).
    \item[] Answer: \answerYes{} 
    \item[] Justification: Our code is available at \url{https://github.com/blgpb/FedIIH}.
    
    \item[] 20. All source code required for conducting and analyzing the experiments is included in a code appendix. (yes/partial/no)
    \item[] Answer: \answerYes{} 
    \item[] Justification: Our code is available at \url{https://github.com/blgpb/FedIIH}.
    
    \item[] 21. All source code required for conducting and analyzing the experiments will be made publicly available upon publication of the paper with a license that allows free usage for research purposes. (yes/partial/no)
    \item[] Answer: \answerYes{} 
    \item[] Justification: All source code will be made publicly available upon publication of the paper with a license that allows free usage for research purposes.

    \item[] 22. All source code implementing new methods have comments detailing the implementation, with references to the paper where each step comes from (yes/partial/no)
    \item[] Answer: \answerYes{} 
    \item[] Justification: We provide them in this Appendix.

    \item[] 23. If an algorithm depends on randomness, then the method used for setting seeds is described in a way sufficient to allow replication of results. (yes/partial/no/NA)
    \item[] Answer: \answerYes{} 
    \item[] Justification: Our code is available at \url{https://github.com/blgpb/FedIIH}. We also provide a README.md file to describe the detailed instructions on how to replicate the results.

    \item[] 24. This paper specifies the computing infrastructure used for running experiments (hardware and software), including GPU/CPU models; amount of memory; operating system; names and versions of relevant software libraries and frameworks. (yes/partial/no)
    \item[] Answer: \answerYes{} 
    \item[] Justification: We provide them in this Appendix.
    
    \item[] 25. This paper formally describes evaluation metrics used and explains the motivation for choosing these metrics. (yes/partial/no)
    \item[] Answer: \answerYes{} 
    \item[] Justification: We provide them in this Appendix.

    \item[] 26. This paper states the number of algorithm runs used to compute each reported result. (yes/no)
    \item[] Answer: \answerYes{} 
    \item[] Justification: We provide them in the paper.

    \item[] 27. Analysis of experiments goes beyond single-dimensional summaries of performance (e.g., average; median) to include measures of variation, confidence, or other distributional information. (yes/no)
    \item[] Answer: \answerYes{} 
    \item[] Justification: We provide them in this Appendix.

    \item[] 28. The significance of any improvement or decrease in performance is judged using appropriate statistical tests (e.g., Wilcoxon signed-rank). (yes/partial/no)
    \item[] Answer: \answerYes{} 
    \item[] Justification: We provide them in the paper.
    
    \item[] 29. This paper lists all final (hyper-)parameters used for each model/algorithm in the paper’s experiments. (yes/partial/no/NA)
    \item[] Answer: \answerYes{} 
    \item[] Justification: We provide them in this Appendix.
    
    \item[] 30. This paper states the number and range of values tried per (hyper-) parameter during development of the paper, along with the criterion used for selecting the final parameter setting. (yes/partial/no/NA)
    \item[] Answer: \answerYes{} 
    \item[] Justification: We provide them in this Appendix.

\end{enumerate}
\end{document}